\documentclass[11pt,letter,twoside,fleqn,USenglish]{article}
\usepackage{graphicx}
\usepackage{cite}
\usepackage{amsmath}
\usepackage{amsthm}
\usepackage{amssymb}
\usepackage{enumerate}
\usepackage{txfonts}
\usepackage{hyperref}
\usepackage{bm}
\usepackage{color}
\usepackage[margin=1in]{geometry}
\usepackage{array}
\usepackage{multirow}

\usepackage{tikz}
\usepackage[ruled]{algorithm2e}
\usepackage[capitalize,noabbrev]{cleveref}
\usepackage{subfig}
\usepackage{./ucb-titlepage}
\usepackage{./fancyheadings}

\numberwithin{equation}{section}

\newcounter{ecount}
  {\end{list}}

\newenvironment{icompact}{
  \vspace*{-4pt}
  \begin{list}{$\bullet$}{
    \parsep 1pt
    \partopsep 1pt
    \topsep 3pt
    \itemsep 0pt plus 1pt}}%
  {\end{list}}

  {\end{list}}

  {\end{list}}

  {\end{list}}

\makeatletter
\long\def\unmarkedfootnote#1{{\long\def\@makefntext##1{##1}\footnotetext{#1}}}
\makeatother

\newtheorem{theorem}{Theorem}
\newtheorem{lemma}[theorem]{Lemma}

\newtheorem{cor}[theorem]{Corollary}

\newcommand{\ssq}{\blacktriangleleft}
\newcommand{\R}{\mathbb{R}}
\newcommand{\F}{\mathrm{F}}
\newcommand{\GL}{\mathrm{GL}}
\newcommand{\DV}{\mathrm{DV}}
\newcommand{\DM}{\mathrm{DM}}

\newcommand{\WDS}{\mathrm{WDS}}
\newcommand{\MM}{\mathrm{MDM}}
\newcommand{\row}{\mathrm{row}\,}
\newcommand{\col}{\mathrm{col}\,}
\newcommand{\cols}{\mathrm{col}}
\newcommand{\rank}{\mathrm{rk}\,}
\newcommand{\Null}{\mathrm{null}\,}
\newcommand{\nulls}{\mathrm{null}}

\newcommand{\image}{\mathrm{image}\,}
\newcommand{\trace}{\mathrm{trace}\,}
\newcommand{\Span}{\mathrm{span}\,}
\newcommand{\proj}{\mathrm{proj}}
\newcommand{\sign}{\mathrm{sign}\,}
\newcommand{\SM}{\mathrm{SM}}
\newcommand{\St}{\mathrm{St}}
\newcommand{\di}{\mathrm{d}}
\newcommand{\dmu}{\mathrm{d}\mu}
\newcommand{\dell}{\mathrm{d}\ell}
\newcommand{\dL}{\mathrm{d}\mathcal{L}}
\newcommand{\free}{{\mathrm{free}}}
\newcommand{\fiber}{{\mathrm{fiber}}}
\newcommand{\off}{{\mathrm{off}}}
\newcommand{\strat}{{\mathrm{strat}}}
\newcommand{\barc}{{\mathrm{barc}}}
\newcommand{\conn}{{\mathrm{conn}}}
\newcommand{\swap}{{\mathrm{swap}}}
\newcommand{\ThetaO}{\Theta_{\mathrm{O}}}
\newcommand{\ThetaT}{\Theta_{\mathrm{T}}}
\newcommand{\ThetaTO}{\Theta_{\mathrm{T+}}}
\newcommand{\DO}{D_{\mathrm{O}}}
\newcommand{\DT}{D_{\mathrm{T}}}

\newcommand{\lm}{\mbox{\hspace{-.37in}}}  

\title{
  {\color[rgb]{.85,.64,.12}\resizebox*{\unitlength}{\baselineskip}{\rotatebox{90}{$\}$}}}\\
  ~The Geometry of the Set of Equivalent Linear Neural Networks\\[-.1in]
  {\color[rgb]{.85,.64,.12}~~\resizebox*{\unitlength}{\baselineskip}{\rotatebox{-90}{$\}$}}}\\
}

\author{Jonathan Richard Shewchuk and Sagnik Bhattacharya}

\trnumber{}

\coverfig{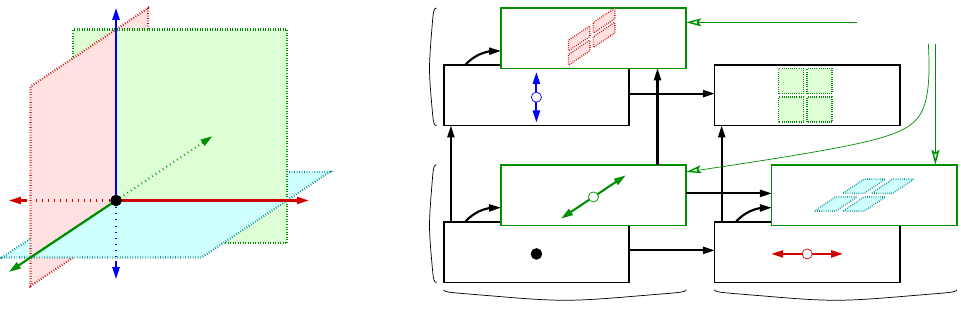}

\citationinfo{
}

\copyrightnotice{
Copyright 2026 Jonathan Richard Shewchuk and Sagnik Bhattacharya 
}

\support{
Supported in part by
the National Science Foundation under Award CCF-1909204
and in part by an Alfred P.\ Sloan Research Fellowship.
The claims in this document are those of the authors.
They are not endorsed by the sponsors or the U.S.\ Government.
}

\keywords{linear neural network, real algebraic variety, stratification,
          multilinear algebra, product of matrices, persistence barcode,
          fundamental subspace, fundamental theorem of linear algebra,
          real algebraic geometry, discrete geometry}

\abstract{
We characterize the geometry of the set of all weight vectors for which
a fully-connected linear neural network computes
the same linear transformation $W$.
Said another way, the set of all sequences of matrices
(of predetermined sizes) whose product is $W$.
This set of weight vectors is
an algebraic variety in the Euclidean {\em weight space}, called
the {\em fiber} of $W$ under the {\em matrix multiplication map}.
Sometimes this variety is a manifold, but usually not.
The {\em rank stratification} of a fiber is a natural partition of the fiber
into smooth manifolds of various dimensions called {\em strata}.
We derive the dimensions of these strata and
characterize how they are connected to each other.
They satisfy the {\em frontier condition}:
if a stratum intersects the closure of another stratum, then
the former stratum is a subset of the closure of the latter stratum.
This subset relationship can be expressed as a partial order with
a single minimal element,
a stratum that is included in the closure of every other stratum.
Hence the strata can be arranged in a hierarchy represented by
a directed acyclic graph that represents that partial order;
we give an algorithm that computes that graph.
A key result describes the relationship between this partial order and
the ranks of certain matrices in the network.
We derive the tangent and normal subspaces of the strata, and
we construct elegant bases for those subspaces.

\vspace{0.1in}

Each stratum represents a different pattern of information flow through
the network, expressed as a {\em barcode},
borrowed from the study of persistent homology.
The topology of a stratum depends solely on its barcode.
So does its geometry, up to a linear transformation in weight space.
We define {\em moves} in weight space, which
replace one weight vector with another on the same fiber, thereby
modifying the neural network's weights
without changing the linear transformation that the network computes.
Some moves move from one stratum to another stratum of the fiber.
Connections among the strata are best understood through
a subset of these moves called {\em barcode moves}, which can be expressed as
simple modifications of the barcodes.

\vspace{0.1in}

Of independent interest is a concept underlying these results:
the {\em fundamental subspaces of a linear neural network}.
These are hidden-layer analogues of the fundamental subspaces of a matrix
(the rowspace, nullspace, columnspace, and left nullspace) that show
how information flows through a linear neural network.
We summarize the properties of these fundamental subspaces in
a {\em Fundamental Theorem of Linear Neural Networks}, analogous to
what Gilbert Strang calls the {\em Fundamental Theorem of Linear Algebra}.
These subspaces are well known in persistent homology in the form of
{\em persistence modules}, but they ought to be more widely known.

\paragraph*{Acknowledgments.}
We thank Marc Khoury for initiating this research collaboration, and
we thank Herbert Edelsbrunner, Jeff Erickson, Marc Glisse, Hubert Wagner, and
Ling Zhou for conversations helping us to understand
the relationship between this work and persistence modules.

\paragraph*{Abbreviated version.}
A short introduction to this paper's most interesting ideas is
available~\cite{shewchuk26a}.
Most readers should read it first.
}

\begin{document}

\maketitle
\cleardoublepage

\renewcommand{\thepage}{\roman{page}}

\tableofcontents

\setlength{\parskip}{\medskipamount}
\setlength{\parindent}{0pt}

\cleardoublepage

\renewcommand{\thepage}{\arabic{page}}
\setcounter{page}{1}

\pagestyle{fancy}
\headheight 14.4pt
\renewcommand{\sectionmark}[1]{\markboth{}{#1}}
\renewcommand{\subsectionmark}[1]{}
\lhead[\rm\thepage]{}
\chead[{\sl Jonathan Richard Shewchuk and Sagnik Bhattacharya}]{{\sl \rightmark}}
\rhead[]{\rm\thepage}
\cfoot[]{}
\thispagestyle{empty}

\section{Introduction}
\label{intro}

In its simplest form, a {\em linear neural network} is
a sequence of matrices whose product is a matrix.
The network takes an {\em input vector} $x$ and produces
an {\em output vector} $y = W_L W_{L-1} \cdots W_2 W_1 x$.
We number the matrices in the order they are applied.
The matrix $W_1$ linearly transforms $x$, producing
a vector that $W_2$ linearly transforms, and so on.
The composition of those transformations is also a linear transformation,
represented by
\[
W = \mu(W_L, W_{L-1}, \ldots, W_2, W_1) = W_L W_{L-1} \cdots W_2 W_1,
\]
where $\mu$ is called the {\em matrix multiplication map}.
Each matrix $W_j$ is interpreted as
a layer of edges (connections) in the network, edge layer number~$j$,
and each component of $W_j$ is interpreted as the {\em weight} of an edge.
For brevity, we omit added terms, 
but we will incorporate them (easily) in Section~\ref{additive}.

This definition is incomplete:  a {\em neural network} also entails
software that computes the sequence of vectors,
an optimization algorithm that {\em trains} the network
by choosing good weights, and more.
To a computer scientist, a {\em linear neural network} is
a neural network in which there are no activation functions.
(Said another way, the activation function at the output of each unit is just
the identity function.)

In this paper, we study $\mu^{-1}(W)$, the set of all factorizations of
a matrix $W$ into a product of matrices of specified sizes.
Said differently, we study
the set of all choices of linear neural network weights such that
the network computes the linear transformation~$W$.
This set is infinite (unless $L = 1$) and
it is a {\em real algebraic variety}---the set of
all real-valued solutions of a system of polynomial equations
(specifically, multilinear equations).
Trager, Kohn, and Bruna~\cite{trager20}, whose paper inspired ours,
call $\mu^{-1}(W)$ the {\em fiber} of~$W$ under the map~$\mu$.
The fiber has a complicated topology and geometry:
it is a union of manifolds of varying dimensions.
Understanding the fiber has applications in understanding
gradient descent algorithms for training neural networks---but
it is also a beautiful mathematical problem in its own right.

\begin{figure}
\centerline{\includegraphics[width=0.4\textwidth]{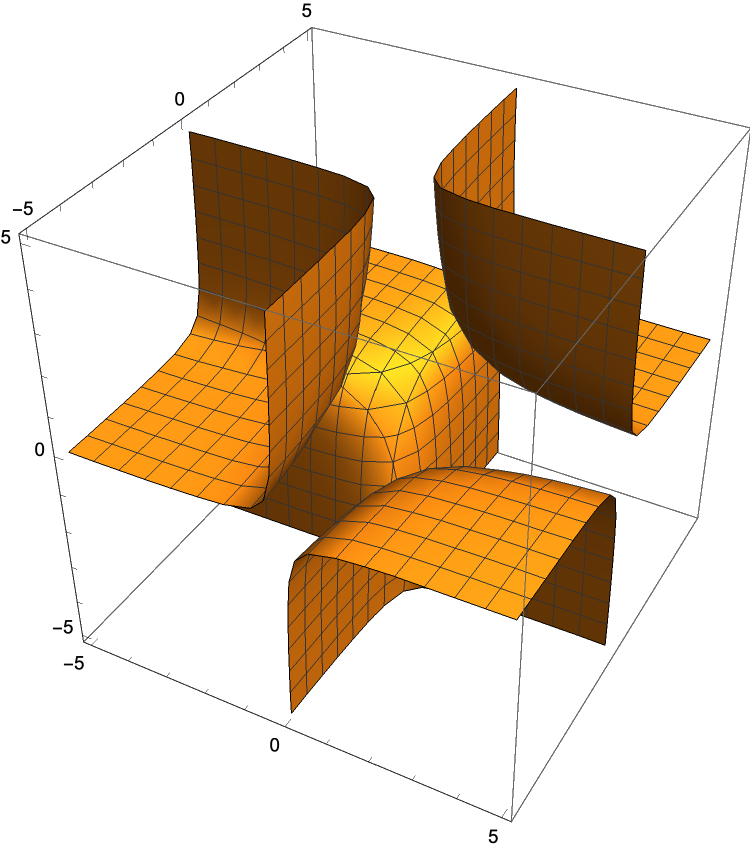}}

\caption{\label{hyperboloid}
The fiber $\mu^{-1}([1])$ for the network
$W_3 W_2 W_1 = [\theta_3] [\theta_2] [\theta_1] = [1] = W$.
}
\end{figure}

Let us give a simple example of a fiber.
Suppose every matrix is square and $W$ is invertible.
Then every factor matrix $W_i$ must also be invertible.
The set of all real, invertible $d \times d$ matrices is called
the {\em general linear group} $\GL(d, \R)$, which is
a $d^2$-dimensional manifold embedded in $\R^{d \times d}$.
$\GL(d, \R)$ has two connected components:  one composed of
matrices with positive determinants and one for negative determinants.
To factor $W$, we can choose each matrix $W_i$ to be
an arbitrary member of $\GL(d, \R)$ except for
one matrix that is uniquely determined by the other choices.
The fiber is
\[
\mu^{-1}(W) =
\{ (W_L, W_{L-1}, \ldots, W_3, W_2, W_2^{-1} W_3^{-1} \cdots W_L^{-1} W) :
W_L, W_{L-1}, \ldots, W_2 \in \GL(d, \R) \}.
\]
This fiber is a smooth, $(d^2 (L-1))$-dimensional manifold with topology
$\GL(d, \R) \times \GL(d, \R) \times \ldots \times \GL(d, \R)$
(with $L - 1$ factors) and hence $2^{L-1}$ connected components
(reflecting the signs of the determinants of the factor matrices).
Figure~\ref{hyperboloid} graphs the fiber $\mu^{-1}([1])$ when
we factor the matrix $[1]$ into three $1 \times 1$ matrices.
Although this graph lacks the complexities of larger matrices,
we see a graceful $2$-dimensional manifold with four components,
as advertised, and we gain an inkling of what the general case might look like.

\begin{figure}
\centerline{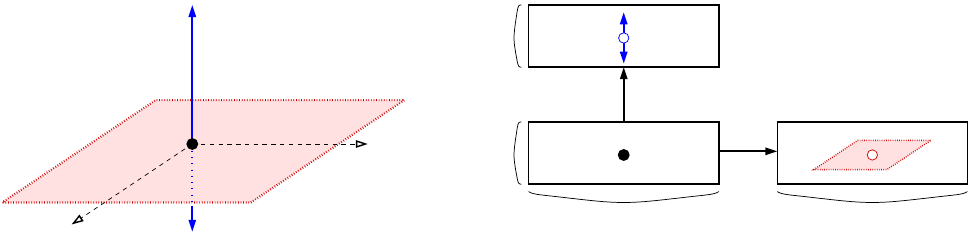}

\caption{\label{strat112}
At left is the fiber $\mu^{-1}([0 ~~~ 0])$ for the network
$W_2 W_1 = [\theta_2] [\theta_1 ~~~ \theta'_1] = [0 ~~~ 0] = W$,
partitioned into three strata:
$S_{00}$ is the origin;
$S_{10}$ (blue) is the $\theta_2$-axis with the origin removed; and
$S_{01}$ (pink) is the plane spanned by the $\theta_1$- and $\theta'_1$-axes
with the origin removed.
At right, the strata are arranged in a {\em stratum dag},
organized as a table indexed by the ranks of $W_1$ and $W_2$.
Each dag vertex specifies the dimension of the stratum (dim),
the number of degrees of freedom of motion on the fiber (dof), and
the number of rank-increasing degrees of freedom (rdof),
which stay on the fiber but move off the stratum and
onto a higher-dimensional stratum.
Always, dof $=$ dim $+$ rdof.
A directed path from one stratum to another implies that
the former is a subset of the closure of the latter.
}
\end{figure}

If the network has a matrix that isn't square or if $W$ does not have full rank,
the fiber is usually no longer a manifold.
But it can be partitioned into smooth manifolds of various dimensions,
called {\em strata}.
The set of strata is called a {\em stratification} of the fiber.
Figure~\ref{strat112} graphs the solutions of
$[\theta_2] [\theta_1 ~~~ \theta'_1] = [0 ~~~ 0]$
(an instantiation of $W_2 W_1 = W$).
The set of solutions is not a manifold, but
it can be partitioned into three strata:
$S_{00}$~is the origin,
$S_{10}$~is the $\theta_2$-axis with the origin removed, and
$S_{01}$~is the $\theta_1$-$\theta'_1$ plane with the origin removed.
The two subscripts of $S_{**}$ are the ranks of $W_2$ and~$W_1$, respectively.
We see that the strata in a fiber can have different dimensions.
The stratum $S_{00}$~lies in the closures of both $S_{10}$ and $S_{01}$;
$S_{00}$ serves to connect $S_{10}$ to $S_{01}$.

Figure~\ref{strat112} illustrates what we call  
a {\em rank stratification} of the fiber $\mu^{-1}(W)$, which has
one stratum for each {\em rank list} defined in Section~\ref{notation};
it is the most natural way to stratify a fiber of the matrix multiplication map.
We will see that each stratum in a rank stratification is
an analytic manifold (hence of class $C^\infty$),\footnote{
Depending on who is defining the term, a ``stratification'' may or may not
require that the strata have some degree of smoothness.
We can speak of a stratification of a topological space without considering
any embedding of the space at all, in which case
it matters only that the strata be manifolds.
In this paper, all the strata are
analytic manifolds embedded in a Euclidean space.
}
though it is not necessarily bounded nor closed nor connected.

The goal of this paper is to use
the rank stratification to study the geometry and topology of the fiber.
We derive the dimension of each stratum and
we characterize the geometry of each stratum in several ways, including
determining the tangent space and the normal space
at each point on each stratum.
We describe how the strata are connected to each other.
The strata satisfy the {\em frontier condition}:
if a stratum intersects the closure of another stratum, then
the former stratum is a subset of the closure of the latter stratum.
This subset relationship induces a partial order with a single minimal element,
a stratum that is included in the closure of every other stratum
(in Figure~\ref{strat112}, that's~$S_{00}$).
We give an algorithm to compute a directed acyclic graph (dag) that
expresses the partial order of the strata; we call it a {\em stratum dag}.
A key result describes the relationship between this partial order and
the ranks of certain matrices in the network.
What makes it interesting is that each stratum represents
a different pattern of information flow through the network.

In 1957, Hassler Whitney \cite{whitney57,whitney65a,whitney65b} proved that
every real algebraic variety can be partitioned into
a finite set of analytic manifolds of various dimensions.
{\L}ojasiewicz~\cite{lojasiewicz65} generalized Whitney's result to
semi-algebraic and semi-analytic sets---a fact we will use in
Section~\ref{sec:affinestrata}.
Thom~\cite{thom62} introduced the terms {\em stratum} and {\em stratification}.
See Lu~\cite[Chapter 5]{lu76} for an excellent exposition.

Linear neural networks compute only linear transformations; they are
far less powerful than networks with nonlinear activation functions such as
rectified linear units (ReLUs, also called ramp functions) and
sigmoid functions (also called logistic functions).
Yet linear networks have become a popular object of study
\cite{arora18acceleration,arora19regularization,choromanska15,lowrankbias24,
  hardtma16,telgarsky18align,kawaguchi2016deep,li21statmech,lukawaguchi17,
  radha20align,trager20,zhang19}.
Why?
We cannot fully understand the training of ReLU-based networks---or probably
any neural networks---if we do not understand linear networks.
Researchers have studied linear neural networks to understand
phenomena observed in ReLU networks such as
implicit regularization in optimizing the training algorithm's cost function
\cite{gunasekar2017implicit,arora19regularization,lowrankbias24},
implicit acceleration of
training by gradient descent~\cite{arora18acceleration}, and
the success of residual networks~\cite{hardtma16}.
Close to our hearts, Trager, Kohn, and Bruna~\cite{trager20} show that
the fiber $\mu^{-1}(W)$ is crucial for understanding
the critical points of the cost functions minimized by
training algorithms for linear neural networks.
We revisit some of their ideas in Section~\ref{critical}.




\clearpage

\section{Notation}
\label{notation}

\begin{figure}
\centerline{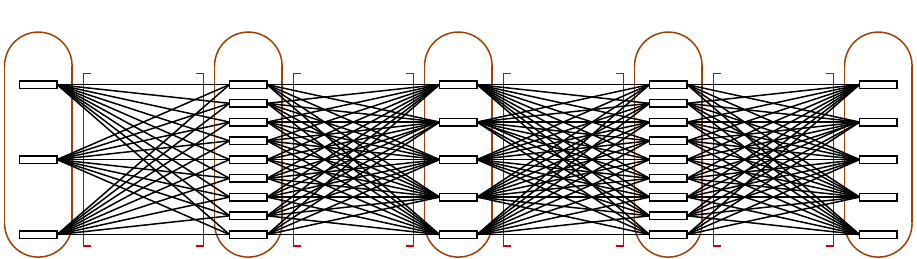}

\caption{\label{network}
Schematic of a linear neural network with four edge layers ($L = 4$).
}
\end{figure}

Let $L$ be the number of matrices---that is,
the number of layers of edges (connections) in the network.
Alternating with the edge layers are $L + 1$ layers of units,
numbered from $0$ to $L$, in which layer~$j$ has
$d_j$~real-valued {\em units} that represent a vector in $\R^{d_j}$.
Unit layer~$0$ is the {\em input layer},
unit layer~$L$ is the {\em output layer}, and
between them are $L - 1$ {\em hidden layers}.
The layers of edges are numbered from $1$ to $L$, and
the edge weights in edge layer $j$ are represented by
a real-valued $d_j \times d_{j-1}$ matrix $W_j$,
as illustrated in Figure~\ref{network}.

We collect all the network's weights in a {\em weight vector}
$\theta = (W_L, W_{L - 1}, \ldots, W_1) \in \R^{d_\theta}$, where
\[
d_\theta = d_L d_{L-1} + d_{L-1} d_{L-2} + \ldots + d_1 d_0
\]
is the number of real-valued weights in the network
(i.e., the number of connections).
Recall the matrix multiplication map $\mu(W_L, W_{L-1}, \ldots, W_2, W_1)$;
we can abbreviate it to $\mu(\theta) = W_L W_{L-1} \cdots W_2 W_1$.
Given a fixed weight vector $\theta$,
the linear neural network takes an {\em input vector} $x \in \R^{d_0}$ and
returns an {\em output vector} $y = W_L W_{L-1} \cdots W_2 W_1 x$,
with $y \in \R^{d_L}$.
Hence, the network implicitly applies a linear transformation specified by
the $d_L \times d_0$ matrix $W = \mu(\theta)$, yielding $y = Wx$.

The map $\mu$ is not bijective (unless $L = 1$), so
it has a preimage set, not an inverse.
Given $W \in \R^{d_L \times d_0}$, let
\[
\mu^{-1}(W) = \{ \theta : \mu(\theta) = W \}
\]
be the set of all factorizations of $W$
for some fixed $d_L, d_{L-1}, \ldots, d_0$.
Call $\mu^{-1}(W)$ the {\em fiber} of~$W$ under~$\mu$;
it is a real algebraic variety in the weight space $\R^{d_\theta}$.
With respect to $\R^{d_\theta}$, $\mu^{-1}(W)$ is a closed point set.
Note that $\mu^{-1}(W)$ is empty if and only if $\rank W > \min_{j=1}^{L-1} d_j$.

Given a weight vector $\theta = (W_L, W_{L-1}, \ldots, W_1)$,
its {\em subsequence matrices} are all the matrices of the form
\[
W_{k \ssq i} = \mu_{k \ssq i}(\theta),
\hspace{.2in}  \mbox{where~}  \hspace{.2in}
\mu_{k \ssq i}(\theta) = W_k W_{k-1} \cdots W_{i+1},
\hspace{.2in} L \geq k \geq i \geq 0.
\]
The notation $W_{k \ssq i}$ (and $\mu_{k \ssq i}$) indicates that this matrix
transforms a vector at unit layer $i$ to produce a vector at unit layer~$k$.
Note that $W = W_{L \ssq 0}$ and $W_j = W_{j \ssq j-1}$.
We use the convention that $W_{k \ssq k} = I_{d_k \times d_k}$,
the $d_k \times d_k$ identity matrix.
Assume that every $W_{k \ssq i}$ in this paper is a function of $\theta$.
We call each $W_j$ a {\em factor matrix}.

The {\em rank list} $\underline{r}$ for
a weight vector $\theta \in \R^{d_\theta}$ is a sequence that lists the rank of
every subsequence matrix $W_{k \ssq i}$ such that $L \geq k \geq i \geq 0$.
The list includes the unit layer sizes $\rank W_{k \ssq k} = d_k$.
For example, for a network with $L = 3$ layers of edges,
the rank list of $\theta$ is
\[
\underline{r} = \langle d_3, d_2, d_1, d_0, \rank W_3, \rank W_2, \rank W_1,
\rank W_3W_2, \rank W_2W_1, \rank W \rangle.
\]

We will use rank lists to partition a fiber into strata.
Sometimes we do not want to specify a particular $\theta$, but rather
we wish to specify some target ranks.
In that case,
we let $r_{k \ssq i}$ denote the target value of $\rank W_{k \ssq i}$ and
we write $\underline{r} = \langle r_{k \ssq i} \rangle_{L \geq k \geq i \geq 0}$.
For example, if we set $r_{3 \ssq 1} = 2$,
we select weight vectors for which $\rank W_3 W_2 = 2$.  
If $L = 2$ and we set $\underline{r} = \langle 9, 8, 7, 2, 3, 1 \rangle$,
we select weight vectors for which $W_2$ is a $9 \times 8$ matrix of rank $2$,
$W_1$ is an $8 \times 7$ matrix of rank $3$, and $W = W_2 W_1$ has rank $1$.

Let $S_{\underline{r}}^W$ denote
the set of points in $W$'s fiber having rank list $\underline{r}$.
That is,
\[
S_{\underline{r}}^W = \{ \theta \in \mu^{-1}(W) :
\mbox{the rank list of~} \theta \mbox{~is~} \underline{r} \}.
\]
If it is nonempty, we call $S_{\underline{r}}^W$ a {\em stratum} in
the rank stratification of $W$'s fiber.
When $W$ is clear from context, we just write $S_{\underline{r}}$.

The {\em rank stratification} of $\mu^{-1}(W)$ is
\[
\mathbb{S} = \{ S_{\underline{r}} :
\underline{r} \mbox{~is the rank list of some weight vector in~} \mu^{-1}(W) \}.
\]
The members of $\mathbb{S}$ are disjoint and
$\mu^{-1}(W) = \bigcup_{S \in \mathbb{S}} S$---that is,
$\mathbb{S}$ is a partition of $\mu^{-1}(W)$.
We have not yet shown that $\mathbb{S}$ is a stratification because
we have not yet shown that the ``strata'' are manifolds;
we do so in Theorem~\ref{stratummanifold}.
Clearly, $\mathbb{S}$~contains finitely many strata.

A rank list $\underline{r}$ is {\em valid} if
there exists some weight vector $\theta \in \R^{d_\theta}$ that has
rank list~$\underline{r}$.
We~show in Appendix~\ref{valid} that if~$W \in \R^{d_L \times d_0}$ and 
$\underline{r}$ is a valid rank list with $r_{L \ssq 0} = \rank W$, then
there exists some $\theta \in \mu^{-1}(W)$ that has rank list $\underline{r}$,
so $S_{\underline{r}}$ is nonempty.

Given two rank lists $\underline{r}$ and $\underline{z}$,
we write $\underline{r} \leq \underline{z}$ to mean that
$r_{k \ssq i} \leq z_{k \ssq i}$ for all $L \geq k \geq i \geq 0$.
We write $\underline{r} < \underline{z}$ to mean that
$\underline{r} \leq \underline{z}$ and $\underline{r} \neq \underline{z}$
(at least one of the inequalities holds strictly).
For example, $\langle 5, 4, 1 \rangle < \langle 5, 4, 3 \rangle$.

Given a set $S \subseteq \R^{d_\theta}$, let $\bar{S}$ denote
the closure of $S$ (with respect to the weight space $\R^{d_\theta}$).
We will see that for two nonempty stratum
$S_{\underline{r}}, S_{\underline{z}} \in \mathbb{S}$,
$S_{\underline{r}} \subseteq \bar{S}_{\underline{z}}$ if and only if
$\underline{r} \leq \underline{z}$.

Given a point (weight vector) $\theta$ on
a smooth manifold $S \subseteq \R^{d_\theta}$,
$T_\theta S$ denotes the {\em tangent space} of $S$ at~$\theta$---the
subspace of~$\R^{d_\theta}$ tangent to $S$ at $\theta$ that has
the same dimension as $S$---and
$N_\theta S$ denotes its orthogonal complement,
the {\em normal space} of $S$ at $\theta$---the
subspace of~$\R^{d_\theta}$ orthogonal to $S$ at $\theta$ that has
dimension $d_\theta - \dim S$.
In our setting, both~$T_\theta S$ and $N_\theta S$ pass through the origin
(they are true subspaces of $\R^{d_\theta}$), not necessarily through $\theta$.
Two vectors $\zeta = (X_L, X_{L-1}, \ldots, X_1)$ and
$\xi = (Y_L, Y_{L-1}, \ldots, Y_1)$ in the weight space $\R^{d_\theta}$
are {\em orthogonal} if $\zeta \cdot \xi = 0$; hence $\zeta \cdot \xi = 0$
for every $\zeta \in T_\theta S$ and $\xi \in N_\theta S$.
This Euclidean inner product can be written
$\zeta \cdot \xi = \sum_{j=1}^L \langle X_j, Y_j \rangle_{\F}$, where
$\langle M, H \rangle_{\F} = \trace (M H^\top) =
\sum_{i=1}^p \sum_{k=1}^q M_{ik} H_{ik}$ denotes
the Frobenius inner product of two $p \times q$ matrices $M$ and $H$
(essentially just a dot product for matrices).

We have a frequent need to multiply a matrix by a subspace.
Given a $p \times q$ matrix $M$ and a subspace $P \in \R^q$, we define
\[
MP = \{ Mv : v \in P \},
\]
which is a subspace of $\R^p$.
For example, $M \R^q$ is the columnspace of $M$.

Tables~\ref{notation1}, \ref{notation2}, and~\ref{notation3} collect
most of the notation used in this paper for easy reference.

\begin{table}[tbp]
\begin{center}
\begin{tabular}{l}
\hline
$A_{kji} = \Null W_{k+1 \ssq j} \cap \col W_{j \ssq i}$
\rule{0pt}{11pt} \hfill  flow subspace of $\R^{d_j}$ for unit layer $j$
(note: $A_{j-1,j,i} = \{ {\bf 0} \} = A_{k,j,-1}$)  \\
$a_{kji} \in A_{kji} \downarrow (A_{k,j,i-1} + A_{k-1,j,i})$  \hfill
prebasis subspace of $\R^{d_j}$ for unit layer $j$  \\
$\mathcal{A}_{kji} = \{ a_{k'ji'} \neq \{ {\bf 0} \}:
k' \in [j, k], i' \in [0, i] \}$  \hfill
prebasis for $A_{kji}$ (set of prebasis subspaces spanning $A_{kji}$)  \\
$\mathcal{A}_j = \mathcal{A}_{Ljj}$  \hfill
prebasis for $\R^{d_j}$ for unit layer $j$
(set of prebasis subspaces spanning $\R^{d_j}$)  \\
$B_{kji} = \row W_{k \ssq j} \cap \Null W_{j \ssq i-1}^\top$  \hfill
flow subspace of $\R^{d_j}$ for transpose network
($B_{L+1,j,i} = \{ {\bf 0} \} = B_{k,j,j+1}$)  \\
$b_{kji} \in B_{kji} \downarrow (B_{k,j,i+1} + B_{k+1,j,i})$  \hfill
prebasis subspace of $\R^{d_j}$ for unit layer $j$ of the transpose network  \\
$\mathcal{B}_{kji} = \{ b_{k'ji'} \neq \{ {\bf 0} \}:
k' \in [k, L], i' \in [i, j] \}$  \hfill
prebasis for $B_{kji}$ (set of prebasis subspaces spanning $B_{kji}$)  \\
$\mathcal{B}_j = \mathcal{B}_{jj0}$  \hfill
prebasis for $\R^{d_j}$ for unit layer $j$ of the transpose network
(set of subspaces spanning $\R^{d_j}$)  \\
$d_j$  \hfill
number of units in unit layer $j$ of the neural network  \\
$d_\theta = d_L d_{L-1} + d_{L-1} d_{L-2} + \ldots + d_1 d_0$  \hspace{.2in}
number of weights in network (dimension of the weight space)  \\
$d = \min_{j=0}^L d_j$  \hfill  the network's smallest unit layer size  \\
$\DO^\off$, $\DO^\fiber$, $\DO^\barc$, $\DO^\strat$, $\DT^\off$, $\DT^\barc$,
$D^\free$, $D^\strat$, $D^\fiber$  \rule{0pt}{11pt}  \hfill
dimension of $\Span \ThetaO^\off$, etc.; see Table~\ref{subspacesets} \\
$e_{lktsih} = b_{lti} \otimes a_{ksh}$  \hfill
subsequence subspace of $\R^{d_t \times d_s}$  \\
$e_{ki} = e_{LkL0i0} = b_{LLi} \otimes a_{k00}$  \hfill
subsequence subspace of $\R^{d_L \times d_0}$  \\
$\mathcal{E}_{ts} = \{ e_{lktsih} \neq \{ 0 \} :
l \in [t, L], k \in [s, L], i \in [0, t], h \in [0, s] \}$  \hfill
subsequence prebasis for $\R^{d_t \times d_s}$  \\
$G = \nabla_W \ell(W)$  \hfill gradient of $\ell$;
the $d_L \times d_0$ matrix with $G_{ij} = \partial \ell / \partial W_{ij}$  \\
$\tilde{I}_j$  \hfill
almost-identity matrix; part of the canonical weight vector
$\tilde{\theta} = (\tilde{I}_L, \tilde{I}_{L-1}, \ldots, \tilde{I}_1)$  \\
$J_{kji}$  \hfill
a $d_j \times \omega_{ki}$ matrix whose columns are a basis for $a_{kji}$  \\
$J_j = [ J_{kji} ]_{k \in [j, L], i \in [0, j]}$  \hfill
invertible $d_j \times d_j$ matrix whose columns are a basis for $\R^{d_j}$
for unit layer $j$  \\
$K_{kji}$  \hfill
a $d_j \times \omega_{ki}$ matrix whose columns are a basis for $b_{kji}$  \\
$K_j = [ K_{kji} ]_{k \in [j, L], i \in [0, j]}$  \hfill
invertible $d_j \times d_j$ matrix whose columns are a basis for $\R^{d_j}$
for unit layer $j$  \\
$L$  \hfill  number of layers of edges in the neural network  \\
$\ell(W)$  \hfill
neural network training cost function, $\R^{d_L \times d_0} \rightarrow \R$  \\
$\mathcal{L}(\theta) = \ell(\mu(\theta))$  \hfill
neural network training cost function, $\R^{d_\theta} \rightarrow \R$  \\
$N_j = \Null W_{L \ssq j} \otimes \R^{d_{j-1}} +
\R^{d_j} \otimes \Null W_{j-1 \ssq 0}^\top$  \hfill
set of displacements of $W_j$ that do not change $W$  \\
$N_\theta S$  \hfill
subspace of $\R^{d_\theta}$ normal to manifold $S$ at $\theta$
(passes through origin, not necessarily through $\theta$)  \\
$o_{lkjih} = a_{lji} \otimes b_{k,j-1,h}$  \hfill
prebasis subspace of $\R^{d_j \times d_{j-1}}$  \\
$\mathcal{O}_j = \{ o_{lkjih} \neq \{ 0 \} :
l \in [j, L], k \in [j-1, L], i \in [0, j], h \in [0, j-1] \}$  \hfill
prebasis for $\R^{d_j \times d_{j-1}}$  \\
$\mathcal{O}^\off_j = \{ o_{lkjih} \in \mathcal{O}_j : l = L$ and $h = 0 \}$
\hfill  \rule{0pt}{11pt}
subspaces in $\mathcal{O}_j$ that are linearly independent of $N_j$  \\
$\mathcal{O}^\fiber_j = \mathcal{O}_j \setminus \mathcal{O}^\off_j$
\hfill  \rule{0pt}{11pt}
subspaces in $\mathcal{O}_j$ that are subspaces of $N_j$;
$\mathcal{O}^\fiber_j$ is a prebasis for $N_j$  \\
$P = \{ (W_L, W_{L-1}, \ldots, W_{j+2}, W_{j+1} (I + \epsilon H),
(I + \epsilon H)^{-1} W_j, W_{j-1}, \ldots, W_1) :
\epsilon \in [0, \hat{\epsilon}] \}$  \hfill  two-matrix path  \\
$\underline{r} = \langle \rank W_{k \ssq i} \rangle_{L \geq k \geq i \geq 0}$ or
$\langle r_{k \ssq i} \rangle_{L \geq k \geq i \geq 0}$  \hfill
rank list; lists the ranks of all subsequence matrices $W_{k \ssq i}$  \\
$r_{k \ssq i}$  \hfill  target value for $\rank W_{k \ssq i}$;
one of the entries in a rank list $\underline{r}$  \\
$S_{\underline{r}} = S_{\underline{r}}^W = \{ \theta \in \mu^{-1}(W) :$
the rank list of $\theta$ is $\underline{r} \}$  \hfill
stratum (of the fiber of $W$) with rank list $\underline{r}$  \\
$\mathbb{S} = \{ S_{\underline{r}} :
\underline{r} \mbox{~is the rank list of some weight vector in~} \mu^{-1}(W) \}$
\hfill  rank stratification of $\mu^{-1}(W)$  \\
$T_\theta S$  \hfill
subspace of $\R^{d_\theta}$ tangent to manifold $S$ at $\theta$
(passes through origin, not necessarily through $\theta$)  \\
$W = \mu(\theta) = W_L W_{L-1} \cdots W_1$  \hfill
product of factor matrices; linear map computed by neural network  \\
$W_j \in \R^{d_j \times d_{j-1}}$  \hfill  factor matrix;
layer $j$ of edges in neural network; connects unit layers $j - 1$ and $j$  \\
$\Delta W_j$  \hfill  displacement applied to factor matrix $W_j$  \\
$W'_j = W_j + \Delta W_j$
\hfill  factor matrix after a move from
$\theta = (W_L, \ldots, W_1)$ to $\theta' = (W'_L, \ldots, W'_1)$  \\
$W_{k \ssq i} = \mu_{k \ssq i}(\theta) = W_k W_{k-1} \cdots W_{i+1}$  \hfill
subsequence matrix; maps unit layer $i$ to unit layer $k$  \\
\hspace{.2in}  Notes:  $W_{L \ssq 0} = W$; $W_{j \ssq j-1} = W_j$;
by convention, $W_{k \ssq k} = I_{d_k \times d_k}$;  
$W_{k \ssq -1} = 0_{d_k \times 1}$; $W_{L+1 \ssq i} = 0_{1 \times d_i}$  \\
$x \in \R^{d_0}$  \hfill  input vector  \\
$y \in \R^{d_L}$
\hfill  output vector; $y = Wx = W_L W_{L-1} \cdots W_2 W_1 x$  \\
$\hat{y}(x) = Wx$  \hfill function mapping an input vector to an output vector,
$\hat{y} : \R^{d_0} \rightarrow \R^{d_L}$  \\
\hline
\end{tabular}
\end{center}

\caption{\label{notation1}
Notation used in this paper.
Continued in Tables~\ref{notation2} and~\ref{notation3}.
See also Table~\ref{subspacesets}.
}
\end{table}

\begin{table}[tbp]
\begin{center}
\begin{tabular}{l}
\hline
$\alpha_{kji} = \dim A_{kji} = \rank W_{j \ssq i} - \rank W_{k+1 \ssq i}
= \sum_{t = j}^k \sum_{s = 0}^i \omega_{ts}$
\hfill  \rule{0pt}{11pt}  dimension of flow subspace $A_{kji}$  \\
$\beta_{kji} = \dim B_{kji} = \rank W_{k \ssq j} - \rank W_{k \ssq i-1}
= \sum_{t = k}^L \sum_{s = i}^j \omega_{ts}$
\hfill  dimension of flow subspace $B_{kji}$  \\
$\eta(M_L, M_{L-1}, \ldots, M_1) =
(J_L M_L J_{L-1}^{-1}, J_{L-1} M_{L-1} J_{L-2}^{-1}, \ldots, J_1 M_1 J_0^{-1})$
\hfill  \rule{0pt}{11pt}  a linear transformation of $\R^{d_\theta}$  \\
$\theta = (W_L, W_{L - 1}, \ldots, W_1) \in \R^{d_\theta}$  \hfill
weight vector representing a neural network  \\
$\Delta \theta = (\Delta W_L, \Delta W_{L - 1}, \ldots, \Delta W_1)
\in \R^{d_\theta}$  \hfill
displacement applied to a weight vector $\theta$  \\
$\theta' = \theta + \Delta \theta = (W'_L, W'_{L - 1}, \ldots, W'_1)$  \hfill
weight vector after a move from $\theta$ to $\theta'$  \\
$\tilde{\theta} = (\tilde{I}_L, \tilde{I}_{L-1}, \ldots, \tilde{I}_1)$
\hfill  \rule{0pt}{11pt}
the canonical weight vector (for some specified rank list $\underline{r}$)  \\
$\ThetaO = \{ \phi_{lkjih} \neq \{ {\bf 0} \} :
L \geq l \geq j \geq i \geq 0$ and $L \geq k \geq j - 1 \geq h \geq 0 \}$
\hfill  the one-matrix prebasis for $\R^{d_\theta}$  \\
$\ThetaT = \{ \tau_{lkjih} \neq \{ {\bf 0} \} : L \geq l > j > h \geq 0$ and
$L \geq k \geq j \geq i \geq 0 \}$  \hfill
restricted set of two-matrix subspaces  \\
$\ThetaTO = \{ \tau_{lkjih} \neq \{ {\bf 0} \} :
L > j > 0, L \geq l \geq j \geq h \geq 0,$ and
$L \geq k \geq j \geq i \geq 0 \}$  \hfill  two-matrix subspaces  \\
$\ThetaO^\off$, $\ThetaO^\fiber$, $\ThetaO^\barc$, $\ThetaO^\strat$,
$\ThetaT^\off$, $\ThetaT^\barc$, $\Theta^\free$, $\Theta^\strat$, $\Theta^\fiber$
\hfill
prebases for subspaces of $\R^{d_\theta}$; see Table~\ref{subspacesets}  \\
$\mu(\theta) = \mu(W_L, W_{L - 1}, \ldots, W_1) = W_L W_{L - 1} \cdots W_1$
\hfill  matrix multiplication map,
$\mu : \R^{d_\theta} \rightarrow \R^{d_L \times d_0}$  \\
$\mu^{-1}(W) = \{ \theta \in \R^{d_\theta} : \mu(\theta) = W \}$
\hfill  the fiber of $W$ under the matrix multiplication map;
$W \in \R^{d_L \times d_0}$  \\
$\dmu(\theta)(\Delta \theta) = \sum_{j=1}^L W_{L \ssq j} \Delta W_j W_{j-1 \ssq 0}$
\hfill  the differential map of $\mu$ at $\theta$ is
$\dmu(\theta) : \R^{d_\theta} \rightarrow \R^{d_L \times d_0}$  \\
$\dmu^\top(\theta)(\Delta W) = (\Delta W W_{L-1 \ssq 0}^\top, \ldots,
W_{L \ssq j}^\top \Delta W W_{j-1 \ssq 0}^\top, \ldots, W_{L \ssq 1}^\top \Delta W)$
\hfill  transpose of the differential map  \\
$\mu_{t \ssq s}(\theta) = W_t W_{t-1} \cdots W_{s+1}$  \hfill
subsequence matrix $W_{t \ssq s}$'s map,
$\mu_{t \ssq s} : \R^{d_\theta} \rightarrow \R^{d_t \times d_s}$  \\
$\dmu_{t \ssq s}(\theta)(\Delta \theta) =
\sum_{j=s+1}^t W_{t \ssq j} \Delta W_j W_{j-1 \ssq s}$
\hfill  the differential map of $\mu_{t \ssq s}$ at $\theta$ is
$\dmu_{t \ssq s}(\theta) : \R^{d_\theta} \rightarrow \R^{d_t \times d_s}$  \\
$\tau_{lkjih} = \{ (0, 0, \ldots, 0, W_{j+1} H, - H W_j, 0, \ldots, 0) :
H \in a_{lji} \otimes b_{kjh} \}$  \hfill
two-matrix subspace of $\R^{d_\theta}$  \\
$\phi_{lkjih} = \{ (0, 0, \ldots, 0, \Delta W_j, 0, \ldots, 0) :
\Delta W_j \in o_{lkjih} = a_{lji} \otimes b_{k,j-1,h} \}$
\hfill  one-matrix subspace of $\R^{d_\theta}$  \\
$\chi_r(M)$  \hfill  maps matrix $M$ to
a vector listing the determinant of every $(r + 1) \times (r + 1)$ minor of $M$
\\
$\psi_{lktsih} = \{ (X_L, X_{L-1}, \ldots, X_1) : M \in b_{lti} \otimes a_{ksh} \}$
where $X_j = W_{t \ssq j}^\top M W_{j-1 \ssq s}^\top$ if $j \in [s + 1, t]$;
else $X_j = 0$  \\
$\psi_{lkih} = \psi_{lklhih} =
\{ (X_L, \ldots,) : M \in b_{lli} \otimes a_{khh} \}$
where $X_j = W_{l \ssq j}^\top M W_{j-1 \ssq h}^\top$ if $j \in [i, k + 1]$;
else $X_j = 0$  \\
$\Psi^\free = \{ \psi_{Lki0} \neq \{ {\bf 0} \} : k, i \in [0, L]$ and
$k + 1 \geq i \}$  \hfill
freedom normal prebasis; spans $\row \dmu(\theta)$  \\
$\Psi^\strat = \Psi^\free \cup
\{ \psi_{lkih} \neq \{ {\bf 0} \}: L \geq l \geq k + 1 \geq i > h \geq 0 \}$
\hfill  stratum normal prebasis; spans $N_\theta S_{\underline{r}}$  \\
$\omega_{ki} = \dim a_{kji} = \dim b_{kji} = \rank W_{k \ssq i} -
\rank W_{k \ssq i-1} - \rank W_{k+1 \ssq i} + \rank W_{k+1 \ssq i-1}$  \hfill
interval count for $[i, k]$  \\
\hline
\end{tabular}
\end{center}

\caption{\label{notation2}
More notation used in this paper.
Continued in Table~\ref{notation3}.
See also Table~\ref{subspacesets}.
}
\end{table}

\begin{table}[tbp]
\begin{center}
\begin{tabular}{l}
\hline
$\underline{r} \leq \underline{z}$  \hfill
$r_{k \ssq i} \leq z_{k \ssq i}$ for all $L \geq k \geq i \geq 0$  \\
$\underline{r} < \underline{z}$  \hfill
$\underline{r} \leq \underline{z}$ and $\underline{r} \neq \underline{z}$
(at least one of the inequalities holds strictly)  \\
$\bar{S}$  \hfill  closure of set $S \subseteq \R^{d_\theta}$
with respect to weight space $\R^{d_\theta}$  \\
$Q^\perp$  \hfill  orthogonal complement of subspace $Q$
(subspace of all vectors orthogonal to all vectors in $Q$)  \\
$MP = \{ Mv : v \in P \}$  \hfill
product of a matrix $M$ and a subspace $P$  \\
$Z = P \oplus Q$  \hfill
direct sum decomposition:
$Z = P + Q$ (vector sum of subspaces) and $P \cap Q = \{ {\bf 0} \}$  \\
$Z = \bigoplus_{i=1}^m P_i$  \hfill
direct sum decomposition:  $Z = \sum_{i=1}^m P_i$ and
for every $i \in [1, m]$, $P_i \cap \sum_{j \neq i} P_j= \{ {\bf 0} \}$  \\
$Z \downarrow Q = \{ P \subseteq Z : Z = P \oplus Q \}$  \hspace{.2in}
subspaces of $Z$ of dimension $\dim Z - \dim Q$ linearly independent of $Q$  \\
$P \otimes Q = \{ M \in \R^{t \times s} :
\col M \subseteq P  \mbox{~and~}  \row M \subseteq Q \}$  \hfill
tensor product of subspaces $P \subseteq \R^t$ and $Q \subseteq \R^s$  \\
\hline
$\col M$; $\col \dmu(\theta)$
\hfill  columnspace/image of matrix $M$ or of linear map $\dmu(\theta)$  \\
$\dim S$  \hfill  dimension of subspace or manifold $S$  \\
$\Null M$; $\Null \dmu(\theta)$
\hfill  nullspace of matrix $M$ or of linear map $\dmu(\theta)$  \\
$\Null M^\top$; $\Null \dmu^\top(\theta)$
\hfill  left nullspace of matrix $M$ or of linear map $\dmu(\theta)$  \\
$\proj_Q P = Q \cap (Q^\perp + P)$  \hfill
orthogonal projection of subspace $P$ onto subspace $Q$  \\
$\rank M = \dim \col M = \dim \row M$;
$\rank \dmu(\theta) = \dim \col \dmu(\theta) = \dim \row \dmu(\theta)$
\hfill  rank of $M$ or of $\dmu(\theta)$  \\
$\row M$; $\row \dmu(\theta)$
\hfill  rowspace of matrix $M$ or of linear map $\dmu(\theta)$  \\
$\Span \Theta = \sum_{\sigma \in \Theta} \sigma$  \hfill
vector sum of the subspaces in a prebasis $\Theta$  \\
\hline
$\DV_r^{p \times q} = \{ M \in \R^{p \times q} : \rank M \leq r \}$  \hfill
\rule{0pt}{11pt}  determinantal variety  \\
$\DM_r^{p \times q} = \{ M \in \R^{p \times q} : \rank M = r \}
= \DV_r^{p \times q} \setminus \DV_{r-1}^{p \times q}$  \hfill
determinantal manifold  \\
$\WDS_{\underline{r}}^{k \ssq i} =
\{ (W_L, \ldots, W_1) \in \R^{d_\theta}: \rank W_{k \ssq i} = r_{k \ssq i} \}$
\hfill  weight-space determinantal (semi-algebraic) set  \\
$\MM_{\underline{r}} = \{ (W_L, \ldots, W_1) \in \R^{d_\theta} :
\rank W_{k \ssq i} = r_{k \ssq i} \mbox{~for all~} L \geq k \geq i \geq 0 \}$
\hfill  multideterminantal manifold  \\
\hspace{.2in}  Notes:
$\MM_{\underline{r}} = \bigcap_{L \ge k > i \ge 0} \WDS_{\underline{r}}^{k \ssq i}$ and
$S_{\underline{r}}^W = \mu^{-1}(W) \cap \MM_{\underline{r}}$  \\
$\GL(d, \R)$  \hfill
the general linear group on real-valued $d \times d$ matrices \\
\hline
\end{tabular}
\end{center}

\caption{\label{notation3}
Still more notation used in this paper.
See also Table~\ref{subspacesets}.
}
\end{table}

\clearpage

\section{A Summary of our Results}
\label{foretaste}

Our central claim is that the fiber $\mu^{-1}(W)$ of a matrix $W$ under
the matrix multiplication map $\mu$ is naturally stratified by
the rank list---that is, the rank list partitions the fiber into manifolds.
The partition (set of manifolds) is
the {\em rank stratification} $\mathbb{S}$ defined in Section~\ref{notation}.
So our most crucial result (Theorem~\ref{stratummanifold}) is that
each $S_{\underline{r}} \in \mathbb{S}$ is
an analytic manifold (without boundary, but
not necessarily closed nor connected nor bounded).\footnote{
The strata are, in the language of topology, {\em manifolds without boundary},
as for every stratum $S$ and every point $\zeta \in S$,
there is an open neighborhood $N \subset S$ that contains $\zeta$ and is
homeomorphic to a ball of the same dimension as the stratum.
Unfortunately, the term ``boundary'' has conflicting meanings in topology:
the term ``manifold without boundary'' is defined in a fashion that takes
$S$ to be the entire topological space, with no larger context.
However, when we consider $S$ as
a point set in the topological space~$\R^{d_\theta}$,
the {\em boundary} of $S$ is defined to be the set of points that lie in both
the closure of $S$ and the closure of $\R^{d_\theta} \setminus S$.
In our setting, the dimension of a stratum $S$ is always less than $d_\theta$,
so the closure of $\R^{d_\theta} \setminus S$ is $\R^{d_\theta}$.
Therefore, the boundary of a stratum~$S$ is its closure~$\bar{S}$.
For example, in Figure~\ref{strat112},
$S_{01}$ is a plane with the origin removed,
the closure of~$S_{01}$ is the whole plane, and
the boundary of~$S_{01}$ also is the whole plane.
So our manifolds without (intrinsic) boundary have (extrinsic) boundaries.
Note that
$S_{01}$ is an example of a stratum that is neither closed nor bounded, and
$S_{10}$ is a stratum that is not closed, bounded, nor connected.
}
As each $S_{\underline{r}} \in \mathbb{S}$ is a manifold,
$\mathbb{S}$ is a stratification and each~$S_{\underline{r}}$ is a stratum.

One of our main results is that the dimension of a stratum $S_{\underline{r}}$ in
the rank stratification
(``dim'' in Figure~\ref{strat112} and several forthcoming figures) is
\[
D^\strat =
d_\theta - \rank W \cdot (d_L + d_0 - \rank W) -  \hspace*{-.1in}
\sum_{L \geq k+1 \geq i > 0} 
  \hspace*{-.1in}
  (\rank W_{k+1 \ssq i} - \rank W_{k+1 \ssq i-1}) \,
  (\rank W_{k \ssq i-1} - \rank W_{k+1 \ssq i-1}),
\]
hence solely a function of its rank list $\underline{r}$.
Whereas showing that each stratum is a manifold is relatively easy,
our derivation of the dimension of a stratum is complicated:
in Section~\ref{counting} we prove that
$D^\strat$ is a lower bound on the dimension of a stratum, and
in Section~\ref{normal2} we prove that it is an upper bound.

We study how the strata are connected to each other in
the rank stratification and use that knowledge to understand
the geometry and topology of the fiber.
Let $S_{\underline{r}} \neq \emptyset$ and $S_{\underline{z}} \neq \emptyset$ be
two strata in $\mathbb{S}$ with
valid rank lists $\underline{r}$ and $\underline{z}$.
Recall that $\bar{S}$ denotes the closure of $S$.  
Our main result about stratum interconnections (Theorem~\ref{rankclosureequiv})
is that the following statements are equivalent (imply each other).
\begin{icompact}
\item[A.]  $S_{\underline{r}} \subseteq \bar{S}_{\underline{z}}$.
\item[B.]  $S_{\underline{r}} \cap \bar{S}_{\underline{z}} \neq \emptyset$.
\item[C.]  $\underline{r} \leq \underline{z}$. 
\item[D.]  There exists a sequence of rank-one abstract barcode moves that
  proceed from $\underline{r}$ to~$\underline{z}$,
  with all the intermediate rank lists being valid.
\end{icompact}

We have not yet defined the terminology in statement~D;
we will do so in Section~\ref{1hierarchy}
(with some hints in Section~\ref{dagdetails}).
A {\em stratum dag} is a directed acyclic graph (dag) with
one vertex for each stratum in $\mathbb{S}$ and
one directed edge for each rank-one abstract barcode move;
an example appears at right in Figure~\ref{strat112}.
For now, just know that these moves are a central new concept in this paper,
which organize the strata in $\mathbb{S}$ in a natural hierarchy.
Read statement~D as ``there is a directed path in the stratum dag
from $S_{\underline{r}}$ to $S_{\underline{z}}$.''

Among our equivalences, the most important is the equivalence of
statements~A and~C:
$S_{\underline{r}} \subseteq \bar{S}_{\underline{z}}$ if and only if
$\underline{r} \leq \underline{z}$.
The rank lists tell you which strata's closures include (or intersect)
which strata.

The stratum dag is acyclic because statements~C and~D are equivalent.  
The equivalence of A~and~D implies that
$S_{\underline{r}} \subseteq \bar{S}_{\underline{z}}$ if and only if
there is a path in the stratum dag from $S_{\underline{r}}$ to $S_{\underline{z}}$.

The fact that B implies~A means that rank stratifications satisfy
the {\em frontier condition} defined in the Introduction:
for every pair of strata $S, T \in \mathbb{S}$,
either $S \cap \bar{T} = \emptyset$ or $S \subseteq \bar{T}$.
(The two possibilities are mutually exclusive, because $S \neq \emptyset$.)
That is, the inclusion of points of~$S$ in $\bar{T}$ is all or nothing,
which helps us to understand how strata are connected to each other.
For example, if $S \subseteq \bar{T}$ and $S \neq T$, then
$\dim S < \dim T$ and from any point on $S$
there is an infinitesimal perturbation that takes us onto~$T$.
A directed path in the stratum dag visits
strata of monotonically increasing dimensions.

As the fiber is a closed point set, the frontier condition implies that
the closure of a stratum in $\mathbb{S}$ is a union of strata in $\mathbb{S}$.
By the equivalence of statements~A and~C,
\[
\bar{S}_{\underline{z}} = \bigcup_{\underline{r} \leq \underline{z}} S_{\underline{r}}.
\]

Figure~\ref{strat1111} depicts another nonmanifold fiber,
its rank stratification, and its stratum dag.
The fiber $\mu^{-1}([0])$ is the variety of solutions to
$[\theta_3] [\theta_2] [\theta_1] = [0]$
(an instantiation of $W_3 W_2 W_1 = W$).
The presence of a directed path from $S_{000}$ to~$S_{011}$ in the dag implies  
that $S_{000} \subset \bar{S}_{011}$ and $\dim S_{000} < \dim S_{011}$.  
Consistent with the frontier condition,
$S_{000} \subset \bar{S}_{011}$ and $S_{010} \subset \bar{S}_{011}$, whereas
$S_{100} \cap \bar{S}_{011} = \emptyset$ and
$S_{011} \cap \bar{S}_{100} = \emptyset$.
Moreover, $\bar{S}_{011} = S_{011} \cup S_{010} \cup S_{001} \cup S_{000}$. 

\begin{figure}
\centerline{\input{strat1111.pdf_t}}

\caption{\label{strat1111}
At left is the variety of solutions to
$W_3 W_2 W_1 = [\theta_3] [\theta_2] [\theta_1] = [0] = W$,
partitioned into seven strata:
$S_{000}$ is the origin;
$S_{001}$, $S_{010}$, and $S_{100}$ are
the three coordinate axes with the origin removed; and
$S_{011}$, $S_{101}$, and $S_{110}$ are the three coordinate planes with
the coordinate axes removed.
At right is the stratum dag, organized as
a three-dimensional table indexed by the ranks of $W_3$, $W_2$, and~$W_1$.
}
\end{figure}

Mather~\cite{mather70} showed that the stratifications constructed by
Whitney~\cite{whitney65a,whitney65b} satisfy the frontier condition.
But there exist stratifications that do not---for instance,
in Figure~\ref{strat112}, if we fuse $S_{00}$ and $S_{10}$ into a line and
leave $S_{01}$ unchanged, we have a stratification with only two strata,
which may seem simpler, but it does not satisfy the frontier condition, and
it is harder to express how the strata are connected to each other.
Fortunately, rank stratifications satisfy the frontier condition.

We will prove that in the rank stratification,
the topology of a stratum is determined solely by its rank list, and
its geometry is determined solely by
its rank list and a linear transformation in weight space.
Specifically, if we consider two strata that come from two different fibers
but have the same rank list, then
there is a linear homeomorphism mapping one stratum to the other and
one fiber to the other (Corollary~\ref{affinestratacor}).
Moreover, there is a sense in which every point on a stratum looks the same:
if we specify two points on the same stratum,
there is a bijective linear transformation mapping the stratum to itself and
mapping one specified point to the other (Corollary~\ref{selfsimilar}).

We devote a lot of writing to deriving
the tangent and normal spaces at each point on a stratum.
(Deriving these spaces is part of our derivation of the dimension of a stratum.)
These spaces help us understand the geometry of the fiber.
Table~\ref{resulttable} tabulates the most important of these results,
including explicit formulae for the space tangent to a stratum at $\theta$,
$T_\theta S$ (derived in Section~\ref{expressions}), and
the space normal to a stratum at $\theta$,
$N_\theta S$ (derived in Section~\ref{normal2}).
While it is nice to be able to write explicit formulae in terms of
nullspaces, rowspaces, and columnspaces,
for practical purposes it is more useful to compute
a basis for each of these subspaces, and we devote even more writing to
that task (Sections~\ref{1hierarchy} through~\ref{normal}).

\begin{table}
\begin{center}
\begin{tabular}{l}
\hline
\vspace*{.2in}
\begin{minipage}{6.3in}
\begin{eqnarray*}
\dim S & = &
d_\theta - \rank W \cdot (d_L + d_0 - \rank W) -  \\  
&   & \sum_{L \geq k+1 \geq i > 0} 
  \hspace*{-.1in}
  (\rank W_{k+1 \ssq i} - \rank W_{k+1 \ssq i-1}) \,
  (\rank W_{k \ssq i-1} - \rank W_{k+1 \ssq i-1}).  \\
T_\theta S & = &
\left\{ (\Delta W_L + W_L H_{L-1},
         \Delta W_{L-1} + W_{L-1} H_{L-2} - H_{L-1} W_{L-1}, \ldots,
\rule{0pt}{10pt}  \right.  \\
&   &    \Delta W_j + W_j H_{j-1} - H_j W_j, \ldots,
         \Delta W_2 + W_2 H_1 - H_2 W_2, \Delta W_1 - H_1 W_1) :  \\
&   &    H_j \in \R^{d_j \times d_j},  \\
&   &    \Delta W_j \in
\sum_{h=1}^{j-1} \col W_{j \ssq h} \otimes \Null W_{j-1 \ssq h-1}^\top +
(\Null W_{L \ssq j} \cap \col W_{j \ssq 0}) \otimes \R^{d_{j-1}} +  \\
&   &    \left. \sum_{l=j}^{L-1} \Null W_{l+1 \ssq j} \otimes \row W_{l \ssq j-1} +
         \R^{d_j} \otimes (\row W_{L \ssq j-1} \cap \Null W_{j-1 \ssq 0}^\top)
\right\}.  \\
\dim T_\theta S & = & \dim S.  \\
N_\theta S & = &
\left\{ (X^{(L0)}_L, X^{(L0)}_{L-1}, \ldots, X^{(L0)}_1) :
        M^{(L0)} \in \R^{d_L \times d_0} \right\} +  \\
& &
\sum_{L \ge t > s \ge 0} \left\{ (X^{(ts)}_L, X^{(ts)}_{L-1}, \ldots, X^{(ts)}_1) :
M^{(ts)} \in \Null W_{t \ssq s}^\top \otimes \Null W_{t \ssq s} \right\}
\hspace{.2in}  \mbox{where}  \\
X^{(ts)}_j & = &
\left\{ \begin{array}{rl}
W_{t \ssq j}^\top M^{(ts)} W_{j-1 \ssq s}^\top, & j \in [ s + 1, t ],  \\
0, & j \not\in [ s + 1, t ].
\end{array} \right.  \\
\dim N_\theta S & = & d_\theta - \dim S.  \\
\rank \dmu(\theta) & = &
\sum_{j=1}^L \rank W_{L \ssq j} \cdot \rank W_{j-1 \ssq 0} -
\sum_{j=1}^{L-1} \rank W_{L \ssq j} \cdot \rank W_{j \ssq 0}.
\hspace{.2in}  \mbox{(Trager et al.~\cite{trager20}, Lemma~3.)}  \\
\Null \dmu(\theta) & = &
\left\{ (\Delta W_L + W_L H_{L-1},
         \Delta W_{L-1} + W_{L-1} H_{L-2} - H_{L-1} W_{L-1}, \ldots,
\rule{0pt}{10pt}  \right.  \\
&   &    \Delta W_j + W_j H_{j-1} - H_j W_j, \ldots,
         \Delta W_2 + W_2 H_1 - H_2 W_2, \Delta W_1 - H_1 W_1) :  \\
&   &    \left. H_j \in \R^{d_j \times d_j},
         \Delta W_j \in \Null W_{L \ssq j} \otimes \R^{d_{j-1}} +
         \R^{d_j} \otimes \Null W_{j-1 \ssq 0}^\top \right\}.  \\
\dim \Null \dmu(\theta) & = & d_\theta - \rank \dmu(\theta).  \\
\row \dmu(\theta)
& = & \{ (X_L, X_{L-1}, \ldots, X_1) : M \in \R^{d_L \times d_0} \}
\hspace{.2in}  \mbox{where~}  \\
X_j & = & W_{L \ssq j}^\top M W_{j-1 \ssq 0}^\top.  \\
\dim \row \dmu(\theta) & = & \rank \dmu(\theta).  \\
\mbox{Notes:}
&   &
N_\theta S = (T_\theta S)^\perp; \row \dmu(\theta) = (\Null \dmu(\theta))^\perp;
\\
&   &
T_\theta S \subseteq \Null \dmu(\theta); N_\theta S \supseteq \row \dmu(\theta).
\end{eqnarray*}
\end{minipage}  \\
\hline
\end{tabular}
\end{center}

\caption{\label{resulttable}
Some of this paper's main results about
subspaces of the weight space $\R^{d_\theta}$ and their dimensions.
$S$~is the stratum of the fiber $\mu^{-1}(W)$ containing the point
$\theta = (W_L, W_{L-1}, \ldots, W_1)$,
$T_\theta S$ is the tangent space of~$S$ at $\theta$
($T_\theta S$ passes through the origin, not through $\theta$),
$N_\theta S$ is the normal space of~$S$ at $\theta$
(the orthogonal complement of $T_\theta S$), and
$\dmu(\theta)$ is the (linear) differential map of
the matrix multiplication map~$\mu$ at $\theta$,
defined in Section~\ref{weightprebases}.
In our illustrations of stratum dags, ``dim'' refers to $\dim S$ and
``dof'' refers to $\dim \Null \dmu(\theta)$.
}

\end{table}

We are particularly interested in points on the fiber where
multiple strata meet.
Given a point (weight vector) $\theta \in \mu^{-1}(W)$,
we describe a {\em fiber basis} composed of linearly independent vectors having
the property that for every stratum~$S$ whose closure contains $\theta$,
some subset of the fiber basis spans the tangent space $T_\theta \bar{S}$.
(This assumes that $\bar{S}$ has a tangent space at $\theta$---if not,
a more subtle characterization is needed.)
Each vector in the fiber basis is tangent to some smooth path leaving $\theta$
on the fiber.
For example, in Figure~\ref{strat1111}, at a point $\theta \in S_{001}$,
the three unit coordinate vectors can serve as the fiber basis:
the tangent plane $T_\theta \bar{S}_{011}$ is
spanned by the $\theta_1$- and $\theta_2$-axes,
and the tangent plane $T_\theta \bar{S}_{101}$ is
spanned by the $\theta_1$- and $\theta_3$-axes.
At a point $\theta \in S_{101}$, the fiber basis has just two vectors---the
$\theta_1$- and $\theta_3$-axes will do.

That example might mislead the reader.
In general, the vectors in the fiber basis cannot always be orthogonal to
each other, because strata do not always meet each other at right angles.
Moreover, every point on $\mu^{-1}(W)$ may need a different fiber basis, because
a typical fiber is curved.
Even the size of the fiber basis is different at different points on the fiber:
in our illustrations of stratum dags,
each stratum's ``degrees of freedom'' (dof) is
the number of vectors in the fiber basis.
But at any one point $\theta \in \mu^{-1}(W)$,
a single fiber basis suffices to describe all the tangencies, so
it deepens our understanding of {\em how} the strata meet at $\theta$.

A key tool for understanding the geometry of the strata and
the interconnections among them is a set of operations we call {\em moves}
that map one weight vector on the fiber to another weight vector,
which may or may not be on the fiber.
Moves that stay on the fiber replace a linear neural network with
another that computes the same function.
Moves have both theoretical and practical motivations.
The theoretical motivation is that moves provide a great deal of intuition about
the geometry and topology of the fiber.
Most moves on the fiber imply smooth paths on the fiber,
which help us determine tangent spaces.
Some moves that we call ``discrete moves'' reveal
how the strata are connected to each other, by exposing
an interesting combinatorial structure that links those connections to
the rank lists of the strata.

The practical motivation for moves is that
although two different neural networks might compute the same transformation,
one might be much more amenable to training than the other.
It is well known that during training, a neural network's weight vector may lie
near a critical point in the cost function that slows down network training but
is ``spurious'' in the sense that
it is not related to the transformation being learned;
it is merely a side effect of how that transformation happens to be encoded
in the network's layers~\cite{trager20}.
Spurious critical points appear to be one of the reasons that
deep neural networks sometimes learn more slowly than shallow ones.
Our original motivation for studying the fiber was to find
ways to move away from spurious critical points, thereby speeding up learning,
without changing the function that a network has learned.
(This paper doesn't solve that problem, but it is a step along the way.)

Each stratum represents a different pattern by which information flows
(or fails to flow) through the neural network.
To understand information flow in a linear neural network, we propose
a \textit{Fundamental Theorem of Linear Neural Networks},
analogous to what Strang~\cite{strang93} calls
the \textit{Fundamental Theorem of Linear Algebra}.
There are four fundamental subspaces of a matrix:
its domain can be decomposed into its rowspace and its nullspace, and
its range can be decomposed into its columnspace and its left nullspace.
In a linear neural network, there is an analogous decomposition of
each layer of units into a set of ``fundamental'' subspaces
that show how information flows through the neural network---in particular,
tracing which information is annihilated at which layers of the network, and
identifying subspaces that carry no information but might start
to carry information as training modifies the network weights.
These decompositions were first investigated in the study of
{\em persistent homology}
\cite{edelsbrunner00,edelsbrunner02,zomorodian05,edelsbrunner10,chazal16,loh23},
a subfield of computational topology, but
we have not previously seen them applied to neural networks.
We summarize these information flows in ``barcodes''
that reveal a rich and occasionally surprising structure.
There is one barcode for each stratum $S$.
We use the fundamental subspaces as building blocks to construct
the fiber basis and the subspaces $T_\theta S$ and $N_\theta S$ at $\theta$.

\subsection{Some Details about Stratum Dags}
\label{dagdetails}

One goal of this paper is to provide an algorithm that generates
stratum dags for any $L \geq 2$, given $L$, the~$d_j$'s, and~$\rank W$ as input.
That algorithm appears in Section~\ref{computedag}.
Here we discuss some finer points of stratum dags and their structure.

Return to Figure~\ref{strat112}, depicting
the rank stratification of the solutions to
$W_2 W_1 = [\theta_2] [\theta_1 ~~~ \theta'_1] = [0 ~~~ 0] = W$.
There are two ways to achieve $W_2 W_1 = [0 ~~~ 0]$:
we can set $W_2 = [0]$ or we can set $W_1 = [0 ~~~ 0]$.
The former solutions lie on the pink plane in Figure~\ref{strat112}, and
the latter solutions lie on the blue line.
Recall that the subscripts of $S_{**}$ are $\rank W_2$ and $\rank W_1$.
If we set both $W_2$ and $W_1$ to zero,
our weight vector $\theta = (W_2, W_1)$ is the origin,
a $0$-dimensional stratum labeled $S_{00}$.
If we set only $W_2$ to zero, our weight vector lies on the stratum $S_{01}$,
the pink plane with the origin removed.
If we set only $W_1$ to zero, our weight vector lies on the stratum $S_{10}$,
the blue line with the origin removed.
 
We have omitted the other ranks in the rank list from
the subscripts of $S$ in Figure~\ref{strat112} because
only $\rank W_2$ and $\rank W_1$ vary.
The omitted ranks, the $d_j$'s and $\rank W$, are fixed for a specific fiber.
In general, a rank list has $(L^2 + 3L) / 2 + 1$ numbers, but
if we leave out the $d_j$'s and $\rank W$, we have $(L^2 + L) / 2 - 1$ numbers.
Hence it is natural to organize the strata in a table or array with
$(L^2 + L) / 2 - 1$ indices or dimensions.

In Figure~\ref{strat1111}
(the rank stratification of the solutions to
$W_3 W_2 W_1 = [\theta_3] [\theta_2] [\theta_1] = [0] = W$),
we arrange the stratum dag in a three-dimensional table,
indexed by $\rank W_3$, $\rank W_2$, and~$\rank W_1$.
Ordinarily, three-matrix fibers ($L = 3$) require five indices to index
the strata, as $\rank W_3W_2$ and $\rank W_2W_1$ can vary as well; but
in this example every matrix is $1 \times 1$, so
those two ranks are uniquely determined by the first three.

\begin{figure}
\begin{center}
\begin{tabular}{r|lr|c|lr|c|lr|c|lr|}
\cline{2-3} \cline{5-6}
\multirow{2}{*}{\hspace{-.1in} $\rank W_2 = 5$}  
    &  dim: 31  &  dof: 34   &   
       \multirow{2}{*}{\makebox[0in]{\hspace*{-.035in} $\rightarrow$}}   &
       dim: 34  &  dof: 34   \\  
    &  $S_{51}$ &  rdof: \; 3  & &
       $S_{52}$ &  rdof: \; 0  \\
\cline{2-3} \cline{5-6}
\multicolumn{1}{c}{}
    &  \multicolumn{2}{c}{$\uparrow$}
    &  \multicolumn{1}{c}{}
    &  \multicolumn{2}{c}{$\uparrow$}  \\
\cline{2-3} \cline{5-6} \cline{8-9}
\multirow{2}{*}{\hspace{-.1in} $\rank W_2 = 4$}  
    &  dim: 29  &  dof: 37   &
       \multirow{2}{*}{\makebox[0in]{\hspace*{-.035in} $\rightarrow$}}   &
       dim: 33  &  dof: 36   &
       \multirow{2}{*}{\makebox[0in]{\hspace*{-.035in} $\rightarrow$}}   &
       dim: 35  &  dof: 35   \\
    &  $S_{41}$ &  rdof: \; 8  & &
       $S_{42}$ &  rdof: \; 3  & &
       $S_{43}$ &  rdof: \; 0  \\
\cline{2-3} \cline{5-6} \cline{8-9}
\multicolumn{1}{c}{}
    &  \multicolumn{2}{c}{$\uparrow$}
    &  \multicolumn{1}{c}{}
    &  \multicolumn{2}{c}{$\uparrow$}
    &  \multicolumn{1}{c}{}
    &  \multicolumn{2}{c}{$\uparrow$}  \\
\cline{2-3} \cline{5-6} \cline{8-9} \cline{11-12}
\multirow{2}{*}{\hspace{-.1in} $\rank W_2 = 3$}  
    &  dim: 25  &  dof: 40   &
       \multirow{2}{*}{\makebox[0in]{\hspace*{-.035in} $\rightarrow$}}   &
       dim: 30  &  dof: 38   &
       \multirow{2}{*}{\makebox[0in]{\hspace*{-.035in} $\rightarrow$}}   &
       dim: 33  &  dof: 36   &
       \multirow{2}{*}{\makebox[0in]{\hspace*{-.035in} $\rightarrow$}}   &
       dim: 34  &  dof: 34   \\
    &  $S_{31}$ &  rdof: 15  & &
       $S_{32}$ &  rdof: \; 8  & &
       $S_{33}$ &  rdof: \; 3  & &
       $S_{34}$ &  rdof: \; 0  \\
\cline{2-3} \cline{5-6} \cline{8-9} \cline{11-12}
\multicolumn{1}{c}{}
    &  \multicolumn{2}{c}{$\uparrow$}
    &  \multicolumn{1}{c}{}
    &  \multicolumn{2}{c}{$\uparrow$}
    &  \multicolumn{1}{c}{}
    &  \multicolumn{2}{c}{$\uparrow$}
    &  \multicolumn{1}{c}{}
    &  \multicolumn{2}{c}{$\uparrow$}  \\
\cline{2-3} \cline{5-6} \cline{8-9} \cline{11-12}
\multirow{2}{*}{\hspace{-.1in} $\rank W_2 = 2$}  
    &  dim: 19  &  dof: 43   &
       \multirow{2}{*}{\makebox[0in]{\hspace*{-.035in} $\rightarrow$}}   &
       dim: 25  &  dof: 40   &
       \multirow{2}{*}{\makebox[0in]{\hspace*{-.035in} $\rightarrow$}}   &
       dim: 29  &  dof: 37   &
       \multirow{2}{*}{\makebox[0in]{\hspace*{-.035in} $\rightarrow$}}   &
       dim: 31  &  dof: 34   \\
    &  $S_{21}$ &  rdof: 24  & &
       $S_{22}$ &  rdof: 15  & &
       $S_{23}$ &  rdof: \; 8   & &
       $S_{24}$ &  rdof: \; 3   \\
\cline{2-3} \cline{5-6} \cline{8-9} \cline{11-12}
\multicolumn{1}{c}{}
    &  \multicolumn{2}{c}{$\uparrow$}
    &  \multicolumn{1}{c}{}
    &  \multicolumn{2}{c}{$\uparrow$}
    &  \multicolumn{1}{c}{}
    &  \multicolumn{2}{c}{$\uparrow$}
    &  \multicolumn{1}{c}{}
    &  \multicolumn{2}{c}{$\uparrow$}  \\
\cline{2-3} \cline{5-6} \cline{8-9} \cline{11-12}
\multirow{2}{*}{\hspace{-.1in} $\rank W_2 = 1$}  
    &  dim: 11  &  dof: 46   &
       \multirow{2}{*}{\makebox[0in]{\hspace*{-.035in} $\rightarrow$}}   &
       dim: 18  &  dof: 42   &
       \multirow{2}{*}{\makebox[0in]{\hspace*{-.035in} $\rightarrow$}}   &
       dim: 23  &  dof: 38   &
       \multirow{2}{*}{\makebox[0in]{\hspace*{-.035in} $\rightarrow$}}   &
       dim: 26  &  dof: 34   \\
    &  $S_{11}$ &  rdof: 35  & &
       $S_{12}$ &  rdof: 24  & &
       $S_{13}$ &  rdof: 15  & &
       $S_{14}$ &  rdof: \; 8  \\
\cline{2-3} \cline{5-6} \cline{8-9} \cline{11-12}
\multicolumn{1}{c}{}
    &  \multicolumn{2}{c}{$\rank W_1 = 1$}
    &  \multicolumn{1}{c}{}
    &  \multicolumn{2}{c}{$\rank W_1 = 2$}
    &  \multicolumn{1}{c}{}
    &  \multicolumn{2}{c}{$\rank W_1 = 3$}
    &  \multicolumn{1}{c}{}
    &  \multicolumn{2}{c}{$\rank W_1 = 4$}
\end{tabular}
\end{center}

\caption{\label{strat564}
Stratum dag representing the stratification of $\mu^{-1}(W)$ for
$W = W_2 W_1$, $W_2 \in \R^{5 \times 6}$, $W_1 \in \R^{6 \times 4}$, and
$\rank W = 1$.
For every pair of strata $S_{ts}$ and $S_{t's'}$
with $t \leq t'$ and $s \leq s'$, $S_{ts} \subseteq \bar{S}_{t's'}$.
}
\end{figure}

Figure~\ref{strat564} depicts the stratum dag representing
the rank stratification of the fiber $\mu^{-1}(W)$
for any $5 \times 4$ matrix~$W$ with rank~$1$ and
a network with $L = 2$, $d_2 = 5$, $d_1 = 6$, and $d_0 = 4$.
The fiber is embedded in a $54$-dimensional weight space, and
the strata have dimensions ranging from $11$ to $35$.
Unfortunately, we cannot visualize the fiber, but
it has multiple strata like in Figure~\ref{strat1111}, and
the strata are curved like in Figure~\ref{hyperboloid}.

In the stratum dags in
Figures~\ref{strat112}, \ref{strat1111}, and~\ref{strat564},
each dag vertex lists the dimension of the stratum (dim $= D^\strat$),
the number of degrees of freedom along which smooth motion on the fiber is
possible (dof), and how many of those degrees of freedom increase at least one
rank in the rank list (rdof, for ``rank-increasing degrees of freedom''), and
thus represent moves off the stratum onto a higher-dimensional stratum.
These numbers always satisfy dof $=$ dim $+$ rdof.
Recall that the number of degrees of freedom (dof) is
the number of vectors in the fiber basis at $\theta$.
More directly, ``dof'' is the dimension of the subspace $\Null \dmu(\theta)$,
called the {\em nullspace of the differential map of $\mu$ at $\theta$},
defined in Section~\ref{diffmap}.
Formulae for $\Null \dmu(\theta)$ and its dimension appear in
Table~\ref{resulttable}.

Figures~\ref{strat112}, \ref{strat1111}, and~\ref{strat564} illustrate
that some fibers have rank stratifications with multiple maximal elements.
Thus, there is no single stratum whose closure is the whole fiber
(equivalently, no dag vertex that is reachable from all the other vertices).
By contrast, every nonempty fiber's rank stratification has
one unique minimal element---a stratum that is
a subset of the closure of every other stratum.
In the stratum dag, it is represented by a dag vertex from which
all the other vertices are reachable.
In the figures, the minimal strata are $S_{00}$, $S_{000}$, and $S_{11}$,
respectively.
The minimal element is the stratum for which
every subsequence matrix's rank is $\rank W$,
except the layer sizes $\rank W_{j \ssq j} = d_j$.

For two-matrix fibers ($L = 2$), the general table shape is
a pentagon like the one illustrated in Figure~\ref{strat564}.
We have drawn the stratum dag so that the horizontal axis specifies $\rank W_1$
and the vertical axis specifies $\rank W_2$ for each stratum.
The left and right boundaries of the table are determined by
$\rank W_1 \in [\rank W, \min \{ d_1, d_0 \}]$, and
the top and bottom boundaries are determined by
$\rank W_2 \in [\rank W, \min \{ d_2, d_1 \}]$.
Sometimes the upper right corner of the table is cut off by a fifth constraint:
by Sylvester's inequality, $\rank W_2 + \rank W_1 \leq d_1 + \rank W$.
In Figure~\ref{strat564}, that means $\rank W_2 + \rank W_1 \leq 7$.
This inequality generates the fifth edge of the pentagon.

When $L > 2$, the table has $(L^2 + L) / 2 - 1$ indices and
its shape is more complicated than a pentagon because then
we have not one inequality, but
many occurrences of the Frobenius rank inequality
$\rank W_{l \ssq i} + \rank W_{k \ssq h} \leq
\rank W_{l \ssq h} + \rank W_{k \ssq i}$ (where $l \geq k \geq i \geq h$),
dictating how the ranks in the rank list constrain each other, and thereby 
dictating which rank lists are valid.
Sections~\ref{enumeratesec} and~\ref{enumeratesec2} describe
two algorithms for determining the set of
valid rank lists---equivalently, the set of strata in the rank stratification.

Figures~\ref{strat112}, \ref{strat1111}, and~\ref{strat564} might give
the false impression that stratum dag edges are always axis-aligned,
representing a change of a single rank.
But in networks with more layers, the stratum dags typically have edges
that represent the increase of multiple ranks simultaneously.
See Section~\ref{dagedges}, and also Figure~\ref{connecting} (bottom).

Recall that by definition, a stratum dag has
a directed edge $(S_{\underline{r}}, S_{\underline{z}})$ if 
there exists a {\em rank-$1$ abstract barcode move} that
transforms the rank list $\underline{r}$ to the rank list $\underline{z}$.
Unfortunately, we won't be able to define those terms until
Section~\ref{abstractmoves}.
We can give a hint:  think of a ``rank-$1$ barcode move'' as
an infinitesimally small move on the fiber from one stratum to another,
along a direction chosen from the fiber basis, that
increases the ranks of one or more subsequence matrices by one
(and decreases the rank of none).
The stratum dag encodes the partial order of the strata induced by $\leq$, but
it is not necessarily the simplest dag that does so (sometimes called
a {\em Hasse diagram} or {\em transitive reduction}\footnote{
To use some more precise terms:
the stratum dag is not necessarily its own transitive reduction;
the edges of the stratum dag do not necessarily express
the covering relation of the partial order.
}),
because sometimes a stratum dag contains
a directed edge~$(S_{\underline{r}}, S_{\underline{z}})$ despite
also containing the edges $(S_{\underline{r}}, S_{\underline{u}})$ and
$(S_{\underline{u}}, S_{\underline{z}})$.
The forthcoming Section~\ref{abstractmoves} and Figure~\ref{multiswap} give
an example of that.
Roughly speaking, an edge $(S_{\underline{r}}, S_{\underline{z}})$ indicates that
from a weight vector $\theta \in S_{\underline{r}}$,
the stratum $S_{\underline{z}}$ permits
one or more degrees of freedom of motion (i.e., directions in the fiber basis)
not ``covered'' by any of the strata $S_{\underline{u}}$ such that
$\underline{r} \leq \underline{u} < \underline{z}$
(i.e., the predecessors of $S_{\underline{z}}$ in the stratum dag
whose closures contain~$\theta$).
In Figure~\ref{strat1111}, the stratum dag has no edge $(S_{000}, S_{011})$
because, although there are two degrees of freedom by which
the weight vector in $S_{000}$ can be perturbed onto $S_{011}$,
they are spanned by the degree of freedom of motion on $S_{001}$ and
the degree of freedom of motion on $S_{010}$.

\clearpage

\section{The Fundamental Subspaces of a Linear Neural Network}
\label{fundsub}

One of the fundamental concepts of linear algebra is that of
the nullspace of a matrix $W$:  the set of vectors~$x$ such that $Wx = 0$.
Given a linear neural network, we can refine the concept by asking,
at what layer does a particular input vector $x$ first {\em disappear}?
Formally, what is the smallest $i$ such that
$W_i W_{i-1} \cdots W_2 W_1 x = {\bf 0}$?
This question is answered by inspecting the nullspaces of
the ``part way there'' matrices $W_{i \ssq 0} = W_i W_{i-1} \cdots W_2 W_1$,
which form a hierarchy of subspaces
\[
\Null W \supseteq \Null W_{L-1 \ssq 0} \supseteq \Null W_{L-2 \ssq 0} \supseteq
\ldots \supseteq \Null W_3 W_2 W_1 \supseteq \Null W_2 W_1 \supseteq \Null W_1.
\]

We can extend the concept further by observing that
linear neural networks can have unused subspaces in hidden layers---subspaces
through which information could flow if it were present, but
the earlier layers are not putting any information into those subspaces.
If the left nullspace $\Null W_{i \ssq 0}^\top$ is not
the trivial subspace $\{ {\bf 0} \}$
(for example, if unit layer $i$ has more units than unit layer $i - 1$),
then the space $\R^{d_i}$ encoded by unit layer $i$ has
one or more ``wasted'' dimensions that carry no information about the input $x$.
We ask:  if information were somehow injected into these left nullspaces,
would it affect the network's output, or
would it be absorbed in the nullspaces of subsequent matrices downstream?
The answer affects gradient descent algorithms for learning,
whose success may depend on injecting information into these channels.

These questions are motivated by both practice and theory.
The main practical motivation comes from neural network training.
Although the wasted dimensions emerging from left nullspaces have no influence
on the linear transformation $W$ that the network computes,
they have tremendous influence on whether a gradient descent algorithm can find
weight updates that improve the network's performance.
To illustrate this fact, consider the neural network with weight vector
$\theta = \left(
\left[ \begin{array}{cc} 1 & 0 \\ 0 & 0 \end{array} \right],
\left[ \begin{array}{cc} 1 & 0 \\ 0 & 0 \end{array} \right],
\left[ \begin{array}{cc} 1 & 0 \\ 0 & 0 \end{array} \right]
\right)$.
This network computes a linear transformation $W$ of rank~$1$ and
standard gradient descent algorithms cannot find
a search direction that increases $W$'s rank above $1$.
In the language of Trager et al.~\cite{trager20},
the cost function has a spurious critical point.
Whereas in the network with weight vector
$\theta = \left(
\left[ \begin{array}{cc} 1 & 0 \\ 0 & 1 \end{array} \right],
\left[ \begin{array}{cc} 1 & 0 \\ 0 & 1 \end{array} \right],
\left[ \begin{array}{cc} 1 & 0 \\ 0 & 0 \end{array} \right]
\right)$,
which computes the same transformation,
gradient descent can easily find a way to update $W_1$ that increases
the rank of~$W$, because
the subspace $\Null W_1^\top$ in hidden layer~1 is already connected to
the network's output layer.

Our main theoretical motivation is the recognition that
the rank stratification of the fiber $\mu^{-1}(W)$ of a matrix $W$ has
one stratum for each rank list, representing
one specific ``state of subspace flow'' through the network.
A network has only finitely many possible such states/strata/rank lists,
even though the subspaces and the transformations those subspaces undergo are
numerical and continuously-valued.
Another theoretical motivation is that we will use the subspaces we discover
to derive the dimensions of the strata and their tangent and normal spaces.
By that means, our characterization of subspace flow will help us to understand
the topology and geometry of the fiber.

Surprisingly to us, many of these observations about
the fundamental subspaces of a linear neural network were first made in
the study of persistent homology
\cite{edelsbrunner00,edelsbrunner02,edelsbrunner10}
(for entirely different reasons)
where they are used to analyze persistence modules
\cite{zomorodian04,zomorodian05,ghrist08,chazal16,loh23},
which are essentially linear neural networks -- but
usually over the field $\mathbb{Z} / 2\mathbb{Z}$,
in contrast to our real-valued weights.
As far as we have seen, the literature on persistence modules has not noted
the connections to neural networks, and
it pays little attention to the subspaces themselves.
We describe these relationships in Section~\ref{modules}.

\subsection{Barcodes and Annotated Barcodes}
\label{barcodes}

\begin{figure}
\centerline{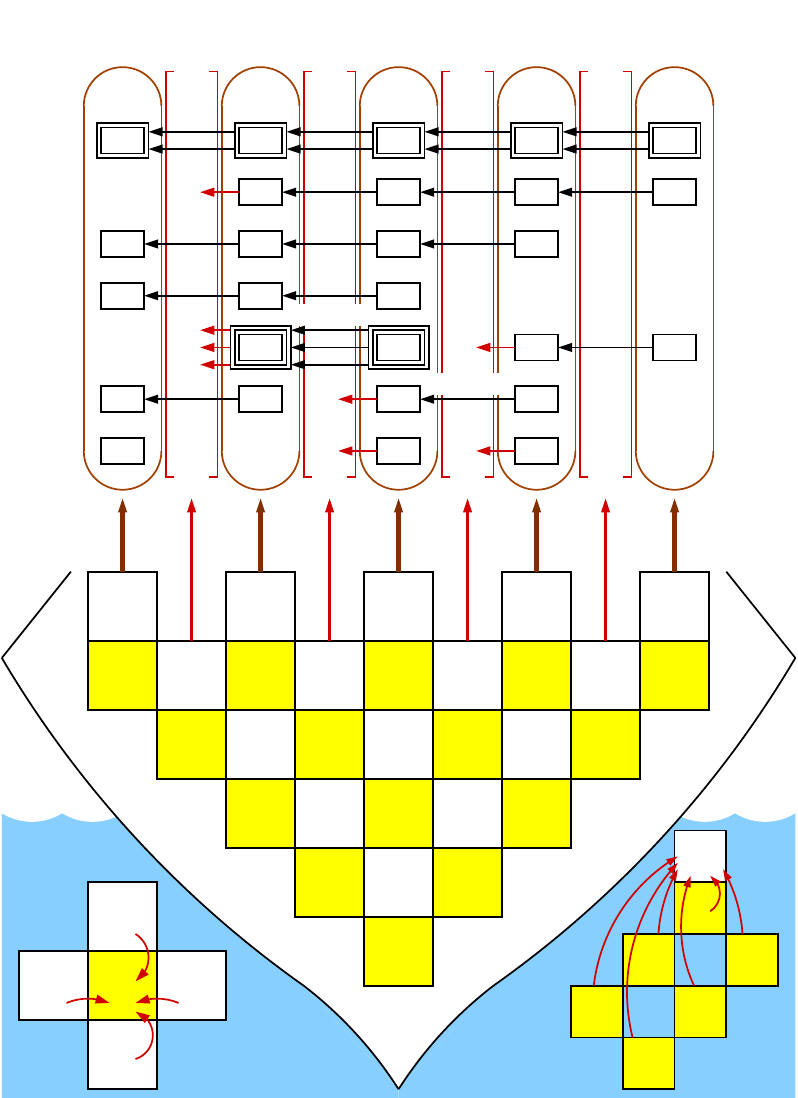}

\caption{\label{interval}
The top half is an annotated barcode that illustrates
the flow of the ``fundamental'' subspaces~$a_{kji}$ through the network.
Double boxes represent subspaces of dimension~2 and
triple boxes represent subspaces of dimension~3.
The bottom half relates
the interval counts, the layer sizes, and the subsequence matrix ranks.
The number of units $d_j$ in unit layer~$j$ is
the sum of the counts~$\omega_{ts}$ of
the intervals that touch layer~$j$
(i.e., the dimensions of the subspaces~$a_{tjs}$).
Each matrix rank $\rank W_{k \ssq i}$ is the sum of the counts of
the intervals that touch both layers~$k$ and~$i$.
}
\end{figure}

We will see that the state of information flow can be represented as
a multiset of intervals called a {\em barcode},
depicted in the top half of Figure~\ref{interval}.
The idea of barcodes originates in persistent homology~\cite{ghrist08};
see Section~\ref{modules}.
An {\em interval} is a set of consecutive integers
$[i, k] = \{ i, i + 1, \ldots, k - 1, k \}$ that signifies
consecutive unit layers. 
Layer~$i$ is called the {\em birth layer} and layer~$k$ the {\em death layer}
for the information that the interval represents.
In a multiset of intervals,
each interval has a {\em multiplicity} $\omega_{ki}$, representing
$\omega_{ki}$ copies of $[i, k]$.
If an interval is absent from the multiset,
its multiplicity is zero ($\omega_{ki} = 0$).
We will call each $\omega_{ki}$ an {\em interval count};
``interval multiplicity,'' though more precise, is too many syllables.
A barcode is fully specified by the interval counts
$\omega_{ki} \geq 0$ for all $k$ and $i$ satisfying $L \geq k \geq i \geq 0$.

Think of an interval $[i, k]$ with multiplicity $\omega_{ki}$ as representing
an $\omega_{ki}$-dimensional subspace that appears
at unit layer $i$ (its {\em birth layer}), being linearly independent of
(but not necessarily orthogonal to) $\col W_i$, then
flows through the weight layers $W_{i+1}, W_{i+2}, \ldots, W_k$,
being linearly transformed at each step until reaching unit layer $k$
(its {\em death layer}) with $\omega_{ki}$ dimensions still intact,
only to disappear into the nullspace of $W_{k+1}$
(unless layer $k$ is the output layer~$L$).

In the simplest example, a one-matrix network ($L = 1$),
the subspaces we speak of are the four fundamental subspaces
$\row W_1$, $\col W_1$, $\Null W_1$, and $\Null W_1^\top$.
The interval $[0, 1]$ represents
the rowspace of~$W_1$ (at the input layer),
the columnspace of $W_1$ (at the output layer), and
the fact that applying $W_1$ to the former yields the latter.
Both $\row W_1$ and $\col W_1$ have dimension $\omega_{10} = \rank W_1$.
The interval $[0, 0]$ represents the nullspace of~$W_1$, which
disappears into $W_1$, sending no information to the output layer.
The dimension of $\Null W_1$ is~$\omega_{00}$.
The interval $[1, 1]$ represents the left nullspace of $W_1$,
the unused dimensions of the output layer.
The dimension of $\Null W_1^\top$ is $\omega_{11}$.
The barcode is the multiset indicated by
the multiplicities $\omega_{10}$, $\omega_{00}$, and~$\omega_{11}$.

There is a second interpretation governing
the transpose network $W^\top = W_1^\top W_2^\top \cdots W_{L-1}^\top W_L^\top$:
the interval $[i, k]$ represents
a (different!)\ $\omega_{ki}$-dimensional subspace that appears
at unit layer~$k$, being linearly independent of $\row W_{k+1}$, then
flows through the weight layers $W_k^\top, W_{k-1}^\top, \ldots, W_{i+1}^\top$
being linearly transformed at each step until reaching
unit layer $i$ with $\omega_{ki}$ dimensions still intact,
only to disappear into the left nullspace of $W_i$ (unless $i = 0$).
We will need both interpretations to derive
tangent and normal spaces on the fiber~$\mu^{-1}(W)$.

The {\em annotated barcode} in the top half of Figure~\ref{interval} is
a barcode annotated with {\em flow prebasis subspaces}
of the form $a_{kji} \subseteq \R^{d_j}$, with $L \geq k \geq j \geq i \geq 0$.
We think of these flow prebasis subspaces as
``fundamental subspaces'' of a linear neural network
(although they are not always unique), akin to
the fundamental subspaces of a matrix.
They have two important properties.
First, they satisfy linear {\em flow relationships}:
the black arrows in Figure~\ref{interval} indicate that
$a_{kji} = W_j a_{k,j-1,i}$ if $k \geq j > i$, and
the red arrows indicate that $W_{k+1} a_{kki} = 0$, which is why
layer $k$ is called the death layer of the interval $[i, k]$.
Second, for each unit layer $j \in [0, L]$,
the flow prebasis subspaces of the form $a_{kji}$ with matching $j$ are
linearly independent and the sum of their dimensions is $d_j$.
One way to think about this is that we can form a basis for $\R^{d_j}$
(unit layer $j$) by specifying, for each interval $[i, k]$ that contains~$j$,
$\omega_{ki}$ basis vectors spanning $a_{kji}$.

A barcode is {\em valid} for a specified network if
it satisfies the constraint that for each unit layer $j \in [0, L]$,
the sum of the multiplicities of the intervals that contain $j$ is $d_j$.
That is,
\begin{equation}
d_j = \sum_{t = j}^L \sum_{s = 0}^j \omega_{ts}
\hspace{.2in}
\mbox{for all~} j \in [0, L].
\label{intervallayer}
\end{equation}
We emphasize a visual interpretation of this identity in
Figure~\ref{interval}:
you can see that verifying whether a barcode is valid is
a simple matter of counting multiplicities in each unit layer~$j$ and
comparing the count with~$d_j$ (i.e., counting basis vectors).

Recall the {\em subsequence matrices} $W_{k \ssq i} = W_k W_{k-1} \cdots W_{i+1}$.
We will see in Section~\ref{ranksbases} that a barcode gives us
an easy way to determine the rank of any subsequence matrix:
the rank of $W_{k \ssq i}$ is the total multiplicity of the intervals that
contain both $i$ and $k$.
That is,
\[
\rank W_{k \ssq i} = \sum_{t = k}^L \sum_{s = 0}^i \omega_{ts}.
\]
Again, study Figure~\ref{interval}:  you can easily read
the rank of each subsequence matrix off the intervals.

In particular,
\begin{equation}
\rank W = \omega_{L0}.
\label{Wrank}
\end{equation}
That is, the interval $[0, L]$ always has multiplicity $\rank W$.
This interval represents the fact that
the subspace $\row W$ at the input layer is mapped by $W$
to $\col W$ at the output layer, and both subspaces have dimension $\rank W$.
When we consider the fiber $\mu^{-1}(W)$ of a specific matrix $W$,
$\rank W$ and $\omega_{L0}$ are fixed, and
so is \mbox{$d_j = \rank W_{j \ssq j}$} for all $j \in [0, L]$.
The other ranks and multiplicities may vary across
different points on $\mu^{-1}(W)$.

Each valid barcode is associated with a particular rank list,
a particular annotated barcode like Figure~\ref{interval} (top), and
a particular stratum of the fiber $\mu^{-1}(W)$ for
any~$W \in \R^{d_L \times d_0}$ with rank $\omega_{L0}$.
Recall that a rank list~$\underline{r}$ is {\em valid} if
there exists some weight vector $\theta \in \R^{d_\theta}$ that has
rank list $\underline{r}$.
In Section~\ref{ranksbases}
we show that there is a bijection from
valid rank lists to valid barcodes:
given a rank list, we can easily determine the interval counts, and
given the interval counts, we can easily determine the ranks.
In~Appendix~\ref{valid} we elaborate on why
every valid barcode maps to a valid rank list.

We will see in Section~\ref{barcmoves} that we can understand
the hierarchy of strata in terms of simple modifications of the barcode.
(We can also understand it in terms of modifications of the
rank list, but those modifications are not as simple.)

\subsection{Flow Subspaces and their Hierarchies}
\label{flowspaces}

In this section, we identify subspaces in each unit layer's space $\R^{d_j}$
that represent information flowing through the linear neural network
(or through
the transpose network $W^\top = W_1^\top W_2^\top \cdots W_{L-1}^\top W_L^\top$),
with special attention to information that does not reach the output layer.
We are aided by the fact that, at a unit layer~$j$ in the network,
the fundamental subspaces associated with the subsequence matrices are
nested in hierarchies (filtrations) as follows.
\begin{align*}
& \R^{d_j} = \row W_{j \blacktriangleleft j} \supseteq \row W_{j + 1} \supseteq
  \row W_{j+2} W_{j+1} \supseteq \ldots \supseteq \row W_{L \blacktriangleleft j} \supseteq
  \row W_{L+1 \blacktriangleleft j} = \{ {\bf 0} \}, \\
& \{ {\bf 0} \} = \Null W_{j \blacktriangleleft j} \subseteq \Null W_{j+1} \subseteq
  \Null W_{j+2} W_{j+1} \subseteq \ldots \subseteq \Null W_{L \blacktriangleleft j} \subseteq
  \Null W_{L+1 \blacktriangleleft j} = \R^{d_j},  \\
& \R^{d_j} = \col W_{j \blacktriangleleft j} \supseteq \col W_j \supseteq \col W_j W_{j-1}
  \supseteq \ldots \supseteq \col W_{j \blacktriangleleft 0} \supseteq
  \col W_{j \blacktriangleleft -1} = \{ {\bf 0} \},  \hspace{.2in}  \mbox{and}  \\
& \{ {\bf 0} \} = \Null W_{j \blacktriangleleft j}^\top \subseteq \Null W_j^\top \subseteq
  \Null (W_j W_{j-1})^\top \subseteq \ldots \subseteq \Null W_{j \blacktriangleleft 0}^\top
  \subseteq \Null W_{j \blacktriangleleft -1}^\top = \R^{d_j}.
\end{align*}
By the Fundamental Theorem of Linear Algebra,
the subspaces in the first row are orthogonal complements of
the corresponding subspaces in the second row, and
the subspaces in the third row are orthogonal complements of
the corresponding subspaces in the fourth row.
Here, we are using the following conventions for subsequence matrices.
\begin{align*}
& W_{j \blacktriangleleft j} = I_{d_j \times d_j}
  \hspace{.2in}  \mbox{(the $d_j \times d_j$ identity matrix)}.  \\
& \mbox{Hence,~} \row W_{j \blacktriangleleft j} = \R^{d_j} = \col W_{j \blacktriangleleft j} \mbox{~~and~~}
  \Null W_{j \blacktriangleleft j} = \{ {\bf 0} \} = \Null W_{j \blacktriangleleft j}^\top.  \\
& W_{j \blacktriangleleft -1} = 0_{d_j \times 1}  \hspace{.2in}  \mbox{and}  \hspace{.2in}
  W_{L+1 \blacktriangleleft j} = 0_{1 \times d_j}  \hspace{.2in}  \mbox{(zero matrices)}.  \\
& \mbox{Hence,~} \row W_{L+1 \blacktriangleleft j} = \{ {\bf 0} \} = \col W_{j \blacktriangleleft -1}
  \hspace{.2in}  \mbox{and}  \hspace{.2in}
  \Null W_{L+1 \blacktriangleleft j} = \R^{d_j} = \Null W_{j \blacktriangleleft -1}^\top.
\end{align*}
(For ease of remembering,
note that the last two lines are consistent with imagining that
the network $W_L W_{L-1} \cdots W_1$ is sandwiched between
two extra matrices $W_{L+1} = 0$ and $W_0 = 0$.)

From these four hierarchies,
we define two hierarchies of {\em flow subspaces} that give us insight about
how information flows, and sometimes fails to flow, through the network.
The flow subspaces of $\R^{d_j}$ at unit layer $j \in [0, L]$ are
\begin{eqnarray}
A_{kji} = \Null W_{k+1 \blacktriangleleft j} \cap \col W_{j \blacktriangleleft i},
& &  k \in [j - 1, L], i \in [-1, j],
\hspace{.2in}  \mbox{and}  \label{A}  \\
B_{kji} = \row W_{k \blacktriangleleft j} \cap \Null W_{j \blacktriangleleft i-1}^\top,
& &  k \in [j, L + 1], i \in [0, j + 1].
\label{B}
\end{eqnarray}
For example, $A_{320} = \Null W_4 W_3 \cap \col W_2 W_1$ and
$B_{320} = \row W_3 \cap \Null W_{2 \ssq -1}^\top = \row W_3$.
We will use commas to separate the subscripts when necessary for
clarity; e.g., $A_{t-1,s+1,-1}$.
Intuitively, $A_{kji} \in \R^{d_j}$ is the subspace that carries
information in unit layer $j$ that was born in layer $i$ or earlier,
but will die at layer $k$ or earlier.
In the transpose network $W^\top = W_1^\top W_2^\top \cdots W_{L-1}^\top W_L^\top$,
$B_{kji} \in \R^{d_j}$ is the subspace that carries
information in unit layer $j$ that has come at least as far as from layer $k$,
but will not survive farther than layer~$i$.

We need a notation for the dimensions of the flow subspaces.
Let
\[
\alpha_{kji} = \dim A_{kji}  \hspace{.2in}  \mbox{and}  \hspace{.2in}
\beta_{kji} = \dim B_{kji}.
\]
It is easy to see that
\begin{align*}
& A_{kji} \supseteq A_{k'ji'}  \mbox{~and~}  \alpha_{kji} \geq \alpha_{k'ji'}
  \mbox{~if~} k \geq k' \mbox{~and~} i \geq i',
  \mbox{~assuming~} j \in [0, L], k, k' \in [j - 1, L],
  i, i' \in [-1, j].  \\
& B_{kji} \subseteq B_{k'ji'}  \mbox{~and~}  \beta_{kji} \leq \beta_{k'ji'}
  \mbox{~if~} k \geq k' \mbox{~and~} i \geq i',
  \mbox{~assuming~} j \in [0, L], k, k' \in [j, L + 1],
  i, i' \in [0, j + 1].
\end{align*}
Table~\ref{nesting} depicts this relationship and
the partial order it imposes on the flow subspaces.

\begin{table}
\begin{center}
\begin{tabular}{ccr|ccccccc}
$\R^{d_2}$      & & $k = 4$ & $A_{4,2,-1} = \{ {\bf 0} \}$ & $\subseteq$ &
  $A_{420}$ & $\subseteq$ & $A_{421}$ & $\subseteq$ & $A_{422} = \R^{d_2}$  \\
\rotatebox[origin=c]{90}{$\subseteq$} & & & & &
\rotatebox[origin=c]{90}{$\subseteq$} & &
\rotatebox[origin=c]{90}{$\subseteq$} & &
\rotatebox[origin=c]{90}{$\subseteq$}  \\
$\Null W_4 W_3$ & & $k = 3$ & $A_{3,2,-1} = \{ {\bf 0} \}$ & $\subseteq$ &
  $A_{320}$ & $\subseteq$ & $A_{321}$ & $\subseteq$ & $A_{322}$  \\
\rotatebox[origin=c]{90}{$\subseteq$} & & & & &
\rotatebox[origin=c]{90}{$\subseteq$} & &
\rotatebox[origin=c]{90}{$\subseteq$} & &
\rotatebox[origin=c]{90}{$\subseteq$}  \\
$\Null W_3$     & & $k = 2$ & $A_{2,2,-1} = \{ {\bf 0} \}$ & $\subseteq$ &
  $A_{220}$ & $\subseteq$ & $A_{221}$ & $\subseteq$ & $A_{222}$  \\
\rotatebox[origin=c]{90}{$\subseteq$} & & & & &
\rotatebox[origin=c]{90}{$\subseteq$} & &
\rotatebox[origin=c]{90}{$\subseteq$} & &
\rotatebox[origin=c]{90}{$\subseteq$}  \\
$\{ {\bf 0} \}$ & & $k = 1$ & $A_{1,2,-1} = \{ {\bf 0} \}$ &             &
  $A_{120} = \{ {\bf 0} \}$ & & $A_{121} = \{ {\bf 0} \}$ &             &
  $A_{122} = \{ {\bf 0} \}$  \\
\hline
$\Null W_{k+1 \ssq 2}$
                & &         & $i = -1$                    &             &
  $i = 0$  &             & $i = 1$   &             & $i = 2$  \\
                & &  \\
$A_{k2i} \nearrow$
  & & $\col W_{2 \ssq i}$ & $\{ {\bf 0} \}$  & $\subseteq$ &
  $\col W_2 W_1$ & $\subseteq$ & $\col W_2$ & $\subseteq$ & $\R^{d_2}$
\end{tabular}

\vspace{.3in}

\begin{tabular}{ccr|ccccccc}
$\{ {\bf 0} \}$ & & $k = 5$ & $B_{520} = \{ {\bf 0} \}$ & &
  $B_{521} = \{ {\bf 0} \}$ & & $B_{522} = \{ {\bf 0} \}$ & &
  $B_{523} = \{ {\bf 0} \}$  \\
\rotatebox[origin=c]{90}{$\supseteq$} & & &
\rotatebox[origin=c]{90}{$\supseteq$} & &
\rotatebox[origin=c]{90}{$\supseteq$} & &
\rotatebox[origin=c]{90}{$\supseteq$} & &  \\
$\row W_4 W_3$  & & $k = 4$ & $B_{420}$ & $\supseteq$ &
  $B_{421}$ & $\supseteq$ & $B_{422}$ & $\supseteq$ &
  $B_{423} = \{ {\bf 0} \}$  \\
\rotatebox[origin=c]{90}{$\supseteq$} & & &
\rotatebox[origin=c]{90}{$\supseteq$} & &
\rotatebox[origin=c]{90}{$\supseteq$} & &
\rotatebox[origin=c]{90}{$\supseteq$} & &  \\
$\row W_3$      & & $k = 3$ & $B_{320}$ & $\supseteq$ &
  $B_{321}$ & $\supseteq$ & $B_{322}$ & $\supseteq$ &
  $B_{323} = \{ {\bf 0} \}$  \\
\rotatebox[origin=c]{90}{$\supseteq$} & & &
\rotatebox[origin=c]{90}{$\supseteq$} & &
\rotatebox[origin=c]{90}{$\supseteq$} & &
\rotatebox[origin=c]{90}{$\supseteq$} & &  \\
$\R^{d_2}$       & & $k = 2$ & $B_{220} = \R^{d_2}$ & $\supseteq$ &
  $B_{221}$ & $\supseteq$ & $B_{222}$ & $\supseteq$ &
  $B_{223} = \{ {\bf 0} \}$  \\
\hline
$\row W_{k \ssq 2}$
                & &         & $i = 0$             &             &
  $i = 1$  &             & $i = 2$   &             & $i = 3$  \\
                & &  \\
$B_{k2i} \nearrow$
  & & $\Null W_{2 \ssq i-1}$ & $\R^{d_2}$ & $\supseteq$ &
  $\Null (W_2 W_1)^\top$ & $\supseteq$ & $\Null W_2^\top$ & $\supseteq$ &
  $\{ {\bf 0} \}$
\end{tabular}
\end{center}

\caption{\label{nesting}
The hierarchical nesting of the flow subspaces
at unit layer $j = 2$ of a network with $L = 4$ matrices.
Top:  $A_{k2i} = \Null W_{k+1 \ssq 2} \cap \col W_{2 \ssq i}$ for each $k, i$.
Bottom:  $B_{k2i} = \row W_{k \ssq 2} \cap \Null W_{2 \ssq i-1}$ for each $k, i$.
}
\end{table}

\begin{figure}
\centerline{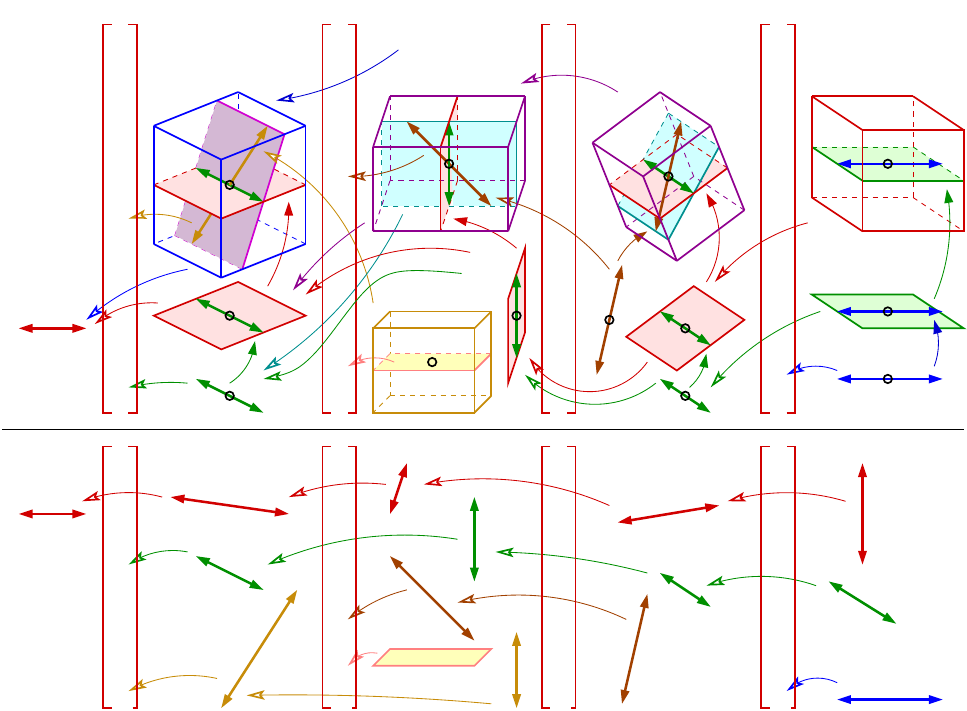}

\caption{\label{aflow}
Top:  an example of flow subspaces $A_{kji}$.
Observe that $A_{000}$ is annihilated in the nullspace of $W_1$, whereas
$A_{300}$ is not entirely annihilated until it reaches the nullspace of $W_4$.
Observe that $A_{311}$ and $A_{410}$ meet at an oblique angle;  
flow subspaces do not always meet orthogonally.
Note that the subspace $A_{322}$ is five-dimensional, so
we cannot draw it complete.
Instead, we draw a three-dimensional subspace of $A_{322}$ labeled
$A_{322} \downarrow A_{321}$ such that
$A_{322}$ is the vector sum of the plane $A_{321}$ and
the space labeled $A_{322} \downarrow A_{321}$.
Similarly, $A_{222}$ is a three-dimensional subspace, but
we draw a two-dimensional subspace of $A_{222}$ labeled
$A_{222} \downarrow A_{221}$ such that
$A_{222}$ is the vector sum of the line $A_{221}$ and
the plane labeled $A_{222} \downarrow A_{221}$.
Bottom:~an example of corresponding prebasis subspaces~$a_{kji}$,
forming flow prebases.
The prebasis subspaces for layer~$j$ span $\R^{d_j}$.
The subspaces $a_{211}$, $a_{410}$, and $a_{310}$ cannot all be
pairwise orthogonal because $A_{311}$ and $A_{410}$ are not orthogonal.
}
\end{figure}

Flow subspaces have relationships across layers.  
Recall from Section~\ref{notation} that
given a matrix $M$ and a subspace $Z$, we define $MZ = \{ Mv : v \in Z \}$,
which is also a subspace.
The simplest {\em flow relationships} are that
$A_{kji} = W_j A_{k,j-1,i}$ (assuming $k \geq j - 1 \geq i$) and
$B_{k,j-1,i} = W_j^\top B_{kji}$ (assuming $k \geq j \geq i$), which exposes
why we call them {\em flow subspaces}:
you may imagine the $A$-subspaces flowing through the network, being
linearly transformed layer by layer (and sometimes shrinking in dimension); and
you may imagine the $B$-subspaces flowing through the transpose network
$W^\top = W_1^\top W_2^\top \cdots W_{L-1}^\top W_L^\top$,
also being transformed at each layer.
Figure~\ref{aflow} depicts the $A$-subspaces at each unit layer of
a linear neural network.
The following lemma expresses the flow relationships in
a slightly more general way.

\begin{lemma}
\label{subspaceflow}
$A_{kji} = W_{j \ssq s} A_{ksi}$ for all $k$, $j$, $i$, and $s$ that satisfy
$L \geq k \geq j - 1$ and $j \geq s \geq i \geq 0$.

Symmetrically,
$B_{kji} = W_{t \ssq j}^\top B_{kti}$ for all $k$, $j$, $i$, and $t$ that satisfy
$L \geq k \geq t \geq j$ and $j + 1 \geq i \geq 0$.
\end{lemma}

\begin{proof}
By definition, $A_{kii} = \Null W_{k+1 \ssq i} \cap \R^{d_i} = \Null W_{k+1 \ssq i}$.
Hence $W_{k+1 \ssq i} A_{kii} = \{ {\bf 0} \}$.
For every $z \in [i, k + 1]$, $W_{k+1 \ssq z} W_{z \ssq i} A_{kii} = \{ {\bf 0} \}$
and thus $W_{z \ssq i} A_{kii} \subseteq \Null W_{k+1 \ssq z}$.
Obviously, $W_{z \ssq i} A_{kii} \subseteq \col W_{z \ssq i}$.
Hence $W_{z \ssq i} A_{kii} \subseteq \Null W_{k+1 \ssq z} \cap \col W_{z \ssq i} =
A_{kzi}$.

We now show that the reverse inclusion also holds:
$W_{z \ssq i} A_{kii} \supseteq A_{kzi}$.
Consider a vector $v \in A_{kzi}$.
As $v \in \col W_{z \ssq i}$, there is a vector $w \in \R^{d_i}$ such that
$v = W_{z \ssq i} w$.
As $v \in \Null W_{k+1 \ssq z}$, we have ${\bf 0} = W_{k+1 \ssq z} v =
W_{k+1 \ssq z} W_{z \ssq i} w = W_{k+1 \ssq i} w$, so
$w \in \Null W_{k+1 \ssq i} = A_{kii}$ and thus $v \in W_{z \ssq i} A_{kii}$.
Hence $W_{z \ssq i} A_{kii} \supseteq A_{kzi}$;
hence $W_{z \ssq i} A_{kii} = A_{kzi}$ for every $z \in [i, k + 1]$.

It follows that
\[
A_{kji} = W_{j \ssq i} A_{kii} = W_{j \ssq s} W_{s \ssq i} A_{kii} =
W_{j \ssq s} A_{ksi}
\]
as claimed.
A symmetric proof for the transpose network shows that
$B_{kji} = W_{t \ssq j}^\top B_{kti}$.
\end{proof}

\subsection{Prebases:  Like Bases, but Made of Subspaces}
\label{basisspaces}

In this section, we show how to decompose each unit layer's space $\R^{d_j}$
into a {\em prebasis} of subspaces of $\R^{d_j}$.
We assume the reader is familiar with
the standard idea from linear algebra of a {\em basis} for~$\R^d$,
comprising $d$ linearly independent {\em basis vectors}.
A~prebasis is like a basis, but it is made up of subspaces rather than vectors;
see below for a definition.
Our prebasis for $\R^{d_j}$ includes (as a subset) a prebasis for
every flow subspace $A_{kji}$
(where the index $j$ matches $\R^{d_j}$ but $k$ and $i$ vary freely).
We also define a second prebasis for $\R^{d_j}$ that includes a prebasis for
every flow subspace $B_{kji}$ (with matching $j$);
this prebasis represents subspaces that flow through layer $j$ of
the transpose network $W^\top = W_1^\top W_2^\top \cdots W_{L-1}^\top W_L^\top$.

Given two subspaces $P, Q \subseteq \R^d$, their vector sum is
$P + Q = \{ p + q : p \in P$ and $q \in Q \}$.
If~$P$ and $Q$ are linearly independent---that is,
if $P \cap Q = \{ {\bf 0} \}$---then $P + Q$ is called a {\em direct sum},
sometimes written\footnote{
The notation $P \oplus Q$ is unconventional, because
as an operator it produces $P + Q$, but it also implies
a constraint on the subspaces $P$ and $Q$:  that $P \cap Q = \{ {\bf 0} \}$.
If~$P \cap Q \neq \{ {\bf 0} \}$, then $P \oplus Q$ is undefined.
}
$P \oplus Q$.
Likewise, given a set of subspaces $\mathcal{P} = \{ P_1, P_2, \ldots, P_m \}$,
the direct sum notation $P_1 \oplus P_2 \oplus \ldots \oplus P_m$ implies that
the subspaces in $\mathcal{P}$ are linearly independent---meaning that
for every $i \in [1, m]$, $P_i \cap \sum_{j \neq i} P_j= \{ {\bf 0} \}$.

If $\R^d = P_1 \oplus P_2 \oplus \ldots \oplus P_m$, then
$\mathcal{P} = \{ P_1, P_2, \ldots, P_m \}$ is
known as a {\em direct sum decomposition} of~$\R^d$.
That's too many syllables, so
we call $\mathcal{P}$ a {\em prebasis} for $\R^d$,
and we say that $\mathcal{P}$~{\em spans}~$\R^d$.  
We call each $P_i$ a {\em prebasis subspace}---a multidimensional analogue of
a basis vector.
The linear independence of the prebasis subspaces implies that
for every vector $v \in \R^d$, there is one and only one way to express $v$ as
a sum of vectors $v = \sum_{i=1}^m v_i$ such that $v_i \in P_i$.
It also implies that $d = \sum_{i=1}^m \dim P_i$.
If desired, it is conceptually easy to convert a prebasis into
a traditional vector basis:  just choose a basis for each $P_i$, then
pool the $d$~vectors together to form a basis for~$\R^d$---hence
the name ``prebasis.''
We will do that in Section~\ref{flowbases}, but we delay that step
because details like the choice of basis for each prebasis subspace and
the length of each basis vector
are irrelevant here and would make our presentation more complicated.

When we choose a prebasis, we typically must make some arbitrary choices.
Given two vector subspaces $Q \subseteq Z$,
a subspace $P \subseteq Z$ is a {\em direct complement} of $Q$ if
$Z = P \oplus Q$---that is,
$Z = P + Q$ and $P \cap Q = \{ {\bf 0} \}$.
Equivalently, $\dim Z = \dim P + \dim Q$ and $P$ is linearly independent of $Q$.
(Note that ``direct complement'' does not imply orthogonality, but
the orthogonal complement of $Q$ in $Z$ is
one of the direct complements of~$P$.)
We denote the set of all direct complements of $Q$ within $Z$ with
a custom operator,
\[
Z \downarrow Q = \{ P \subseteq Z : Z = P \oplus Q \}.
\]
There are two special cases where $Z \downarrow Q$ contains only one element:
if $Q = \{ {\bf 0} \}$ then $Z \downarrow Q = \{ Z \}$, and
if $Q = Z$ then $Z \downarrow Q = \{ \{ {\bf 0} \} \}$.
Otherwise, $Z \downarrow Q$ contains infinitely many subspaces.

Recall the flow subspaces $A_{kji}$ and $B_{kji}$ from Section~\ref{flowspaces},
both of them subspaces of $\R^{d_j}$, and recall that
$A_{k,j,i-1} \subseteq A_{kji}$ and $A_{k-1,j,i} \subseteq A_{kji}$, assuming
$L \geq k \geq j \geq i \geq 0$.
It follows that $A_{k,j,i-1} + A_{k-1,j,i} \subseteq A_{kji}$.
Symmetrically, $B_{k,j,i+1} + B_{k+1,j,i} \subseteq B_{kji}$.
For all indices $k$, $j$, and $i$ satisfying $L \geq k \geq j \geq i \geq 0$,
we choose {\em prebasis subspaces}
\begin{eqnarray*}
a_{kji} & \in & A_{kji} \downarrow (A_{k,j,i-1} + A_{k-1,j,i})
\hspace{.2in}  \mbox{and}  \\
b_{kji} & \in & B_{kji} \downarrow (B_{k,j,i+1} + B_{k+1,j,i}).
\end{eqnarray*}
When applying these definitions, recall that
$A_{k,j,-1} = A_{j-1,j,i} = B_{k,j,j+1} = B_{L+1,j,i} = \{ {\bf 0} \}$.
(So~for example, $a_{jj0} = A_{jj0}$ and $b_{Ljj} = B_{Ljj}$.)
The bottom half of Figure~\ref{aflow} shows examples of
prebasis subspaces chosen from these sets.
It is common that some of the prebasis subspaces are simply
the trivial subspace $\{ {\bf 0} \}$; these can be omitted from any prebasis.

We now define prebases that span the subspaces $A_{kji}$ and $B_{kji}$.
Let
\begin{eqnarray*}
\mathcal{A}_{kji} & = & \{ a_{k'ji'} \neq \{ {\bf 0} \}:
                           k' \in [j, k], i' \in [0, i] \}
\hspace{.2in}  \mbox{and}  \\
\mathcal{B}_{kji} & = & \{ b_{k'ji'} \neq \{ {\bf 0} \}:
                           k' \in [k, L], i' \in [i, j] \}.
\end{eqnarray*}

\begin{lemma}[Prebasis Lemma]
\label{bases}
Given that $L \geq k \geq j \geq i \geq 0$,
$\mathcal{A}_{kji}$ is a prebasis for $A_{kji}$ and
$\mathcal{B}_{kji}$ is a prebasis for~$B_{kji}$.
That is,
\[
A_{kji} = \bigoplus_{k' \in [j, k]} \, \bigoplus_{i' \in [0, i]} a_{k'ji'}
\hspace{.2in}  \mbox{and}  \hspace{.2in}
B_{kji} = \bigoplus_{k' \in [k, L]} \, \bigoplus_{i' \in [i, j]} b_{k'ji'}.
\]
\end{lemma}

\begin{proof}
We prove the first claim by induction on increasing values of $k$ and $i$.
For the base cases, recall our convention that
$A_{k,j,-1} = \{ {\bf 0} \}$ and $A_{j-1,j,i} = \{ {\bf 0} \}$.
The empty set is a prebasis for the subspace $\{ {\bf 0} \}$, so
we establish a convention that
$\mathcal{A}_{k,j,-1} = \emptyset$ and $\mathcal{A}_{j-1,j,i} = \emptyset$.

For the inductive case---showing that $\mathcal{A}_{kji}$ is
a prebasis for $A_{kji}$---we assume the inductive hypothesis that
$\mathcal{A}_{k,j,i-1}$ is a prebasis for $A_{k,j,i-1}$,
$\mathcal{A}_{k-1,j,i}$ is a prebasis for $A_{k-1,j,i}$, and
$\mathcal{A}_{k-1,j,i-1}$ is a prebasis for $A_{k-1,j,i-1}$.
Most of the work in this proof is to show that
$\mathcal{A}_{k,j,i-1} \cup \mathcal{A}_{k-1,j,i}$ is
a prebasis for $A_{k,j,i-1} + A_{k-1,j,i}$.
The inductive hypothesis implies that
$A_{k,j,i-1} + A_{k-1,j,i}$ equals the vector sum of
the subspaces in $\mathcal{A}_{k,j,i-1} \cup \mathcal{A}_{k-1,j,i}$.
But we must also show that
the subspaces in $\mathcal{A}_{k,j,i-1} \cup \mathcal{A}_{k-1,j,i}$ are
linearly independent of each other.

Suppose for the sake of contradiction that they are linearly dependent.
Then there exists a nonempty set~$V$ of nonzero vectors in $\R^{d_j}$
with sum zero such that each vector in $V$ comes from
a different subspace in $\mathcal{A}_{k,j,i-1} \cup \mathcal{A}_{k-1,j,i}$.
Partition $V$ into two disjoint subsets $V'$ and $V''$ such that
each vector in $V'$ comes from a different subspace in $\mathcal{A}_{k,j,i-1}$
and each vector in $V''$ comes from a different subspace in
$\mathcal{A}_{k-1,j,i} \setminus \mathcal{A}_{k,j,i-1}$.
Let $w$ be the sum of the vectors in $V'$.
The sum of the vectors in $V''$ is $-w$.
As $V$ is nonempty, at least one of $V'$ or $V''$ is nonempty.
As the vectors in $V'$ come from a prebasis ($\mathcal{A}_{k,j,i-1}$) and
the vectors in $V''$ come from a prebasis ($\mathcal{A}_{k-1,j,i}$),
$w \neq {\bf 0}$ and both $V'$ and $V''$ are nonempty.
The vectors in~$V'$ are all in the subspace $A_{k,j,i-1}$, so $w \in A_{k,j,i-1}$;
and the vectors in $V''$ are all in $A_{k-1,j,i}$, so $w \in A_{k-1,j,i}$.
Therefore, $w \in A_{k,j,i-1} \cap A_{k-1,j,i} =
\Null W_{k \ssq j} \cap \col W_{j \ssq i-1} = A_{k-1,j,i-1}$.
This implies that $w$ is a linear combination of vectors that come from
subspaces in $\mathcal{A}_{k-1,j,i-1}$, which is
a subset of $\mathcal{A}_{k-1,j,i}$.
Therefore, we can write the nonzero vector $w$ both as
a linear combination of vectors from subspaces in $\mathcal{A}_{k-1,j,i-1}$ and
as a linear combination of vectors from subspaces in
$\mathcal{A}_{k-1,j,i} \setminus \mathcal{A}_{k,j,i-1}$,
which are two disjoint subsets of $\mathcal{A}_{k-1,j,i}$.
But this contradicts the fact that $\mathcal{A}_{k-1,j,i}$ is a prebasis.
It follows from this contradiction that
all the subspaces in $\mathcal{A}_{k,j,i-1} \cup \mathcal{A}_{k-1,j,i}$ are
linearly independent of each other.
Therefore, $\mathcal{A}_{k,j,i-1} \cup \mathcal{A}_{k-1,j,i}$ is
a prebasis for $A_{k,j,i-1} + A_{k-1,j,i}$.

Recall that $A_{kji} \supseteq A_{k,j,i-1} + A_{k-1,j,i}$ and
$a_{kji} \in A_{kji} \downarrow (A_{k,j,i-1} + A_{k-1,j,i})$.
As $\mathcal{A}_{k,j,i-1} \cup \mathcal{A}_{k-1,j,i}$ is
a prebasis for $A_{k,j,i-1} + A_{k-1,j,i}$, $\mathcal{A}_{kji} =
\mathcal{A}_{k,j,i-1} \cup \mathcal{A}_{k-1,j,i} \cup \{ a_{kji} \}$ is
a prebasis for $A_{kji}$.

A symmetric argument (by induction on {\em decreasing} values of $k$ and $i$)
shows that $\mathcal{B}_{kji}$ is a prebasis for $B_{kji}$.
\end{proof}

Let $\mathcal{A}_j = \mathcal{A}_{Ljj}$;
then $\mathcal{A}_j$ is a prebasis for $\R^{d_j}$
(because $A_{Ljj} = \R^{d_j}$), and it is a superset of
all the other \mbox{$\mathcal{A}$-prebases} for unit layer $j$.
Moreover, $\mathcal{A}_{Lji}$ is a~prebasis for $\col W_{j \ssq i}$
(because $A_{Lji} = \col W_{j \ssq i}$) and
$\mathcal{A}_{kjj}$~is a prebasis for $\Null W_{k+1 \ssq j}$
(because $A_{kjj} = \Null W_{k+1 \ssq j}$).
So we have found a single prebasis $\mathcal{A}_j$
whose elements simultaneously span many of the subspaces we are interested in.

Similarly, let $\mathcal{B}_j = \mathcal{B}_{jj0}$.
Then $\mathcal{B}_j$ is a prebasis for $\R^{d_j}$ and a superset of
all the other \mbox{$\mathcal{B}$-prebases} for layer $j$.
$\mathcal{B}_{kj0}$ is a prebasis for $\row W_{k \ssq j}$ and
$\mathcal{B}_{jji}$ is a prebasis for $\Null W_{j \ssq i-1}^\top$.

Lemma~\ref{subspacedim} below states that $\dim a_{kji} = \dim b_{kji}$,
a crucial relationship that surprised us when we stumbled upon it.
This establishes a pleasing symmetry between
flow through a linear neural network and
flow through its transpose network, even though
the prebases $\mathcal{A}_j$ and $\mathcal{B}_j$ can be very different.
In Figure~\ref{interval}, we could depict the flow of prebasis subspaces
through the transpose neural network simply by
replacing each $a_{kji}$ with~$b_{kji}$ and
reversing the directions of the arrows in the top half of the figure.
The bottom half of the figure would not change.
Lemma~\ref{subspacedim} also shows how to compute
the dimensions of the prebasis subspaces from
the dimensions of the flow subspaces.
(The dimensions of the prebasis subspaces do not depend on
which ones we choose.)
Lemma~\ref{hierarchydim} below prepares us for Lemma~\ref{subspacedim}.

Two subspaces $P, Q \subseteq \R^d$ are {\em orthogonal} if
for every vector $p \in P$ and every $q \in Q$, $p^\top q = 0$.
The {\em orthogonal complement} of a subspace $Q \subseteq \R^d$,
denoted $Q^\perp$, is the subspace of~$\R^d$ containing
every vector that is orthogonal to every vector in $Q$.
Orthogonal complements have complementary dimensions:
$\dim Q + \dim Q^\perp = d$.
Linear algebra furnishes two classic examples:
$(\row W)^\perp = \Null W$ and $(\col W)^\perp = \Null W^\top$.

\begin{lemma}
\label{hierarchydim}
Consider subspaces $P \subseteq Q \subseteq \R^d$ and
$Y \subseteq Z \subseteq \R^d$.
Then
\begin{align*}
& \dim (Q \cap Z) - \dim (Q \cap Y + P \cap Z)  \hspace*{-2.2in}  \\
& =
\dim (P^\perp \cap Y^\perp) - \dim (Q^\perp \cap Y^\perp + P^\perp \cap Z^\perp)  \\
& = \dim (Q \cap Z) - \dim (Q \cap Y) - \dim (P \cap Z) + \dim (P \cap Y)  \\
& = \dim (P^\perp \cap Y^\perp) - \dim (Q^\perp \cap Y^\perp)
    - \dim (P^\perp \cap Z^\perp) + \dim (Q^\perp \cap Z^\perp).
\end{align*}
\end{lemma}

\begin{proof}
It is a property of vector subspaces that
$\dim (E + F) = \dim E + \dim F - \dim (E \cap F)$.
Letting $E = Q \cap Y$ and $F = P \cap Z$, we have $E \cap F = P \cap Y$,
which explains why the lemma's first expression equals the third one.
Letting $E = Q^\perp \cap Y^\perp$ and $F = P^\perp \cap Z^\perp$,
we have $E \cap F = Q^\perp \cap Z^\perp$,
which explains why the lemma's second expression equals the fourth one.

To verify that the fourth expression equals the third one,
we use De~Morgan's laws
$(E + F)^\perp = E^\perp \cap F^\perp$ and $(E \cap F)^\perp = E^\perp + F^\perp$,
as well as $\dim (E \cap F) = \dim E + \dim F - \dim (E + F)$.
\begin{align*}
& \dim (P^\perp \cap Y^\perp) - \dim (Q^\perp \cap Y^\perp)
  - \dim (P^\perp \cap Z^\perp) + \dim (Q^\perp \cap Z^\perp)  \\
& = \dim P^\perp + \dim Y^\perp - \dim (P^\perp + Y^\perp)
    - \dim Q^\perp - \dim Y^\perp + \dim (Q^\perp + Y^\perp)  \\
& ~~~~ - \dim P^\perp - \dim Z^\perp + \dim (P^\perp + Z^\perp)
    + \dim (Q^\perp \cap Z^\perp)  \\
& = - \dim (P \cap Y)^\perp - \dim Q^\perp + \dim (Q \cap Y)^\perp
    - \dim Z^\perp + \dim (P \cap Z)^\perp + \dim (Q + Z)^\perp  \\
& = - d + \dim (P \cap Y) - d + \dim Q + d - \dim (Q \cap Y)
    - d + \dim Z + d - \dim (P \cap Z) + d - \dim (Q + Z)  \\
& = \dim (Q \cap Z) - \dim (Q \cap Y) - \dim (P \cap Z) + \dim (P \cap Y).
\qedhere
\end{align*}
\end{proof}

\begin{lemma}
\label{subspacedim}
For $L \geq k \geq j \geq i \geq 0$,
\[
\dim a_{kji} = \dim b_{kji} =
\alpha_{kji} - \alpha_{k,j,i-1} - \alpha_{k-1,j,i} + \alpha_{k-1,j,i-1} =
\beta_{kji} - \beta_{k,j,i+1} - \beta_{k+1,j,i} + \beta_{k+1,j,i+1}
\]
(recalling that $\alpha_{kji} = \dim A_{kji}$ and $\beta_{kji} = \dim B_{kji}$).
\end{lemma}

\begin{proof}
As $a_{kji} \in A_{kji} \downarrow (A_{k,j,i-1} + A_{k-1,j,i})$,
it follows from the definition of the operator $\downarrow$ that
$\dim a_{kji} = \dim A_{kji} - \dim (A_{k,j,i-1} + A_{k-1,j,i})$.
Similarly, $\dim b_{kji} = \dim B_{kji} - \dim (B_{k,j,i+1} + B_{k+1,j,i})$.
The result follows from Lemma~\ref{hierarchydim} by substituting
$Q = \Null W_{k+1 \ssq j}$, $P = \Null W_{k \ssq j}$,
$Z = \col W_{j \ssq i}$, and $Y = \col W_{j \ssq i-1}$.
(Then $A_{kji} = Q \cap Z$, $A_{k,j,i-1} = Q \cap Y$,
$A_{k-1,j,i} = P \cap Z$, $A_{k-1,j,i-1} = P \cap Y$, 
$B_{kji} = P^\perp \cap Y^\perp$, $B_{k,j,i+1} = P^\perp \cap Z^\perp$,
$B_{k+1,j,i} = Q^\perp \cap Y^\perp$, and $B_{k+1,j,i+1} = Q^\perp \cap Z^\perp$.)
\end{proof}

%
%

One element in $Z \downarrow Q$ is $Z \cap Q^\perp$, the subspace containing
every vector in $Z$ that is orthogonal to every vector in $Q$, and
it is tempting to always choose
$a_{kji} = A_{kji} \cap (A_{k,j,i-1} + A_{k-1,j,i})^\perp$ and
$b_{kji} = B_{kji} \cap (B_{k,j,i+1} + B_{k+1,j,i})^\perp$,
yielding what we call the {\em standard prebases}.
However, in Section~\ref{flowrelations} we exploit the flexibility
that $Z \downarrow Q$ gives us to choose {\em flow prebases} instead, so
the prebasis subspaces ($a$'s and $b$'s) flow through the network
as the flow subspaces ($A$'s~and~$B$'s) do, and
as Figures~\ref{interval} and~\ref{aflow} depict.

We warn that the prebasis $\mathcal{A}_j$ {\em cannot}, in general,
be chosen so its subspaces are mutually orthogonal.
(Nor can $\mathcal{B}_j$.)
An orthogonal prebasis is ruled out whenever there is some $\Null W_{k+1 \ssq j}$
and some $\col W_{j \ssq i}$ that meet each other at an oblique angle;
see $A_{311}$ and $A_{410}$ in Figure~\ref{aflow},
which imply that $a_{211}$, $a_{410}$, and $a_{310}$ cannot all be
pairwise orthogonal.
Even if $\mathcal{A}_j$ is the standard prebasis
(i.e., every subspace we choose from a set of the form $Z \downarrow Q$ is
fully orthogonal to~$Q$),
we cannot force {\em all} the prebasis subspaces in $\mathcal{A}_j$ to be
pairwise orthogonal.

Although our proposed fundamental subspaces of a linear neural network are
flow prebases, not the standard prebases,
it is satisfying to write explicit formulae for the standard prebasis subspaces.
Given two subspaces $P, Q \subseteq \mathbb{R}^d$,
let $\proj_Q \, P$ denote the orthogonal projection of $P$ onto $Q$ and
note that $\proj_Q P = Q \cap (Q^\perp + P)$.
Recall our convention that $W_{L+1 \ssq j} = 0$ and $W_{j \ssq -1} = 0$.

\begin{lemma}[Standard Prebases]
\label{standard}
For $L \geq k \geq j \geq i \geq 0$, the standard prebasis subspaces are
\begin{eqnarray}
a_{kji} & = & \proj_{\col W_{j \ssq i}} \, \row W_{k \ssq j} \cap
              \proj_{\Null W_{k+1 \ssq j}} \, \Null W_{j \ssq i-1}^\top  \nonumber  \\
       & = & \col W_{j \ssq i} \cap
              (\row W_{k \ssq j} + \Null W_{j \ssq i}^\top) \cap
              \Null W_{k+1 \ssq j} \cap
              (\row W_{k+1 \ssq j} + \Null W_{j \ssq i-1}^\top)
              \hspace{.2in}  \mbox{and}  \label{standarda}  \\
b_{kji} & = & \proj_{\row W_{k \ssq j}} \, \col W_{j \ssq i} \cap
              \proj_{\Null W_{j \ssq i-1}^\top} \, \Null W_{k+1 \ssq j}  \nonumber  \\
        & = & \row W_{k \ssq j} \cap (\Null W_{k \ssq j} + \col W_{j \ssq i}) \cap
              \Null W_{j \ssq i-1}^\top \cap
              (\Null W_{k+1 \ssq j} + \col W_{j \ssq i-1}).  \label{standardb}
\end{eqnarray}
\end{lemma}

\begin{proof}
In the standard prebasis, from each set of the form $Z \downarrow Q$
we choose the element $Z \cap Q^\perp$.
Observe that for two subspaces $P$ and $Q$,
$Q \cap (Q \cap P)^\perp = Q \cap (Q^\perp + P^\perp) = \proj_Q P^\perp$.
Hence
\begin{eqnarray*}
a_{kji} & = & A_{kji} \cap (A_{k,j,i-1} + A_{k-1,j,i})^\perp  \\
        & = & A_{kji} \cap A_{k,j,i-1}^\perp \cap A_{k-1,j,i}^\perp  \\
        & = & \Null W_{k+1 \ssq j} \cap \col W_{j \ssq i} \cap
              (\Null W_{k+1 \ssq j} \cap \col W_{j \ssq i-1})^\perp \cap
              (\Null W_{k \ssq j} \cap \col W_{j \ssq i})^\perp  \\
        & = & \proj_{\Null W_{k+1 \ssq j}} \, (\col W_{j \ssq i-1})^\perp \cap
              \proj_{\col W_{j \ssq i}} \, (\Null W_{k \ssq j})^\perp  \\
        & = & \proj_{\Null W_{k+1 \ssq j}} \, \Null W_{j \ssq i-1}^\top \cap
              \proj_{\col W_{j \ssq i}} \, \row W_{k \ssq j}.
\end{eqnarray*}
The third line and De Morgan's laws imply~(\ref{standarda}).
Symmetrically,
\begin{eqnarray*}
b_{kji} & = & B_{kji} \cap (B_{k,j,i+1} + B_{k+1,j,i})^\perp  \\
        & = & B_{kji} \cap B_{k,j,i+1}^\perp \cap B_{k+1,j,i}^\perp  \\
        & = & \row W_{k \ssq j} \cap \Null W_{j \ssq i-1}^\top \cap
              (\row W_{k \ssq j} \cap \Null W_{j \ssq i}^\top)^\perp \cap
              (\row W_{k+1 \ssq j} \cap \Null W_{j \ssq i-1}^\top)^\perp  \\
        & = & \proj_{\row W_{k \ssq j}} \, (\Null W_{j \ssq i}^\top)^\perp \cap
              \proj_{\Null W_{j \ssq i-1}^\top} \, (\row W_{k+1 \ssq j})^\perp  \\
        & = & \proj_{\row W_{k \ssq j}} \, \col W_{j \ssq i} \cap
              \proj_{\Null W_{j \ssq i-1}^\top} \, \Null W_{k+1 \ssq j}.
\end{eqnarray*}
The third line and De Morgan's laws imply~(\ref{standardb}).
\end{proof}

See Appendix~\ref{standardapp} for additional discussion of
the standard prebases.

\subsection{Flow Prebases:
            The Fundamental Subspaces of a Linear Neural Network}
\label{flowrelations}

We have flexibility in choosing each prebasis subspace
$a_{kji} \in A_{kji} \downarrow (A_{k,j,i-1} + A_{k-1,j,i})$.
To understand information flow in neural linear networks,
we choose {\em flow prebasis subspaces},
which satisfy the {\em flow relationships} $a_{kji} = W_j a_{k,j-1,i}$ and
$b_{k,j-1,i} = W_j^\top b_{kji}$ when $k \geq j > i$.
We imagine these subspaces as information flowing through the network
from specific starting layers to specific stopping layers,
as expressed by an annotated barcode such as
Figure~\ref{interval} (top) or Figure~\ref{aflow} (bottom),
thereby outlining how information propagates
(or would propagate, if it was there).

The easiest way to achieve that is to choose
each $a_{kii}$ arbitrarily from $A_{kii} \downarrow (A_{k,i,i-1} + A_{k-1,i,i})$ and
each $b_{kki}$ arbitrarily from
$B_{kki} \downarrow (B_{k,k,i+1} + B_{k+1,k,i})$---e.g.,
we could choose~(\ref{standarda}) for $a_{kii}$
and~(\ref{standardb}) for~$b_{kki}$---then obtain all the other subspaces
from the flow relationships.
Lemma~\ref{prebasisflow}, below, shows that
this construction always yields valid prebases.
It~also shows that
the dimension of $a_{kji}$ is the same for every~$j \in [i, k]$
(even if we choose prebases that don't flow, like the standard prebases).
Unlike a flow subspace, a flow prebasis subspace doesn't shrink in dimension
until it reaches its death layer---after which it vanishes.

\begin{lemma}
\label{dimpreserved}
Given that $L \geq k \geq j \geq s \geq i \geq 0$,
$W_{j \ssq s} a_{ksi}$ has the same dimension as $a_{ksi}$.
Symmetrically, given that $L \geq k \geq t \geq j \geq i \geq 0$,
$W_{t \ssq j}^\top b_{kti}$ has the same dimension as $b_{kti}$.
\end{lemma}

\begin{proof}
By construction, $a_{ksi}$ is linearly independent of
$A_{k-1,s,i} = \Null W_{k \ssq s} \cap \col W_{s \ssq i}$.
(That is, $a_{ksi} \cap A_{k-1,s,i} = \{ {\bf 0} \}$.)
But $a_{ksi} \subseteq A_{ksi} \subseteq \col W_{s \ssq i}$.
Hence, every nonzero vector in $a_{ksi}$ is in $\col W_{s \ssq i}$ but
not in $\Null W_{k \ssq s} \cap \col W_{s \ssq i}$;
thus no nonzero vector in $a_{ksi}$ is in $\Null W_{k \ssq s}$;
thus no nonzero vector in $a_{ksi}$ is in $\Null W_{j \ssq s}$.
Therefore, $W_{j \ssq s} a_{ksi}$ has the same dimension as $a_{ksi}$.

A symmetric argument shows that
$W_{t \ssq j}^\top b_{kti}$ has the same dimension as $b_{kti}$.
\end{proof}

\begin{lemma}[Prebasis Flow]
\label{prebasisflow}
If $L \geq k \geq j > s \geq i \geq 0$ and
$a_{ksi} \in A_{ksi} \downarrow (A_{k,s,i-1} + A_{k-1,s,i})$, then
$W_{j \ssq s} a_{ksi} \in A_{kji} \downarrow (A_{k,j,i-1} + A_{k-1,j,i})$.
(Hence, we can choose to set $a_{kji} = W_{j \ssq s} a_{ksi}$.)
Moreover, every subspace in $A_{kji} \downarrow (A_{k,j,i-1} + A_{k-1,j,i})$ has
the same dimension as~$a_{ksi}$.

If $L \geq k \geq t > j \geq i \geq 0$ and
$b_{kti} \in B_{kti} \downarrow (B_{k,t,i+1} + B_{k+1,t,i})$, then
$W_{t \ssq j}^\top b_{kti} \in B_{kji} \downarrow (B_{k,j,i+1} + B_{k+1,j,i})$.
(Hence, we can choose to set $b_{kji} = W_{t \ssq j}^\top b_{kti}$.)
Moreover, every subspace in $B_{kji} \downarrow (B_{k,j,i+1} + B_{k+1,j,i})$ has
the same dimension as~$b_{kti}$.
\end{lemma}

\begin{proof}
By definition, the notation
$W_{j \ssq s} a_{ksi} \in A_{kji} \downarrow (A_{k,j,i-1} + A_{k-1,j,i})$ is
equivalent to saying that
$A_{kji} = W_{j \ssq s} a_{ksi} + A_{k,j,i-1} + A_{k-1,j,i}$ and
$W_{j \ssq s} a_{ksi} \cap (A_{k,j,i-1} + A_{k-1,j,i}) = \{ {\bf 0} \}$.
We will prove both statements.

To confirm the first statement, observe that by Lemma~\ref{subspaceflow},
$A_{kji} = W_{j \ssq s} A_{ksi}$, $A_{k,j,i-1} = W_{j \ssq s} A_{k,s,i-1}$, and
$A_{k-1,j,i} = W_{j \ssq s} A_{k-1,s,i}$.
By assumption, $a_{ksi} \in A_{ksi} \downarrow (A_{k,s,i-1} + A_{k-1,s,i})$, so
$A_{ksi} = a_{ksi} + A_{k,s,i-1} + A_{k-1,s,i}$.
Pre-multiplying both sides of this identity by $W_{j \ssq s}$ confirms that
$A_{kji} = W_{j \ssq s} a_{ksi} + A_{k,j,i-1} + A_{k-1,j,i}$.

To show that
$W_{j \ssq s} a_{ksi} \cap (A_{k,j,i-1} + A_{k-1,j,i}) = \{ {\bf 0} \}$,
let $v$ be a vector in $W_{j \ssq s} a_{ksi} \cap (A_{k,j,i-1} + A_{k-1,j,i})$.
Then $v \in W_{j \ssq s} a_{ksi} \cap W_{j \ssq s} (A_{k,s,i-1} + A_{k-1,s,i})$.
So there exists a vector $u \in a_{ksi}$ such that $v = W_{j \ssq s} u$,
and there exist a vector $p \in A_{k,s,i-1}$ and a vector $q \in A_{k-1,s,i}$
such that $v = W_{j \ssq s} (p + q)$.
Thus $W_{j \ssq s} (u - p - q) = {\bf 0}$, so
$W_{k \ssq j} W_{j \ssq s} (u - p - q) = {\bf 0}$ and thus
$u - p - q \in \Null W_{k \ssq s}$.
Recall that $A_{k-1,s,i} = \Null W_{k \ssq s} \cap \col W_{s \ssq i}$.
So $q \in \Null W_{k \ssq s}$, hence $u - p \in \Null W_{k \ssq s}$.
Moreover, $u$ and $p$ are both in $\col W_{s \ssq i}$, so
$u - p \in \Null W_{k \ssq s} \cap \col W_{s \ssq i} = A_{k-1,s,i}$ and hence
$u \in A_{k,s,i-1} + A_{k-1,s,i}$.
But $u \in a_{ksi}$ which is linearly independent of $A_{k,s,i-1} + A_{k-1,s,i}$,
so $u = {\bf 0}$ and thus $v = {\bf 0}$.
We have thus shown that every vector in
$W_{j \ssq s} a_{ksi} \cap (A_{k,j,i-1} + A_{k-1,j,i})$ is ${\bf 0}$.

Therefore, $W_{j \ssq s} a_{ksi} \in A_{kji} \downarrow (A_{k,j,i-1} + A_{k-1,j,i})$,
as claimed.
To see that every subspace in $A_{kji} \downarrow (A_{k,j,i-1} + A_{k-1,j,i})$ has
the same dimension as $a_{ksi}$, observe that
$W_{j \ssq s} a_{ksi}$ has the same dimension as $a_{ksi}$
by Lemma~\ref{dimpreserved}, and the subspaces in
$A_{kji} \downarrow (A_{k,j,i-1} + A_{k-1,j,i})$
all have the same dimension as each other.

A symmetric argument shows that
$W_{t \ssq j}^\top b_{kti} \in B_{kji} \downarrow (B_{k,j,i+1} + B_{k+1,j,i})$ and
that every subspace in $B_{kji} \downarrow (B_{k,j,i+1} + B_{k+1,j,i})$ has
the same dimension as $b_{kti}$.
\end{proof}

We say that $\mathcal{A}_0, \mathcal{A}_1, \ldots, \mathcal{A}_L$ are
{\em flow prebases at $\theta$} if
\begin{icompact}
\item
$\mathcal{A}_j = \{ a_{kji} \neq \{ {\bf 0} \}: k \in [j, L], i \in [0, j] \}$
for all $j \in [0, L]$, where
\item
$a_{kii} \in A_{kii} \downarrow (A_{k,i,i-1} + A_{k-1,i,i})$
for all indices satisfying $L \geq k \geq i \geq 0$, and
\item
the flow relationships hold:
for all indices satisfying $j \in [1, L]$ and
$L \geq k \geq j - 1 \geq i \geq 0$,
\begin{icompact}
\item
$W_j a_{k,j-1,i} = a_{kji}$ if $k \geq j$, whereas
\item
$W_j a_{k,j-1,i} = \{ {\bf 0} \}$ if $k = j - 1$.
\end{icompact}
\end{icompact}

Symmetrically, we say that
$\mathcal{B}_0, \mathcal{B}_1, \ldots, \mathcal{B}_L$ are
{\em flow prebases for the transpose network at~$\theta$} if
\begin{icompact}
\item
$\mathcal{B}_j = \{ b_{kji} \neq \{ {\bf 0} \}: k \in [j, L], i \in [0, j] \}$
for all $j \in [0, L]$, where
\item
$b_{kki} \in B_{kki} \downarrow (B_{k,k,i+1} + B_{k+1,k,i})$
for all indices satisfying $L \geq k \geq i \geq 0$ and
\item
the flow relationships hold:
for all indices satisfying $j \in [1, L]$ and $L \geq k \geq j \geq i \geq 0$,
\begin{icompact}
\item
$W_j^\top b_{kji} = b_{k,j-1,i}$ if $j > i$, whereas
\item
$W_j^\top b_{kji} = \{ {\bf 0} \}$ if $j = i$.
\end{icompact}
\end{icompact}
Then Lemma~\ref{prebasisflow} ensures that
each $a_{kji} \in A_{kji} \downarrow (A_{k,j,i-1} + A_{k-1,j,i})$ and  
each $b_{kji} \in B_{kji} \downarrow (B_{k,j,i+1} + B_{k+1,j,i})$.
Lemma~\ref{bases} ensures that
each $\mathcal{A}_j$ is a prebasis for $\R^{d_j}$ and
each $\mathcal{B}_j$ is a prebasis for $\R^{d_j}$.

Observe that even if two prebasis subspaces
$a_{kji}$ and $a_{k'ji'}$ at layer $j$ are orthogonal to each other,
the subspaces $W_{j+1} a_{kji}$ and $W_{j+1} a_{k'ji'}$
generally are not orthogonal.
Choosing flow prebases entails sacrificing the desire
to choose each layer's prebasis to be as close to orthogonal as possible
(i.e., the standard prebasis).
But as we explained in Section~\ref{basisspaces},
a fully orthogonal prebasis is not generally possible anyway.

However, it is always possible to choose flow prebases that have
the nice property that $a_{kii} = W_{k \ssq i}^+ a_{kki}$, where
$M^+$~denotes the Moore--Penrose pseudoinverse of a matrix $M$.
This is nice because the flow relationship $a_{kki} = W_{k \ssq i} a_{kii}$ is
matched by an inverse flow relationship expressed by the pseudoinverse.
Any choice of $a_{kii} \in \row W_{k \ssq i}$ has that property, and
there is always at least one such choice with that property,
namely~(\ref{standarda}) (as
$a_{kii} \subseteq \row W_{k \ssq i} + \Null W_{i \ssq i}^\top = \row W_{k \ssq i}$).
Symmetrically, we can choose flow prebases such that
$b_{kki} = W_{k \ssq i}^{+\top} b_{kii}$
where $M^{+\top}$~denotes the transpose of $M^+$.
Any choice of $b_{kki} \in \col W_{k \ssq i}$ has that property,
and~(\ref{standardb}) is one such choice.


\subsection{Relationships between Matrix Ranks and Subspace Dimensions}
\label{ranksbases}

This section examines the relationship between
the ranks of the subsequence matrices,
the dimensions of the flow subspaces $A_{kji}$ and $B_{kji}$, and
the dimensions of the prebasis subspaces $a_{kji}$ and $b_{kji}$.
A key insight is that if we know all the subsequence matrix ranks,
the dimensions of the prebasis subspaces are uniquely determined, and vice versa
(as illustrated at the bottom of Figure~\ref{interval}).
To say it another way, given fixed layer sizes $d_0, d_1, \ldots, d_L$,
there is a bijection between valid rank lists and valid barcodes
(with ``valid'' defined as in Sections~\ref{notation} and~\ref{barcodes}).

Lemma~\ref{prebasisflow} establishes that
the dimension of $a_{kji}$ is the same for every $j \in [i, k]$.
By Lemma~\ref{subspacedim}, the dimension of $b_{kji}$ is the same too.
So we omit the index $j$ as we now name this dimension.
Let
\[
\omega_{ki} = \dim a_{kji} = \dim b_{kji},  \hspace{.2in}
\mbox{for all~} k, j, i \mbox{~satisfying~} L \geq k \geq j \geq i \geq 0.
\]

We have already seen this notation, $\omega_{ki}$, in Section~\ref{barcodes},
where it denotes the count of an interval $[i, k]$.
Section~\ref{flowrelations} substantiates that connection.
The count $\omega_{ki}$ signifies a prebasis subspace $a_{kii}$
of dimension $\omega_{ki}$ that is born at unit layer $i$,
flows through the network being linearly transformed into
a sequence of subspaces $a_{k,i+1,i}, a_{k,i+2,i}, \ldots$,
all of dimension~$\omega_{ki}$, reaches layer $k$ in the form $a_{kki}$,
still having $\omega_{ki}$ dimensions, and
proceeds no farther---either because $a_{kki} \subseteq \Null W_{k+1}$
or because layer $k$ is the output layer,
as illustrated in Figures~\ref{interval} and~\ref{aflow}.
These subspaces carry information from the input layer only if
$i = 0$ and reach the output layer only if $k = L$; otherwise,
they represent unused potential.

The count $\omega_{ki}$ also signifies a prebasis subspace $b_{kki}$
of dimension $\omega_{ki}$ that is born at layer~$k$ and flows through
the transpose network to die at layer~$i$ in the form $b_{kii}$.
This symmetry surprises us, as sometimes
the prebasis subspaces $a_{kji}$ and $b_{kji}$ are
necessarily unrelated to each other, except that they have the same dimension.
However, the symmetry seems less surprising and even inevitable when
you consider that
the fibers $\mu^{-1}(W)$ and $\mu^{-1}(W^\top)$ must be identical.

\begin{figure}
\centerline{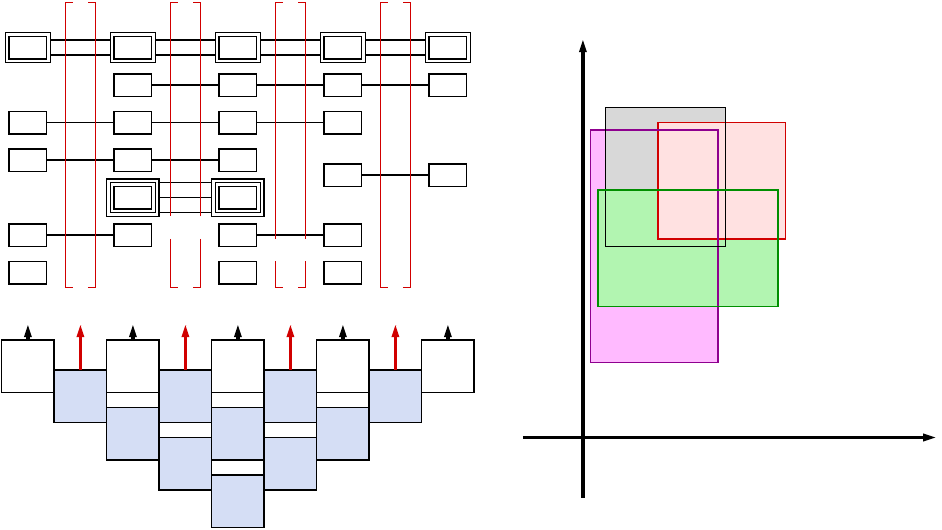}
\centerline{\begin{minipage}{5.4in}
\[
d_j = \sum_{t = j}^L \sum_{s = 0}^j \omega_{ts}
\hspace{.02in}
\left|
\hspace{.03in}
\rank W_{k \ssq i} = \sum_{t = k}^L \sum_{s = 0}^i \omega_{ts}
\hspace{.02in}
\right|
\hspace{.02in}
\alpha_{kji} = \sum_{t = j}^k \sum_{s = 0}^i \omega_{ts}
\hspace{.02in}
\left|
\hspace{.03in}
\beta_{kji} = \sum_{t = k}^L \sum_{s = i}^j \omega_{ts}
\right.
\]
\end{minipage}}

\caption{\label{omega}
At left, we reprise the annotated barcode from Figure~\ref{interval}.
At right, we tabulate the values of the interval counts $\omega_{ts}$
with boxes that illustrate how the four summations compute
$d_1$, $\alpha_{322}$, $\beta_{321}$, and $\rank W_{3 \ssq 1} = \rank W_3W_2$.
At the bottom of the figure, we reprise the four summations for reference.
}
\end{figure}

Figure~\ref{omega} gives a preview of four of the five identities
proven in this section---summations that express
$d_j$, $\rank W_{k \ssq i}$, $\alpha_{kji}$, and $\beta_{kji}$ in terms of
interval counts $\omega_{ts}$---and
a visual interpretation of those summations.
The bottom of Figure~\ref{interval} gives a visual interpretation of
the fifth identity, which expresses $\omega_{ts}$ in terms of matrix ranks,
and a second visual interpretation of the summation for $\rank W_{k \ssq i}$.
It might be helpful to know that the counts $\omega_{ts}$
in Figure~\ref{omega} are the same as in Figure~\ref{interval}, but
they are rotated $135^\circ$.
The five identities hold even if the prebases are not flow prebases
(e.g., they hold for the standard prebases).

By Lemma~\ref{bases}, $\mathcal{A}_{kji}$ is a prebasis for $A_{kji}$.
Hence, the dimension of $A_{kji}$ is  the sum of
the dimensions of the prebasis subspaces in $\mathcal{A}_{kji}$---every
subspace $a_{tjs}$ with $t \leq k$ and $s \leq i$.

\begin{lemma}
\label{alphabeta}
For $L \geq k \geq j \geq i \geq 0$,
the dimensions $\alpha_{kji}$ of the flow subspaces $A_{kji}$,
the dimensions $\beta_{kji}$ of the flow subspaces $B_{kji}$, and
the dimensions $\omega_{ts}$ of the prebasis subspaces
$a_{tjs}$ and $b_{tjs}$ are related by the identities
\begin{equation}
\label{alphabetaid}
\alpha_{kji} = \dim A_{kji} = \sum_{t = j}^k \sum_{s = 0}^i \omega_{ts}
\hspace{.2in}  \mbox{and}  \hspace{.2in}
\beta_{kji} = \dim B_{kji} = \sum_{t = k}^L \sum_{s = i}^j \omega_{ts}.
\end{equation}
\end{lemma}

\begin{proof}
We prove the first claim by induction on increasing values of $k$ and $i$.
For the base cases, recall our convention that
$A_{k,j,-1} = \{ {\bf 0} \}$ and $A_{j-1,j,i} = \{ {\bf 0} \}$; hence
$\alpha_{k,j,-1} = \alpha_{j-1,j,i} = \alpha_{j-1,j,-1} = 0$.

For the inductive case---establishing the identity for $\alpha_{kji}$---we
assume the inductive hypothesis that the identity holds for
$\alpha_{k,j,i-1}$, $\alpha_{k-1,j,i}$, and $\alpha_{k-1,j,i-1}$.
By Lemma~\ref{subspacedim}, $\alpha_{kji} =
\dim a_{kji} + \alpha_{k,j,i-1} + \alpha_{k-1,j,i} - \alpha_{k-1,j,i-1}$.
By Lemma~\ref{prebasisflow}, $\dim a_{kji}$ does not depend on the index~$j$, so
we can write $\alpha_{kji} =
\omega_{ki} + \alpha_{k,j,i-1} + \alpha_{k-1,j,i} - \alpha_{k-1,j,i-1}$.
By substituting~(\ref{alphabetaid}) into the right-hand side,
we obtain~(\ref{alphabetaid}) for $\alpha_{kji}$ on the left-hand side.

A symmetric argument (by induction on {\em decreasing} values of $k$ and $i$),
with the identity
$\beta_{kji} = \omega_{ki} + \beta_{k,j,i+1} + \beta_{k+1,j,i} - \beta_{k+1,j,i+1}$
from Lemma~\ref{subspacedim},
establishes the identity~(\ref{alphabetaid}) for $\beta_{kji}$.
\end{proof}

The following corollary states that, as we would expect,
the number of units $d_j$ in unit layer~$j$ equals
the sum of the dimensions of the subspaces in a prebasis for $\R^{d_j}$.
It implies that the flow prebases at any weight vector $\theta \in \R^{d_\theta}$
are represented by a valid barcodes
(as we defined a barcode to be {\em valid} if
it satisfies the following identity).

\begin{cor}
\label{dj}
The number of units in unit layer~$j$ is, as formula~(\ref{intervallayer}) says,
\[
d_j = \sum_{t = j}^L \sum_{s = 0}^j \omega_{ts}.
\]
\end{cor}

\begin{proof}
As $\R^{d_j} = A_{Ljj} = B_{jj0}$, $d_j = \alpha_{Ljj} = \beta_{jj0}$.
The summation follows by identity~(\ref{alphabetaid}).
\end{proof}

Recall that a rank list
$\underline{r} = \langle \rank W_{k \ssq i} \rangle_{L \geq k \geq i \geq 0}$
is a list of the ranks of all the subsequence matrices,
including those of the form $\rank W_{j \ssq j} = d_j$.
The following lemma shows how to map a rank list to a barcode
(expressed as a list of interval counts $\omega_{ki}$) and vice versa.
The bottom half of Figure~\ref{interval} depicts
the identities~(\ref{ranksum}) and~(\ref{omegarank}) visually.

\begin{lemma}
\label{rankomega}
For $L \geq k \geq j \geq i \geq 0$,
the ranks of the subsequence matrices are related to
the dimensions of the prebasis subspaces and the flow subspaces by
the identities
\begin{eqnarray}
\rank W_{k \ssq i}
  & = & \alpha_{Lki} = \beta_{ki0} = \sum_{t=k}^L \sum_{s=0}^i \omega_{ts},
  \label{ranksum}  \\
\omega_{ki}
  & = & \rank W_{k \ssq i} - \rank W_{k \ssq i-1}
        - \rank W_{k+1 \ssq i} + \rank W_{k+1 \ssq i-1},  \label{omegarank}  \\
\alpha_{kji} & = & \rank W_{j \ssq i} - \rank W_{k+1 \ssq i},
  \hspace{.2in}  \mbox{and}  \label{alpharank}  \\
\beta_{kji} & = & \rank W_{k \ssq j} - \rank W_{k \ssq i-1},  \label{betarank}
\end{eqnarray}
recalling the conventions that $\rank W_{j \ssq j} = d_j$,
$\rank W_{L+1 \ssq i} = 0$, and $\rank W_{k \ssq -1} = 0$.
\end{lemma}

\begin{proof}
We use the Rank-Nullity Theorem to connect the rank of $W_{k \ssq i}$ to
the dimensions of the flow subspaces, and
the formulae~(\ref{alphabetaid}) to connect those to
the interval counts.
Recall that $A_{Lii} = \R^{d_i}$ and $A_{k-1,i,i} = \Null W_{k \ssq i}$.
As $W_{k \ssq i}$ is a $d_k \times d_i$ matrix,
\begin{eqnarray*}
\rank W_{k \ssq i} & = & d_i - \dim \Null W_{k \ssq i}  \nonumber  \\
                  & = & \dim A_{Lii} - \dim A_{k-1,i,i}  \nonumber  \\
                  & = & \alpha_{Lii} - \alpha_{k-1,i,i}  \nonumber  \\
                  & = & \sum_{t=k}^L \sum_{s=0}^i \omega_{ts}  \\
                  & = & \alpha_{Lki} = \beta_{ki0}  \nonumber
\end{eqnarray*}
as claimed.\footnote{
\label{eureka}
Symmetrically, we could obtain the summation~(\ref{ranksum}) by instead
starting from $\rank W_{k \ssq i} = d_k - \dim \Null W_{k \ssq i}^\top$ and
recalling that $B_{kk0} = \R^{d_k}$ and
$B_{k,k,i+1} = 
\Null W_{k \ssq i}^\top$.
That is how we originally stumbled on the fact that
$\dim a_{kji} = \dim b_{kji}$.
Eureka!
Later, we derived Lemma~\ref{hierarchydim} to understand why.
}

We can verify the identity~(\ref{omegarank}) by substituting
the summation~(\ref{ranksum}) into it.
The identities~(\ref{alpharank}) and~(\ref{betarank}) follow
from~(\ref{alphabetaid}) and~(\ref{ranksum}).
\end{proof}

Ferdinand Georg Frobenius~\cite{frobenius11} proved in 1911 that
$\rank W_{k \ssq i} - \rank W_{k \ssq i-1}
- \rank W_{k+1 \ssq i} + \rank W_{k+1 \ssq i-1} \geq 0$,
a statement called the {\em Frobenius rank inequality}.
This confirms that every~$\omega_{ts}$ is nonnegative---a fact we already know,
as every prebasis subspace has a nonnegative dimension.
(An interval count $\omega_{ts}$ can also be thought of as
a {\em Frobenius slack}.)
Our derivations deepen the geometric interpretation of
the Frobenius rank inequality by connecting
the slack~(\ref{omegarank}) in the inequality to
the dimension of the subspaces~(\ref{standarda}) and~(\ref{standardb}).
To~put it in a simpler notation, for any product $RSTU$ of four matrices,
\begin{eqnarray*}
\rank ST - \rank STU - \rank RST + \rank RSTU
& = &
\dim \, (\proj_{\col T} \, \row S \cap \proj_{\Null RS} \, \Null (TU)^\top)  \\
& = & \dim \, (\proj_{\row S} \, \col T \cap \proj_{\Null (TU)^\top} \, \Null RS).
\end{eqnarray*}
Thome~\cite{thome16} offers some generalizations of
the Frobenius rank inequality to linear neural networks.
He proves them by induction, but they also follow easily from~(\ref{ranksum}).

Readers familiar with the properties of {\em summed-area tables} might
find it helpful to know that~(\ref{ranksum}) shows that
the rank list is a summed-area table of the interval counts.  
The identities~(\ref{omegarank}), (\ref{alpharank}), and~(\ref{betarank}) are
three natural uses of a summed-area table.
Summed-area tables
were introduced by Crow~\cite{crow84} 
to speed up texture mapping in computer graphics;
a~well known use is in the Viola--Jones image processing algorithm
for object detection~\cite{viola04}.

\subsection{The Canonical Weight Vector}
\label{canonical}

\begin{figure}[p!]
\centerline{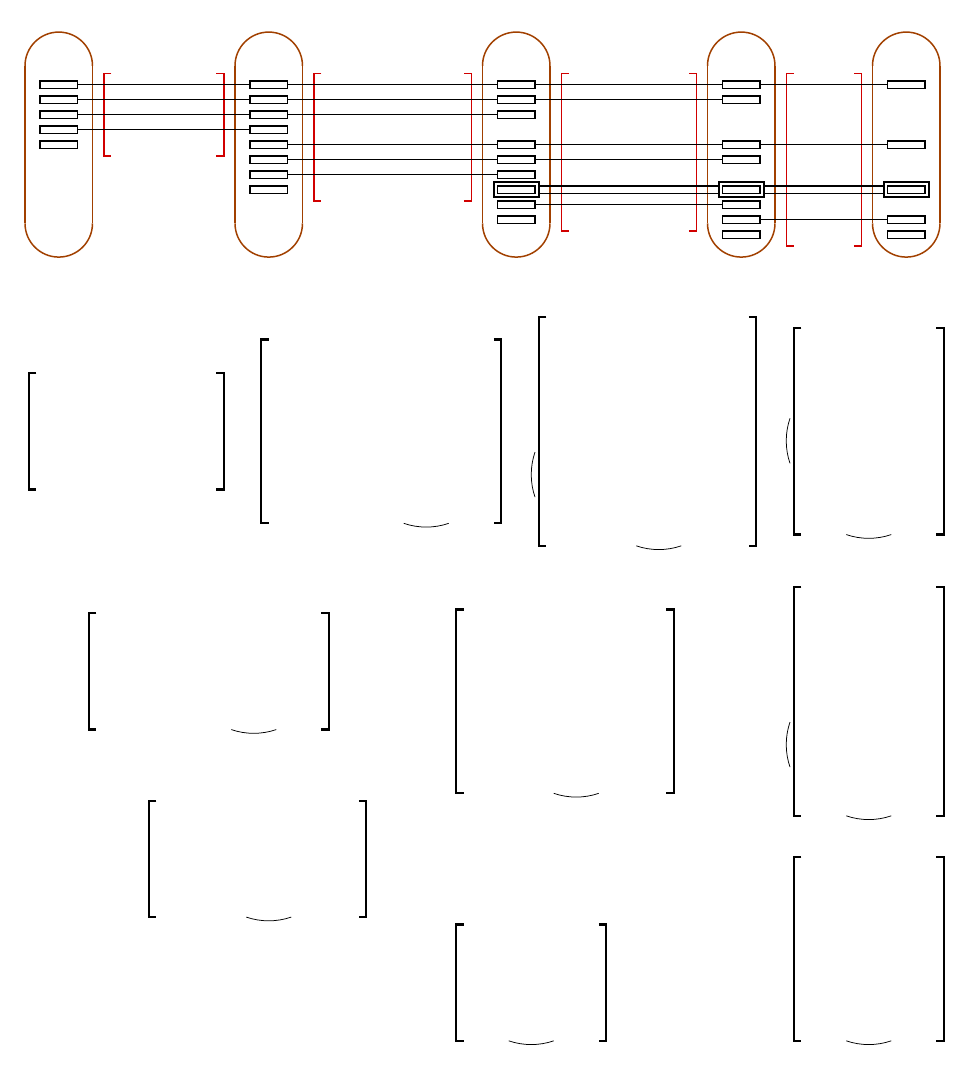}

\caption{\label{ident}
The canonical weight vector
$\tilde{\theta} = (\tilde{I}_4, \tilde{I}_3, \tilde{I}_2, \tilde{I}_1)$ and
its subsequence matrices when
every interval count is $\omega_{ki} = 1$ except $\omega_{20} = 2$.
Each row of $\tilde{I}_j$ is associated with some interval $[i, l]$
that contains~$j$, marked in shorthand with indices $li$, and
each column of $\tilde{I}_j$ is associated with some interval $[h, k]$
that contains $j - 1$, marked in shorthand with indices $kh$.
Blocks where the row indices match the column indices are identity matrices;
all other component are zeros.
Every rank-$1$ matrix $W \in \R^{5 \times 6}$ can be factored as $W =
J_4 \tilde{I}_4 \tilde{I}_3 \tilde{I}_2 \tilde{I}_1 J_0^{-1}$
where the matrices $\tilde{I}_j$ have the values depicted
(determined solely by the barcode)
but $J_4$ and $J_0$ depend on~$W$ and are invertible.
This gives us a weight vector
$(J_4 \tilde{I}_4, \tilde{I}_3, \tilde{I}_2, \tilde{I}_1 J_0^{-1})$ that lies on
the fiber $\mu^{-1}(W)$ and has the same rank list as $\tilde{\theta}$.
}
\end{figure}

Figure~\ref{ident} depicts a sequence of matrices we call
{\em almost-identity matrices}, which we define to be
any matrix obtained by taking an identity matrix,
inserting additional rows of zeros, and
appending additional columns of zeros on the right.
Products of almost-identity matrices are also almost-identity matrices
(see the figure for examples).
For any given valid rank list $\underline{r}$,
we define one unique {\em canonical weight vector}
\[
\tilde{\theta} =
(\tilde{I}_L, \tilde{I}_{L-1}, \ldots, \tilde{I}_1)
\]
whose rank list is $\underline{r}$ and whose matrices $\tilde{I}_j$ are
almost-identity matrices, as illustrated in Figure~\ref{ident}.
It is ``canonical'' in the sense that it depends solely on $\underline{r}$.
The identity of $\tilde{\theta}$ can be inferred from
$\underline{r}$'s barcodes, with
the interval counts $\omega_{ki}$ computed from $\underline{r}$
by Equation~(\ref{omegarank}):  as the figure illustrates,
each $\tilde{I}_j$ is an adjacency matrix for vertices of the flow diagram.

To precisely specify the matrices $\tilde{I}_j$ that constitute
$\tilde{\theta}$, we number their rows and columns in a non-standard way,
illustrated in Figure~\ref{ident}.
For all indices satisfying $L \geq l \geq j \geq i \geq 0$,
the matrix $\tilde{I}_j$ has $\omega_{li}$ rows representing
the interval $[i, l]$; these rows are labeled $li$ in the figure.
We order the rows of~$\tilde{I}_j$ primarily in order of decreasing~$l$ and
secondarily in order of increasing~$i$
(as illustrated---observe that the label $L0$ always comes first).
Similarly, for all indices satisfying $L \geq k \geq j - 1 \geq h \geq 0$,
$\tilde{I}_j$ has $\omega_{kh}$ columns representing the interval $[h, k]$;
these columns are labeled $kh$ in the figure.
We order these columns primarily in order of decreasing~$k$ and
secondarily in order of increasing $h$
(as illustrated).  

Now we can specify which components of $\tilde{I}_j$ are $1$'s.
For every interval $[i, k]$ that contains both $j - 1$ and~$j$
and has count $\omega_{ki} > 0$,
$\tilde{I}_j$ has an $\omega_{ki} \times \omega_{ki}$ block that is
an identity matrix,
located at the rows and columns associated with the interval $[i, k]$.
All the other components of $\tilde{I}_j$
(where the column and the row do not represent the same interval) are zeros.
It is straightforward to verify that
$\tilde{\theta} = (\tilde{I}_L, \tilde{I}_{L-1}, \ldots, \tilde{I}_1)$
has the specified interval counts, and therefore
has the specified rank list $\underline{r}$.

Therefore, for any valid barcode,
we can exhibit a corresponding weight vector.
In Appendix~\ref{valid}, we use this fact to argue that
for every valid barcode,
the identity~(\ref{ranksum}) produces a valid rank list.
See Appendix~\ref{standardapp} for some comments about the
standard prebases associated with $\tilde{\theta}$.

Let $\tilde{I} = \mu(\tilde{\theta}) =
\tilde{I}_L \tilde{I}_{L-1} \cdots \tilde{I}_1$, which is
a $d_L \times d_0$ matrix (like $W$) illustrated at
the bottom of Figure~\ref{ident}.
Our construction of $\tilde{\theta}$ implies that
$\tilde{I} =
\left[ \begin{smallmatrix} I & 0 \\ 0 & 0 \end{smallmatrix} \right]$
with an $r_{L \ssq 0} \times r_{L \ssq 0}$ identity matrix at upper left.

\subsection{Flow Bases and their Relationship to the Canonical Weight Vector}
\label{flowbases}

This section is devoted to showing that
there is a straightforward relationship between
a weight vector $\theta$, the canonical weight vector $\tilde{\theta}$ with
the same rank list $\underline{r}$ as $\theta$, and
any sequence of flow prebases
$\mathcal{A}_0, \mathcal{A}_1, \ldots, \mathcal{A}_L$ associated with $\theta$.
Specifically, the flow prebases induce an invertible linear transformation of
weight space that maps $\theta$ to $\tilde{\theta}$.
It also maps fibers to fibers and strata to strata
(see Section~\ref{sec:affinestrata}).
We will also explore some implications of constructing bases from prebases, or
prebases from bases.

A benefit of this relationship is that it reveals a way to convert
flow prebases $\mathcal{A}_j$ to flow prebases~$\mathcal{B}_j$ for
the transpose network, or vice versa (see Section~\ref{basestranspose}).
More importantly, in Section~\ref{sec:affinestrata} we will use
the canonical weight vector to show that
any two strata (from different fibers) with the same rank list are
related by an invertible linear transformation.
Hence, the topology of a stratum depends solely on its rank list.
The canonical weight vector also gives us intuition
about how strata adjoin each other (see Section~\ref{1canon}).
Lastly, for any valid rank list (or any valid barcode),
the canonical weight vector $\tilde{\theta}$ is an explicit example of
a weight vector having that rank list (or barcode),
from which it is easy to generate
all the other weight vectors having the same rank list.

Consider a weight vector $\theta = (W_L, W_{L-1}, \ldots, W_1) \in \R^{d_\theta}$.
Lemma~\ref{prebasisflow} guarantees that we can choose flow prebases
$\mathcal{A}_0, \mathcal{A}_1, \ldots, \mathcal{A}_L$ for $\theta$
as defined in Section~\ref{flowrelations}.
Each $\mathcal{A}_j$ is a prebasis for~$\R^{d_j}$, meaning that
$\R^{d_j} = \bigoplus_{k = j}^L \bigoplus_{i = 0}^j a_{kji}$ 
($\R^{d_j}$ is the vector sum of the subspaces in~$\mathcal{A}_j$, which are
linearly independent).
We chose the name ``prebasis'' because
we can refine the prebasis~$\mathcal{A}_j$ into a basis for~$\R^{d_j}$.
We do so now, expressing that basis in the form of
a square, invertible matrix~$J_j$ whose columns are the basis vectors.
Constructing $J_j$ is a matter of picking basis vectors for the
subspaces $a_{kji}$---but we want the basis vectors in $J_0, J_1, \ldots, J_L$
to {\em flow}, just like the subspaces do.

For all indices satisfying $L \geq k \geq j \geq i \geq 0$,
let $J_{kji}$ be a $d_j \times \omega_{ki}$ matrix whose columns are
a basis for $a_{kji}$.
(If $\omega_{ki} = 0$, then $a_{kji} = \{ {\bf 0} \}$ and $J_{kji}$ is not used.)
In the case $j = i$ we may choose the columns of $J_{kji}$ to be
any linearly independent vectors spanning $a_{kji}$.
But in the case $j > i$ we set
\[
J_{kji} = W_{j \ssq i} J_{kii}.
\]
This identity is a {\em flow relationship} analogous to
the flow relationship $a_{kji} = W_{j \ssq i} a_{kii}$.
The fact that $J_{kii}$'s columns are a basis for $a_{kii}$ implies that
$J_{kji}$'s columns are a basis for~$a_{kji}$ (because
by Lemma~\ref{dimpreserved}, $a_{kji}$ has the same dimension as $a_{kii}$).
Note that this flow relationship implies another flow relationship:
$J_{kji} = W_j J_{k,j-1,i}$.
Moreover, the flow relationship $W_j a_{j-1,j-1,i} = \{ {\bf 0} \}$ implies
a flow relationship $W_j J_{j-1,j-1,i} = 0$.

We construct a $d_j \times d_j$ matrix $J_j$ whose columns are
a basis for $\R^{d_j}$ by grouping together
all the matrices $J_{kji}$ with matching $j$.
For example, if $L = 4$ then
\[
J_2 = [ J_{420} ~~~~ J_{421} ~~~~ J_{422} ~~~~ J_{320} ~~~~ J_{321} ~~~~ J_{322}
        ~~~~ J_{220} ~~~~ J_{221} ~~~~ J_{222} ].
\]
Note that some of the blocks in this matrix may be empty; for instance,
if $\omega_{31} = 0$, then $J_{321}$ contributes nothing to $J_2$.
In each $J_j$, we order the matrices $J_{kji}$
primarily in order of decreasing $k$ and secondarily in order of increasing~$i$,
to match the order of the rows and columns of
the almost-identity matrices depicted in Figure~\ref{ident}.
As $\mathcal{A}_j$ is a prebasis for~$\R^{d_j}$,
the $a_{kji}$'s are linearly independent (for a fixed $j$ and varying $k$, $i$),
so the columns of $J_j$ are linearly independent.
Hence the columns of $J_j$ are a basis for $\R^{d_j}$ and $J_j$ is invertible.

A {\em J-vector} is a sequence $\lambda = (J_L, J_{L-1}, \ldots, J_0)$ of 
matrices such that $J_j \in \R^{d_j \times d_j}$ for each $j \in [0, L]$.
We say that a J-vector represents a sequence of {\em flow bases at $\theta$} if
\begin{icompact}
\item
$J_j$ is invertible for all $j \in [0, L]$, and
\item
the flow relationships hold:  for all indices satisfying
$j \in [1, L]$ and $L \geq k \geq j - 1 \geq i \geq 0$,
\begin{icompact}
\item
$W_j J_{k,j-1,i} = J_{kji}$ if $k \geq j$, whereas
\item
$W_j J_{k,j-1,i} = 0$ if $k = j - 1$.
\end{icompact}
\end{icompact}

The following lemma shows that these flow relationships can be collected
in a more concise flow relationship
\[
J_j \tilde{I}_j = W_j J_{j-1}  \hspace{.2in}  \mbox{for~all~} j \in [1, L].
\]
The~purpose of $\tilde{I}_j$ in this identity is to match
columns of $J_j$ with corresponding columns of~$J_{j-1}$.

\begin{lemma}
\label{basesflow}
Let $\theta = (W_L, W_{L-1}, \ldots, W_1) \in \R^{d_\theta}$ be
a weight vector with rank list $\underline{r}$ and
let $\tilde{\theta} = (\tilde{I}_L, \tilde{I}_{L-1}, \ldots, \tilde{I}_1)$ be
the canonical weight vector for $\underline{r}$.
Let $J_0 \in \R^{d_0 \times d_0}, J_1 \in \R^{d_1 \times d_1}, \ldots,
J_L \in \R^{d_L \times d_L}$ be invertible matrices.

Then $J_j \tilde{I}_j = W_j J_{j-1}$
(equivalently, $\tilde{I}_j = J_j^{-1} W_j J_{j-1}$;
equivalently, $W_j = J_j \tilde{I}_j J_{j-1}^{-1}$) for all $j \in [1, L]$
if and only if $(J_L, J_{L-1}, \ldots, J_0)$ represents
a sequence of flow bases at $\theta$.
\end{lemma}

\begin{proof}
For indices with $k = j - 1$ (in full, $L \geq k + 1 = j > i \geq 0$),
the columns of $J_j \tilde{I}_j$ corresponding to
the interval $[i, k]$ (at unit layer $j - 1$) are zero because
the same columns of~$\tilde{I}_j$ are zero by construction.
Hence, if $J_j \tilde{I}_j = W_j J_{j-1}$ then
the same columns of~$W_j J_{j-1}$ are zero and therefore $W_j J_{k,j-1,i} = 0$.

For indices with $k \geq j$ (in full, $L \geq k \geq j > i \geq 0$),
the columns of $J_j \tilde{I}_j$ corresponding to
the interval $[i, k]$ (at unit layer $j - 1$) are the columns of $J_{kji}$
by the construction of~$\tilde{I}_j$.
The columns of $W_j J_{j-1}$ corresponding to~$[i, k]$ are
the columns of $W_j J_{k,j-1,i}$.
Hence, if $J_j \tilde{I}_j = W_j J_{j-1}$ then $W_j J_{k,j-1,i} = J_{kji}$.

Therefore, if $J_j \tilde{I}_j = W_j J_{j-1}$ for all $j \in [1, L]$, then
all the flow relationships hold.
Each $J_j$ is invertible by assumption, so
$(J_L, J_{L-1}, \ldots, J_0)$ represents a sequence of flow bases at $\theta$,
as claimed.

The converse holds too:
if $(J_L, J_{L-1}, \ldots, J_0)$ represents a sequence of flow bases at $\theta$,
then all the flow relationships hold.
Hence for each $j \in [1, L]$, we can equate any column of $J_j \tilde{I}_j$
with the same column of $W_j J_{j-1}$ by applying one of the flow relationships.
Therefore, $J_j \tilde{I}_j = W_j J_{j-1}$ for all $j \in [1, L]$ as claimed.
\end{proof}

At a weight vector $\theta$, there is at least one possible sequence of
flow prebases~$\mathcal{A}_j$---usually there are infinitely many---and
there always are
infinitely many possible sequences of flow basis matrices~$J_j$.
Throughout Section~\ref{fundsub} we have explained why this is so, but
for completeness, the following lemma summarizes the reasons.
By Lemma~\ref{basesflow}, every choice of flow bases satisfies
the flow relationships $J_j \tilde{I}_j = W_j J_{j-1}$.

\begin{lemma} 
\label{basisexists}
Every weight vector $\theta \in \R^{d_\theta}$ has
at least one sequence of flow prebases at $\theta$ and 
infinitely many J-vectors that represent flow bases at $\theta$.
\end{lemma}

\begin{proof}
For all indices $k$ and $i$ satisfying $L \geq k \geq i \geq 0$,
let $a_{kii} = A_{kii} \cap (A_{k,i,i-1} + A_{k-1,i,i})^\perp$.
Clearly, $a_{kii} \in A_{kii} \downarrow (A_{k,i,i-1} + A_{k-1,i,i})$ because
$a_{kii} \subseteq A_{kii}$ and
$a_{kii}$ is linearly independent of $A_{k,i,i-1} + A_{k-1,i,i}$.
Let $\omega_{ki}$ be the dimension of $a_{kii}$ and
let $J_{kii}$ be a $d_i \times \omega_{ki}$ matrix whose columns form
a basis for $a_{kii}$.
If $\omega_{ki} \neq 0$ then there are infinitely many such matrices, as
each column can be scaled by an arbitrary nonzero factor.

For all indices $k$, $j$, and $i$ satisfying $L \geq k \geq j > i \geq 0$,
let $a_{kji} = W_{j \ssq i} a_{kii}$ and $J_{kji} = W_{j \ssq i} J_{kii}$.
As the columns of $J_{kii}$ span $a_{kii}$,
the columns of $J_{kji}$ span $a_{kji}$.
By Lemma~\ref{dimpreserved}, the dimension of $a_{kji}$ is $\omega_{ki}$, so
the columns of $J_{kji}$ are linearly independent and form a basis for $a_{kji}$.
These choices satisfy the flow conditions
$W_j a_{k,j-1,i} = a_{kji}$ and $W_j J_{k,j-1,i} = J_{kji}$.
They also satisfy the flow conditions
$W_j a_{k,j-1,i} = \{ {\bf 0} \}$ and $W_j J_{k,j-1,i} = 0$
for indices satisfying $L \geq k = j - 1 \geq i \geq 0$, because
$a_{j-1,j-1,i} \subseteq A_{j-1,j-1,i} \subseteq \Null W_j$.
By Lemma~\ref{prebasisflow},
each $a_{kji} \in A_{kji} \downarrow (A_{k,j,i-1} + A_{k-1,j,i})$.

For $j \in [0, L]$, define $\mathcal{A}_j$ and $J_j$ as usual
(by collecting all the $a_{kji}$'s with matching $j$ into $\mathcal{A}_j$ and
concatenating all the $J_{kji}$'s with matching $j$ into $J_j$).
By Lemma~\ref{bases}, each $\mathcal{A}_j$ is a prebasis for $\R^{d_j}$.
In particular, the subspaces $a_{kji}$ are linearly independent
for each fixed~$j$ (with $k$ and $i$ varying), so
the columns of $J_j$ are linearly independent and $J_j$ is invertible.
By definition, the prebases
$\mathcal{A}_0, \mathcal{A}_1$, \ldots, $\mathcal{A}_L$ together are
flow prebases at $\theta$ and
the basis matrices $J_0, J_1, \ldots, J_L$ represent flow bases at $\theta$.
\end{proof}

Conversely, any J-vector $\lambda$ whose matrices are all invertible
can be interpreted as a sequence of flow bases for some weight vector.
The identity of that weight vector depends also on a choice of rank list.
Let $\underline{r}$ be a valid rank list and
let $\tilde{\theta} = (\tilde{I}_L, \tilde{I}_{L-1}, \ldots, \tilde{I}_1)$
be the canonical weight vector for $\underline{r}$.
Lemma~\ref{basesflow} implies that we can map a sequence of flow bases to
their weight vector with the function
\[
\theta_{\underline{r}}(\lambda) = \theta_{\underline{r}}(J_L, J_{L-1}, \ldots, J_0) =
(J_L \tilde{I}_L J_{L-1}^{-1}, J_{L-1} \tilde{I}_{L-1} J_{L-2}^{-1}, \ldots
J_1 \tilde{I}_1 J_0^{-1}).
\]
The following lemma shows that
$\theta_{\underline{r}}(\lambda)$ has rank $\underline{r}$ and that
$\lambda$ represents flow bases for $\theta_{\underline{r}}(\lambda)$.

\begin{lemma} 
\label{Jtoweight}
Let $\underline{r}$ be a rank list and let $\lambda$ be
a J-vector in which every matrix is invertible.
Then $\theta_{\underline{r}}(\lambda)$ has rank list $\underline{r}$ and
$\lambda$ represents
a sequence of flow bases for $\theta_{\underline{r}}(\lambda)$.
Moreover, $\mu(\theta_{\underline{r}}(\lambda)) = J_L \tilde{I} J_0^{-1}$, where
$\tilde{I} = \mu(\tilde{\theta}) =
\tilde{I}_L \tilde{I}_{L-1} \cdots \tilde{I}_1$.
\end{lemma}

\begin{proof}
Let $\theta = \theta_{\underline{r}}(\lambda)$.
Then $\mu(\theta) =
J_L \tilde{I}_L J_{L-1}^{-1} J_{L-1} \tilde{I}_{L-1} J_{L-2}^{-1} \cdots
J_1 \tilde{I}_1 J_0^{-1} =
J_L \tilde{I}_L \tilde{I}_{L-1} \cdots \tilde{I}_1 J_0^{-1} =
J_L \tilde{I} J_0^{-1}$ as claimed.
Writing $\theta = (W_L, W_{L-1}, \ldots, W_1)$,
the subsequence matrices of $\theta$ can be written
$W_{k \ssq i} = J_k \tilde{I}_{k \ssq i} J_i^{-1}$.
A matrix's rank is invariant under multiplication by an invertible matrix, so
$\rank W_{k \ssq i} = \rank \tilde{I}_{k \ssq i}$.
Hence $\theta$ has the same rank list as $\tilde{\theta}$,
namely,~$\underline{r}$.

For all $j \in [0, L]$, $W_j = J_j \tilde{I}_j J_{j-1}^{-1}$, so
by Lemma~\ref{basesflow}, $\lambda$ represents
a sequence of flow bases for $\theta$.
\end{proof}

The last three lemmas have
two immediate corollaries that characterize
some weight vectors with rank list~$\underline{r}$.

\begin{cor}
\label{ranklistgenerateall}
Let $\underline{r}$ be a valid rank list and
let $\tilde{\theta} = (\tilde{I}_L, \tilde{I}_{L-1}, \ldots, \tilde{I}_1)$
be the canonical weight vector for $\underline{r}$.
The set of all weight vectors with rank list  $\underline{r}$ is
\begin{equation}
\{ \theta_{\underline{r}}(J_L, J_{L-1}, \ldots, J_0) :
   J_j \in \GL(d_j, \R) \mbox{~for~} j \in [0, L] \}.
\label{mm}
\end{equation}
\end{cor}

\begin{proof}
Every weight vector in this set has rank list $\underline{r}$ by
Lemma~\ref{Jtoweight}.
Every weight vector with rank list $\underline{r}$ is in this set
because every weight vector $\theta$ has a J-vector $\lambda$ that represents
a sequence of flow bases for $\theta$ by Lemma~\ref{basisexists}, and
if $\theta$ has rank list $\underline{r}$ then
$\theta = \theta_{\underline{r}}(\lambda)$ by Lemma~\ref{basesflow}.
\end{proof}

We will discuss the set~(\ref{mm}) again in Section~\ref{rankstrata},
where we call it a {\em multideterminantal manifold}.
Note that over this domain, $\theta_{\underline{r}}$ is not a bijection;
there are infinitely many J-vectors $\lambda$ for which
$\theta_{\underline{r}}(\lambda)$ is the same.

\begin{cor}
\label{factorWI}
Let $\theta \in \R^{d_\theta}$ be a weight vector with rank list $\underline{r}$
and let $\tilde{\theta} = (\tilde{I}_L, \tilde{I}_{L-1}, \ldots, \tilde{I}_1)$ be
the canonical weight vector for $\underline{r}$.
Then $W = \mu(\theta)$ can be factored into $L + 2$ matrices of the form
\begin{equation}
W = J_L \tilde{I}_L \tilde{I}_{L-1} \cdots \tilde{I}_1 J_0^{-1}
\label{Wfactorcanon}
\end{equation}
and thus the weight vector
$(J_L \tilde{I}_L, \tilde{I}_{L-1}, \tilde{I}_{L-2}, \ldots,
\tilde{I}_2, \tilde{I}_1 J_0^{-1})$
lies on the fiber $\mu^{-1}(W)$ and on the stratum $S_{\underline{r}}$.
\end{cor}

\begin{proof}
By Lemma~\ref{basisexists}, there exists
a J-vector $(J_L, J_{L-1}, \ldots, J_0)$ that represents flow bases at $\theta$.
By Lemma~\ref{basesflow},
\[
J_L \tilde{I}_L \tilde{I}_{L-1} \cdots \tilde{I}_1 J_0^{-1} =
J_L J_L^{-1} W_L J_{L-1} J_{L-1}^{-1} W_{L-1} J_{L-2} \cdots
J_1 W_1 J_0 J_0^{-1} = W_L W_{L-1} \cdots W_1 = W.
\qedhere
\]
\end{proof}

This factorization includes two matrices that are not canonical,
$J_L$ and $J_0^{-1}$ (which depend on the value of $W$ and
on some arbitrary choices of prebasis subspaces and basis vectors).

Above, we have discussed refining a sequence of flow prebases to obtain
a sequence of flow bases.
Now we reverse the process.
Given a J-vector $\lambda = (J_L, J_{L-1}, \ldots, J_0)$,
we would like to infer a sequence of
prebases $\mathcal{A}_0, \mathcal{A}_1, \ldots, \mathcal{A}_L$, but
a J-vector carries no information about
the rank list or interval counts, whereas the prebases do.
If we are also given a rank list $\underline{r}$,
we can calculate the corresponding interval counts $\omega_{ki}$
specified by~(\ref{omegarank}) and use them to subdivide
each matrix $J_j$ into blocks $J_{kji}$ of width $\omega_{ki}$.
We say that the prebases $\mathcal{A}_0, \mathcal{A}_1, \ldots, \mathcal{A}_L$
{\em induced} by $\lambda$ and $\underline{r}$ are
the prebases obtained by letting each $a_{kji}$ be
the subspace spanned by the columns of~$J_{kji}$.
(Note that if a $J_j$ is not invertible,
$\mathcal{A}_j$ is not a prebasis for~$\R^{d_j}$.)

\begin{lemma}
\label{basesprebases}
Let $\theta = (W_L, W_{L-1}, \ldots, W_1) \in \R^{d_\theta}$ be a weight vector
and let $\underline{r}$ be its rank list.
Let $\lambda = (J_L, J_{L-1}, \allowbreak \ldots, J_0)$ be
a J-vector that represents a sequence of flow bases at $\theta$.
Then the prebases $\mathcal{A}_0, \mathcal{A}_1, \ldots, \mathcal{A}_L$
induced by $\lambda$ and $\underline{r}$ are flow prebases at $\theta$.
\end{lemma}

\begin{proof}
Each $J_j$ is an invertible $d_j \times d_j$ matrix, so
the columns of each $J_j$ are linearly independent, and
so are the columns of each $J_{kji}$.
As $J_{kji}$ has $\omega_{ki}$ columns taken from~$J_j$,
each $a_{kji}$ is an $\omega_{ki}$-dimensional subspace of~$\R^{d_j}$.
As $J_j$ has $d_j$ linearly independent columns,
$\mathcal{A}_j$~is a prebasis for $\R^{d_j}$.
By the definition of flow bases, (for the specified indices)
$W_j J_{k,j-1,i} = 0$ if $k = j - 1$ whereas
$W_j J_{k,j-1,i} = J_{kji}$ if $k \geq j$;
this implies that $W_j a_{k,j-1,i} = \{ {\bf 0} \}$ if $k = j - 1$ whereas
$W_j a_{k,j-1,i} = a_{kji}$ if $k \geq j$.

To show that $a_{kji} \subseteq A_{kji} \downarrow (A_{k,j,i-1} + A_{k-1,j,i})$,
observe that the flow relationships imply that
$W_{k+1 \ssq j} a_{kji} = \{ {\bf 0} \}$ and $a_{kji} = W_{j \ssq i} a_{kii}$, so
$a_{kji} \subseteq \Null W_{k+1 \ssq j} \cap \col W_{j \ssq i} = A_{kji}$.
Let $\mathcal{A}_{kji} = \{ a_{k'ji'} : k' \in [j, k], i' \in [0, i] \}$.
By the same logic, for each member of $\mathcal{A}_{kji}$,
$a_{k'ji'} \subseteq A_{k'ji'} \subseteq A_{kji}$; hence
$A_{kji} \supseteq \bigoplus_{k'=j}^k \bigoplus_{i'=0}^i a_{k'ji'}$.
(We can write the direct sum because $\mathcal{A}_j$ is a prebasis.)
By Lemma~\ref{alphabeta},
\[
\dim A_{kji} = \sum_{k'=j}^k \sum_{i'=0}^i \omega_{k'i'} =
\sum_{k'=j}^k \sum_{i'=0}^i \dim a_{k'ji'} =
\dim \bigoplus_{k'=j}^k \bigoplus_{i'=0}^i a_{k'ji'},
\]
so $A_{kji} = \bigoplus_{k'=j}^k \bigoplus_{i'=0}^i a_{k'ji'}$.
For each member of $\mathcal{A}_{kji}$ except $a_{kji}$,
$a_{k'ji'} \subseteq A_{k'ji'} \subseteq A_{k,j,i-1} + A_{k-1,j,i}$.
It follows that $A_{kji} \subseteq A_{k,j,i-1} + A_{k-1,j,i} + a_{kji}$; therefore,
$a_{kji}$ is a superset of some subspace in
$A_{kji} \downarrow (A_{k,j,i-1} + A_{k-1,j,i})$.
By Lemma~\ref{prebasisflow}, every subspace in
$A_{kji} \downarrow (A_{k,j,i-1} + A_{k-1,j,i})$ has dimension $\omega_{ki}$,
the same dimension as $a_{kji}$; therefore,
$a_{kji}$ is a subspace in $A_{kji} \downarrow (A_{k,j,i-1} + A_{k-1,j,i})$.
\end{proof}

\subsection{The Transpose Network and Complementary Flow Bases}
\label{basestranspose}

Lemma~\ref{basesflow} describes a relationship between a weight vector $\theta$,
the canonical weight vector~$\tilde{\theta}$ with the same rank list, and
a sequence of flow prebases $\mathcal{A}_0, \ldots, \mathcal{A}_L$
associated with $\theta$.
Can we use flow prebases $\mathcal{B}_0, \ldots, \mathcal{B}_L$ associated with
the transpose network $W_1^\top W_2^\top \cdots W_L^\top$ instead?
We could apply Lemma~\ref{basesflow} to the transpose network.
But the canonical weight vector for the transpose network does not correspond to
the canonical weight vector for the original network,
because the layer indices $k$ and $i$ that determine
the ordering of the rows and columns of the matrices are swapped.
Retaining the layer numbers of the original network reveals
a connection between the two sets of flow prebases.
This connection will enable us to generate the flow prebases~$\mathcal{B}_j$
from the flow prebases $\mathcal{A}_j$, or vice versa.

Mirroring the construction in Section~\ref{flowbases},
for all indices satisfying $L \geq k \geq j \geq i \geq 0$,
let $K_{kji}$ be a $d_j \times \omega_{ki}$ matrix whose columns are
a basis for $b_{kji}$.
If $k = j$ we may choose the columns of $K_{kji}$ to be
any linearly independent vectors spanning $b_{kji}$; whereas
if $k > j$ we set
\[
K_{kji} = W_{k \ssq j}^\top K_{kki},
\]
thereby establishing a flow relationship analogous to
$b_{kji} = W_{k \ssq j}^\top b_{kki}$.
Then we collect matrices together to build a basis matrix $K_j$
for each $\R^{d_j}$, exactly as we built $J_j$.
For example, if $L = 4$ then
\[
K_2 = [ K_{420} ~~~~ K_{421} ~~~~ K_{422} ~~~~ K_{320} ~~~~ K_{321} ~~~~ K_{322}
        ~~~~ K_{220} ~~~~ K_{221} ~~~~ K_{222} ].
\]
(Note that this is a different matrix than we would create if
we simply applied Lemma~\ref{basesflow} to the transpose network, because
the latter approach would swap the indices $k$ and $i$,
ordering the blocks differently.)

To formalize this construction, consider a J-vector
$\lambda = (K_L, K_{L-1}, \ldots, K_0)$.
We say that a J-vector represents a sequence of
{\em flow bases for the transpose network at $\theta$} if
\begin{icompact}
\item
$K_j$ is invertible for all $j \in [0, L]$
(hence its columns form a basis for $\R^{d_j}$), and
\item
the flow relationships hold:  for all indices satisfying
$j \in [1, L]$ and $L \geq k \geq j \geq i \geq 0$,
\begin{icompact}
\item
$W_j^\top K_{kji} = K_{k,j-1,i}$ if $j > i$, whereas
\item
$W_j^\top K_{kji} = 0$ if $j = i$.
\end{icompact}
\end{icompact}

The following lemma shows that these flow relationships can be collected
in a more concise flow relationship
\[
K_{j-1} \tilde{I}_j^\top = W_j^\top K_j 
\hspace{.2in}  \mbox{for~all~} j \in [1, L].
\]
We omit the proof, as
it is a straightforward adaptation of the proof of Lemma~\ref{basesflow}.
(But note that if we simply applied Lemma~\ref{basesflow} unchanged to
the transpose network, $\tilde{I}_j$ would be replaced by
a matrix with differently ordered rows and columns.)

\begin{lemma}
\label{basesflowtranspose}
Let $\theta = (W_L, W_{L-1}, \ldots, W_1) \in \R^{d_\theta}$ be
a weight vector with rank list $\underline{r}$ and
let $\tilde{\theta} = (\tilde{I}_L, \tilde{I}_{L-1}, \ldots, \tilde{I}_1)$ be
the canonical weight vector for $\underline{r}$.
Let $K_0 \in \R^{d_0 \times d_0}, K_1 \in \R^{d_1 \times d_1}, \ldots,
K_L \in \R^{d_L \times d_L}$ be invertible matrices.

Then $K_{j-1} \tilde{I}_j^\top = W_j^\top K_j$
(equivalently, $\tilde{I}_j = K_j^\top W_j K_{j-1}^{-\top}$;
equivalently, $W_j = K_j^{-\top} \tilde{I}_j K_{j-1}^{\top}$)
for all $j \in [1, L]$ if and only if $(K_L, K_{L-1}, \ldots, K_0)$ represents
a sequence of flow bases for the transpose network at $\theta$.
\end{lemma}

The following lemma is analogous to Lemma~\ref{basesprebases} for
the transpose network.
Again we omit the proof, which is
almost identical to that of Lemma~\ref{basesprebases}.

\begin{lemma}
\label{basesprebasestranspose}
Let $\theta = (W_L, W_{L-1}, \ldots, W_1) \in \R^{d_\theta}$ be a weight vector
and let $\underline{r}$ be its rank list.
Let $\lambda = (K_L, \allowbreak K_{L-1}, \allowbreak \ldots, K_0)$ be
a J-vector that represents
a sequence of flow bases for the transpose network at $\theta$.
Then the prebases $\mathcal{B}_0, \mathcal{B}_1, \ldots, \mathcal{B}_L$
induced by $\lambda$ and $\underline{r}$ are
flow prebases for the transpose network at $\theta$.
\end{lemma}

Now we come to the main idea of this section.
By comparing Lemmas~\ref{basesflow} and~\ref{basesflowtranspose},
we see that if you have flow basis matrices $J_0, J_1, \ldots, J_L$ for
the original network at $\theta$, then
you can find flow bases for the transpose network by choosing
\begin{equation}
K_j = J_j^{-\top}.
\label{compbases}
\end{equation}
Symmetrically, if you have flow basis matrices $K_0, K_1, \ldots, K_L$ for
the transpose network, you can obtain flow bases for the original network
by choosing $J_j = K_j^{-\top}$.

\begin{lemma}
\label{compbaseslemma}
$J_0, J_1, \ldots, J_L$ represent a sequence of flow bases at $\theta$
if and only if
$K_0 = J_0^{-\top}, K_1 = J_1^{-\top}, \ldots, \allowbreak K_L = J_L^{-\top}$
represent a sequence of flow bases for the transpose network at~$\theta$.
\end{lemma}

\begin{proof}
By Lemma~\ref{basesflow},
$J_0, J_1, \ldots, J_L$ represent a sequence of flow bases at $\theta$
if and only if
$J_j \tilde{I}_j = W_j J_{j-1}$ for all $j \in [1, L]$; equivalently, 
$K_{j-1} \tilde{I}_j^\top = W_j^\top K_j$ for all $j \in [1, L]$.
By Lemma~\ref{basesflowtranspose},
$K_{j-1} \tilde{I}_j^\top = W_j^\top K_j$ for all $j \in [1, L]$ if and only if
$K_0, K_1, \ldots, K_L$ represent
a sequence of flow bases for the transpose network at $\theta$.
The result follows.
\end{proof}

If you choose $K_j = J_j^{-\top}$ or $J_j = K_j^{-\top}$, then
$J_j^\top K_j = I$, hence
\begin{equation}
J_{kji}^\top K_{k'ji'} =
\left\{ \begin{array}{ll}
I, & k = k' \mbox{~and~} i = i',  \\
0, & \mbox{otherwise}.
\end{array} \right.
\label{JK}
\end{equation}
Therefore, (for this choice of flow bases)
if $k \neq k'$ or $i \neq i'$, every vector in the subspace $a_{kji}$ is
orthogonal to every vector in $b_{k'ji'}$.
Hence if you have flow prebases
for the original network,
then you can find flow prebases
for the transpose network by choosing each $b_{kji}$ to be
the unique subspace of dimension $\omega_{ki}$ that satisfies
this orthogonality property.

\begin{lemma}
\label{compprebaseslemma}
Given flow prebases $\mathcal{A}_0, \mathcal{A}_1, \ldots, \mathcal{A}_L$ for
the original network at $\theta$, the following prebasis subspaces form
flow prebases $\mathcal{B}_0, \mathcal{B}_1, \ldots, \mathcal{B}_L$
for the transpose network at $\theta$.
\begin{equation}
b_{kji} =
\left( \sum_{a_{k'ji'} \in \mathcal{A}_j \setminus \{ a_{kji} \}} a_{k'ji'} \right)^\perp
\hspace{.2in}  \mbox{for all~} L \geq k \geq j \geq i \geq 0.
\label{complementaryprebases}
\end{equation}
This transformation serves as its own inverse:  it can convert
flow prebases $\mathcal{B}_j$ for the transpose network to
flow prebases $\mathcal{A}_j$ for the original network.
\end{lemma}

\begin{proof}
Let $J_0, J_1, \ldots, J_L$ be flow basis matrices for
the flow prebases $\mathcal{A}_0, \mathcal{A}_1, \ldots, \mathcal{A}_L$.
Let $K_j = J_j^{-\top}$ for all $j \in [0, L]$.
By Lemma~\ref{compbaseslemma},
the $K_j$'s are flow basis matrices for the transpose network at $\theta$.
Let $\mathcal{B}_0, \mathcal{B}_1, \ldots, \mathcal{B}_L$ be  
the prebases induced by the basis matrices $K_0, K_1, \ldots, K_L$.
By Lemma~\ref{basesprebasestranspose},
these are flow prebases for the transpose network at $\theta$.

For each $j \in [0, L]$ and
each flow prebasis subspace $b_{kji} \in \mathcal{B}_j$,
it follows from the identity~(\ref{JK}) that $b_{kji}$~is orthogonal to
every subspace in $\mathcal{A}_j$ except $a_{kji}$.
The right-hand side of~(\ref{complementaryprebases}) is
the subspace containing all vectors that are orthogonal to
every subspace in~$\mathcal{A}_j$ except~$a_{kji}$; hence
$b_{kji}$ is a subspace of the right-hand side.
As $\mathcal{A}_j$ is a prebasis that spans $\R^{d_j}$,
the subspace on the right-hand side of~(\ref{complementaryprebases}) has
the same dimension as $a_{kji}$, which has the same dimension as $b_{kji}$
by Lemma~\ref{subspacedim}.
Hence the subspace on the right-hand side of~(\ref{complementaryprebases}) is
$b_{kji}$.
\end{proof}

Note that the expression~(\ref{complementaryprebases}) does not depend on
the bases $J_j$ chosen to refine the prebases~$\mathcal{A}_j$.

We say that the flow prebases
$\mathcal{B}_0, \mathcal{B}_1, \ldots, \mathcal{B}_L$
generated by~(\ref{complementaryprebases})---or, equivalently,
by~(\ref{compbases})---are {\em complementary} to the flow prebases
$\mathcal{A}_0, \mathcal{A}_1, \ldots, \mathcal{A}_L$.
We say that the flow basis matrices $K_0, K_1, \ldots, K_L$
generated by~(\ref{compbases}) are {\em complementary} to
the flow basis matrices $J_0, J_1, \ldots, J_L$.
Note that complementary flow bases imply complementary flow prebases, but
not vice versa.

\subsection{A Fundamental Theorem of Linear Neural Networks}
\label{fundthm}

Matrices have crucial properties that Gilbert Strang~\cite{strang93} summarizes
as the {\em Fundamental Theorem of Linear Algebra}.
These properties are relationships between
the four fundamental subspaces of a matrix:
the rowspace (denoted $\row M$), the nullspace ($\Null M$),
the columnspace ($\col M$), and the left nullspace ($\Null M^\top$).
For any $p \times q$ matrix $M$,
the rowspace of $M$ is the orthogonal complement of the nullspace of $M$ and
the sum of their dimensions is~$q$.
(The latter fact is called the {\em Rank-Nullity Theorem}.)
The same observation applies to $M^\top$, so
the columnspace of~$M$ is the orthogonal complement of
the left nullspace of $M$ and the sum of their dimensions is~$p$.
The dimensions of $\row M$ and $\col M$ are the same, and
we call that dimension the {\em rank} of $M$.
(Strang's ``Fundamental Theorem'' also addresses
properties of the singular value decomposition, not treated here.)

Here, we propose a candidate for
an analogous {\em Fundamental Theorem of Linear Neural Networks}.
Theorem~\ref{fundnn}, below, takes parts of
Lemmas~\ref{bases}, \ref{subspacedim}, \ref{dimpreserved}, \ref{prebasisflow},
\ref{basisexists}, and~\ref{compprebaseslemma},
and the formulae~(\ref{intervallayer}),
(\ref{alphabetaid})--(\ref{betarank}), and~(\ref{complementaryprebases}),
and reframes them around the decomposition of each layer of units into
``fundamental'' flow prebasis subspaces akin to
$\row M$, $\col M$, $\Null M$, and $\Null M^\top$
(though our decomposition is not generally unique).

\begin{theorem}[Fundamental Theorem of Linear Neural Networks]
\label{fundnn}
Let $\theta = (W_L, W_{L-1}, \ldots, W_1)$ be
a weight vector representing a linear neural network with
$L$ layers of edges and unit layer sizes $d_L, d_{L-1}, \ldots, d_0$.
For all indices $k$ and $i$ satisfying $L \geq k \geq i \geq 0$, let
\[
\omega_{ki} = \rank W_{k \ssq i} - \rank W_{k \ssq i-1} -
              \rank W_{k+1 \ssq i} + \rank W_{k+1 \ssq i-1},
\]
using the conventions that $\rank W_{j \ssq j} = d_j$,
$\rank W_{L+1 \ssq i} = 0$, and $\rank W_{k \ssq -1} = 0$.

Then there exist subspaces $a_{kji} \subseteq \R^{d_j}$ and
$b_{kji} \subseteq \R^{d_j}$ of dimension $\omega_{ki}$
for all indices satisfying $L \geq k \geq j \geq i \geq 0$, such that
for all $j \in [0, L]$,
\begin{icompact}
\item
$\mathcal{A}_j = \{ a_{kji} \neq \{ {\bf 0} \} : k \in [j, L], i \in [0, j] \}$
is a direct sum decomposition of $\R^{d_j}$ and
\item
$\mathcal{B}_j = \{ b_{kji} \neq \{ {\bf 0} \} : k \in [j, L], i \in [0, j] \}$
is also a direct sum decomposition of $\R^{d_j}$;
\end{icompact}
and the subspaces satisfy the {\em flow relationships}
\begin{eqnarray*}
W_j a_{k,j-1,i} & = &
\left\{ \begin{array}{ll}
a_{kji}, & k \geq j,  \\
\{ {\bf 0} \}, & k = j - 1,
\end{array} \right.
\hspace{.2in}  \forall j \in [1, L], k \in [j-1, L], i \in [0, j-1],
\hspace{.2in}  \mbox{and}  \\
W_j^\top b_{kji} & = &
\left\{ \begin{array}{ll}
b_{k,j-1,i}, & j > i,  \\
\{ {\bf 0} \}, & j = i,
\end{array} \right.
\hspace{.24in}  \forall j \in [1, L], k \in [j, L], i \in [0, j].
\end{eqnarray*}

This decomposition into subspaces $a_{kji}$ (or $b_{kji}$)
is not necessarily unique, but
the subspace dimensions $\dim a_{kji} = \dim b_{kji} = \omega_{ki}$
are the same for all such decompositions.
Moreover,
\[
d_j = \sum_{t = j}^L \sum_{s = 0}^j \omega_{ts},
\hspace{.1in}  \forall j \in [0, L],
\hspace{.2in}  \mbox{and}  \hspace{.2in}
\rank W_{k \ssq i} = \sum_{t=k}^L \sum_{s=0}^i \omega_{ts},
\hspace{.1in}  L \geq k \geq i \geq 0.
\]

Moreover,
\begin{eqnarray*}
\bigoplus_{t = j}^k \bigoplus_{s = 0}^i a_{tjs} & = &
A_{kji} \stackrel{\mathrm{def}}{=}
\Null W_{k+1 \ssq j} \cap \col W_{j \ssq i},
\hspace{.2in}  \forall j \in [1, L], k \in [j - 1, L], i \in [-1, j],
\hspace{.2in}  \mbox{and}  \\
\bigoplus_{t = k}^L \bigoplus_{s = i}^j b_{tjs} & = &
B_{kji} \stackrel{\mathrm{def}}{=}
\row W_{k \ssq j} \cap \Null W_{j \ssq i-1}^\top,
\hspace{.2in}  \forall j \in [1, L], k \in [j, L + 1], i \in [0, j + 1].
\end{eqnarray*}
The dimensions of $A_{kji}$ and $B_{kji}$ are, respectively,
\[
\alpha_{kji} = \sum_{t = j}^k \sum_{s = 0}^i \omega_{ts},
\hspace{.2in}  \mbox{and}  \hspace{.2in}
\beta_{kji} = \sum_{t = k}^L \sum_{s = i}^j \omega_{ts}.
\]

Moreover, we obtain subspaces of the specified dimensions satisfying
the flow constraints by arbitrarily choosing
$a_{kii} \in A_{kii} \downarrow (A_{k,i,i-1} + A_{k-1,i,i})$ and
$b_{kki} \in B_{kki} \downarrow (B_{k,k,i+1} + B_{k+1,k,i})$
for all indices satisfying $L \geq k \geq i \geq 0$ and
letting the flow relationships determine the remaining subspaces.


Moreover, given prebases $\mathcal{A}_0, \mathcal{A}_1, \ldots, \mathcal{A}_L$
that satisfy the constraints above ({\em flow prebases}),
we can obtain prebases $\mathcal{B}_0, \mathcal{B}_1, \ldots, \mathcal{B}_L$
that satisfy them by choosing
\[
b_{kji} = \left( \sum_{a_{k'ji'} \in \mathcal{A}_j \setminus \{ a_{kji} \}} a_{k'ji'}
\right)^\perp.
\]
We say these $\mathcal{B}$-prebases are {\em complementary} to
the $\mathcal{A}$-prebases.
By a symmetric formula, given
prebases $\mathcal{B}_0, \mathcal{B}_1, \ldots, \mathcal{B}_L$ that satisfy
the constraints above ({\em flow prebases for the transpose network}),
we can obtain (complementary) prebases
$\mathcal{A}_0, \mathcal{A}_1, \ldots, \mathcal{A}_L$ that satisfy them.
\end{theorem}

\subsection{Connections with Persistence Modules and (Algebraic) Modules}
\label{modules}

After we wrote Section~\ref{fundsub} (excepting this subsection),
we learned that most of the relationships we describe there
were previously derived in the study of persistent homology, which is
the best known method for topological data analysis.
One of the key abstractions of persistent homology is called
a {\em persistence module}, which is essentially a linear neural network---but
so far as we can tell,
nobody has pointed out the applications to neural networks.
We have retroactively adopted some terms from that literature, including
{\em barcode}, {\em birth}, and {\em death}.

In this section, we explain the connections between
the results in Section~\ref{fundsub}, abstract algebra, and
the persistent homology literature (leaving out the homology).
In particular, most of Theorem~\ref{fundnn} can be derived from
a classic result in algebra called the {\em structure theorem
for finitely generated modules over a principal ideal domain}~\cite{goodman15},
which we henceforth will call the Structure Theorem.
Persistent homology and persistence modules as
tools for topological data analysis were introduced by a pioneering paper of
Edelsbrunner, Letscher, and Zomorodian~\cite{edelsbrunner00,edelsbrunner02}.
A~classic paper of Zomorodian and Carlsson~\cite{zomorodian05}
links persistence modules with the Structure Theorem and
uses the latter to extract barcodes.
An advantage of knowing this connection is that the Structure Theorem is
a general tool that supports extensions of
the results in Section~\ref{fundsub}.
In particular, our Theorem~\ref{fundnn} extends to neural networks whose
weights have complex values with no difficulty.
One could also apply the Structure Theorem to
linear neural networks with skip connections.

The identity~(\ref{ranksum}), expressing the subsequence matrix ranks
in terms of the barcode, can be found in
the persistent homology literature, albeit specialized for homology.
In that context, the book {\em Computational Topology:  An Introduction} by
Edelsbrunner and Harer~\cite{edelsbrunner10} calls~(\ref{ranksum})
the {\em Fundamental Lemma of Persistent Homology}, and
calls each $\rank W_{k \ssq i}$ a {\em persistent Betti number}
(a term that may originate with Robins~\cite{robins99}).
The identity~(\ref{omegarank}) also appears on the same page (p.~152) of
Edelsbrunner and Harer, and appeared earlier in
a paper by Cohen-Steiner, Edelsbrunner, and Harer~\cite{cohensteiner07}.

While most of Section~\ref{fundsub} represents
a rediscovery of already known mathematics
(from entirely different motivations---our wish to characterize
the geometry of the fiber),
let us point out some distinctions between our presentation and
the prior literature on persistence modules.
Our target audience is neural network researchers, and
we do not ask readers to learn abstract algebra nor homology.
We have provided complete, elementary proofs that are accessible to many who
would be unwilling to undertake the time-consuming challenge of
learning a proof of the Structure Theorem and applying it to neural networks;
and we think the direct proofs are in some respects more revealing.
We place much emphasis on the identities of
the fundamental subspaces $a_{kji}$ and $b_{kji}$, as
we use them extensively in the rest of this report.
The topological persistence literature typically
shows little interest in the fundamental subspaces, and much interest in
the barcodes, ranks, Betti numbers, and other topological invariants.

The remainder of this section explains the connections between
persistence modules, (algebraic) modules, the Structure Theorem, and
the fundamental subspaces of a linear neural network.
We must assume the reader is familiar with
groups, fields, rings, ideals, and modules, but if you are not, have no fear;
none of this is necessary for understanding the rest of the report, and
you may freely skip to Section~\ref{rankstrata}.
We hope our take on the subject will provide a helpful alternative view to
customary treatments in the persistence literature, and
help students of persistent homology to better understand persistence modules.
Our presentation below moves in the opposite of the usual order:
whereas Zomorodian and Carlsson~\cite{zomorodian05} start with
the Structure Theorem and use it to move from a persistence module to
a unique barcode, our narrative starts with
the ``fundamental'' flow prebasis subspaces of Theorem~\ref{fundnn},
shows how they can be used to decompose a persistence module into submodules,
then explains their (rather obscure, in our opinion) relationship to
the Structure Theorem.

For our purposes, a {\em persistence module} is
a set of vector spaces $V_i$ defined over some or all integers $i$,
together with a set of linear maps $\mathcal{W}_i : V_{i-1} \rightarrow V_i$.
The vector spaces can be defined over any field,\footnote{
Zomorodian and Carlsson~\cite{zomorodian05} define {\em persistence module}
more generally so that the spaces $V_i$ can be modules over a ring, and
each $\mathcal{W}_i$ is a linear homomorphism.
Depending on the choice of ring, the Structure Theorem might or might not apply.
}
and in principle a persistence module can have infinitely many layers, though
typical applications use indices~$i$ in a finite range. 
In our setting of linear neural networks,
the vectors spaces are $V_i = \R^{d_i}$ and
the linear maps are represented by arbitrary real matrices~$W_i$.
To express the homology groups associated with geometric constructions,
Edelsbrunner, Letscher, and Zomorodian~\cite{edelsbrunner02} use
vector spaces over the field $\mathbb{Z} / 2\mathbb{Z}$, and
the linear maps $\mathcal{W}_i$ are inclusion maps
(represented by 0-1 matrices $W_i$ in which
every column contains exactly one $1$ and every row contains at most one $1$).

The name ``persistence module'' is a bit misleading as
it is not a module in the algebraic sense, but
Zomorodian and Carlsson~\cite{zomorodian05} show us how to express
a persistence module as an algebraic module and
use the Structure Theorem to extract its barcode.
For our setting of neural networks, where
$V_i = \R^{d_i}$ and the indices $i$ are in $[0, L]$,
define an abelian group
$V = V_0 \times V_1 \times \cdots \times V_L$
whose group operator is standard vector addition
(we will soon promote $V$ to be a module), and
let $d_V = \sum_{j=0}^L d_j$ denote the dimension of $V$.
For instance, if $L = 3$, we can write an element of $V$ as a vector
\[
v = \left[ \begin{array}{c} v_0 \\ v_1 \\ v_2 \\ v_3 \end{array} \right] \in
V = V_0 \times V_1 \times V_2 \times V_3 =
\left[ \begin{array}{c} V_0 \\ V_1 \\ V_2 \\ V_3 \end{array} \right] = \R^{d_V},
\]
where $v_j \in V_j$ specifies values for the units in layer~$j$.
We define a $d_V \times d_V$ matrix $E$ that
expresses a weight vector~$\theta$ and encodes all its matrices $W_i$
in a manner that can be applied to vectors in $V$.
If $L = 3$,
\[
E = \left[ \begin{array}{cccc}
0   & 0   & 0   & 0 \\
W_1 & 0   & 0   & 0 \\
0   & W_2 & 0   & 0 \\
0   & 0   & W_3 & 0
\end{array} \right],
E^2 = \left[ \begin{array}{cccc}
0       & 0       & 0   & 0 \\
0       & 0       & 0   & 0 \\
W_2 W_1 & 0       & 0   & 0 \\
0       & W_3 W_2 & 0   & 0
\end{array} \right],
E^3 = \left[ \begin{array}{cccc}
0           & 0       & 0   & 0 \\
0           & 0       & 0   & 0 \\
0           & 0       & 0   & 0 \\
W_3 W_2 W_1 & 0       & 0   & 0
\end{array} \right],
\]
and $E^4 = E^5 = \ldots = 0$.
It should be clear how to generalize $E$ to other values of $L$.
(In general, $E^{L+1} = 0$.)
Observe that all the subsequence matrices appear in powers of $E$.

Let $\R[E]$ be the polynomial ring in $E$ over $\R$.
We can write it as
\[
\R[E] = \{ c_0 I + c_1 E + c_2 E^2 + \ldots + c_L E^L : c_i \in \R \}.
\]

Now we define $V$ as a left $\R[E]$-module
whose elements are $\R^{d_V}$, whose group operation is vector addition, and
whose multiplication operation is the customary product of
a matrix in $\R[E]$ with a vector in $V$.
$V$ is an $\R[E]$-module because $(V, +)$ is an abelian group,
the matrix-vector product always produces an element of~$V$, and
other properties are satisfied.\footnote{
The module's multiplication operation $\cdot : \R[E] \times V \rightarrow V$ has
the property that for all $M, M' \in \R[E]$ and all $v, v' \in V$,
(1)~$M \cdot (v + v') = M \cdot v + M \cdot v'$,
(2)~$(M + M') \cdot v = M \cdot v + M' \cdot v$,
(3)~$(MM') \cdot v = M \cdot (M' \cdot v)$, and
(4)~$I \cdot v = v$, where $I$ is the identity element of the ring $\R[E]$.
}

The Structure Theorem describes how a module can be decomposed into submodules.
A {\em submodule} of $V$ is a subgroup of $(V, +)$ that is also
a module---equivalently, a subgroup such that the product of
any matrix in $\R[E]$ and any vector in the submodule is also in the submodule.
Theorem~\ref{fundnn} and
the flow prebasis subspaces defined in Section~\ref{flowrelations} offer us
building blocks for decomposing $V$ into submodules.
For a three-layer network, examples of submodules of $V$ include
\begin{equation}
V_{30} = \left[ \begin{array}{c} a_{300} \\ a_{310} \\ a_{320} \\ a_{330}
                \end{array} \right],
V_{21} = \left[ \begin{array}{c} {\bf 0} \\ a_{211} \\ a_{221} \\ {\bf 0}
                \end{array} \right],
\mbox{~and~}
V_{33} = \left[ \begin{array}{c} {\bf 0} \\ {\bf 0} \\ {\bf 0} \\ a_{333}
                \end{array} \right].
\label{intervalmodules}
\end{equation}

We call these submodules {\em interval modules}, as each $V_{ki}$ corresponds to
the interval $[i, k]$ in the barcode defined in Section~\ref{barcodes}.
Observe that each interval module is a subspace of $V$, as
each $a_{kji}$ is a subspace of $V_j = \R^{d_j}$.
Hence it is clear why each interval module is a subgroup of $(V, +)$.
To see that each interval module is a module, we must verify that
for every matrix $M \in \R[E]$ and every vector $v$ in the interval module,
$Mv$ is also in the interval module.
For the submodule $V_{30}$ in~(\ref{intervalmodules}),
we see that this is true because the flow conditions imply that
\begin{equation}
E \left[ \begin{array}{c} a_{300} \\ {\bf 0} \\ {\bf 0} \\ {\bf 0} \end{array}
  \right] =
  \left[ \begin{array}{c} {\bf 0} \\ a_{310} \\ {\bf 0} \\ {\bf 0} \end{array}
  \right],
E \left[ \begin{array}{c} {\bf 0} \\ a_{310} \\ {\bf 0} \\ {\bf 0} \end{array}
  \right] =
  \left[ \begin{array}{c} {\bf 0} \\ {\bf 0} \\ a_{320} \\ {\bf 0} \end{array}
  \right],
E \left[ \begin{array}{c} {\bf 0} \\ {\bf 0} \\ a_{320} \\ {\bf 0} \end{array}
  \right] =
  \left[ \begin{array}{c} {\bf 0} \\ {\bf 0} \\ {\bf 0} \\ a_{330} \end{array}
  \right],
\mbox{~and~}
E \left[ \begin{array}{c} {\bf 0} \\ {\bf 0} \\ {\bf 0} \\ a_{330} \end{array}
  \right] = {\bf 0},
\label{moduleflow}
\end{equation}
so for every $v \in V_{30}$ and $M \in \R[E]$, $Mv \in V_{30}$.
We see here that the action of $E$ on neural network units is
to advance the information one layer forward in the neural network.
Hence each $\R[E]$-submodule naturally encodes
how some of the information is transformed as it moves through the network.

With this background,
the central idea of Theorem~\ref{fundnn} can be expressed in terms of modules.
For every weight vector~$\theta$, let $E$ be the corresponding matrix and
let $V$ be the corresponding $\R[E]$-module.
Then we can decompose $V$ into submodules as
\begin{equation}
V = \bigoplus_{L \geq k \geq i \geq 0} V_{ki}
\label{submodules}
\end{equation}
where each submodule $V_{ki}$ of $V$ follows
the pattern of~(\ref{intervalmodules}) and~(\ref{moduleflow}) and has
entries satisfying $\dim a_{kji} = \dim a_{kj'i}$ for all $j, j' \in [i, k]$.
This decomposition of $V$ into submodules shows both
how each layer of the network is decomposed into flow prebasis subspaces and
how those subspaces flow through the network.

We can further decompose each interval module into cyclic interval modules.
A module is {\em cyclic} if it is generated by a single element.
Recall that the dimension of $a_{kji}$ is $\omega_{ki}$, and
in Section~\ref{flowbases} we define a matrix $J_{kji}$ with
$\omega_{ki}$ columns that form a basis for $a_{kji}$.
An interval module $V_{ki}$ of the form~(\ref{intervalmodules}) can be written
as a direct sum of $\omega_{ki}$ cyclic modules,
one for each basis vector of $a_{kii}$.
(If $\omega_{ki} = 1$, then $V_{ki}$ is itself cyclic.)
For example, consider the case where $L = 3$ and $\omega_{30} = 3$, and
let $p$, $p'$, and~$p''$ be the columns of $J_{300}$ (a basis for $a_{300}$).
Then
\begin{equation}
V_{30} =
\left[ \begin{array}{c} a_{300} \\ a_{310} \\ a_{320} \\ a_{330}
       \end{array} \right] =
\left[ \begin{array}{c} \R p \\ \R W_1 p \\ \R W_2 W_1 p \\
                        \R W_3 W_2 W_1 p
       \end{array} \right] \oplus
\left[ \begin{array}{c} \R p' \\ \R W_1 p' \\ \R W_2 W_1 p' \\
                        \R W_3 W_2 W_1 p'
       \end{array} \right] \oplus
\left[ \begin{array}{c} \R p'' \\ \R W_1 p'' \\ \R W_2 W_1 p'' \\
                        \R W_3 W_2 W_1 p'' \end{array} \right].
\label{subsubmodules}
\end{equation}
The submodules on the right-hand side are cyclic because they are generated by
the elements
\[
\left[ \begin{array}{c} p \\ 0 \\ 0 \\ 0 \\ \end{array} \right],
\left[ \begin{array}{c} p' \\ 0 \\ 0 \\ 0 \\ \end{array} \right],
\mbox{~and~}
\left[ \begin{array}{c} p'' \\ 0 \\ 0 \\ 0 \\ \end{array} \right].
\]

Some authors use the term ``interval module'' to mean
what we call ``cyclic interval modules''~\cite{chazal16}, but
we find it useful to extend the definition of ``interval module'' to include
sums of cyclic interval submodules represented by the same interval.

The upshot is that we can refine the submodule decomposition~(\ref{submodules})
to obtain a decomposition of $V$ into cyclic interval submodules,
\begin{equation}
V = \bigoplus_{L \geq k \geq i \geq 0} \bigoplus_{z=1}^{\omega_{ki}} V_{ki}^z,
\label{cyclicsubmodules}
\end{equation}
where each $V_{ki}^z$ is a cyclic submodule of $V_{ki}$.
Note that this decomposition is not generally unique,
for the same reasons that the $a_{kji}$'s and $J_{kji}$'s are not unique.

For our purposes, the Structure Theorem is a statement that
every $\R[E]$-module has a decomposition into
cyclic $\R[E]$-submodules of the form~(\ref{cyclicsubmodules}).
We wish we could end our discussion here, but
the Structure Theorem is written not in terms of submodules of $V$,
but rather in terms of quotient rings derived from the ring $\R[E]$.
The Structure Theorem implies the existence of
the decomposition~(\ref{cyclicsubmodules}) in an opaque manner,
which will occupy three more pages of writing.
(You're still welcome to skip to Section~\ref{rankstrata}.)

To apply the Structure Theorem to persistence modules,
we need only quotient rings of the form $\R[E] / (E^t)$, where
$t$~is a positive integer and $(E^t)$ denotes the ideal generated by~$E^t$.
Recall that the ideal generated by $E^t$ is the set
(representing an equivalence class)
\[
(E^t) = \R[E] E^t = \{ M E^t : M \in \R[E] \} =
\{ c_t E^t + c_{t+1} E^{t+1} + c_{t+2} E^{t+2} + \ldots :
   c_t, c_{t+1}, c_{t+2}, \ldots \in \R \}.
\]
Each quotient ring is a set of equivalence classes of elements in $\R[E]$,
denoted
\[
\R[E] / (E^t) =
\{ c_0 I + c_1 E + c_2 E^2 + \ldots + c_{t-1} E^{t-1} + (E^t) :
   c_0, c_1, \ldots, c_{t-1} \in \R \}.
\]
This quotient ring is an $\R[E]$-module, as the product of
any element of $\R[E]$ with any element of $\R[E] / (E^t)$ is
in $\R[E] / (E^t)$.\footnote{
This is a potential point of confusion because one might think that
the product of $2$ with $(E^t)$ would be $(2 E^t)$, or
the product of $E^s$ with $(E^t)$ would be $(E^{s+t})$.
But multiplication in a quotient ring module is defined so that
every product also has the term~$(E^t)$.
We write this footnote because
one coauthor lost many days trying to figure this out.
}

Each cyclic interval module $V_{ki}^z$ in~(\ref{cyclicsubmodules}) is
isomorphic to the quotient ring $\R[E] / (E^{k-i+1})$.
We write that as
\[
V_{ki}^z \cong \R[E] / (E^{k-i+1}).
\]

References about persistent homology are
often miserly about specifying the isomorphism.
We will be generous and describe an isomorphism $\varphi$ that maps
$V_{ki}^z$ to $\R[E] / (E^{k-i+1})$.
Let $v$ be an element of $V_{ki}^z$ and
recall that a cyclic module $V_{ki}^z$ is generated by some single element
$u \in V$; therefore, there is an element $M \in \R[E]$ such that $v = Mu$.
We define~$\varphi$ so it maps $v$ to the equivalence class $M + (E^{k-i+1})$.
This map is motivated by the fact that
we can replace~$M$ with any other member of $M + (E^{k-i+1})$ and
still have $v = Mu$.
Why?
Because $E^{k-i+1} u = 0$:  the submodule~$V_{ki}$ represents some information in
unit layers~$i$ through $k$ of the neural network that dies at layer~$k$
(never reaching layer~$k + 1$).
From Theorem~\ref{fundnn}, one can show that $\varphi$ is a bijection
from elements of $V_{ki}^z$ to elements of $\R[E] / (E^{k-i+1})$.

When we call $\varphi$ an {\em isomorphism},
we mean not only that $\varphi$ is a bijection, but also that
the bijection preserves the module structure, in the sense that
$\varphi(Mv) = M \varphi(v)$
for every $v \in V_{ki}^z$ and every $M \in \R[E] / (E^{k-i+1})$.
(We know that $\varphi(Mv) = M \varphi(v)$ simply because
$M(M'u) = (MM')u$, where $M'$ is the matrix such that $v = M'u$.)

The decomposition~(\ref{cyclicsubmodules}) implies the isomorphism
\[
V \cong
\varprod_{L \geq k \geq i \geq 0} \varprod_{z=1}^{\omega_{ki}} \R[E] / (E^{k-i+1}),
\]
which is very nearly the Structure Theorem.
Again, the isomorphism symbol $\cong$ means not only that
there is a bijection from the $\R[E]$-module $V$ to
the Cartesian product of quotient spaces on the right-hand side, but also
that the bijection preserves the $\R[E]$-module structure.
This raises the question of how we can interpret the right-hand side as
an $\R[E]$-module.
The right-hand side is a product of quotient rings, and
each of those quotient rings is an $\R[E]$-module.
To make the right-hand side be an $\R[E]$-module,
we define the product of an element $M \in \R[E]$ and
a Cartesian product of $\R[E]$-quotient spaces
$(q_1, q_2, q_3, \ldots)$ to be $(M q_1, M q_2, M q_3, \ldots)$.

What the Structure Theorem actually says is that
for every $\R[E]$-module $V$, there exists an isomorphism\footnote{
The Structure Theorem is usually written
$V \cong \bigoplus_z \R[E] / (r_z)$.
Here, the symbol $\bigoplus$ represents an {\em external direct sum}, equivalent
to $\varprod_z \R[E] / (r_z)$, in contrast to
the {\em internal direct sums} we have been using throughout this report.
We reserve $\bigoplus$ for the latter.
}
\begin{equation}
V \cong \varprod_z \R[E] / (r_z)
\label{structuretheorem}
\end{equation}
where each $r_z \in R[E]$ generates a proper ideal $(r_z)$ of $R[E]$ and
$r_1$ divides $r_2$, $r_2$ divides $r_3$, and so on
(hence $(r_1) \supseteq (r_2) \supseteq (r_3) \ldots$).
Moreover, this decomposition is unique
(although the isomorphism itself is generally not unique,
for the same reasons that the $a_{kji}$'s and $J_{kji}$'s are not unique).

The Structure Theorem does not specify the values of $r_z$ in
the right-hand side of~(\ref{structuretheorem}), nor
how many factors are on the right-hand side,
nor the birth and death layers of the intervals,
and it says even less about how to decompose $V$ into submodules;
it leaves out most of the information
on the right-hand side of~(\ref{cyclicsubmodules}).
But standard proofs of the Structure Theorem suggest
a {\em reduction algorithm} for answering these questions.

The Structure Theorem~(\ref{structuretheorem}) implies that
$V$ has a decomposition into cyclic submodules, but
the form of that decomposition is not necessarily~(\ref{cyclicsubmodules}).
Consider a linear neural network with {\em skip connections}---for instance,
some connections that go directly from layer~2 to layer~4.
The Structure Theorem applies and the isomorphism~(\ref{structuretheorem})
implies that a submodule decomposition exists, but in general,
the ``barcode'' might no longer be made up of intervals.
There is a stronger variant of the Structure Theorem that applies only
to networks with no skip connections; it implies that
there is a submodule decomposition of the form~(\ref{cyclicsubmodules}).

Zomorodian and Carlsson~\cite{zomorodian05} use a {\em graded} version of
the Structure Theorem that encodes the birth and death layers
by expressing $\R[E]$ as a graded ring, and
$V$ and $\R[E] / (E^t)$ as graded modules.
The graded Structure Theorem provides an explicit guarantee that every
persistence module has a decomposition of the form~(\ref{cyclicsubmodules}), and
every such decomposition has the same unique barcode.

We won't provide a general definition of {\em graded ring}, but
we will write $\R[E]$ as a graded ring.
Observe that the polynomial ring $\R[E]$ with indeterminate matrix $E$
can be written as a direct sum of abelian groups
$\R[E] = \bigoplus_{j=0}^\infty R_j$ where $R_j = \{ c E^j : c \in \R \}$.
That is, $R_j$ contains all the monomials of degree $j$.
What makes this decomposition a {\em grading} is
the property that $R_i \cdot R_j \subseteq R_{i+j}$ for all $i, j \geq 0$.
(In this case, $R_i \cdot R_j = R_{i+j}$.)
This decomposition of $\R[E]$ by monomial degree is called
the {\em canonical grading} of the polynomial ring.

Likewise, $V = V_0 \times V_1 \times \cdots \times V_L$ can be written as
a {\em graded module}.
Let $\bar{V}_j =
\{ {\bf 0} \} \times \cdots \times \{ {\bf 0} \} \times V_j \times
\{ {\bf 0} \} \times \cdots \times \{ {\bf 0} \}$, so
$\bar{V}_j$ is the vector space $V_j$ padded out with zeros to be
a subspace of the larger space $V$.
We can write $V = \bigoplus_{j=0}^L \bar{V}_j$.
What makes this decomposition a {\em grading} is the property that
\[
R_i \cdot \bar{V}_j \subseteq \bar{V}_{i+j}  \mbox{~for all~} i, j \geq 0,
\]
because a degree-$i$ monomial in $E$ has the action of advancing information
exactly $i$~layers forward in the neural network; recall~(\ref{moduleflow}).
(This property depends on our suitable choice of~$E$.)
Observe that $\R[E]$ and $V$ are graded in different ways---$\R[E]$ by
the degree of each monomial, and
$V$ by the unit layers of the neural network (or persistence module)---but
this relationship links the grading of $\R[E]$ to the grading of $V$.

A key insight is that by restricting $V$ to be a graded module
(over a graded ring),
we ensure that every cyclic submodule of $V$ is an interval module.
(A cyclic submodule cannot ``skip'' a layer of the network.)

The quotient ring $\R[E] / (E^t)$ can also be written as
a graded $\R[E]$-module.
Use the canonical grading:
$\R[E] / (E^t) = \bigoplus_{j=0}^\infty \bar{R}_j$ where
$\bar{R}_j = \{ c E^j + (E^t) : c \in \R \}$ for $j \in [0, t - 1]$ and
$\bar{R}_j = \{ (E^t) \}$ for $j \geq t$.
This yields a graded module because
$R_i \cdot \bar{R}_j \subseteq \bar{R}_{i+j}$ for all $i, j \geq 0$.

For the graded version of the Structure Theorem,
we need a new notation $\Sigma^i \R[E] / (E^t)$, which has
the effect of shifting the quotient ring $i$ steps upward in the grading,
meaning that the group $\bar{R}_j$ is renamed to be $\bar{R}_{i+j}$.
The groups of the graded module $\Sigma^i \R[E] / (E^t)$ are
\[
\bar{R}_j = \left\{
\begin{array}{ll}
\{ 0 \}, & j < i, \\
\{ c E^{j-i} + (E^t) : c \in \R \}, & j \in [i, i + t - 1], \\
\{ (E^t) \} & j \geq i + t.
\end{array}
\right.
\]
If we ignore the grading, the quotient groups
$\R[E] / (E^t)$ and $\Sigma^i \R[E] / (E^t)$ are identical;
only the indices of the grading have changed.
The shifted quotient ring $\Sigma^i \R[E] / (E^t)$ is interpreted to symbolize
an interval $[i, i + t - 1]$, having
birth layer $i$ and death layer $i + t - 1$.

Recall the isomorphism $V_{ki}^z \cong \R[E] / (E^{k-i+1})$.
We can reformulate this as a {\em graded isomorphism}
\[
V_{ki}^z \cong \Sigma^i \R[E] / (E^{k-i+1}).
\]
Here, the isomorphism $\varphi$ is exactly the same function as before, but now
it maps each element of $\bar{V}_j$ to an element of $\bar{R}_j$
for all $j \geq 0$ (rather than mapping $\bar{V}_j$ into $\bar{R}_{j-i}$), so
we say the isomorphism is ``graded.''

The graded version of the Structure Theorem says that
for every graded $\R[E]$-module $V$, there exists a graded isomorphism
\begin{equation}
V \cong \varprod_z \Sigma^{i_z} \R[E] / (r_z)
\label{gradedstructure}
\end{equation}
where each $r_z$ is {\em homogeneous}, meaning
it belongs to a single group $R_j$ in the grading of $\R[E]$.
Hence we must have $(r_z) = (E^{t_z})$ for some positive integer $t_z$.
The term {\em graded isomorphism} implies that
it maps each $\bar{V}_j$ into $(\bar{R}_j, \bar{R}_j, \ldots, \bar{R}_j)$.
Moreover, the decomposition on the right-hand side of~(\ref{gradedstructure}) is
unique.
(Again, the isomorphism might not be.)

The upshot is that the graded version of the Structure Theorem indirectly states
that there exists a decomposition of $V$ into cyclic interval submodules
of the form~(\ref{cyclicsubmodules}).
Each quotient ring $\Sigma^{i_z} \R[E] / (E^{t_z})$ in
the Cartesian product~(\ref{gradedstructure}) signifies
a cyclic interval submodule with
birth layer $i_z$ and death layer $i_z + t_z - 1$.
In this sense, the Graded Structure Theorem implies
our Fundamental Theorem of Linear Neural Networks (Theorem~\ref{fundnn}).
Although the right-hand side of~(\ref{gradedstructure}) leaves out
much of the information in~(\ref{cyclicsubmodules}),
it does encode the complete barcode
(unlike its ungraded analogue~(\ref{structuretheorem})), and
no other information.
The uniqueness of the right-hand side implies that $V$'s barcode is unique,
even though $V$'s decomposition is not.
This uniqueness appears to be a reason why
mathematicians prefer~(\ref{gradedstructure}) to~(\ref{cyclicsubmodules}).

The Graded Structure Theorem, like the original Structure Theorem,
does not itself specify
how many factors are on the right-hand side, nor the values of $i_z$ or $t_z$,
but its proof suggests a method for finding them.
That proof is similar to the proof of the original Structure Theorem, but
additional effort is needed to show that
the left-hand and right-hand gradings can be aligned.
See L\"{o}h~\cite{loh23} for a helpful elucidation of the
graded version of the Structure Theorem and its proof.

We have used our Theorem~\ref{fundnn} as a starting point to explain
how the Structure Theorem applies to persistence modules, as
we think that's the pedagogically easier direction.
There is much power (albeit more technical difficulty) in
reasoning in the opposite direction---that is,
using the Structure Theorem (and the reduction algorithm suggested by its proof)
to decompose a module into cyclic submodules,
as Zomorodian and Carlsson~\cite{zomorodian05} do.
This is how we know that Theorem~\ref{fundnn} generalizes easily to
matrices over other fields, notably to complex matrices.
One fascinating extension is that the Structure Theorem shows that
one can define ``fundamental subspaces'' for
a linear neural network with an infinite number of layers, or
a linear neural network in which some layers are infinite-dimensional
spaces---but not necessarily for both simultaneously!
For details, see
Chazal, de~Silva, Glisse, and Oudot~\cite{chazal16}, Section~2.5.

\clearpage

\section{Strata in the Rank Stratification}
\label{rankstrata}

In this section, we prove some of the results about rank stratifications
promised in Section~\ref{foretaste}.
\begin{itemize}
\item
Each stratum $S_{\underline{r}}$---the set of points on a fiber $\mu^{-1}(W)$
with rank list $\underline{r}$---is an analytic manifold
(Theorem~\ref{stratummanifold}).
\item
Given two strata from two different fibers that have the same rank list,
there is a linear homeomorphism mapping one stratum to the other
(Corollary~\ref{affinestratacor}).
Hence the topology of a stratum is solely determined by its rank list.
\item
Given two points on the same stratum,
there is a linear homeomorphism mapping the stratum to itself and
mapping one point to the other (Corollary~\ref{selfsimilar}).
\end{itemize}

\subsection{Strata, Semi-Algebraic Sets, and Determinantal Varieties}
\label{sec:detmanifolds}

Let $\mathbb{S}$ be a rank stratification of a fiber $\mu^{-1}(W)$.
Here we will see that although a stratum in $\mathbb{S}$ is
not necessarily an algebraic variety,
it belongs to the closely related class of semi-algebraic sets.
This will help us to see in Section~\ref{sec:affinestrata}
why each stratum in $\mathbb{S}$ is a manifold.
A {\em semi-algebraic set} is a set that can be obtained from finitely many
sets of the form $\{ \theta \in \R^{d_\theta} : f_i(\theta) \geq 0 \}$,
where each $f_i$ is a polynomial, 
by a combination of union, intersection, and complementation operations.
This class of sets include all algebraic varieties,
which can be expressed as intersections of
sets of the form $\{ \theta \in \R^{d_\theta} : f_j(\theta) = 0 \}$,
each of which is the intersection of two sets of the form
$\{ \theta \in \R^{d_\theta} : f_k(\theta) \geq 0 \}$.

To see that a stratum is an algebraic set,
consider first how to constrain the rank of a matrix with algebra.
The property that a matrix $M$ has rank at most $r$ can be written as
a constraint that
the determinant of every $(r + 1) \times (r + 1)$ minor of $M$ is zero.
(The number of polynomial equations specified this way can grow
exponentially with $r$, but typically most of them are redundant.)
Thus the set of all $p \times q$ matrices with rank~$r$ or less is
an algebraic variety,
called the \textit{determinantal variety}~\cite[Lecture 9]{harris95},
which we denote
\[
\DV_r^{p \times q} = \{ M \in \R^{p \times q} : \rank M \leq r \}.
\]

The determinantal variety has singular points and thus is not a manifold
(unless $r$ is zero; $\DV_0^{p \times q}$ contains only the zero matrix).
It is well known that the singular locus of
$\DV_r^{p \times q}$ is $\DV_{r-1}^{p \times q}$---the
matrices with rank strictly less than $r$.
If we omit those matrices,
we obtain a manifold that we call the \textit{determinantal manifold},
\[
\DM_r^{p \times q} = \DV_r^{p \times q} \setminus \DV_{r-1}^{p \times q}
= \{ M \in \R^{p \times q} : \rank M = r \}.
\]
It is not hard to see that
$\DV_r^{p \times q}$ is the closure of $\DM_r^{p \times q}$.\footnote{
The set $\{ \DM_0^{p \times q}, \DM_1^{p \times q}, \ldots, \DM_r^{p \times q} \}$
is a stratification of $\DV_r^{p \times q}$ into manifolds of dimension
$0, 1, \ldots, r$.}
So although $\DM_r^{p \times q}$ is a manifold,
it is not a closed point set with respect to the weight space (unless $r = 0$).
It is well known that $\DV_r^{p \times q}$ has dimension $r (p + q - r)$,
hence so does $\DM_r^{p \times q}$.

$\DM_r^{p \times q}$ is not an algebraic variety (unless $r = 0$).
A $p \times q$ matrix $M$ is in $\DM_r^{p \times q}$ if and only if
the determinant of every $(r + 1) \times (r + 1)$ minor of $M$ is zero but
the determinant of at least one $r \times r$ minor is nonzero.
We cannot express the latter constraint in a system of polynomial equations.
But $\DM_r^{p \times q}$ is a set difference of two algebraic varieties, so
it is a semi-algebraic set.

We define a stratum in $\mathbb{S}$ by fixing
the rank of each subsequence matrix.
If a weight vector $\theta = (W_L, \ldots, W_1)$ has rank list $\underline{r}$,
then $W_1$ lies on the determinantal manifold $\DM_{r_{1 \ssq 0}}$,
$W_2W_1$ lies on $\DM_{r_{2 \ssq 0}}$,
$W_4W_3W_2$ lies on $\DM_{r_{4 \ssq 1}}$, and so on.
These constraints are semi-algebraic, so a stratum is a semi-algebraic set.

For our purposes, we need to express these constraints
in terms of the weight vector $\theta$,
not in terms of $W_2W_1$ and $W_4W_3W_2$.
This motivates what we call a \textit{weight-space determinantal set},
denoted
\[
\WDS_{\underline{r}}^{k \ssq i} =
\{ (W_L, \ldots, W_1) : \rank W_{k \ssq i} = r_{k \ssq i} \}.
\]
One difference between $\WDS_{\underline{r}}^{k \ssq i}$ and
$\DM_r^{d_k \times d_i}$ is that $\WDS_{\underline{r}}^{k \ssq i}$ is
a set of points in weight space, and
$\DM_r^{d_k \times d_i}$ is a set of points in matrix space.
For example, in the case where $k = i + 1$,
$\WDS_{\underline{r}}^{i+1 \ssq i}$ is the manifold
\[
\R^{d_L \times d_{L-1}} \times \cdots \times \R^{d_{i+2} \times d_{i+1}} \times 
\DM_{r_{i+1 \ssq i}}^{d_{i+1} \times d_i} \times \R^{d_i \times d_{i-1}} \times
\cdots \times \R^{d_1 \times d_0}.
\]
Another difference is that
the weight-space determinantal set is {\em not} always a manifold,
for the same reason that $\mu^{-1}(W)$ is not always a manifold.\footnote{
For example, if $r_{k \ssq i} = 0$ then
$\WDS_{\underline{r}}^{k \ssq i} =
\mu_{k \ssq i}^{-1}(0)$, where
$\mu_{k \ssq i}(W_L, W_{L-1}, \ldots, W_1) = W_k W_{k-1} \cdots W_{i+1}$.
The variety $\mu_{k \ssq i}^{-1}(0)$ is not generally a manifold, as
Figures~\ref{strat112} and~\ref{strat1111} remind us.
}
But the subsequence matrices are polynomial in~$\theta$, so
$\WDS_{\underline{r}}^{k \ssq i}$ is a semi-algebraic set that can be expressed as
a set difference of two algebraic varieties, just like $\DM_r^{p \times q}$.

Consider partitioning the entire weight space $\R^{d_\theta}$ by rank list
(forgetting briefly about the fiber and the strata).
Consider the set of all points in weight space having
a specified rank list $\underline{r}$.
We call this set a {\em multideterminantal manifold}, denoted
\[
\MM_{\underline{r}} = \{ (W_L, \ldots, W_1) \in \R^{d_\theta} :
\rank W_{k \ssq i} = r_{k \ssq i} \mbox{~for all~} L \geq k \geq i \geq 0 \}.
\]
We gave one characterization of $\MM_{\underline{r}}$ in
Corollary~\ref{ranklistgenerateall}.
The multideterminantal manifold is
the intersection of the weight-space determinantal sets:
\[
\MM_{\underline{r}} = \bigcap_{L \ge k > i \ge 0} \WDS_{\underline{r}}^{k \ssq i}.
\]
It follows that $\MM_{\underline{r}}$ is a semi-algebraic set.
($\MM_{\underline{r}}$ is also a manifold, but we will not prove it, as
we don't need to.
But in Section~\ref{sec:affinestrata} we prove that each stratum is a manifold,
and that proof can easily be adapted to $\MM_{\underline{r}}$.)
The entire weight space $\R^{d_\theta}$ can be partitioned into
a finite set of these multideterminantal manifolds,
one for each valid rank list.

Each stratum in the rank stratification $\mathbb{S}$ is
the intersection of the fiber with a multideterminantal manifold,
\[
S_{\underline{r}}^W = \mu^{-1}(W) \cap \MM_{\underline{r}}.
\]
Hence, $S_{\underline{r}}^W$ is a semi-algebraic set.
We prove in the next section that it is a manifold.

\subsection{The Rank List Solely Determines a Stratum's Topology}
\label{sec:affinestrata}

Here we show that in the rank stratification, each stratum is a smooth manifold;
moreover, the topology of a stratum is determined solely by its rank list.
In particular, if two strata in two different fibers have the same rank list,
a linear transformation maps one stratum to the other.
This linear transformation (restricted to the first stratum)
is a homeomorphism, showing that
the topology of a stratum is fully determined by its rank list.
Moreover, this transformation can be chosen to map
any selected point on one stratum to any selected point on the other.
Therefore, the local topology of a stratum looks the same
from every point on the stratum, and
even the geometry of a stratum looks the same from every point
if we ignore scaling.
A~semi-algebraic set with this property is necessarily a smooth manifold.

Consider a matrix $W$ and a point on its fiber,
$\theta = (W_L, W_{L-1}, \ldots, W_1) \in \mu^{-1}(W)$,
with rank list $\underline{r}$.
In the rank stratification $\mathbb{S}$ of $\mu^{-1}(W)$,
$S^W_{\underline{r}}$~is the stratum that contains $\theta$.
Let $J_0, J_1, \ldots, J_L$ represent flow bases at $\theta$,
as described in Section~\ref{flowbases}, and
recall that for $j \in [1, L]$, $W_j = J_j \tilde{I}_j J_{j-1}^{-1}$ where
\[
\tilde{\theta} = (\tilde{I}_L, \tilde{I}_{L-1}, \ldots, \tilde{I}_1).
\]
is the canonical weight vector for $\underline{r}$, defined in
Section~\ref{canonical}.
Recall that the value of $\tilde{\theta}$ depends solely on $\underline{r}$;
it~does not otherwise depend on $\theta$ nor $W$.

Consider the linear transformations
\begin{align}
\eta & : \R^{d_\theta} \rightarrow \R^{d_\theta}, (M_L, M_{L-1}, \ldots, M_1) \mapsto
(J_L M_L J_{L-1}^{-1}, J_{L-1} M_{L-1} J_{L-2}^{-1}, \ldots, J_1 M_1 J_0^{-1})
\hspace{.2in}  \mbox{~and}  \label{eta}  \\
\eta^{-1} & : \R^{d_\theta} \rightarrow \R^{d_\theta},
(M_L, M_{L-1}, \ldots, M_1) \mapsto
(J_L^{-1} M_L J_{L-1}, J_{L-1}^{-1} M_{L-1} J_{L-2}, \ldots, J_1^{-1} M_1 J_0).
\nonumber
\end{align}
Observe that
$\eta(\tilde{\theta}) = \theta$ and $\eta^{-1}(\theta) = \tilde{\theta}$.
As the $J_i$'s are invertible,
$\eta$ and $\eta^{-1}$ are linear bijections from weight space to weight space.
Crucially, it is easy to see that
$\xi$, $\eta(\xi)$, and $\eta^{-1}(\xi)$ all have the same rank list
for any $\xi \in \R^{d_\theta}$, as
a matrix's rank is invariant under multiplication by an invertible matrix.

Let $\tilde{I} = \mu(\tilde{\theta}) =
\tilde{I}_L \tilde{I}_{L-1} \cdots \tilde{I}_1$, depicted in Figure~\ref{ident}.
Consider the fiber $\mu^{-1}(\tilde{I})$ of $\tilde{I}$ and
some stratum $S^{\tilde{I}}_{\underline{z}}$ in
the rank stratification of that fiber.
The following lemma shows that the linear transformation $\eta$ maps
the fiber $\mu^{-1}(\tilde{I})$ to the fiber $\mu^{-1}(W)$, and
likewise maps the stratum $S^{\tilde{I}}_{\underline{z}}$ to
the stratum $S^W_{\underline{z}}$.
So both fibers have the same topology, and both strata have the same topology.
Given a set $Z \subseteq \R^{d_\theta}$, let $\eta(Z)$ denote
the set obtained by applying $\eta$ to every weight vector in $Z$.

\begin{lemma}
$\eta(\mu^{-1}(\tilde{I})) = \mu^{-1}(W)$ and
$\eta \left( S^{\tilde{I}}_{\underline{z}} \right) = S^W_{\underline{z}}$ for
every stratum $S^{\tilde{I}}_{\underline{z}}$ in
the rank stratification of $\mu^{-1}(\tilde{I})$.
\label{affinestratalem}
\end{lemma}

\begin{proof}
For any point $\xi = (X_L, X_{L-1}, \ldots, X_1) \in \R^{d_\theta}$,
\begin{eqnarray*}
\mu(\eta(\xi)) & = &
J_L X_L J_{L-1}^{-1} J_{L-1} X_{L-1} J_{L-2}^{-1} \cdots J_1 X_1 J_0^{-1} =
J_L X_L X_{L-1} \cdots X_1 J_0^{-1} = J_L \, \mu(\xi) \, J_0^{-1}
\hspace{.2in}  \mbox{~and}  \\
\mu(\eta^{-1}(\xi)) & = &
J_L^{-1} X_L J_{L-1} J_{L-1}^{-1} X_{L-1} J_{L-2} \cdots J_1^{-1} X_1 J_0 =
J_L^{-1} X_L X_{L-1} \cdots X_1 J_0 = J_L^{-1} \mu(\xi) \, J_0.
\end{eqnarray*}
From~(\ref{Wfactorcanon}), we have $W = J_L \tilde{I} J_0^{-1}$, and thus
$\tilde{I} = J_L^{-1} W J_0$.
To see that $\eta$~maps any point on $\mu^{-1}(\tilde{I})$ to
a point on $\mu^{-1}(W)$,
observe that for any point $\xi \in \mu^{-1}(\tilde{I})$,
$\mu(\eta(\xi)) = J_L \, \mu(\xi) J_0^{-1} = J_L \tilde{I} J_0^{-1} = W$.
Conversely, to see that $\eta^{-1}$ maps any point on $\mu^{-1}(W)$ to
a point on $\mu^{-1}(\tilde{I})$,
observe that for any point $\xi \in \mu^{-1}(W)$,
$\mu(\eta^{-1}(\xi)) = J_L^{-1} \mu(\xi) J_0 = J_L^{-1} W J_0 = \tilde{I}$.
Hence, $\eta(\mu^{-1}(\tilde{I})) = \mu^{-1}(W)$ as claimed.

Recall that $S^W_{\underline{z}}$ contains
the points on $\mu^{-1}(W)$ with rank list $\underline{z}$ and
$S^{\tilde{I}}_{\underline{z}}$ contains
the points on $\mu^{-1}(\tilde{I})$ with rank list $\underline{z}$.
As $\xi$, $\eta(\xi)$, and $\eta^{-1}(\xi)$ have
the same rank list for any $\xi \in \R^{d_\theta}$,
$\eta$ maps any point on~$S^{\tilde{I}}_{\underline{z}}$ to
a point on~$S^W_{\underline{z}}$,
$\eta^{-1}$ maps any point on $S^W_{\underline{z}}$ to
a point on $S^{\tilde{I}}_{\underline{z}}$, and
$\eta \left( S^{\tilde{I}}_{\underline{z}} \right) = S^W_{\underline{z}}$ as claimed.
\end{proof}

See Appendix~\ref{transnulldmu} for another lemma related to
Lemma~\ref{affinestratalem}, applied to
a subspace we will introduce in Section~\ref{diffmap}.

It is a short extra step to see that
for any other matrix $W' \in \R^{d_L \times d_0}$ of the same rank,
there exists a linear transformation that likewise maps
$\mu^{-1}(W)$ to $\mu^{-1}(W')$ and $S^W_{\underline{z}}$ to $S^{W'}_{\underline{z}}$.
Therefore all strata with the same rank list have essentially the same geometry,
up to a linear transformation in weight space.

\begin{cor}
\label{affinestratacor}
Consider two matrices $W$ and $W'$ (not necessarily distinct)
such that $\rank W = \rank W'$, and
two points $\theta \in \mu^{-1}(W)$ and $\theta' \in \mu^{-1}(W')$ that
both have the same rank list. 
Then there is a linear bijection from $\R^{d_\theta}$ to $\R^{d_\theta}$ that
maps $\theta$ to $\theta'$, $\mu^{-1}(W)$ to $\mu^{-1}(W')$, and
$S^W_{\underline{z}}$ to $S^{W'}_{\underline{z}}$ for
every stratum $S^W_{\underline{z}}$ in the rank stratification of $\mu^{-1}(W)$.
\end{cor}

\begin{proof}
Let $J_0, J_1, \ldots, J_L$ represent flow bases at $\theta$, and
define $\eta$ as described above:
a linear bijection from $\R^{d_\theta}$ to $\R^{d_\theta}$
that maps $\tilde{\theta}$ to $\theta$ and, by Lemma~\ref{affinestratalem},
maps $\mu^{-1}(\tilde{I})$ to $\mu^{-1}(W)$ and
$S^{\tilde{I}}_{\underline{z}}$ to~$S^W_{\underline{z}}$.

Let $J'_0, J'_1, \ldots, J'_L$ represent flow bases at $\theta'$, and
define $\eta'$ similarly to $\eta$, replacing each $J_j$ with $J'_j$.
Then $\eta'$~is a linear bijection from $\R^{d_\theta}$ to $\R^{d_\theta}$ that maps
$\tilde{\theta}$ to $\theta'$,
$\mu^{-1}(\tilde{I})$ to $\mu^{-1}(W')$, and
$S^{\tilde{I}}_{\underline{z}}$ to~$S^{W'}_{\underline{z}}$.

Hence $\eta' \circ \eta^{-1}$ is
a linear bijection from $\R^{d_\theta}$ to $\R^{d_\theta}$ that maps
$\theta$ to~$\theta'$, $\mu^{-1}(W)$ to $\mu^{-1}(W')$, and
$S^W_{\underline{z}}$ to~$S^{W'}_{\underline{z}}$.
\end{proof}

If we choose $W = W'$,
Corollary~\ref{affinestratacor} describes several automorphisms:
a bijection that maps $\mu^{-1}(W)$ to itself and, for
every stratum $S^W_{\underline{z}}$ in the rank stratification of $\mu^{-1}(W)$,
a bijection that maps $S^W_{\underline{z}}$ to itself.
The following corollary follows by letting $\underline{z}$ be
the rank list $\underline{r}$ of $\theta$.

\begin{cor}
\label{selfsimilar}
For any two points $\theta, \theta'$ in the same stratum $S_{\underline{r}}$,
there is a linear homeomorphism from $S_{\underline{r}}$ to itself that maps
$\theta$ to $\theta'$.
\label{affinestrataself}
\end{cor}

Corollary~\ref{selfsimilar} implies that
the stratum looks topologically the same at every point---more precisely,
given an open neighborhood of one point on the stratum,
every point on the stratum has an open neighborhood homeomorphic to that one.
As a consequence, there are only two possibilities:
every point on $S_{\underline{r}}$ has an open neighborhood homeomorphic to
an open ball of some fixed dimension---hence
$S_{\underline{r}}$ is a manifold---or no point on $S_{\underline{r}}$ has
such a neighborhood.
The latter possibility is ruled out because every nonempty semi-algebraic set
has at least one point with an open neighborhood homeomorphic to a ball.

We give some background on stratifications to justify this claim.
Every semi-algebraic set can be stratified---that is, partitioned into
strata that are manifolds without boundary---in a manner that satisfies
three criteria.
First, every stratum is analytic and thus smooth---specifically,
$C^\infty$-differentiable.
Second, the stratification is {\em locally finite}, meaning that
every point of the semi-algebraic set has
an open neighborhood that intersects only finitely many strata.
Third, the stratification satisfies the frontier condition we defined in
Section~\ref{foretaste}:
for every pair of distinct strata $S$, $T$ in the stratification,
either $S \cap \bar{T} = \emptyset$ or $S \subseteq \bar{T}$.
See Benedetti and Risler~\cite{benedetti90} for a proof.
Here we use this result to stratify each stratum in
the rank stratification---that is,
$S_{\underline{r}}$~has a stratification satisfying the three criteria.

%

Let $Z$ be a semi-algebraic set.
Let $\mathbb{S}_Z$ be a locally finite stratification of~$Z$
satisfying the frontier condition and having analytic strata.
Let $d$ be the maximum dimension among the strata in $\mathbb{S}_Z$.
The {\em dimension} of~$Z$ is $d$.
(This is by definition, though there are competing, equivalent definitions.)
Let $S \in \mathbb{S}_Z$ be a stratum of dimension $d$, and
let $\theta$ be any point on~$S$.
As $S$ is a $d$-manifold,
there is an open neighborhood $N \subset S$ of~$\theta$ homeomorphic to
an open $d$-ball.
The frontier condition implies that no point in $N$ lies in
the closure of any stratum besides~$S$ (as $S$ has the maximum dimension).
The fact that $\mathbb{S}_Z$ is locally finite implies that
no point in $N$ lies in the closure of $Z \setminus S$.
Therefore, $N$ is an open set in $Z$
(which is a stronger statement than it being an open set in $S$).
Hence $\theta$~has
an open neighborhood $N \subset Z$ homeomorphic to a $d$-ball.

This reasoning applies to $Z = S_{\underline{r}}$, so
some point on $S_{\underline{r}}$ has
an open neighborhood homeomorphic to a ball.
Hence Corollary~\ref{selfsimilar} implies that
every point on $S_{\underline{r}}$ has
an open neighborhood homeomorphic to a ball of the same dimension.
We conclude that each stratum $S_{\underline{r}}$ in a rank stratification is
a manifold.

Moreover, as $\theta \in N \subset S \subseteq S_{\underline{r}}$ and
$S$ is an analytic manifold,
the manifold $S_{\underline{r}}$ is locally analytic at $\theta$.
Corollary~\ref{affinestratacor} states that for every point
$\theta' \in S_{\underline{r}}$, there is a linear homeomorphism from
$S_{\underline{r}}$ to itself that maps $\theta$ to $\theta'$, so
$S_{\underline{r}}$ is locally analytic everywhere, hence analytic.
Thus we have proven our main theorem.

\begin{theorem}
\label{stratummanifold}
Each stratum $S_{\underline{r}}$ in the rank stratification of
a fiber $\mu^{-1}(W)$ is an analytic manifold.
\end{theorem}


Unfortunately, this reasoning tells us nothing about
the dimension of that manifold.
Determining that dimension will require a good deal more work,
which we undertake in Sections~\ref{weightprebases} and~\ref{normal}.

\clearpage

\section{Moves in Weight Space}
\label{movesec}

Imagine you are standing at a point $\theta$ on a fiber $\mu^{-1}(W)$.
A {\em move} $(\theta, \theta')$ is a step you take from $\theta$
to another point~$\theta'$, which may or may not be on the fiber.
Let $\Delta \theta = \theta' - \theta$ be the {\em displacement} of the move.
We write
\[
\theta'         =  (W'_L, W'_{L - 1}, \ldots, W'_1) \in \R^{d_\theta}
\hspace{.2in}  \mbox{and}  \hspace{.2in}
\Delta \theta   =  (\Delta W_L, \Delta W_{L - 1}, \ldots, \Delta W_1)
                    \in \R^{d_\theta}.
\]
We use analogous notation for the product $W' = \mu(\theta')$,
its displacement $\Delta W = W' - W$,
the modified subsequence matrices $W'_{t \ssq s} = W'_t W'_{t-1} \cdots W'_{s+1}$,
and their displacements $\Delta W_{t \ssq s} = W'_{t \ssq s} - W_{t \ssq s}$.
All prebasis subspaces below are defined at $\theta$.
(They are different at $\theta'$!)

Moves on the fiber replace a linear neural network with
another that computes the same function.
The latter might be superior in some respects (such as not being near
a spurious critical point of the cost function used to train the network).
Moves on the fiber are also a tool for gaining intuition about
the geometry and topology of the fiber.
In Section~\ref{1hierarchy}, we use moves to understand how
the strata in the rank stratification are connected to each other.
In Section~\ref{weightprebases}, we study the subspace tangent to a stratum
(in weight space) and build a revealing basis for that tangent space;
the moves described here will guide that study.

Two classes of move suffice to characterize strata, their interconnections, and
their connected components:  one-matrix moves and two-matrix moves.
Any move from any point on the fiber to any other can be broken down into
a sequence of one-matrix and two-matrix moves.
\begin{itemize}
\item
A {\em one-matrix move} has
at most one nonzero displacement matrix $\Delta W_j$.
That is, $W'_z = W_z$ for all $z \neq j$.
One-matrix moves are linear in two senses.
First, the displacement $\Delta W$ is linear in the displacement~$\Delta W_j$.
Second, as a consequence,
if a one-matrix move stays on the fiber ($W' = W$), then
the line through $\theta$ and $\theta'$ is a subset of the fiber.
A one-matrix move stays on the fiber if
$W_{L \ssq j} \Delta W_j W_{j-1 \ssq 0} = 0$; otherwise, it moves off the fiber.

A {\em discrete move} is a one-matrix move that changes
the ranks of one or more subsequence matrices.
A~discrete move implies that $\theta'$ has
a different rank list (and a different barcode) than $\theta$.
Some discrete moves stay on the fiber, and some move off of it, but
all discrete moves move off of the stratum---that is,
$\theta'$ does not lie on the same stratum as $\theta$.

A {\em barcode move} is a special type of discrete, one-matrix move
from a stratum to an adjoining higher-dimensional stratum.
Recall that each edge of the stratum dag represents a barcode move;
barcode moves suffice to characterize all the connections among strata.
Barcode moves are also a useful tool for determining
the connected components of the fiber.
(A one-matrix move cannot move
from one connected component of the fiber to another.)
We discuss one-matrix moves further in Section~\ref{1matrix}, and
we study discrete moves in detail in Sections~\ref{effects}--\ref{hierarchy}.


\item
A {\em two-matrix move} has
two nonzero displacement matrices $\Delta W_{j+1}$ and $\Delta W_j$,
which are always consecutive.
That is, \mbox{$W'_z = W_z$} for all $z \not\in \{ j, j + 1 \}$.
We specify a two-matrix move by
selecting an invertible $d_j \times d_j$ matrix $M$ and
setting $W'_{j+1} = W_{j+1} M$ and $W'_j = M^{-1} W_j$.

Clearly, all two-matrix moves stay on the fiber:
$W' = \mu(\theta') = \mu(\theta) = W$.
Moreover, all two-matrix moves stay on the same stratum, as
a two-matrix move does not change the rank of any subsequence matrix.
(There are no discrete two-matrix moves.)
One way to think of this move:  an invertible linear transformation
changes the coordinate system with which layer~$j$ of hidden units represents
information, without otherwise changing anything that the network computes.

Unlike in a one-matrix move,
often there is no straight path on the fiber from $\theta$ to $\theta'$.
But if $M$ has a positive determinant, usually
they are connected by a smooth, curved path on the stratum that contains them.
We call that path a {\em two-matrix path}.
By ``smooth,'' we mean that every point on the path has
a well-defined tangent line.
These tangent lines are also tangent to the stratum, so
they help us identify the stratum's tangent spaces.
Two-matrix paths also help us show
which points on a single stratum are path-connected.

Two-matrix moves where $M$ has a negative determinant provide a way to move
from one connected component of the fiber to another
(in which case there is no path on the fiber connecting $\theta$ to $\theta'$).
Section~\ref{2matrix} discusses two-matrix moves.
\end{itemize}

\subsection{One-Matrix Moves}
\label{1matrix}

In a one-matrix move, we start at a point $\theta$, let $W = \mu(\theta)$,
choose one finite displacement $\Delta W_j$, and
set $\Delta W_z = 0$ for all $z \neq j$.
(A valid choice is $\Delta W_j = 0$, so
our moves include a ``move'' that doesn't move.)
Thus, we move from a point $\theta \in \mu^{-1}(W)$ to
\[
\theta' = (W_L, W_{L-1}, \ldots, W_{j+1}, W'_j, W_{j-1}, \ldots, W_1)
\]
where $W'_j = W_j + \Delta W_j$.
Then
\[
W' = \mu(\theta') = W_{L \ssq j} (W_j + \Delta W_j) W_{j-1 \ssq 0} =
\mu(\theta) + W_{L \ssq j} \Delta W_j W_{j-1 \ssq 0} =
W + W_{L \ssq j} \Delta W_j W_{j-1 \ssq 0}.
\]
Therefore, $\theta'$ lies on the fiber $\mu^{-1}(W)$ if and only if
$\Delta W_j$ satisfies $W_{L \ssq j} \Delta W_j W_{j-1 \ssq 0} = 0$.
The set of displacements that satisfy this identity is the subspace
(of $\R^{d_j \times d_{j-1}}$)
\[
N_j = \Null W_{L \ssq j} \otimes \R^{d_{j-1}} +
      \R^{d_j} \otimes \Null W_{j-1 \ssq 0}^\top.
\]
Here, the symbol ``$\otimes$'' denotes a tensor product.
For subspaces $Y \subseteq \R^t$ and $Z \subseteq \R^s$,
\[
Y \otimes Z = \{ M \in \R^{t \times s} :
                 \col M \subseteq Y  \mbox{~and~}  \row M \subseteq Z \}.
\]
That is, a $t \times s$ matrix $M$ is in $Y \otimes Z$ if
$M$ maps all points in $\R^s$ into $Y$ and
$M^\top$~maps all points in $\R^t$ into $Z$.

The dimension of $N_j$ is
\begin{eqnarray}
\dim N_j
& = & \dim \, (\Null W_{L \ssq j} \otimes \R^{d_{j-1}}) +
      \dim \, (\R^{d_j} \otimes \Null W_{j-1 \ssq 0}^\top) -
      \dim \, (\Null W_{L \ssq j} \otimes \Null W_{j-1 \ssq 0}^\top)
      \nonumber  \\
& = & (d_j - \rank W_{L \ssq j}) \cdot d_{j-1} +
      d_j \cdot (d_{j-1} - \rank W_{j-1 \ssq 0}) -
      (d_j - \rank W_{L \ssq j}) \cdot (d_{j-1} - \rank W_{j-1 \ssq 0})
      \nonumber  \\
& = & d_j d_{j-1} - \rank W_{L \ssq j} \cdot \rank W_{j-1 \ssq 0}.
\label{dimNj}
\end{eqnarray}



The set of all points reachable from $\theta$ by
a one-matrix move with $\Delta W_j \in N_j$,
\[
N^\theta_j =
\{ (W_L, W_{L-1}, \ldots, W_{j+1}, W_j + \Delta W_j, W_{j-1}, \ldots, W_1) :
   \Delta W_j \in N_j \},
\]
is an {\em affine subspace} of $\R^{d_{\theta}}$
(which is like a subspace, but it does not necessarily pass through the origin)
satisfying $N^\theta_j \subseteq \mu^{-1}(W)$.
That is, $N^\theta_j$ is a subset of the fiber
(but not necessarily a subset of any stratum).
This gives us some insight into the geometry of the fiber, but
it tells us nothing about directions along which the fiber is curved.
The curvature is revealed by the two-matrix moves.

\subsection{Two-Matrix Moves and Two-Matrix Paths}
\label{2matrix}

Recall that a {\em two-matrix move} is a move that, for some $j \in [1, L - 1]$,
selects an invertible $d_j \times d_j$ matrix~$M$ and
sets $W'_{j+1} = W_{j+1} M$ and $W'_j = M^{-1} W_j$
(but no other factor matrix changes).
A two-matrix move always places $\theta'$
on the same fiber and the same stratum as $\theta$.
To see that, let $W = \mu(\theta)$ and
let $S$ be the stratum containing $\theta$ in
the rank stratification of the fiber $\mu^{-1}(W)$.
After the move, the subsequence matrices are
\[
W'_{k \ssq i} = \left \{ \begin{array}{ll}
W_{k \ssq j} M,       &  i = j \mbox{~and~} k > i,  \\
M^{-1} W_{j \ssq i},  &  k = j \mbox{~and~} k > i,  \\
W_{k \ssq i}          &  \mbox{otherwise}.
\end{array} \right.
\]
In particular, $W' = W$ (the move stays on the fiber).
Matrix rank is invariant to multiplication by an invertible matrix, so
$\rank W'_{k \ssq i} = \rank W_{k \ssq i}$ for all the subsequence matrices,
$\theta'$~has the same rank list as $\theta$, and $\theta'$~lies on $S$
(the move stays on the stratum).
There are no discrete two-matrix moves.

Unlike a one-matrix move, a two-matrix move doesn't reveal a line on the fiber,
but it can reveal a smooth, curved path on the fiber and on $S$, which tells us
a direction tangent to $S$ at $\theta$.
Consider a two-matrix move where $M = I + \epsilon H$
for an arbitrary, nonzero $d_j \times d_j$ matrix $H$ and a real $\epsilon > 0$.
For a sufficiently small $\epsilon$, $M$ is invertible, so
we can draw a smooth path on the fiber, with endpoint $\theta$,
by varying $\epsilon$ from zero to some small value.
Thus we define the {\em two-matrix path}
\begin{eqnarray}
P & = & \{ \theta'(\epsilon) : \epsilon \in [0, \hat{\epsilon}] \}
\hspace{.2in}  \mbox{where}  \nonumber  \\
\theta'(\epsilon) & = & (W_L, W_{L-1}, \ldots, W_{j+2}, W_{j+1} (I + \epsilon H),
                        (I + \epsilon H)^{-1} W_j, W_{j-1}, \ldots, W_1).
\label{2matrixpath}
\end{eqnarray}

The curved grid lines in Figure~\ref{hyperboloid} are examples of such paths.
Every point on $P$ is the target of some two-matrix move from $\theta$, so
$P \subseteq S$.
To find the line tangent to $P$ at~$\theta$,
observe that for a small $\epsilon$, $(I + \epsilon H)^{-1} =
I - \epsilon H + \epsilon^2 H^2 - \epsilon^3 H^3 + \ldots$, so
\begin{eqnarray*}
\frac{\di}{\di\epsilon} W'_{j+1}
& = & \frac{\di}{\di\epsilon} \left( W_{j+1} (I + \epsilon H) \right) = W_{j+1} H
\hspace{.2in}  \mbox{and}  \hspace{.2in}  \\
\left. \frac{\di}{\di\epsilon} W'_j \right|_{\epsilon = 0}
& = & \left. \frac{\di}{\di\epsilon}
      \left( (I + \epsilon H)^{-1} W_j \right) \right|_{\epsilon = 0}
  =   \left. \frac{\di}{\di\epsilon} \left(
      (I - \epsilon H + \epsilon^2 H^2 - \epsilon^3 H^3 + \ldots) W_j
      \right) \right|_{\epsilon = 0} = - H W_j.
\end{eqnarray*}
Hence $(\di\theta'/\di\epsilon)|_{\epsilon = 0} =
(0, \ldots, 0, W_{j+1} H, - H W_j, 0, \ldots, 0)$.

Let $T_\theta P$ be $P$'s tangent line at $\theta$ and
let $T_\theta S$ be $S$'s tangent space at $\theta$---both defined so that
$T_\theta P$ and $T_\theta S$ pass through the origin,
not necessarily through~$\theta$.
Then
\begin{equation}
\label{TPtheta}
T_\theta P =
\{ (0, 0, \ldots, 0, \underbrace{\gamma W_{j+1} H}_{\Delta W_{j+1}},
   \underbrace{- \gamma H W_j}_{\Delta W_j}, 0, \ldots, 0) :
   \gamma \in \R \}
\end{equation}
and $T_\theta P \subseteq T_\theta S$, because
$P \subseteq S$ and both $P$ and $S$ are smooth at $\theta$.

In Section~\ref{weightprebases}, we build a basis for $T_\theta S$ from
one-matrix and two-matrix tangent directions.

\clearpage

\section{The One-Matrix Prebasis and the Hierarchy of Strata}
\label{1hierarchy}

In Section~\ref{foretaste}, we introduced
a theorem characterizing how the strata are connected to each other;
in this section, we prove it.
For two strata $S_{\underline{r}} \neq \emptyset$ and
$\bar{S}_{\underline{z}} \neq \emptyset$ in a rank stratification,
the following statements are equivalent.
\begin{icompact}
\item[A.]  $S_{\underline{r}} \subseteq \bar{S}_{\underline{z}}$.
\item[B.]  $S_{\underline{r}} \cap \bar{S}_{\underline{z}} \neq \emptyset$.
\item[C.]  $\underline{r} \leq \underline{z}$. 
\item[D.]  There exists a sequence of rank-one abstract barcode moves that
  proceed from $\underline{r}$ to~$\underline{z}$,
  with all the intermediate rank lists being valid.
\end{icompact}
We discussed some implications of these equivalences in Section~\ref{foretaste}.
(Recall that the frontier condition is the claim that B implies A.)
In Sections~\ref{smallmoves}--\ref{abstractmoves} we define  
the terms in statement~D and explain what the stratum dag really means---recall
that its edges represent the barcode moves of statement~D.
There is a partial order on the strata, imposed by
the transitive relationship $\underline{r} \leq \underline{z}$ and also by
the transitive relationship $S_{\underline{r}} \subseteq \bar{S}_{\underline{z}}$.
The stratum dag diagrams that partial order, but
it also encodes some additional information about barcode moves
(see Section~\ref{abstractmoves});
our phrase ``the hierarchy of strata'' encompasses all of that.

Our proof shows the four implications
A $\rightarrow$ B $\rightarrow$ C $\rightarrow$ D $\rightarrow$ A.
Clearly A~implies~B, as we assume $S_{\underline{r}} \neq \emptyset$.
To see that B implies C, observe that
an infinitesimal perturbation of a weight vector $\theta \in S_{\underline{r}}$
may increase the ranks of some subsequence matrices but
it cannot decrease any of their ranks, so
a stratum $S_{\underline{z}}$ can have $\theta$ in its closure only
if~$\underline{r} \leq \underline{z}$.
We flesh out this argument in Section~\ref{smallmoves}.
Showing that D implies A is not difficult,
once the terms in statement~D are defined.
(An abstract move implies arbitrarily small geometric moves).
The fact that C implies D was by far the hardest of
the four implications to prove; for many months we did not know if
C~implies any of the other statements.

While statements A and B are statements about geometry,  
statements C and D are purely combinatorial.  
Our proof that C implies D (Corollary~\ref{dagarrows}) is
also purely combinatorial.
The proof is constructive:  we specify an algorithm that finds
a suitable sequence of barcode moves (thereby showing one exists).

To understand the connections among strata,
we use discrete one-matrix moves from one stratum to another.
To specify the directions of those moves,
we decompose the weight space $\R^{d_\theta}$ into a prebasis,
just as we decomposed each unit layer's space $\R^{d_j}$ into
a prebasis in Section~\ref{basisspaces}.
First we use the prebases from Section~\ref{basisspaces} to decompose
each edge layer's space $\R^{d_j \times d_{j-1}}$ into
a prebasis of subspaces of $\R^{d_j \times d_{j-1}}$; then
we decompose $\R^{d_\theta}$ into a prebasis of subspaces of $\R^{d_\theta}$,
which we call the {\em one-matrix prebasis at $\theta$}.

The one-matrix prebasis helps us to distinguish
moves that stay on the fiber from those that do not, to distinguish 
moves that stay on the stratum from those that do not, to identify
discrete moves, and to classify moves according to which stratum they move to.
Each member of the one-matrix prebasis is a {\em one-matrix subspace}, 
a set of move displacements that have largely the same effect combinatorially.
(This equivalent is not perfect.
Some displacements in a one-matrix subspace may {\em decrease} 
the rank of a subsequence matrix, but most of them do not.)
The one-matrix prebasis and its subspaces are
different at each weight vector~$\theta \in \mu^{-1}(W)$.

Discrete moves suffice to characterize all the interconnections among strata,
as we will see in Section~\ref{hierarchy}.
Sections~\ref{effects}--\ref{1canon} examine the effects of discrete moves
on the rank list, which tells us which stratum the moves move to.
Each directed edge of the stratum dag represents
one or more one-matrix subspaces associated with discrete moves.
All the strata that meet at a point $\theta$ can be accessed from $\theta$
by a sequence of tiny, discrete, one-matrix moves,
each corresponding to an edge in the stratum~dag.

Another purpose of the one-matrix prebasis is to help identify
the space tangent to a stratum~$S$ at a point $\theta \in S$,
denoted $T_\theta S$.
In Section~\ref{1matrixstratum} we determine
which one-matrix subspaces are subspaces of $T_\theta S$, but
those are not all the subspaces we need to span $T_\theta S$.
To account for directions along which $S$ is curved, we will define
some {\em two-matrix subspaces} in Section~\ref{2matrixsubspaces}, and
use them to derive $T_\theta S$ in Section~\ref{expressions}.

\subsection{Small Moves}
\label{smallmoves}

To understand the interconnections among strata,
it is helpful to think of infinitesimal moves, because
an infinitesimal perturbation of a matrix can increase its rank, but 
cannot decrease its rank.

Infinitesimals have a complicated status in the history of mathematical rigor.
To strip away everything that is not essential,
we define a {\em small move} to be any move that does not decrease
the rank of any subsequence matrix.
This definition happens to include a lot of large moves, but
it meets our needs and it fits our intuitions about
which strata an infinitesimal move can reach.

This definition has a counterintuitive consequence:
the inverse of a small move is {\em not} necessarily small.
If a small move is discrete, its inverse is not small.
For example, in Figure~\ref{strat1111},
a move from $S_{010}$ to $S_{011}$ is small, but a move back is not.
As statements~C and~D are equivalent,
a small move that stays on the fiber always ends at a stratum that is
reachable in the stratum dag from the starting stratum.
Small moves can move up the hierarchy, but not down.

The observation that a sufficiently small perturbation of a matrix cannot
decrease its rank gives us a proof that B implies C.

\begin{lemma}
\label{stratatouch}
If two strata in a rank stratification satisfy
$S_{\underline{r}} \cap \bar{S}_{\underline{z}} \neq \emptyset$, then
$\underline{r} \leq \underline{z}$.
\end{lemma}

\begin{proof}
By assumption, there exists
a point $\theta \in S_{\underline{r}} \cap \bar{S}_{\underline{z}}$.
Therefore, every open neighborhood $N \subset \R^{d_\theta}$ of~$\theta$
intersects $S_{\underline{z}}$.
But a sufficiently small perturbation of~$\theta$ cannot reduce
the rank of any subsequence matrix, so
there exists an open neighborhood $N \subset \R^{d_\theta}$ of $\theta$
such that for every weight vector $\theta_N \in N$,
the rank list $\underline{u}$ of $\theta_N$ satisfies
$r_{k \ssq i} \leq u_{k \ssq i}$ for all $L \geq k \geq i \geq 0$.
As $N$ intersects $S_{\underline{z}}$,
there exists a point $\theta' \in N \cap S_{\underline{z}}$.
The rank list of $\theta'$ is $\underline{z}$, so
$r_{k \ssq i} \leq z_{k \ssq i}$ for all $L \geq k \geq i \geq 0$.
\end{proof}

\subsection{The One-Matrix Subspaces and the One-Matrix Prebasis}
\label{1matrixprebasis}

Recall from Section~\ref{basisspaces} that we decompose
each unit layer's space $\R^{d_j}$ into a {\em prebasis}---a ``basis''
made up of subspaces---and in Section~\ref{flowbases}
we decompose it further into a basis of vectors (a more familiar concept).
Here we construct prebases for the factor matrix spaces
$\R^{d_j \times d_{j-1}}$ and the weight space~$\R^{d_\theta}$.
Given a starting point $\theta \in \R^{d_\theta}$,
one of our goals is that our prebases should separate
one-matrix moves that stay on the fiber from those that do not.
Moreover, for every stratum $S$ whose closure contains $\theta$, 
our prebases should separate 
one-matrix moves that stay on $\bar{S}$ from those that do not. 
We saw a hint about how to achieve the former in Section~\ref{1matrix}, where
we expressed the set of displacements $\Delta W_j$ that stay on the fiber as
a subspace, $N_j = \Null W_{L \ssq j} \otimes \R^{d_{j-1}} +
\R^{d_j} \otimes \Null W_{j-1 \ssq 0}^\top$.

First, we construct a prebasis $\mathcal{O}_j$ for $\R^{d_j \times d_{j-1}}$.
Like the $\mathcal{A}_j$'s and $\mathcal{B}_j$'s,
the $\mathcal{O}_j$'s are defined at a specific weight vector $\theta$, and
there is usually some flexibility in choosing them, inherited from
the choice of the subspaces $a_{kji}$ and $b_{kji}$ defined in
Section~\ref{basisspaces} and~\ref{flowrelations}.  
For indices satisfying
$L \geq l \geq j \geq i \geq 0$ and $L \geq k \geq j - 1 \geq h \geq 0$,
define the prebasis subspace
\[
o_{lkjih} = a_{lji} \otimes b_{k,j-1,h} =
\{ M \subseteq \R^{d_j \times d_{j-1}} :
   \col M \subseteq a_{lji}  \mbox{~and~}  \row M \subseteq b_{k,j-1,h} \}.
\]
For our results here, we need flow subspaces
(i.e., subspaces satisfying the flow conditions
$a_{kji} = W_{j \ssq s} a_{ksi}$ and $b_{kji} = W_{t \ssq j}^\top b_{kti}$).
Theorem~\ref{fundnn} guarantees that such flow subspaces exist.

Recall from Sections~\ref{flowbases} and~\ref{basestranspose}
the matrix $J_{lji}$ whose $\omega_{li}$ columns are a basis for $a_{lji}$ and
the matrix $K_{k,j-1,h}$ whose $\omega_{kh}$ columns are
a basis for $b_{k,j-1,h}$.
We can easily construct a basis for $o_{lkjih}$, namely, the set
\[
\{ uv^\top : u \mbox{~is a column of~} J_{lji} \mbox{~and~}
             v \mbox{~is a column of~} K_{k,j-1,h} \}.
\]
This basis is composed of $\omega_{li} \, \omega_{kh}$ rank-$1$ matrices.
Hence another definition of the subspace $o_{lkjih}$ is
\[
o_{lkjih} =
\{ J_{lji} C K_{k,j-1,h}^\top : C \in \R^{\omega_{li} \times \omega_{kh}} \},
\]
where the matrix $C$ holds the coefficients of the basis matrices.
It is clear that
\[
\dim o_{lkjih} = \dim a_{lji} \cdot \dim b_{k,j-1,h} = \omega_{li} \, \omega_{kh}.
\]

For each $j \in [1, L]$, define the prebasis
\[
\mathcal{O}_j = \{ o_{lkjih} \neq \{ 0 \} :
                   l \in [j, L], k \in [j-1, L], i \in [0, j], h \in [0, j-1]
                \}.
\]
$\mathcal{O}_j$ is a prebasis for the weight layer $\R^{d_j \times d_{j-1}}$, as
it pairs every subspace in the prebasis $\mathcal{A}_j$ (spanning $\R^{d_j}$)
with every subspace in the prebasis $\mathcal{B}_{j-1}$ (spanning $\R^{d_{j-1}}$).

\begin{lemma}
\label{Ojprebasis}
$\mathcal{O}_j$ is a prebasis for $\R^{d_j \times d_{j-1}}$.
In particular, the subspaces in $\mathcal{O}_j$ are linearly independent.
\end{lemma}

\begin{proof}
The vector sum of the one-matrix subspaces in $\mathcal{O}_j$ is
\begin{eqnarray*}
\sum_{o_{lkjih} \in \mathcal{O}_j} o_{lkjih} & = &
\sum_{a_{lji} \in \mathcal{A}_j} \sum_{b_{k,j-1,h} \in \mathcal{B}_{j-1}}
  a_{lji} \otimes b_{k,j-1,h} =
\left( \sum_{a_{lji} \in \mathcal{A}_j} a_{lji} \right) \otimes
\left( \sum_{b_{k,j-1,h} \in \mathcal{B}_{j-1}} b_{k,j-1,h} \right)  \\
& = & \R^{d_j} \otimes \R^{d_{j-1}} = \R^{d_j \times d_{j-1}}.
\end{eqnarray*}
(The second line follows from Lemma~\ref{bases}.)
Hence, $\mathcal{O}_j$ spans $\R^{d_j \times d_{j-1}}$,
which has dimension $d_j d_{j-1}$.
The sum of the dimensions of these subspaces is
\[
\sum_{o_{lkjih} \in \mathcal{O}_j} \omega_{li} \, \omega_{kh} =
\sum_{a_{lji} \in \mathcal{A}_j} \sum_{b_{k,j-1,h} \in \mathcal{B}_{j-1}}
  \omega_{li} \, \omega_{kh} =
\left( \sum_{a_{lji} \in \mathcal{A}_j} \omega_{li} \right) \cdot
\left( \sum_{b_{k,j-1,h} \in \mathcal{B}_{j-1}} \omega_{kh} \right) =
d_j d_{j-1}.
\]
As $\R^{d_j \times d_{j-1}}$ is the vector sum of the subspaces and
its dimension equals the sum of the dimensions of the subspaces,
the subspaces are linearly independent.
Hence $\mathcal{O}_j$ is a prebasis.
\end{proof}

Now we construct a prebasis $\ThetaO$ for the weight space $\R^{d_\theta}$ that
we call the {\em one-matrix prebasis}.
The subspaces in $\ThetaO$ are called the {\em one-matrix subspaces} and
have the form
\[
\phi_{lkjih} = \{ (0, \ldots, 0, M, 0, \ldots, 0) : M \in o_{lkjih} \}
\]
with $M$ in position $j$ from the right
(the same position as $W_j$ in $\theta$).
The one-matrix prebasis is
\[
\ThetaO = \{ \phi_{lkjih} \neq \{ {\bf 0} \} :
             L \geq l \geq j \geq i \geq 0 \mbox{~and~}
             L \geq k \geq j - 1 \geq h \geq 0 \}.
\]
It follows from Lemma~\ref{Ojprebasis} that
$\ThetaO$ is a prebasis for $\R^{d_\theta}$.

\begin{cor}
\label{ThetaOprebasis}
$\ThetaO$ is a prebasis for $\R^{d_\theta}$.
In particular, the subspaces in $\ThetaO$ are linearly independent.
\end{cor}

\begin{proof}
For any fixed $j \in [1, L]$, $\ThetaO$ contains a subspace
$\phi_{lkjih} = (0, \ldots, 0, o_{lkjih}, 0, \ldots 0)$
for every choice of the four indices
$l \in [j, L]$, $k \in [j-1, L]$, $i \in [0, j]$, and $h \in [0, j-1]$, where
$o_{lkjih}$ occurs at position~$j$ in the weight vector.
The set $\mathcal{O}_j$ contains one subspace $o_{lkjih}$ for every
$l \in [j, L]$, $k \in [j-1, L]$, $i \in [0, j]$, and $h \in [0, j-1]$.
By Lemma~\ref{Ojprebasis},
$\mathcal{O}_j$ is a prebasis for $\R^{d_j \times d_{j-1}}$.
Hence, for a fixed $j$ and varying $l$, $k$, $i$, and $h$,
$\sum_{\phi_{lkjih} \in \ThetaO} \phi_{lkjih} =
(0, \ldots, 0, \R^{d_j \times d_{j-1}}, 0, \ldots 0)$.
If we sum over $j \in [1, L]$ as well, we have
$\sum_{\phi_{lkjih} \in \ThetaO} \phi_{lkjih} = \R^{d_\theta}$.
The sum of the dimensions of the subspaces in $\ThetaO$ is
$\sum_{j=1}^L d_j d_{j-1} = d_\theta$, matching the dimension of $\R^{d_\theta}$, so
the subspaces in $\ThetaO$ are linearly independent.
Therefore, $\ThetaO$ is a prebasis for $\R^{d_\theta}$.
\end{proof}

\subsection{Basic One-Matrix Moves and their Effects on
            the Subsequence Matrices}
\label{effects}

Recall that a move is {\em discrete} if it changes
the rank of some subsequence matrix.
Following a discrete move,
$\theta'$ has a different rank list than $\theta$ (by definition),
$\theta'$ has a different barcode than $\theta$, and crucially,
$\theta'$ is in a different stratum than $\theta$,
usually of a different dimension.

In this section, we study {\em basic one-matrix moves}, for which
$\Delta \theta \in \phi_{lkjih}$ for some $\phi_{lkjih} \in \ThetaO$.
(``Basic'' means ``from the prebasis,'' not ``simple.'')
Equivalently, a basic move replaces $W_j$ with $W'_j = W_j + \Delta W_j$, where
$\Delta W_j \in o_{lkjih}$ and $o_{lkjih} \in \mathcal{O}_j$.
Once we understand basic moves,
other moves can be understood as sequences of basic moves.
We are particularly interested in basic moves that change the rank list and
move from one stratum to another.

\begin{figure}
\centerline{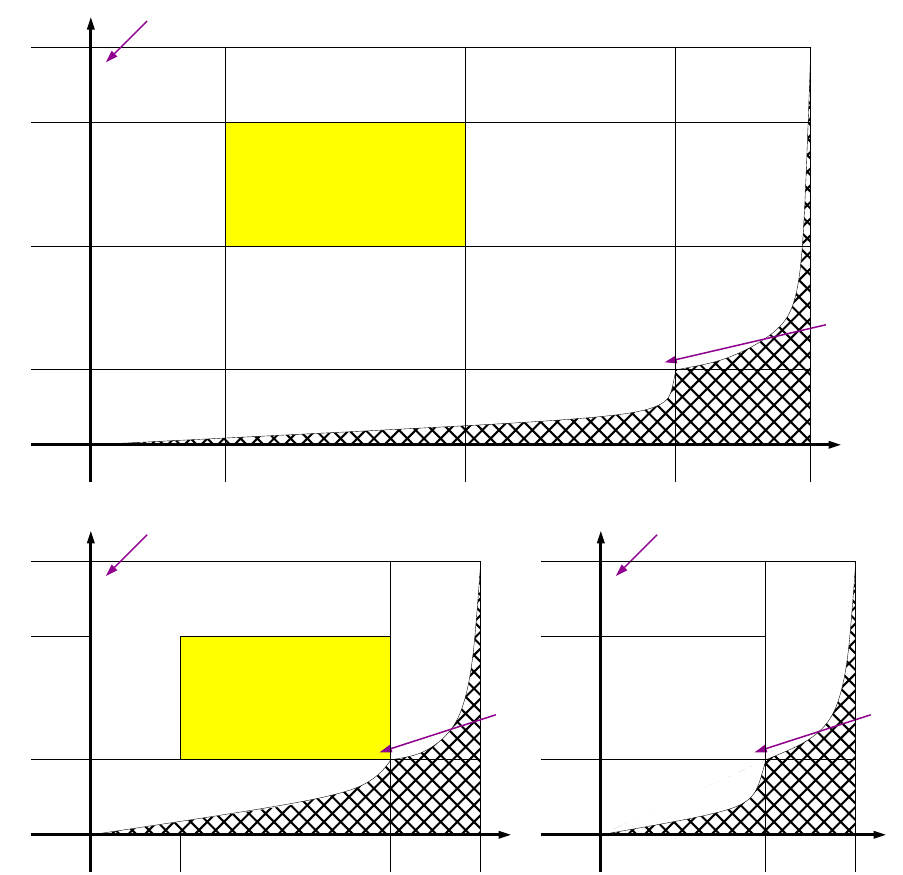}

\caption{\label{matchange}
The influence of a basic move in which a factor matrix $W_j$ undergoes
a nonzero displacement $\Delta W_j \in o_{lkjih}$.
The effects on the subsequence matrix $W_{t \ssq s}$ are listed
for every $t$ and~$s$ with $L \geq t \geq s \geq 0$.
These tables are triangular, though it's not obvious at first:
the hatched regions represent unused zones where $t < s$.
A yellow rectangle indicates which subsequence matrices increase in rank,
constituting a barcode move if no rank decreases.
The black font indicates where $W'_{t \ssq s} \neq W_{t \ssq s}$.
The red font indicates where $W'_{t \ssq s} = W_{t \ssq s}$ because
the matrix $W_j$ is not a factor in~$W_{t \ssq s}$.
The blue font indicates where $W'_{t \ssq s} = W_{t \ssq s}$ by
properties of the flow relationships.
(a)~Table for the case where $L > l > k \geq j > i > h > 0$.
An example of a swapping move (if no rank decreases).
(b)~The third row disappears if $k + 1 = j$, and
the third column disappears if $i = j$.
When $k + 1 = j = i$, the move is a connecting move.
(c)~The second row disappears if $k \geq l$, and
the second column disappears if $i \leq h$.
If either inequality holds, no subsequence matrix rank increases;
the move is not a barcode move.
The first column disappears if $h = 0$.
(The first row disappears if $l = L$, though we don't depict that case here.
If $h = 0$ and $l = L$, then $W' \neq W$ and we move off the fiber.)
}
\end{figure}

For each subspace $o_{lkjih} \in \mathcal{O}_j$, we ask:
does a move with displacement $\Delta W_j \in o_{lkjih}$ change $W$?
Which subsequence matrices change?
Which subsequence matrices change rank?
Which subsequence matrices undergo a change in rowspace or columnspace?
The three lemmas below answer these questions, and
Figure~\ref{matchange} summarizes the answers.
(Figure~\ref{matchange} assumes the displacement $\Delta W_j$ is nonzero;
the lemmas do not.)
These answers explain why we chose subspaces of the form
$o_{lkjih} = a_{lji} \otimes b_{k,j-1,h}$.


\begin{lemma}
\label{Wsame}
Given $L \geq t \geq s \geq 0$,
$W'_{t \ssq s} = W_{t \ssq s}$ if and only if
$\Delta W_j = 0$ or $j \not\in [s + 1, t]$ or $t > l$ or $s < h$.
Moreover, if none of those four conditions holds, then
$\rank \Delta W_{t \ssq s} = \rank \Delta W_j$.
\end{lemma}

\begin{proof}
If $j \not\in [s + 1, t]$, then $W'_j$ is not one of the matrices constituting
$W'_{t \ssq s}$, so $W'_{t \ssq s} = W_{t \ssq s}$ as claimed.

Otherwise, $\Delta W_{t \ssq s} = W_{t \ssq j} \Delta W_j W_{j-1 \ssq s}$.
Observe that
$\col \Delta W_j \subseteq a_{lji} \subseteq A_{lji} \subseteq
\Null W_{l+1 \ssq j}$ and
$\row \Delta W_j \subseteq b_{k,j-1,h} \subseteq B_{k,j-1,h} \subseteq
\Null W^\top_{j-1 \ssq h-1}$.
Therefore, if $t > l$ then $W_{t \ssq j} \Delta W_j = 0$;
symmetrically, if $s < h$ then $\Delta W_j W_{j-1 \ssq s} = 0$.
Thus if $t > l$ or $s < h$ or $\Delta W_j = 0$, then
$\Delta W_{t \ssq s} = 0$ and $W'_{t \ssq s} = W_{t \ssq s}$.

Now consider the case where none of the four conditions holds---the case where
$\Delta W_j \neq 0$ and $j \in [s + 1, t]$ and $t \leq l$ and $s \geq h$.
By Lemma~\ref{dimpreserved},
$W_{t \ssq j} a_{lji}$ has the same dimension as $a_{lji}$.
Thus the linear transformation $v \mapsto W_{t \ssq j} v$ is
a bijection from $a_{lji}$ to $W_{t \ssq j} a_{lji}$, which implies that
$W_{t \ssq j} \col \Delta W_j$ has the same dimension as $\col \Delta W_j$
(as $\col \Delta W_j \subseteq a_{lji}$).
Symmetrically,
$W_{j-1 \ssq s}^\top b_{k,j-1,h}$ has the same dimension as $b_{k,j-1,h}$
by Lemma~\ref{dimpreserved}, thus
$W_{j-1 \ssq s}^\top \row \Delta W_j$ has the same dimension as $\row \Delta W_j$
(as $\row \Delta W_j \subseteq b_{k,j-1,h}$).
Hence
\begin{eqnarray*}
\rank (W_{t \ssq j} \Delta W_j) & = & \dim \col (W_{t \ssq j} \Delta W_j) =
\dim (W_{t \ssq j} \, \col \Delta W_j) = \dim \col \Delta W_j = \rank \Delta W_j
\hspace{.2in}  \mbox{and}  \\
\rank (\Delta W_j W_{j-1 \ssq s}) & = & \dim \row (\Delta W_j W_{j-1 \ssq s}) =
\dim (W_{j-1 \ssq s}^\top \, \row \Delta W_j) = \dim \row \Delta W_j =
\rank \Delta W_j.
\end{eqnarray*}
By the Frobenius rank inequality,
\[
\rank (W_{t \ssq j} \Delta W_j W_{j-1 \ssq s}) \geq
\rank (W_{t \ssq j} \Delta W_j) + \rank (\Delta W_j W_{j-1 \ssq s})
- \rank \Delta W_j = \rank \Delta W_j.
\]
The rank of $W_{t \ssq j} \Delta W_j W_{j-1 \ssq s}$ cannot exceed
the rank of $\Delta W_j$, so
$\rank \Delta W_{t \ssq s} = \rank \Delta W_j$ as claimed.
By assumption, $\Delta W_j \neq 0$, so $\Delta W_{t \ssq s} \neq 0$ and
$W'_{t \ssq s} \neq W_{t \ssq s}$ as claimed.
\end{proof}



The next lemma identifies which subsequence matrices do or do not have
new vectors appear in their row\-spaces or columnspaces.

\begin{lemma}
\label{colrow}
Given $L \geq t \geq s \geq 0$,
$\col \Delta W_{t \ssq s} \subseteq \col W_{t \ssq s}$
(equivalently, $\col W'_{t \ssq s} \subseteq \col W_{t \ssq s}$) if and only if
$\Delta W_j = 0$ or $j \not\in [s+1, t]$ or $t > l$ or $s < h$ or $s \geq i$.
Moreover, if none of those five conditions holds, then
$\col \Delta W_{t \ssq s} \cap \col W_{t \ssq s} = \{ {\bf 0} \}$.

Symmetrically,
$\row \Delta W_{t \ssq s} \subseteq \row W_{t \ssq s}$
(equivalently,  $\row W'_{t \ssq s} \subseteq \row W_{t \ssq s}$) if and only if
$\Delta W_j = 0$ or $j \not\in [s+1, t]$ or $t > l$ or $s < h$ or $t \leq k$.
Moreover, if none of those five conditions holds, then
$\row \Delta W_{t \ssq s} \cap \row W_{t \ssq s} = \{ {\bf 0} \}$.
\end{lemma}

\begin{proof}
If $\Delta W_j = 0$ or $j \not\in [s + 1, t]$ or $t > l$ or $s < h$, then
$\Delta W_{t \ssq s} = 0$ by Lemma~\ref{Wsame} and the results follow.

So assume that
$\Delta W_j \neq 0$ and $j \in [s + 1, t]$ and $t \leq l$ and $s \geq h$.
By Lemma~\ref{Wsame}, $\rank \Delta W_{t \ssq s} = \rank \Delta W_j \neq 0$.
Therefore, $\col \Delta W_{t \ssq s} \neq \{ {\bf 0} \}$ and
$\row \Delta W_{t \ssq s} \neq \{ {\bf 0} \}$.

Recall that $\Delta W_{t \ssq s} = W_{t \ssq j} \Delta W_j W_{j-1 \ssq s}$
and $\col \Delta W_j \subseteq a_{lji}$.
Thus
\begin{equation}
\col \Delta W_{t \ssq s} \subseteq \col (W_{t \ssq j} \Delta W_j) =
W_{t \ssq j} \, \col \Delta W_j \subseteq W_{t \ssq j} a_{lji} \subseteq
W_{t \ssq j} A_{lji} \subseteq W_{t \ssq j} \col W_{j \ssq i} =
\col W_{t \ssq i}.
\label{colincol}
\end{equation}
If $s \geq i$ then $\col W_{t \ssq i} \subseteq \col W_{t \ssq s}$, so
$\col \Delta W_{t \ssq s} \subseteq \col W_{t \ssq s}$ as claimed.

Symmetrically, as $\row \Delta W_j \subseteq b_{k,j-1,h}$,
\begin{align}
& \row \Delta W_{t \ssq s} \subseteq \row (\Delta W_j W_{j-1 \ssq s}) =
W_{j-1 \ssq s}^\top \, \row \Delta W_j \subseteq
W_{j-1 \ssq s}^\top b_{k,j-1,h} \subseteq
W_{j-1 \ssq s}^\top B_{k,j-1,h}  \nonumber  \\
& \subseteq W_{j-1 \ssq s}^\top \row W_{k \ssq j-1} = \row W_{k \ssq s}.
\label{rowinrow}
\end{align}
If $t \leq k$ then $\row W_{k \ssq s} \subseteq \row W_{t \ssq s}$, so
$\row \Delta W_{t \ssq s} \subseteq \row W_{t \ssq s}$ as claimed.

It remains to consider the cases where $s < i$ or $t > k$;
consider the former first.
Recall that
$\col W_{j \ssq s} = A_{Ljs} = \sum_{z = j}^L \sum_{y = 0}^s a_{zjy}$.
If $s < i$, this summation does not include the term $a_{lji}$; so
by Lemma~\ref{bases},
$a_{lji}$ is linearly independent of the subspaces in the summation
(because all these subspaces are members of the prebasis~$\mathcal{A}_j$),
hence $a_{lji} \cap \col W_{j \ssq s} = \{ {\bf 0} \}$.
Premultiplying both sides by $W_{t \ssq j}$ gives
$W_{t \ssq j} a_{lji} \cap \col W_{t \ssq s} = \{ {\bf 0} \}$.
By~(\ref{colincol}), $\col \Delta W_{t \ssq s} \subseteq W_{t \ssq j} a_{lji}$.
We conclude that if $s < i$,
$\col \Delta W_{t \ssq s} \cap \col W_{t \ssq s} = \{ {\bf 0} \}$ and
$\col \Delta W_{t \ssq s} \not\subseteq \col W_{t \ssq s}$ as claimed.

Symmetrically,
$\row W_{t \ssq j-1} = B_{t,j-1,0} = \sum_{z=t}^L \sum_{y=0}^{j-1} b_{z,j-1,y}$.
If $t > k$, this summation does not include the term~$b_{k,j-1,h}$; so
by Lemma~\ref{bases},
$b_{k,j-1,h}$ is linearly independent of the subspaces in the summation
(because all these subspaces are members of the prebasis $\mathcal{B}_{j-1}$),
hence $b_{k,j-1,h} \cap \row W_{t \ssq j-1} = \{ {\bf 0} \}$.
Premultiplying both sides by $W_{j-1 \ssq s}^\top$ gives
$W_{j-1 \ssq s}^\top b_{k,j-1,h} \cap \row W_{t \ssq s} = \{ {\bf 0} \}$.
By~(\ref{rowinrow}),
$\row \Delta W_{t \ssq s} \subseteq W_{j-1 \ssq s}^\top b_{k,j-1,h}$.
We conclude that if $t > k$,
$\row \Delta W_{t \ssq s} \cap \row W_{t \ssq s} = \{ {\bf 0} \}$ and
$\row \Delta W_{t \ssq s} \not\subseteq \row W_{t \ssq s}$ as claimed.
\end{proof}

The next lemma addresses the crucial question of which basic moves can change
the rank of a subsequence matrix---that is, which basic moves are discrete.

\begin{lemma}
\label{rankincrease}
%
Given $L \geq t \geq s \geq 0$, if $t \in [j, k]$ and $s \in [i, j - 1]$, then
$\rank W'_{t \ssq s} \leq \rank W_{t \ssq s}$ and
if $\| \Delta W_j \|$ is sufficiently small (in any matrix norm),
$\rank W'_{t \ssq s} = \rank W_{t \ssq s}$.

If $\Delta W_j \neq 0$ and $t \in [k + 1, l]$ and $s \in [h, i - 1]$, then
$\rank W'_{t \ssq s} = \rank W_{t \ssq s} + \rank \Delta W_j$. 

In all other cases,  $\rank W'_{t \ssq s} = \rank W_{t \ssq s}$.
\end{lemma}

\begin{proof}
By Lemma~\ref{Wsame},
if $\Delta W_j = 0$ or $j \not\in [s + 1, t]$ or $t > l$ or $s < h$, then 
$W'_{t \ssq s} = W_{t \ssq s}$ and thus $\rank W'_{t \ssq s} = \rank W_{t \ssq s}$.
Henceforth, we assume the converse:
$\Delta W_j \neq 0$ and $t \in [j, l]$ and $s \in [h, j - 1]$,
which limits us to the middle four squares in Figure~\ref{matchange}(a);
we will consider the four cases corresponding to those four squares.
By Lemma~\ref{Wsame}, $\rank \Delta W_{t \ssq s} = \rank \Delta W_j$.

If $t \leq k$ (corresponding to the bottom two of the four middle squares), then
$\row W'_{t \ssq s} \subseteq \row W_{t \ssq s}$ by Lemma~\ref{colrow}, so
$\rank W'_{t \ssq s} \leq \rank W_{t \ssq s}$; whereas  
if $t > k$ (corresponding to the top two of the four middle squares), then 
$\row \Delta W_{t \ssq s} \cap \row W_{t \ssq s} = \{ {\bf 0} \}$ 
by Lemma~\ref{colrow}, so $\rank W'_{t \ssq s} \geq \rank W_{t \ssq s}$. 

Symmetrically, if $s \geq i$
(corresponding to the right two of the four middle squares), then 
$\col W'_{t \ssq s} \subseteq \col W_{t \ssq s}$ by Lemma~\ref{colrow}, and
again $\rank W'_{t \ssq s} \leq \rank W_{t \ssq s}$; whereas   
if $s < i$ (corresponding to the left two of the four middle squares), then 
$\col \Delta W_{t \ssq s} \cap \col W_{t \ssq s} = \{ {\bf 0} \}$ 
by Lemma~\ref{colrow}, so $\rank W'_{t \ssq s} \geq \rank W_{t \ssq s}$. 

Hence, if $t \leq k$ and $s \geq i$
(corresponding to the bottom right of the four middle squares), then 
$\rank W'_{t \ssq s} \leq \rank W_{t \ssq s}$.
Moreover, if $\| W_j \|$ is sufficiently small (in any matrix norm), then
$\rank W'_{t \ssq s} = \rank W_{t \ssq s}$, as 
decreasing the rank of a matrix requires some finite displacement.
This confirms the lemma's first claim. 

If $t > k$ and $s < i$
(corresponding to the top left of the four middle squares), then 
$\row \Delta W_{t \ssq s} \cap \row W_{t \ssq s} = \{ {\bf 0} \}$ and
$\col \Delta W_{t \ssq s} \cap \col W_{t \ssq s} = \{ {\bf 0} \}$, so
$\rank W'_{t \ssq s} = \rank W_{t \ssq s} + \rank \Delta W_{t \ssq s}
= \rank W_{t \ssq s} + \rank \Delta W_j$. 
This confirms the second claim. 

In the remaining two cases
(corresponding to the top right and bottom left of the four middle squares),
we have $\rank W'_{t \ssq s} \leq \rank W_{t \ssq s}$ and
$\rank W'_{t \ssq s} \geq \rank W_{t \ssq s}$,
confirming the final claim.
\end{proof}

\subsection{Moves on and off the Fiber}
\label{onofffiber}

As $W = W_{L \ssq 0}$, Lemma~\ref{Wsame} tells us that a basic move moves off
the fiber $\mu^{-1}(W)$ if and only if $\Delta W_j$ is
a nonzero displacement from a subspace $o_{lkjih}$ with $l = L$ and $h = 0$.
This motivates us to partition $\mathcal{O}_j$ into two sets,
$\mathcal{O}^\off_j$~and~$\mathcal{O}^\fiber_j$, and
to partition $\ThetaO$ into two sets, $\ThetaO^\off$~and~$\ThetaO^\fiber$.
\begin{eqnarray*}
\mathcal{O}^\off_j & = &
  \{ o_{lkjih} \in \mathcal{O}_j : l = L  \mbox{~and~}  h = 0 \} =
  \{ o_{Lkji0} \neq \{ 0 \} :  k \in [j-1, L], i \in [0, j] \},  \\
\ThetaO^\off & = & \{ \phi_{lkjih} \in \ThetaO : l = L \mbox{~and~} h = 0 \}
                  = \{ \phi_{Lkji0} \neq \{ {\bf 0} \} :
                       L \geq j \geq 1, L \geq k \geq j - 1, \mbox{~and~}
                       j \geq i \geq 0 \},  \\
\mathcal{O}^\fiber_j & = & \mathcal{O}_j \setminus \mathcal{O}^\off_j =
  \{ o_{lkjih} \in \mathcal{O}_j : L > l  \mbox{~or~}  h > 0 \},
  \hspace{.1in}  \mbox{and}  \\
\ThetaO^\fiber & = & \ThetaO \setminus \ThetaO^\off
                 =   \{ \phi_{lkjih} \in \ThetaO : L > l \mbox{~or~} h > 0 \}.
\end{eqnarray*}
Observe that each $\mathcal{O}^\off_j$ and each $\mathcal{O}^\fiber_j$ is defined
for a fixed value of $j \in [1, L]$, whereas
$\ThetaO^\off$ and $\ThetaO^\fiber$ collect all values of $j$.
To give an example, in the two-matrix case ($L = 2$),
\begin{eqnarray*}
\mathcal{O}^\off_1 & = & \{ o_{20100}, o_{20110}, o_{21100}, o_{21110},
                            o_{22100}, o_{22110} \} \setminus \{ \{ 0 \} \},  \\
\mathcal{O}^\off_2 & = & \{ o_{21200}, o_{21210}, o_{21220}, o_{22200},
                            o_{22210}, o_{22220} \} \setminus \{ \{ 0 \} \},  \\
\ThetaO^\off   & = & \{ \phi_{20100}, \phi_{20110}, \phi_{21100}, \phi_{21110},
                        \phi_{22100}, \phi_{22110},
                        \phi_{21200}, \phi_{21210}, \phi_{21220}, \phi_{22200},
                        \phi_{22210}, \phi_{22220} \}
                        \setminus \{ \{ {\bf 0} \} \},  \\
\mathcal{O}^\fiber_1 & = & \{ o_{10100}, o_{10110}, o_{11100}, o_{11110},
                            o_{12100}, o_{12110} \} \setminus \{ \{ 0 \} \},  \\
\mathcal{O}^\fiber_2 & = & \{ o_{21201}, o_{21211}, o_{21221}, o_{22201},
                            o_{22211}, o_{22221} \} \setminus \{ \{ 0 \} \},  \\
\ThetaO^\fiber & = & \{ \phi_{10100}, \phi_{10110}, \phi_{11100}, \phi_{11110},
                        \phi_{12100}, \phi_{12110},
                        \phi_{21201}, \phi_{21211}, \phi_{21221}, \phi_{22201},
                        \phi_{22211}, \phi_{22221} \}
                        \setminus \{ \{ {\bf 0} \} \}.
\end{eqnarray*}


$\ThetaO^\fiber$ contains the one-matrix subspaces that ``stay on the fiber,''
whereas
$\ThetaO^\off$ contains the one-matrix subspaces that ``move off the fiber,''
in the following sense.

\begin{cor}
\label{ofiber}
For every subspace $\phi_{lkjih} \in \ThetaO^\fiber$ and
every displacement $\Delta \theta \in \phi_{lkjih}$,
$\mu(\theta + \Delta \theta) = W$.
For every subspace $\phi_{lkjih} \in \ThetaO^\off$ and
every displacement $\Delta \theta \in \phi_{lkjih} \setminus \{ {\bf 0} \}$,
$\mu(\theta + \Delta \theta) \neq W$.
\end{cor}

\begin{proof}
As $W = W_{L \ssq 0}$, both statements follow directly from Lemma~\ref{Wsame}.
\end{proof}

%

Corollary~\ref{ofiber} gives us a hint about the geometry of the fiber.
Given a subspace $\phi_{lkjih} \in \ThetaO$,
let $\theta + \phi_{lkjih}$ denote the {\em affine subspace} produced
by translating $\phi_{lkjih}$ to pass through $\theta$.
Formally,
\[
\theta + \phi_{lkjih} =
\{ \theta + \Delta \theta : \Delta \theta \in \phi_{lkjih} \}.
\]
Corollary~\ref{ofiber} says that
for any subspace $\phi_{lkjih} \in \ThetaO^\fiber$,
$\theta + \phi_{lkjih} \subseteq \mu^{-1}(W)$---that is,
the translated subspace is a subset of the fiber---whereas
for any subspace $\phi_{lkjih} \in \ThetaO^\off$,
$\theta + \phi_{lkjih}$ intersects $\mu^{-1}(W)$ at only one point,~$\theta$.

Related to Corollary~\ref{ofiber} is the fact that
$\mathcal{O}^\fiber_j$ is a prebasis for $N_j$,
the subspace of displacements $\Delta W_j$ that satisfy
$W_{L \ssq j} \Delta W_j W_{j-1 \ssq 0} = 0$, defined in Section~\ref{1matrix}.
We will prove that in Section~\ref{diffmap}, Lemma~\ref{Ofiberjprebasis}.

\subsection{Barcode Moves:  Walking the Hierarchy of Strata}
\label{barcmoves}

Recall that a basic move replaces $\theta$ with
$\theta' = \theta + \Delta \theta$, where
$\Delta \theta \in \phi_{lkjih}$ and $\phi_{lkjih} \in \ThetaO$.
Equivalently, it replaces $W_j$ with $W'_j = W_j + \Delta W_j$, where
$\Delta W_j \in o_{lkjih}$ and $o_{lkjih} \in \mathcal{O}_j$.
Recall that a discrete move is a move that changes
the ranks of one or more subsequence matrices.

Lemma~\ref{rankincrease} and Figure~\ref{matchange} tell us
two ways a basic move can be discrete.
Every subsequence matrix $W_{t \ssq s}$ 
with $t \in [k + 1, l]$ and $s \in [h, i - 1]$ increases in rank, and
their ranks all increase by the same amount:  the rank of~$\Delta W_j$.
A subsequence matrix $W_{t \ssq s}$
with $t \in [j, k]$ and $s \in [i, j - 1]$ {\em might} decrease in rank, but
not if $\| \Delta W_j \|$ is sufficiently small.
If a basic move is small (i.e., no subsequence rank decreases),
we can ignore the latter case.
Hence, a small, basic move with displacement $\Delta \theta \in \phi_{lkjih}$ is
discrete if and only if $\Delta \theta \neq {\bf 0}$, $l > k$, and $i > h$
(because then $[k + 1, l] \neq \emptyset$ and $[h, i - 1] \neq \emptyset$). 

We define a set of one-matrix subspaces that signify basic moves that increase
the rank of some subsequence matrix.
\[
\ThetaO^\barc  = \{ \phi_{lkjih} \in \ThetaO : l > k \mbox{~and~} i > h \}
               = \{ \phi_{lkjih} \neq \{ {\bf 0} \} :
                     L \geq l \geq k + 1 \geq j \geq i > h \geq 0 \}.
\]
                         
To understand how the strata are connected to each other in
the rank stratification of a fiber,
it suffices to study small moves with displacements from subspaces in
$\ThetaO^\fiber \cap \ThetaO^\barc$---moves that stay on the fiber but
change the stratum.
Given a subspace $\phi_{lkjih} \in \ThetaO^\fiber \cap \ThetaO^\barc$,
by Corollary~\ref{ofiber}, $\theta + \phi_{lkjih} \subseteq \mu^{-1}(W)$, but
by Lemma~\ref{rankincrease}, no point in $\theta + \phi_{lkjih}$
except $\theta$ lies on the stratum $S$ that contains~$\theta$.\footnote{
By contrast, for
a subspace $\phi_{lkjih} \in \ThetaO^\fiber \setminus \ThetaO^\barc$,
$\theta + \phi_{lkjih} \subseteq \bar{S}$
(see Lemma~\ref{1strat} in Section~\ref{1matrixstratum}).
}
These facts give us some hints about the connections among the strata and
the shape of the stratification (albeit without addressing its curvature).

Interestingly, although a small, basic move may change the ranks of
many subsequence matrices, at most four interval counts change.
Recall the identity~(\ref{omegarank}):
$\omega_{ts} = \rank W_{t \ssq s} - \rank W_{t \ssq s-1} -
\rank W_{t+1 \ssq s} + \rank W_{t+1 \ssq s-1}$.
If all four ranks increase by $\rank \Delta W_j$, or
exactly two ranks with opposite signs do, then
$\omega_{ts}$ does not change.

For a small displacement $\Delta \theta \in \phi_{lkjih} \in \ThetaO^\barc$,
it is straightforward to check that
$\omega_{kh}$ and $\omega_{li}$ decrease by $\rank \Delta W_j$,
$\omega_{lh}$ and $\omega_{ki}$ increase by $\rank \Delta W_j$, and
no other interval count changes.
The barcode encodes the changes made by small, basic, discrete moves
more elegantly than the rank list does.
(Recall that the rank list is a {\em summed-area table}~\cite{crow84} of
the interval counts.)

\begin{figure}
\centerline{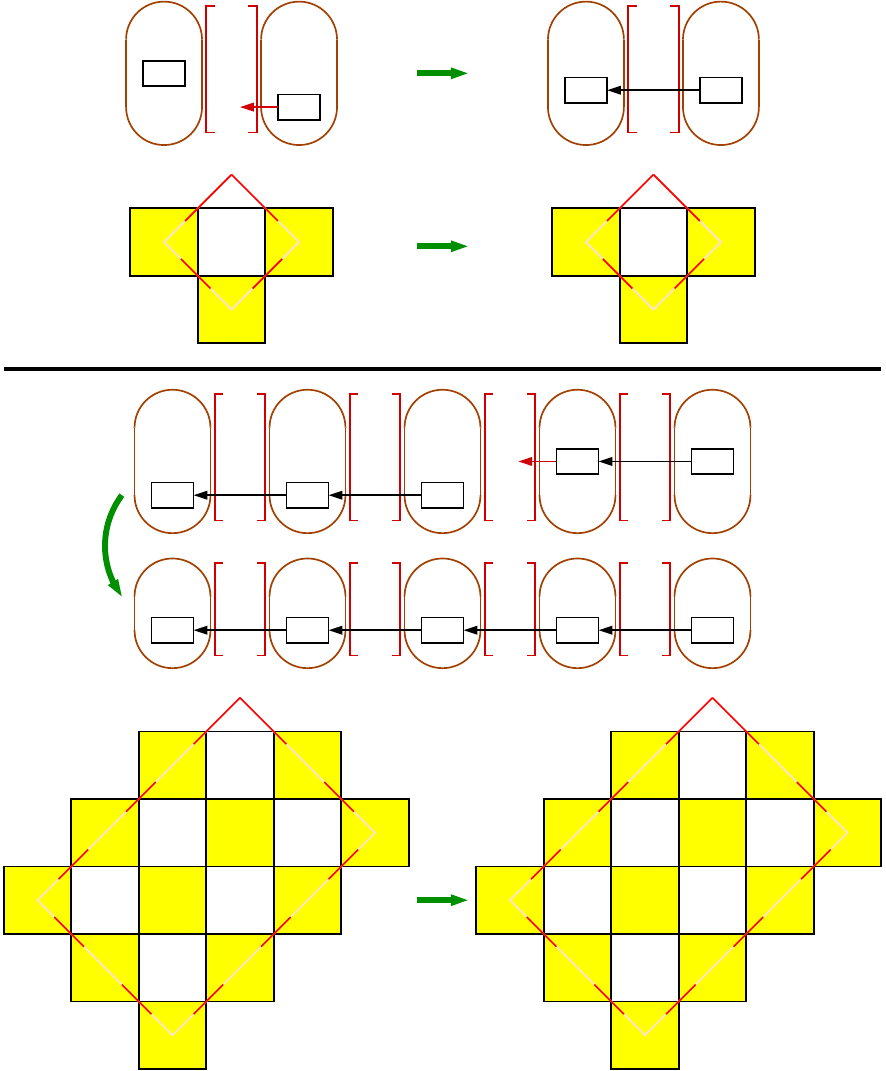}

\caption{\label{connecting}
Two examples of connecting moves.
The top example is the simplest example possible:
$W_1$ has been perturbed to increase its rank from zero to one.
In the bottom example, $W_2$ has been perturbed.
In both examples, the perturbation of $W_j$ replaces
two intervals $[h, j - 1]$ and $[j, l]$ with
a single interval~$[h, l]$.
Three interval counts change,
at three of the four corners of the red rectangle:
$\omega_{lj}$ and $\omega_{j-1,h}$ decrease by one, and
$\omega_{lh}$ increases by one.
The ranks of the subsequence matrices $W_{t \ssq s}$ increase by one
for all $t \in [j, l]$ and $s \in [h, j - 1]$
(the ranks inside the red rectangle, including~$\rank W_j$).
Outside the red rectangle, all interval counts and matrix ranks are unchanged.
}
\end{figure}

A {\em barcode move} is a small, basic, discrete move.
There are two types of barcode moves:  connecting moves and swapping moves.
Every subspace $\phi_{lkjih} \in \ThetaO$ has
indices satisfying $k + 1 \geq j \geq i$.
A~{\em connecting move} is a small move with a nonzero displacement
$\Delta \theta \in \phi_{lkjih} \in \ThetaO^\barc$
in the case where $k + 1 = j = i$.
In a connecting move, $\omega_{ki}$ does not exist (as $k < i$) and
only three interval counts change.
Figure~\ref{connecting} illustrates two examples of connecting moves and
offers an interpretation in terms of a changing barcode:
a connecting move deletes $\rank \Delta W_j$ copies of the interval~$[h, k]$ and
$\rank \Delta W_j$ copies of the interval $[i, l]$, and replaces them
with $\rank \Delta W_j$ copies of the interval $[h, l]$.
We think of this as connecting the intervals $[h, k]$ and~$[i, l]$ together
with an added link $[j - 1, j] = [k, i]$ to create an interval~$[h, l]$;
hence the name ``connecting move.''
(There is much intuition that can be gleaned from a careful study of the figure
that is hard to explain in words.)
The {\em rank} of a connecting move is $\rank \Delta W_j$.

\begin{figure}
\centerline{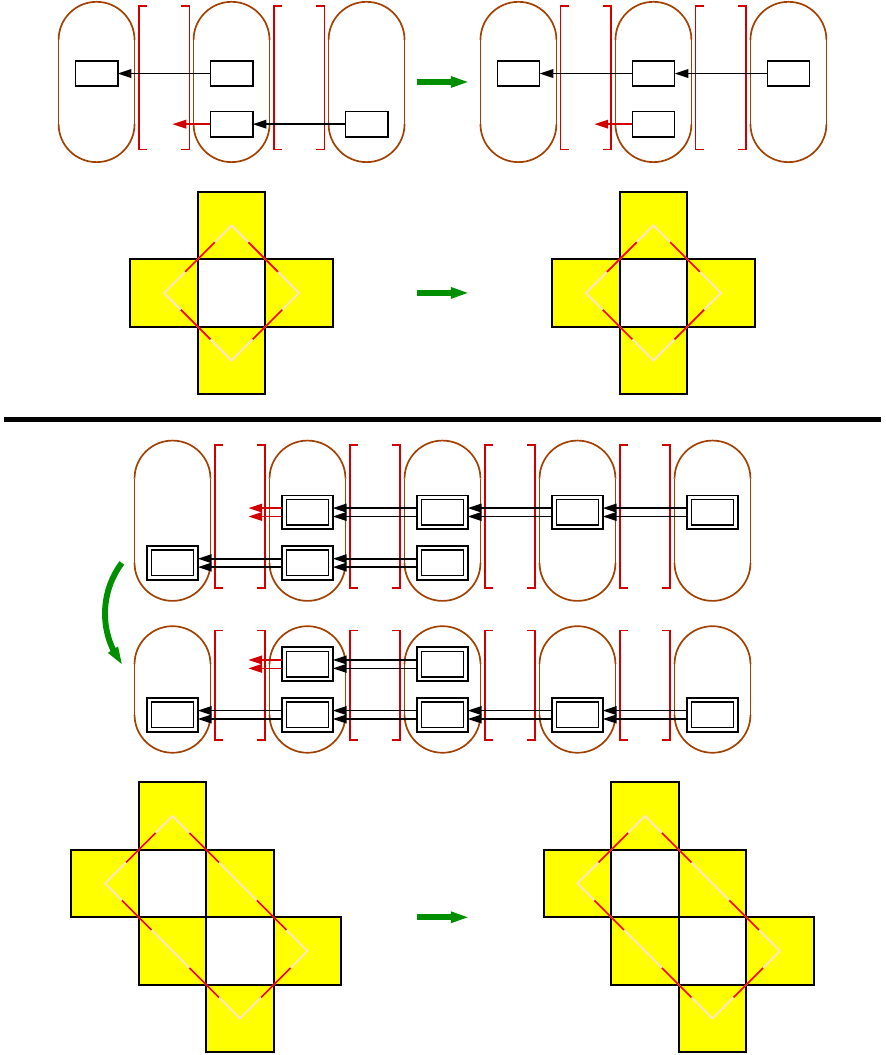}

\caption{\label{swapping}
Two examples of swapping moves.
In the top example---the simplest example possible---either
$W_1$ or $W_2$ may be perturbed to cause the move.
In the bottom example, any of $W_2$, $W_3$, or~$W_4$ may be perturbed.
The intervals $[h, k]$ and $[i, l]$ are replaced by
an interval $[h, l]$, longer than both original intervals, and
an interval $[i, k]$, shorter than both.
Four interval counts change, at the four corners of the red rectangle:
$\omega_{kh}$ and $\omega_{li}$ decrease, and
$\omega_{lh}$ and $\omega_{ki}$ increase.
The ranks of the subsequence matrices $W_{t \ssq s}$ increase
for all $t \in [k + 1, l]$ and $s \in [h, i - 1]$
(the ranks inside the red rectangle).
Outside the red rectangle, all interval counts and matrix ranks are unchanged.
}
\end{figure}

A {\em swapping move} is a small move with a nonzero displacement
$\Delta \theta \in \phi_{lkjih} \in \ThetaO^\barc$ in the case where $k \geq i$.
A swapping move changes four interval counts.
Figure~\ref{swapping} illustrates two examples.
A swapping move splices $\rank \Delta W_j$ copies of the interval~$[h, k]$ with
$\rank \Delta W_j$ copies of the interval~$[i, l]$, replacing them with
$\rank \Delta W_j$ copies of the interval~$[h, l]$
(which is longer than both of the replaced intervals) and
$\rank \Delta W_j$ copies of the interval~$[i, k]$ (which is shorter than both).
In effect, the interval endpoints are swapped.
(Again, careful study of the figure is rewarding.)
The {\em rank} of a swapping move is $\rank \Delta W_j$.

The edges of stratum dags signify barcode moves with $\rank \Delta W_j = 1$.
Barcode moves, along with
Figures~\ref{connecting} and~\ref{swapping}, expose
much intuition about how the strata are connected to each other.
Barcode moves give us some simple ways that
an infinitesimal perturbation of a point in weight space
can move you from one stratum to another stratum (always of higher dimension).
Not all small discrete moves are basic, but in Section~\ref{hierarchy}
we will see that every small discrete move that stays on the fiber
can be broken up into a sequence of barcode moves, signified by
a directed path through the stratum dag.

We define the following sets of one-matrix subspaces that
correspond to connecting moves, swapping moves, and barcode moves.
We also define a fourth set for one-matrix subspaces that signify
moves that stay on the same stratum, or at least its closure---some
subsequence matrix ranks might decrease.
That set omits all the barcode moves.
\begin{eqnarray*}
\ThetaO^\conn  & = & \{ \phi_{lkjih} \in \ThetaO : l \geq k + 1 = j = i > h \}
                 =   \{ \phi_{l,j-1,j,j,h} \neq \{ {\bf 0} \} :
                         L \geq l \geq j > h \geq 0 \},  \\
\ThetaO^\swap  & = & \{ \phi_{lkjih} \in \ThetaO : l > k \geq i > h \}
                 =   \{ \phi_{lkjih} \neq \{ {\bf 0} \} :
                         L \geq l > k \geq i > h \geq 0 \mbox{~and~}
                         k + 1 \geq j \geq i \},  \\
\ThetaO^\barc  & = & \ThetaO^\conn \cup \ThetaO^\swap  \\
               & = & \{ \phi_{lkjih} \in \ThetaO : l > k \mbox{~and~} i > h \}
                 =   \{ \phi_{lkjih} \neq \{ {\bf 0} \} :
                         L \geq l \geq k + 1 \geq j \geq i > h \geq 0 \},  \\
\ThetaO^\strat & = & \ThetaO^\fiber \setminus \ThetaO^\barc
                 = \{ \phi_{lkjih} \in \ThetaO :
                      (L > l \mbox{~or~} h > 0) \mbox{~and~}
                      (l \leq k \mbox{~or~} i \leq h) \}.
\end{eqnarray*}
For example, in the two-matrix case ($L = 2$),
\begin{eqnarray*}
\ThetaO^\conn  & = & \{ \phi_{10110}, \phi_{20110}, \phi_{21220}, \phi_{21221} \}
                        \setminus \{ \{ {\bf 0} \} \},  \\
\ThetaO^\swap  & = & \{ \phi_{21110}, \phi_{21210} \}
                        \setminus \{ \{ {\bf 0} \} \},  \\
\ThetaO^\barc  & = & \{ \phi_{10110}, \phi_{20110}, \phi_{21110}, \phi_{21210},
                        \phi_{21220}, \phi_{21221} \}
                        \setminus \{ \{ {\bf 0} \} \},  \\
\ThetaO^\strat & = & \{ \phi_{10100}, \phi_{11100}, \phi_{11110},
                        \phi_{12100}, \phi_{12110},
                        \phi_{21201}, \phi_{21211}, \phi_{22201},
                        \phi_{22211}, \phi_{22221} \}
                        \setminus \{ \{ {\bf 0} \} \}.
\end{eqnarray*}

\subsection{Abstract Barcode Moves}
\label{abstractmoves}

Each barcode move has an effect on the rank list (and the barcode) that depends
solely on the indices $h$, $i$, $k$, and $l$ and the rank of $\Delta W_j$.
This motivates the idea of an {\em abstract move} that maps
one rank list to another rank list, and one barcode to another barcode,
divorced entirely from any geometry.
A~{\em rank-$c$ abstract barcode move} takes
a valid rank list $\underline{r}$ and an index tuple $(l, k, i, h)$ satisfying
$L \geq l \geq k + 1 \geq i > h \geq 0$, and yields
the modified rank list $\underline{z}$ produced by setting
$z_{t \ssq s} = r_{t \ssq s} + c$ for $t \in [k + 1, l], s \in [h, i - 1]$;
for the other ranks, $z_{t \ssq s} = r_{t \ssq s}$.
Equivalently, it takes the barcode for $\underline{r}$,
decreases $\omega_{li}$ and~$\omega_{kh}$ by~$c$, and
increases $\omega_{lh}$ and $\omega_{ki}$ by $c$---except that
if $k + 1 = i$, then $\omega_{ki}$ does not exist and
only three interval counts change (an abstract connecting move).
An abstract barcode move produces a valid rank list if and only if
$\omega_{li} \geq c$ and $\omega_{kh} \geq c$ prior to the move.

In Section~\ref{hierarchy}, we will see that for valid rank lists satisfying
$\underline{r} < \underline{z}$, there exists
a sequence of rank-$1$ abstract barcode moves that proceed
from $\underline{r}$ to $\underline{z}$.
For that reason, in Section~\ref{dagdetails} we introduced the convention that
each edge of the stratum dag represents
a rank-$1$ abstract barcode move.

We think one of the most interesting aspects of this paper is that
the hierarchy of strata is captured by
these simple discrete operations on barcodes---barcodes that indicate
the flow of information through the neural network.
Barcode moves create farther-reaching flows and
may offer network training algorithms more opportunities to succeed.

\begin{figure}
\centerline{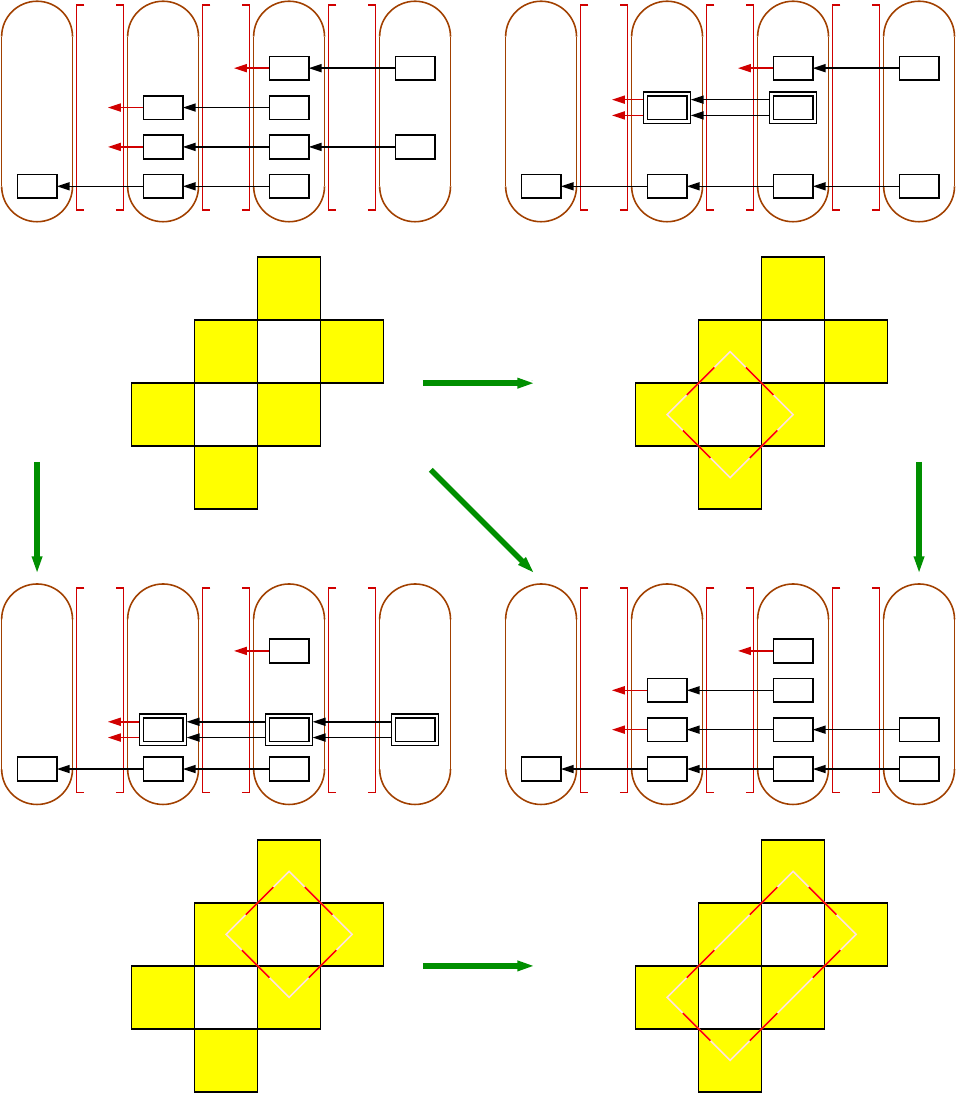}

\caption{\label{multiswap}
Annotated barcodes for four strata in a stratum dag.
(Assume that $W_4 = 0$, although $W_4$ is not illustrated, so
all four strata are on the fiber of $W = 0$.)
Green arrows are edges of the dag; in this example
they all representing swapping moves.
This dag includes the directed edge $(S_{\underline{r}}, S_{\underline{z}})$,
even though there are directed paths from $S_{\underline{r}}$ to $S_{\underline{z}}$
passing through $S_{\underline{u}}$ or $S_{\underline{v}}$.
Hence the dag is not a Hasse diagram.
The edge $(S_{\underline{r}}, S_{\underline{z}})$ represents swapping moves
whose displacements come from the subspaces $\phi_{31210}$ or $\phi_{31110}$.
}
\end{figure}

In Section~\ref{dagdetails} we noted that sometimes a stratum dag contains
a directed edge $(S_{\underline{r}}, S_{\underline{z}})$ despite also containing
the edges $(S_{\underline{r}}, S_{\underline{u}})$ and
$(S_{\underline{u}}, S_{\underline{z}})$---that is, the dag is not a Hasse diagram.
An example illustrated in Figure~\ref{multiswap} explains why.
Consider a stratum dag that includes the four strata whose
annotated barcodes are depicted.
Suppose that the four strata also have a fourth matrix $W_4 = 0$,
which is omitted from the figure---we need $W_4$ to have
an example in which all four of these strata are on the same fiber.
From the stratum $S_{\underline{r}}$, depicted at upper left,
there are rank-$1$ swapping moves that move directly onto
the stratum $S_{\underline{z}}$ at lower right.
But it is also possible to move from $S_{\underline{r}}$ to $S_{\underline{z}}$
through a sequence of two rank-$1$ swapping moves, passing through
$S_{\underline{u}}$ or $S_{\underline{v}}$ along the way.
A Hasse diagram would omit $(S_{\underline{r}}, S_{\underline{z}})$.
But we include $(S_{\underline{r}}, S_{\underline{z}})$ in the stratum dag
because the one-matrix subspaces $\phi_{31210}$ and $\phi_{31110}$
(as they are defined at a point $\theta \in S_{\underline{r}}$) represent
degrees of freedom of direct motion
from $S_{\underline{r}}$ into $S_{\underline{z}}$ that are
linearly independent of the one-matrix subspaces that represent
moves into $S_{\underline{u}}$ or $S_{\underline{v}}$
($\phi_{32310}$, $\phi_{32210}$, $\phi_{32110}$, $\phi_{21210}$, and $\phi_{21110}$).
These degrees of freedom of motion into $S_{\underline{z}}$ are not represented by
the dag's indirect paths from $S_{\underline{r}}$ to $S_{\underline{z}}$, but
they are important for understanding how the strata meet each other
geometrically.

\subsection{One-Matrix Moves on a Stratum}
\label{1matrixstratum}

At the end of Section~\ref{barcmoves}, we defined
a set of subspaces $\ThetaO^\strat$.
Let $\theta$ be the weight vector for which $\ThetaO^\strat$ is defined, and
let $S$ be the stratum that contains $\theta$.
The significance of $\ThetaO^\strat$ is that
for any subspace $\phi_{lkjih} \in \ThetaO^\strat$,
$\theta + \phi_{lkjih} \subseteq \bar{S}$.
That is, a~move from $\theta$ with
a displacement $\Delta \theta \in \phi_{lkjih}$ stays on~$\bar{S}$.
This implies that the move stays on the fiber and
does not increase any subsequence matrix rank.
It is possible that the rank of some subsequence matrix decreases
(i.e., it is not a small move), so the move might not stay on $S$, but
the following lemma shows that it stays on $\bar{S}$.
If $\Delta \theta$ is sufficiently small, the move does stay on $S$.
Therefore, $\phi_{lkjih} \subseteq T_\theta S$---a fact
we will use later to determine the tangent space.

Henceforth, for any set of subspaces $\Theta$, we define
the {\em span} of $\Theta$ to be $\Span \Theta = \sum_{\sigma \in \Theta} \sigma$,
the vector sum of the subspaces in $\Theta$.

\begin{lemma}
\label{1strat}
Every $\phi_{lkjih} \in \ThetaO^\strat$ satisfies
$\theta + \phi_{lkjih} \subseteq \bar{S}$
and $\phi_{lkjih} \subseteq T_\theta S$.
Moreover, $\Span \ThetaO^\strat \subseteq T_\theta S$.
\end{lemma}

\begin{proof}
Consider a subspace $\phi_{lkjih} \in \ThetaO^\strat$ and
a displacement $\Delta \theta \in \phi_{lkjih}$.
Recall that $\ThetaO^\strat = \ThetaO^\fiber \setminus \ThetaO^\barc$.
As $\phi_{lkjih} \in \ThetaO^\fiber$, by Corollary~\ref{ofiber},
$\mu(\theta + \Delta \theta) = W$.
Hence the translated subspace $\theta + \phi_{lkjih}$ is
a subset of the fiber $\mu^{-1}(W)$.

Consider the line though $\theta$ and $\theta + \Delta \theta$.
The parametrized point $\theta' = \theta + \epsilon \, \Delta \theta$ lies on
this line for every $\epsilon \in \R$.
Let $W'_{t \ssq s}$ denote a subsequence matrix derived from $\theta'$.
As $\phi_{lkjih} \not\in \ThetaO^\barc$, either $l \leq k$ or $i \leq h$.
By Lemma~\ref{rankincrease}, $\rank W'_{t \ssq s} \leq \rank W_{t \ssq s}$
for every subsequence matrix.
Let $\underline{r}$ be the rank list of $\theta$ and
let $\underline{r}'$ be the rank list of $\theta'$; thus
$\underline{r}' \leq \underline{r}$.
We will show that on the line though $\theta$ and $\theta + \Delta \theta$,
there are only finitely many points where $\underline{r}' < \underline{r}$,
and thus almost all the points on the line are in the stratum $S$, so
the line is a subset of $\bar{S}$.

A subsequence matrix $W_{t \ssq s}$ derived from $\theta$ has rank $r_{t \ssq s}$.
Thus there is some $r_{t \ssq s} \times r_{t \ssq s}$ minor of $W_{t \ssq s}$ that
has a nonzero determinant.
Consider the same minor of $W'_{t \ssq s}$ as a function of $\epsilon$.
The determinant of that minor is a degree-$r_{t \ssq s}$ polynomial in $\epsilon$
that is nonzero at $\epsilon = 0$.
Hence it has at most $r_{t \ssq s}$ zeros (with respect to $\epsilon$).
Hence on the line though $\theta$ and $\theta + \Delta \theta$,
there are at most $r_{t \ssq s}$ points where
$\rank W'_{t \ssq s} < \rank W_{t \ssq s}$.
At all other points on the line, $\rank W'_{t \ssq s} = \rank W_{t \ssq s}$.
As the number of subsequence matrices is finite, there are
only finitely many points on the line where $\underline{r}' < \underline{r}$,
as claimed; the line is a subset of $\bar{S}$.
This reasoning holds for every nonzero displacement
$\Delta \theta \in \phi_{lkjih}$, so
$\theta + \phi_{lkjih} \subseteq \bar{S}$ as claimed.

By Theorem~\ref{stratummanifold}, $S$ is a smooth manifold, hence
a tangent space $T_\theta S$ exists.
The line though $\theta$ and $\theta + \Delta \theta$ described above is
tangent to $S$ at $\theta$, so
$\Delta \theta$ is a subset of~$T_\theta S$.
This reasoning holds for every displacement $\Delta \theta \in \phi_{lkjih}$, so
$\phi_{lkjih} \subseteq T_\theta S$ as claimed.

It follows that $\Span \ThetaO^\strat \subseteq T_\theta S$ because
$T_\theta S$ is a subspace of $\R^{d_\theta}$ and
$\phi_{lkjih} \subseteq T_\theta S$ for every $\phi_{lkjih} \in \ThetaO^\strat$.
\end{proof}

\subsection{Some Intuition for the One-Matrix Subspaces}
\label{1canon}

We can gain some intuition about the one-matrix prebasis by
inspecting the one-matrix subspaces for the canonical weight vector
$\tilde{\theta} = (\tilde{I}_L, \tilde{I}_{L-1}, \ldots, \tilde{I}_1)$
defined in Section~\ref{canonical}.
At the canonical weight vector, every one-matrix subspace is {\em axis-aligned},
meaning that it is spanned by a subset of the coordinate axes of $\R^{d_\theta}$.
The one-matrix subspace $\phi_{lkjih}$ at $\tilde{\theta}$ is
the set of weight displacements that are zero everywhere except in
a single block of $\Delta \tilde{I}_j$, whose weights can take on any value.
That block is the one whose rows are associated with the interval $[i, l]$ and
whose columns are associated with the interval $[h, k]$.

\begin{figure}
\centerline{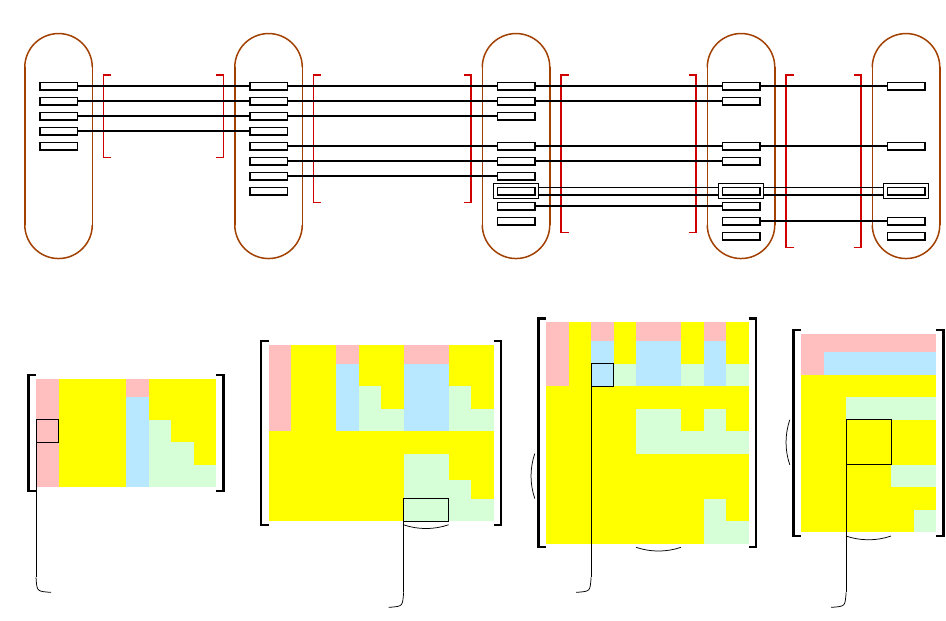}

\caption{\label{identfactor}
One-matrix subspaces for the canonical weight vector
$\tilde{\theta} = (\tilde{I}_4, \tilde{I}_3, \tilde{I}_2, \tilde{I}_1)$ when
every $\omega_{ki} = 1$ except $\omega_{20} = 2$.
Yellow and green backgrounds indicate basic moves that stay on the fiber
(moves with displacements from subspaces in $\ThetaO^\fiber$ such as
$\phi_{22100}$ and $\phi_{32330}$), whereas
pink and blue backgrounds indicate basic moves off the fiber
(moves from subspaces in $\ThetaO^\off$ such as $\phi_{44420}$ and $\phi_{43220}$,
satisfying $l = 4$ and $h = 0$).
Green and blue backgrounds indicate discrete moves
that increase the rank of some subsequence matrix
(moves from subspaces in $\ThetaO^\barc$ such as $\phi_{32330}$ and $\phi_{43220}$,
satisfying $l > k$ and $i > h$), whereas
yellow and pink backgrounds indicate moves that do not increase any rank.
Hence, yellow indicates moves that stay on the closure of the stratum
(moves from subspaces in $\ThetaO^\strat$ such as $\phi_{22100}$).
Green indicates discrete moves to a different stratum
that stay on the fiber (moves from
subspaces in $\ThetaO^\fiber \cap \ThetaO^\barc$ such as $\phi_{32330}$);
these induce edges in the stratum dag.
Pink indicates moves off the fiber in which
no subsequence matrix's rank increases.
Blue indicates discrete moves off the fiber.
}
\end{figure}

To clarify this idea, see Figure~\ref{identfactor}, which reprises
part of Figure~\ref{ident} with some background colors added.
The one-matrix subspace $\phi_{22100}$ is
the set of displacements $\Delta \theta$ that can be obtained
by writing arbitrary numbers in
the $2 \times 2$ block of~$\Delta \tilde{I}_1$ whose rows and columns are
associated with the interval $[0, 2]$ (labeled ``20'' in the figure), while
setting $\Delta \tilde{I}_4$, $\Delta \tilde{I}_3$, $\Delta \tilde{I}_2$, and
all the other components of $\Delta \tilde{I}_1$ to zero.

In the figure, we have given each component of each factor matrix
a background color to distinguish the effects of a one-matrix move.
In any single factor matrix $\tilde{I}_j$,
the components with a yellow background can be changed
without changing the product $\tilde{I}_4 \tilde{I}_3 \tilde{I}_2 \tilde{I}_1$
nor increasing the rank of any subsequence matrix.
A~change to a yellow component (or a yellow block with common indices)
is a basic move whose displacement is in a subspace in $\ThetaO^\strat$,
such as $\phi_{22100}$.
This move stays on the fiber by Corollary~\ref{ofiber} and,
by Lemma~\ref{1strat}, it also stays on the stratum's closure.
(If the move is small, it stays on the stratum proper.
If the move is not small, it might not stay on the stratum, because
some subsequence matrix's rank might decrease; but
no subsequence matrix's rank can increase.)

A green background marks components whose change increases
the rank of some subsequence matrix but does not change
$\tilde{I}_4 \tilde{I}_3 \tilde{I}_2 \tilde{I}_1$---hence
staying on the fiber but moving to a different stratum.
That is, a change to a green component is a move whose displacement is in
a subspace in $\ThetaO^\fiber \cap \ThetaO^\barc$, such as $\phi_{32330}$.
The edges of the stratum dag signify small changes to green components.

Pink and blue mark components whose change changes
$\tilde{I}_4 \tilde{I}_3 \tilde{I}_2 \tilde{I}_1$---hence moving off the fiber.
A change to a pink component increases no rank;  
it is a move whose displacement is in
a subspace in $\ThetaO^\off \setminus \ThetaO^\barc$, such as~$\phi_{44420}$.
A change to a blue component not only changes
$\tilde{I}_4 \tilde{I}_3 \tilde{I}_2 \tilde{I}_1$ but also increases its rank
(and possibly the ranks of other subsequence matrices).
It is a discrete move off the fiber, whose displacement is in
a subspace in $\ThetaO^\off \cap \ThetaO^\barc$, such as $\phi_{43220}$.


This intuition can be extended from the canonical weight vector
$\tilde{\theta}$ for a rank list $\underline{r}$ to
any other weight vector~$\theta$ that also has rank list $\underline{r}$.
The main observation is that there exists some linear transformation that maps
the axis-aligned subspaces illustrated in Figure~\ref{identfactor} to  
the basic subspaces at $\theta$.  
Let $W = \mu(\theta)$ and
recall from Section~\ref{sec:affinestrata} the linear function
$\eta: (X_L, X_{L-1}, \ldots, X_1) \mapsto
(J_L X_L J_{L-1}^{-1}, J_{L-1} X_{L-1} J_{L-2}^{-1}, \ldots, J_1 X_1 J_0^{-1})$
that maps the fiber of $\tilde{I}$ to the fiber of $W$.
An invertible linear transformation of weight space preserves tangencies, so
the function $\eta$ maps
the one-matrix subspaces at $\tilde{\theta}$ (which are axis-aligned) to
the one-matrix subspaces at $\theta$ (which are not).

\subsection{The Hierarchy of Strata}
\label{hierarchy}

We are now ready to prove the claim about the stratum interconnections made
in Section~\ref{foretaste}.

\begin{theorem}
\label{rankclosureequiv}
Let $\underline{r}$ and $\underline{z}$ be two valid rank lists
for the same linear neural network
(i.e., $r_{j \ssq j} = z_{j \ssq j} = d_j$ for all $j \in [0, L]$).
Let $\mu^{-1}(W)$ be the fiber of a matrix $W \in \R^{d_L \times d_0}$
whose rank satisfies $\rank W = r_{L \ssq 0} = z_{L \ssq 0}$.
Let $S_{\underline{r}}$ and $S_{\underline{z}}$ be
the strata with rank lists $\underline{r}$ and $\underline{z}$ in
the rank stratification $\mathbb{S}$ of $\mu^{-1}(W)$, and
observe that
$S_{\underline{r}} \neq \emptyset$ and $S_{\underline{z}} \neq \emptyset$
by Lemma~\ref{nonemptystratum}.
Then the following statements are equivalent (imply each other).
\begin{enumerate}[A.]
\item  $S_{\underline{r}} \subseteq \bar{S}_{\underline{z}}$.
\item  $S_{\underline{r}} \cap \bar{S}_{\underline{z}} \neq \emptyset$.
\item  $\underline{r} \leq \underline{z}$.
\item  There exists a sequence of rank-one abstract barcode moves that proceed
  from $\underline{r}$ to~$\underline{z}$, with all the intermediate rank lists
  being valid.
\end{enumerate}
\end{theorem}

\begin{proof}
Clearly A implies B, as we assume $S_{\underline{r}} \neq \emptyset$.
Lemma~\ref{stratatouch} states that B implies C.
The forthcoming Corollary~\ref{dagarrows}
states that C implies D.
The forthcoming Lemma~\ref{lem:moves-imply-closure}
states that D implies~A.
\end{proof}

We now prove that D implies A.
The proof is more opaque than we would like, but
the essence of the proof is that every abstract barcode move can be instantiated
as an actual move on the geometry of the fiber---a move that can be
arbitrarily small, but is not merely an abstract or infinitesimal move.

\begin{lemma}\label{lem:moves-imply-closure}
Consider a sequence of valid rank lists
$\underline{r}^0, \underline{r}^1, \ldots, \underline{r}^t$ such that
each successive rank list can be reached from the previous rank list by
an abstract barcode move (of any rank).
Let $W$ be a matrix, and suppose that
$r^i_{L \ssq 0} = \rank W$ for all $i \in [0, t]$.
Let $S_{\underline{r}^0}$ and $S_{\underline{r}^t}$ be
strata in the rank stratification of the fiber $\mu^{-1}(W)$.
Then $S_{\underline{r}^0} \subseteq \bar{S}_{\underline{r}^t}$.
\end{lemma}

\begin{proof}
If $S_{\underline{r}^0} = \emptyset$, the result follows immediately.
Otherwise, let $\theta_0$ be any point in $S_{\underline{r}^0}$.
For each $m \in [1, t]$, we will identify a point
$\theta_m \in S_{\underline{r}^m}$ that can be reached from $\theta_{m-1}$ by
a (not abstract, not infinitesimal) barcode move.
By induction, we obtain a point $\theta_t \in S_{\underline{r}^t}$ that is
as close to $\theta_0$ as we like.

Assume for the sake of induction that there is
a point $\theta_{m-1} \in S_{\underline{r}^{m-1}}$.
By assumption there is a rank-$c$ abstract move
from $\underline{r}^{m-1}$ to $\underline{r}^m$;
it is specified by an index tuple $(l, k, i, h)$ and the rank $c$.
The fact that $\underline{r}^m$ is a valid rank list implies that
the previous rank list $\underline{r}^{m-1}$ has
interval counts $\omega_{li} \geq c$ and $\omega_{kh} \geq c$
(as the move decreases both counts by~$c$).
Choose any $j \in [i, k + 1]$ and consider
the one-matrix subspace $\phi_{lkjih}$ defined at the point $\theta_{m-1}$.
Its dimension is
$\dim \phi_{lkjih} = \dim o_{lkjih} = \omega_{li} \, \omega_{kh} \geq c^2$, and
some matrices in $o_{lkjih}$ have rank~$c$, because
$o_{lkjih} = a_{lji} \otimes b_{k,j-1,h}$ is a tensor product of
two subspaces of dimension $c$ or greater.
Let $\Delta \theta \in \phi_{lkjih}$ be
a displacement such that $\Delta W_j \in o_{lkjih}$ has rank $c$ and is
sufficiently small that the move decreases no subsequence matrix rank.
By Lemma~\ref{rankincrease},
$\theta_{m-1} + \Delta \theta$ has rank list $\underline{r}^m$.
We set $\theta_m = \theta_{m-1} + \Delta \theta$.

Each move in the sequence is a small move, so
the sequence of moves constitutes a small move.
As the final point $\theta_t$ lies on $S_{\underline{r}^t}$,
it follows that $\theta_0 \in \bar{S}_{\underline{r}^t}$.

This reasoning applies to every point $\theta_0 \in S_{\underline{r}^0}$, so
$S_{\underline{r}^0} \subseteq \bar{S}_{\underline{r}^t}$.
\end{proof}

We return to the hard part of Theorem~\ref{rankclosureequiv},
the claim that statement~C implies statement~D:
if~$\underline{r} \leq \underline{z}$,
there exists a sequence of rank-one abstract barcode moves that
takes us from $\underline{r}$ to~$\underline{z}$.
(Clearly D implies~C---abstract moves never decrease ranks.)
This claim is a bridge to showing that
$S_{\underline{r}} \subseteq \bar{S}_{\underline{z}}$ if and only if
$\underline{r} \leq \underline{z}$.

\begin{figure}
\centerline{\mbox{
\begin{minipage}{6in}
\begin{tabbing}
MMM \= MM \= MM \= MM \= MM \= MM \= MMMMMMMMMMMM \= \kill
{\sc FindLastMove}$(\underline{r}, \underline{z})$  \\
// Given valid rank lists $\underline{r}$ and $\underline{z}$ such that
   $\underline{r} < \underline{z}$, returns a valid rank list $\underline{u}$
   such that $\underline{r} \leq \underline{u} < \underline{z}$ and  \\
// a rank-one abstract
   barcode move takes us from $\underline{u}$ to $\underline{z}$.  \\
1  \> {\bf for all} $t$ and $s$ satisfying $L \geq t \geq s \geq 0$  \\
   \> \> // From $\underline{r}$ and $\underline{z}$, compute all the interval
            counts with~(\ref{omegarank}).  \\
2  \> \> $\omega^r_{ts} \leftarrow
         r_{t \ssq s} - r_{t \ssq s-1} - r_{t+1 \ssq s} + r_{t+1 \ssq s-1}$
         \hspace{.2in}
         // by convention, $r_{L+1 \ssq s} = r_{t \ssq -1} = 0$  \\
3  \> \> $\omega^z_{ts} \leftarrow
         z_{t \ssq s} - z_{t \ssq s-1} - z_{t+1 \ssq s} + z_{t+1 \ssq s-1}$  \\
4  \> \> $\Delta \omega_{ts} \leftarrow \omega^z_{ts} - \omega^r_{ts}$  \\
5  \> \> $\Delta r_{t \ssq s} \leftarrow z_{t \ssq s} - r_{t \ssq s}$
         \hspace{.2in}  // every $\Delta r_{t \ssq s}$ is nonnegative, as
                           $\underline{r} < \underline{z}$  \\
   \> \> // Initialize $\underline{u}$ to be the same as $\underline{z}$.  \\
6  \> \> $u_{t \ssq s} \leftarrow z_{t \ssq s}$  \\
7  \> $[h, l] \leftarrow$ the longest interval (i.e., maximize $l - h$)
      such that $\Delta r_{l \ssq h} > 0$  \\
8  \> $i' \leftarrow$ the smallest index in $[h + 1, l]$ such that
      $i' = l$ or $\Delta \omega_{l-1,i'} > 0$ or $\Delta r_{l \ssq i'} = 0$  \\
9  \> {\bf if} $i' = l$ {\bf or} $\Delta \omega_{l-1,i'} > 0$  \\
10 \> \> $i \leftarrow i'$; $k \leftarrow l - 1$  \\
11 \> {\bf else}  \hspace{.2in}  // $\Delta r_{l \ssq i'} = 0$  \\
12 \> \> $k \leftarrow$ the greatest index in $[i' - 1, l - 2]$ such that
         $k = i' - 1$ or $\Delta \omega_{ki} > 0$ for some $i \in [h + 1, i']$
         \\
13 \> \> $i \leftarrow$ the smallest index in $[h + 1, i']$ such that
         $i = k + 1$ or $\Delta \omega_{ki} > 0$  \\
   \> // Update $\underline{u}$ to perform a reverse rank-$1$ barcode move with
         index tuple $(l, k, i, h)$.  \\
14 \> {\bf for} $t \leftarrow k + 1$ {\bf to} $l$  \\
15 \> \> {\bf for} $s \leftarrow h$ {\bf to} $i - 1$  \\
16 \> \> \> $u_{t \ssq s} \leftarrow z_{t \ssq s} - 1$  \\
17 \> {\bf return} $\underline{u}$  \\
\\
{\sc FindDAGPath}$(\underline{r}, \underline{z})$  \\
// Given rank lists $\underline{r}$ and $\underline{z}$ such that
   $\underline{r} \leq \underline{z}$, finds and writes a sequence of rank lists
   proceeding  \\
// from $\underline{r}$ to $\underline{z}$
   by a sequence of rank-one abstract barcode moves.  \\
18 \> {\bf if} $\underline{r} \neq \underline{z}$  \\
19 \> \> $\underline{u} \leftarrow$
         {\sc FindLastMove}$(\underline{r}, \underline{z})$  \\
20 \> \> {\sc FindDAGPath}$(\underline{r}, \underline{u})$  \\
21 \> Write $\underline{z}$
\end{tabbing}
\end{minipage}
}}

\caption{\label{findmoves}
Algorithm to find a sequence of rank-$1$ abstract barcode moves
that proceed from a rank list $\underline{r}$ to a rank list $\underline{z}$,
assuming that $\underline{r} \leq \underline{z}$.
{\sc FindDAGPath} operates recursively, repeatedly invoking
{\sc FindLastMove} to identify the last move in the sequence.
}
\end{figure}

Given rank lists $\underline{r}$ and $\underline{z}$
with $\underline{r} < \underline{z}$,
finding a sequence of abstract moves that takes us
from $\underline{r}$ to $\underline{z}$ is
a puzzle that took us three months to solve.
The constructive part of our proof is the algorithm {\sc FindLastMove} in
Figure~\ref{findmoves}, which finds a rank list $\underline{u}$ such that
$\underline{r} \leq \underline{u} < \underline{z}$ and
a single rank-one abstract barcode move takes us from
$\underline{u}$ to~$\underline{z}$.
Building on this step, a simple recursive algorithm,
{\sc FindDAGPath} in Figure~\ref{findmoves}, finds
a sequence of abstract barcode moves that take us
from $\underline{r}$ to $\underline{z}$
(computing the sequence in reverse order,
from $\underline{z}$ to $\underline{r}$).
The proof of correctness of {\sc FindDAGPath}, and
thus the proof that C implies D, follows by induction.

{\sc FindLastMove} and our proofs use differences between
the rank lists $\underline{r}$ and $\underline{z}$.
Let the interval counts associated with~$\underline{r}$ be
$\omega^r_{ts}$ and
the interval counts associated with~$\underline{z}$ be $\omega^z_{ts}$.
Let
\[
\Delta r_{t \ssq s} = z_{t \ssq s} - r_{t \ssq s}
\hspace{.2in}  \mbox{and}  \hspace{.2in}
\Delta \omega_{ts} = \omega^z_{ts} - \omega^r_{ts}
\hspace{.2in}  \mbox{for all $t$ and $s$ satisfying~}  L \geq t \geq s \geq 0.
\]
As we assume that $\underline{r} \leq \underline{z}$,
no $\Delta r_{t \ssq s}$ is negative.
(No such constraint applies to $\Delta \omega_{ts}$.)

\begin{figure}
\centerline{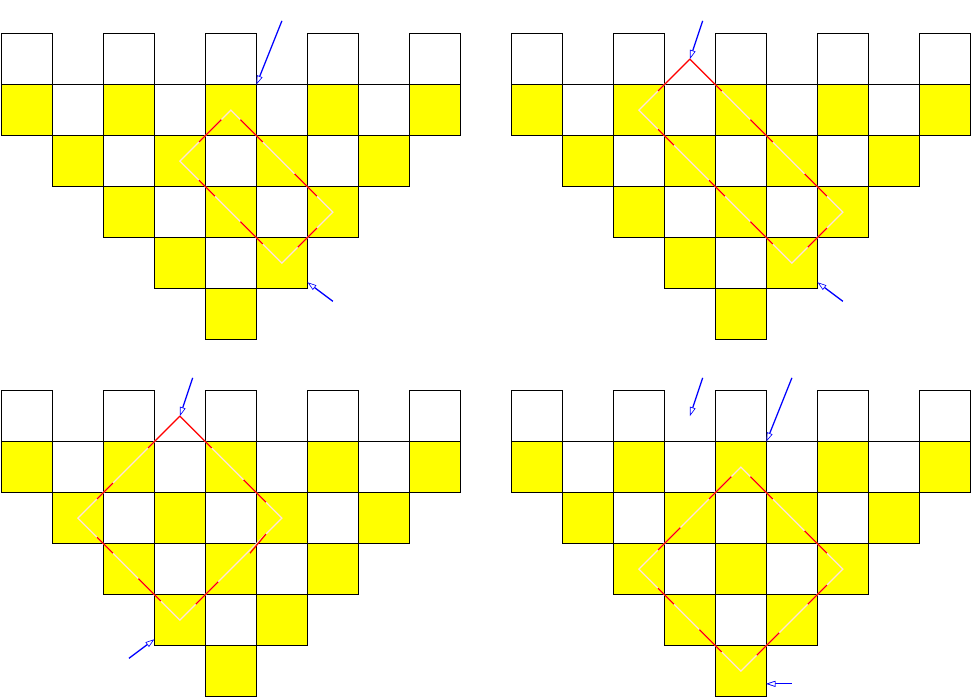}

\caption{\label{reverse}
Four examples of reverse barcode moves found by the algorithm {\sc FindLastMove}
(shown prior to performing the reverse move).
In each example, the algorithm finds a red rectangle such that 
its bottom corner lies at a bottommost positive $\Delta \omega_{lh}$,
its top corner lies at a positive $\Delta \omega_{ki}$ or  
between two $\Delta d$'s, and  
every~$\Delta r$ inside the rectangle is positive.
In (b)~and~(c) the algorithm finds a reverse connecting move ($k = i - 1$),
whereas in (a)~and~(d) it finds a reverse swapping move ($k \geq i$),
with~(d) being a move that changes $W$'s rank.
In~(a)~and~(b), Line~10 of {\sc FindLastMove} determines the top corner of  
the rectangle (hence $k = l - 1$), whereas 
in (c)~and~(d), Lines~12 and~13 do (hence~$k \leq l - 2$).
}
\end{figure}

Figure~\ref{reverse} depicts examples of reverse moves found by
{\sc FindLastMove}, and thereby sheds light on Lines~7--13,  
the core of the algorithm.
In each example, a red rectangle shows which reverse move the algorithm
finds---specifically, the index tuple $(l, k, i, h)$ of the move.
Line~7 selects $l$ and $h$ so that the bottom corner of the red rectangle,
$\Delta \omega_{lh}$, is a bottommost positive $\Delta \omega$.
(Equivalently, the bottommost rank inside the rectangle, $\Delta r_{l \ssq h}$,
is a bottommost positive $\Delta r$.)
Lines~8--13 search for values of $k$ and $i$ such that
every~$\Delta r$ inside the rectangle is positive and
the rectangle's top corner is either
at a positive $\Delta \omega_{ki}$ or between two~$\Delta d$'s
($k = i - 1$)---these constraints guarantee that a reverse move yields
a valid rank list.

Lines~8 and~9 check whether such a rectangle exists with $k = l - 1$.
If so, Line~10 sets $i$ to indicate the smallest such rectangle;
see Figures~\ref{reverse}(a) and~(b) for examples.
If not, the obstacle is some $\Delta r_{l \ssq i'}$ equal to zero.
In that case, Lines~12 and~13 find values of $k$ and $i$ that satisfy
the constraints; see Figures~\ref{reverse}(c) and~(d) for examples.  
The hard part of our proof is showing that Lines~12 and~13 always succeed.

Theorem~\ref{rankclosureequiv} assumes that $r_{L \ssq 0} = z_{L \ssq 0}$; but
{\sc FindDAGPath} and the proofs below require {\em only} that
$r_{L \ssq 0} \leq z_{L \ssq 0}$.
The case $\Delta r_{L \ssq 0} > 0$ is not relevant to
the stratum dag of $\mu^{-1}(W)$, on which $\rank W$ is fixed, but 
the lemma still holds, as Figure~\ref{reverse}(d) illustrates.
Perhaps this extra generality will find a use someday.

The following four lemmas derive properties of the algorithms and
prove that {\sc FindDAGPath} always finds a valid sequence of rank lists.

\begin{lemma}
\label{lem:omega-quad}
For indices satisfying $L \geq k \geq i \geq 0$,
\begin{eqnarray*}
\Delta r_{k \ssq i} & = & \sum_{t=k}^L \sum_{s=0}^i \Delta \omega_{ts}
\hspace{.2in}  \mbox{and}  \\
\Delta \omega_{ki} & = & \Delta r_{k \ssq i} - \Delta r_{k \ssq i-1} -
\Delta r_{k+1 \ssq i} + \Delta r_{k+1 \ssq i-1}.
\end{eqnarray*}
For indices satisfying $L \geq l > k \geq i > h \geq 0$,
\[
\Delta r_{k \ssq i} - \Delta r_{k \ssq h} - \Delta r_{l \ssq i} +
\Delta r_{l \ssq h} = \sum_{t=k}^{l-1} \sum_{s=h+1}^i \Delta \omega_{ts}.
\]
\end{lemma}

\begin{proof}
The first identity is obtained from
$\Delta r_{k \ssq i} = z_{k \ssq i} - r_{k \ssq i}$ by
substituting~(\ref{ranksum}) for both $r_{k \ssq i}$ and $z_{k \ssq i}$,
then substituting $\omega^z_{ts} - \omega^r_{ts} = \Delta \omega_{ts}$.
The second and third identities follow directly from the first by substitution.
(Alternatively, the second identity follows from
$\Delta \omega_{ki} = \omega^z_{ki} - \omega^r_{ki}$ by
substituting~(\ref{omegarank}) for both $\omega^r_{ki}$ and $\omega^z_{ki}$.)
\end{proof}

\begin{lemma}
\label{posrank}
Let $\underline{r}$ and $\underline{z}$ be two valid rank lists
for the same linear neural network
(i.e., $r_{j \ssq j} = z_{j \ssq j} = d_j$ for all $j \in [0, L]$) such that
$\underline{r} < \underline{z}$.
Let $h$, $i$, $k$, and $l$ be
the indices chosen by Lines~7--13 of {\sc FindLastMove}.
Then
\begin{icompact}
\item
$\Delta r_{t \ssq s} \geq 1$ for all $t \in [k + 1, l]$ and $s \in [h, i - 1]$,
\item
$\omega^z_{lh} \geq 1$, and
\item
one of these two hold:  $k = i - 1$ or $\omega^z_{ki} \geq 1$.
\end{icompact}
\end{lemma}

\begin{proof}
Line~7 of {\sc FindLastMove} chooses $[h, l]$ to be
the longest interval such that $\Delta r_{l \ssq h} \geq 1$.
(As $\underline{r} < \underline{z}$, there is at least
one choice of $h$ and $l$ for which $\Delta r_{l \ssq h} > 0$ and $l > h$.)
By Lemma~\ref{lem:omega-quad},
$\Delta \omega_{lh} = \Delta r_{l \ssq h} - \Delta r_{l \ssq h-1} -
\Delta r_{l+1 \ssq h} + \Delta r_{l+1 \ssq h-1}$.
The last three terms on the right-hand side represent longer intervals, so
$\Delta \omega_{lh} = \Delta r_{l \ssq h} \geq 1$.
As $\Delta \omega_{lh} = \omega^z_{lh} - \omega^r_{lh}$ and
$\omega^r_{lh}$ is always nonnegative, $\omega^z_{lh} \geq 1$,
verifying the second claim.

Line~8 chooses $i'$ to be the smallest index in $[h + 1, l]$ such that
$i' = l$, $\Delta \omega_{l-1,i'} > 0$, or $\Delta r_{l \ssq i'} = 0$.
Hence $\Delta \omega_{l-1,s} \leq 0$ and $\Delta r_{l \ssq s} \geq 1$
for all $s \in [h + 1, i' - 1]$.
(Otherwise, a smaller $i'$ would have been chosen.
Note that the statement is vacuously true if $i' = h + 1$.)
As we have already shown that $\Delta r_{l \ssq h} \geq 1$,
we have $\Delta r_{l \ssq s} \geq 1$ for all $s \in [h, i' - 1]$.

If $i' = l$ or $\Delta \omega_{l-1,i'} > 0$, then
Line~10 sets $i = i'$ and $k = l - 1$, and all the lemma's claims hold.
The lemma's first claim, that
$\Delta r_{t \ssq s} \geq 1$ for all $t \in [k + 1, l]$ and $s \in [h, i - 1]$,
holds because as we have seen,
$\Delta r_{l \ssq s} \geq 1$ for all $s \in [h, i' - 1]$.
The lemma's third claim holds because if $i' = l$ then $k = i - 1$,
as in Figure~\ref{reverse}(b); whereas  
if $\Delta \omega_{l-1,i'} > 0$, then $\Delta \omega_{ki} \geq 1$ and thus
$\omega^z_{ki} \geq 1$, as in Figure~\ref{reverse}(a).

The case remains where neither $i' = l$ nor $\Delta \omega_{l-1,i'} > 0$.
In that case, $\Delta \omega_{l-1,i'} \leq 0$,
thus $\Delta \omega_{l-1,s} \leq 0$ for all $s \in [h + 1, i']$, and
Line~8 has chosen $i'$ such that $\Delta r_{l \ssq i'} = 0$.
Line~12 chooses $k$ to be the greatest index in $[i' - 1, l - 2]$
such that $k = i' - 1$ or $\Delta \omega_{ki} > 0$ for some $i \in [h + 1, i']$.
Hence $\Delta \omega_{ts} \leq 0$
for all $t \in [k + 1, l - 1]$ and $s \in [h + 1, i']$.
(Otherwise, a greater $k$ would have been chosen.)

We now show that
$\Delta r_{t \ssq s} \geq 1$ for all $t \in [k + 1, l]$ and $s \in [h, i' - 1]$. We have already shown that
$\Delta r_{l \ssq s} \geq 1$ for all $s \in [h, i' - 1]$.
It remains to show $\Delta r_{t \ssq s} \geq 1$ for
$t \in [k + 1, l - 1]$ and $s \in [h, i' - 1]$.
By Lemma~\ref{lem:omega-quad},
\[
\Delta r_{t \ssq i'} - \Delta r_{t \ssq s} - \Delta r_{l \ssq i'} +
\Delta r_{l \ssq s} = \sum_{q=t}^{l-1} \sum_{p=s+1}^{i'} \Delta \omega_{qp}
\leq 0.
\]
By assumption, $\underline{r} < \underline{z}$ and thus
$\Delta r_{t \ssq i'} \geq 0$.
Recall that $\Delta r_{l \ssq i'} = 0$ and $\Delta r_{l \ssq s} \geq 1$, so
\[
\Delta r_{t \ssq s}
\geq \Delta r_{t \ssq i'} - \Delta r_{l \ssq i'} + \Delta r_{l \ssq s}
\geq 0 - 0 + 1 = 1.
\]
Therefore,
$\Delta r_{t \ssq s} \geq 1$ for all $t \in [k + 1, l]$ and $s \in [h, i' - 1]$
as claimed.
As the algorithm chooses $i$ from $[h + 1, i']$,
this confirms the lemma's first claim, that
$\Delta r_{t \ssq s} \geq 1$ for all $t \in [k + 1, l]$ and $s \in [h, i - 1]$.

The lemma's third claim follows because Lines~12 and~13 explicitly choose
$k$ and $i$ such that $i = k + 1$, as in Figure~\ref{reverse}(c), or
$\Delta \omega_{ki} > 0$, as in Figure~\ref{reverse}(d);
in the latter case, $\omega^z_{ki} \geq 1$.
\end{proof}

\begin{lemma}
\label{rvalid}
Let $\underline{r}$ and $\underline{z}$ be two valid rank lists
for the same linear neural network
(i.e., $r_{j \ssq j} = z_{j \ssq j} = d_j$ for all $j \in [0, L]$) such that
$\underline{r} < \underline{z}$.
Then the rank list $\underline{u}$ returned by
{\sc FindLastMove}$(\underline{r}, \underline{z})$ is a valid rank list
that satisfies $\underline{r} \leq \underline{u} < \underline{z}$, and
a single rank-$1$ abstract barcode move changes
$\underline{u}$ to~$\underline{z}$.
\end{lemma}

\begin{proof}
Inspection of {\sc FindLastMove} shows that the indices chosen by Lines~7--13
always satisfy $l > k$, $k + 1 \geq i$, and $i > h$.
Line~6 of {\sc FindLastMove} initializes $\underline{u}$ to be
the same rank list as $\underline{z}$, then
Lines~14-16 reduce the value of $u_{t \ssq s}$ by one
for every $t \in [k + 1, l]$ and $s \in [h, i - 1]$.
At least one rank is reduced, which confirms that
$\underline{u} < \underline{z}$.
By Lemma~\ref{posrank},
$\Delta r_{t \ssq s} = z_{t \ssq s} - r_{t \ssq s} \geq 1$
for every $t \in [k + 1, l]$ and $s \in [h, i - 1]$, so
$u_{t \ssq s} \geq r_{t \ssq s}$ for all $t$ and $s$ satisfying
$L \geq t \geq s \geq 0$, which confirms that
$\underline{r} \leq \underline{u}$.

If $k + 1 = i$, then a transition from $\underline{u}$ to $\underline{z}$ is
a rank-$1$ abstract connecting move that replaces
one copy of $[h, k]$ and one copy of $[i, l]$ with one copy of $[h, l]$,
whereas if $k \geq i$, a transition from $\underline{u}$ to $\underline{z}$ is
a rank-$1$ abstract swapping move that replaces
one copy of $[h, k]$ and one copy of $[i, l]$ with
one copy of $[h, l]$ and one copy of $[i, k]$.
If the algorithm were explicitly computing
the interval counts $\omega^u_{ts}$ associated with $\underline{u}$,
we would have $\omega^u_{li} = \omega^z_{li} + 1$,
$\omega^u_{kh} = \omega^z_{kh} + 1$, $\omega^u_{lh} = \omega^z_{lh} - 1$, and
if $k \geq i$, $\omega^u_{ki} = \omega^z_{ki} - 1$.
All the other interval counts associated with $\underline{u}$ are
the same as those associated with $\underline{z}$.

To show that $\underline{u}$ is a valid rank list,
we show that the barcode associated with $\underline{u}$ is valid,
then apply Lemma~\ref{validranklist}.
Recall that a barcode represented by
interval counts $\omega^u_{ts}$, $L \geq t \geq s \geq 0$, is valid if
$d_j = \sum_{t = j}^L \sum_{s = 0}^j \omega^u_{ts}$ for every $j \in [0, L]$.
It is straightforward to see that if the counts $\omega^z_{ts}$ satisfy
these identities, then the counts $\omega^u_{ts}$ specified above
(following a reverse abstract move) satisfy them too.
We also require that every $\omega^u_{ts}$ is nonnegative---otherwise,
the interval counts do not represent a multiset.
By Lemma~\ref{posrank},
$\omega^z_{lh} \geq 1$ and either $k = i - 1$ or $\omega^z_{ki} \geq 1$,
from which it follows that
$\omega^u_{lh} \geq 0$ and either $k = i - 1$ or $\omega^u_{ki} \geq 0$.
For every other pair of indices $t$ and $s$,
$\omega^u_{ts} \geq \omega^z_{ts} \geq 0$.
Hence every $\omega^u_{ts}$ is nonnegative, and
$\underline{u}$ is a valid rank list by Lemma~\ref{validranklist}.
\end{proof}

\begin{cor}
\label{dagarrows}
Let $\underline{r}$ and $\underline{z}$ be two valid rank lists
for the same linear neural network
(i.e., $r_{j \ssq j} = z_{j \ssq j} = d_j$ for all $j \in [0, L]$) such that
$\underline{r} \leq \underline{z}$.
Then there exists a sequence of valid rank lists that
starts with~$\underline{r}$ and ends with $\underline{z}$ such that
each rank list after $\underline{r}$ can be obtained from
the previous rank list in the sequence by
a single rank-$1$ abstract barcode move.
Moreover, the algorithm {\sc FindDAGPath} in Figure~\ref{findmoves} finds
such a sequence.
\end{cor}

\begin{proof}
The first claim follows from Lemma~\ref{rvalid} by a simple induction.
The second claim follows because {\sc FindDAGPath} implements
this constructive inductive proof as a recursive algorithm.
\end{proof}

Corollary~\ref{dagarrows} completes the proof of Theorem~\ref{rankclosureequiv}.

\clearpage

\clearpage

\section{The Tangent Space, the Nullspace of $\mathrm{d}\mu(\theta)$, and
         Prebases that Span Them}
\label{weightprebases}

Consider a fiber $\mu^{-1}(W)$, a stratum $S$ in its rank stratification, and
a weight vector $\theta \in S$. 
We wish to identify $T_\theta S$, the space tangent to $S$ at $\theta$.

We identified some subspaces of $T_\theta S$
in Lemma~\ref{1strat} (Section~\ref{1matrixstratum})---specifically,
the one-matrix subspaces that are subsets of $T_\theta S$---but
they do not generally suffice to span $T_\theta S$.
In this section, we find a complete prebasis for $T_\theta S$, which we call
the {\em stratum prebasis}, by adding some {\em two-matrix subspaces}.
These subspaces, related to two-matrix moves, represent
some directions along which $S$ is curved.
The prebasis will help us to write an explicit expression for $T_\theta S$ and
to determine the dimension of~$S$ (as $\dim S = \dim T_\theta S$).

We are not satisfied with knowing merely $T_\theta S$.
For every stratum~$S'$ whose closure contains~$\theta$,
we wish to identify the tangent space $T_\theta \bar{S}'$ if it exists.
(Sometimes $T_\theta \bar{S}'$ does not exist because
$\theta$ is a singular point of~$\bar{S}'$.)
Observe that for every stratum~$S'$ with $\theta \in \bar{S}'$,
$\bar{S'} \supseteq S$ because rank stratifications satisfy
the frontier condition (by Lemma~\ref{rankclosureequiv}).
Thus if $T_\theta \bar{S}'$ exists, then $T_\theta \bar{S}' \supseteq T_\theta S$.

Moreover, we want a {\em single} prebasis that can express
{\em all} of these tangent spaces at $\theta$.
The prebasis that does this is called the {\em fiber prebasis} at $\theta$,
described in Section~\ref{fiberbasis}.
This prebasis allows us to identify all the directions by which
a smooth path can leave $\theta$ on the fiber, on any stratum.

\begin{figure}
\centerline{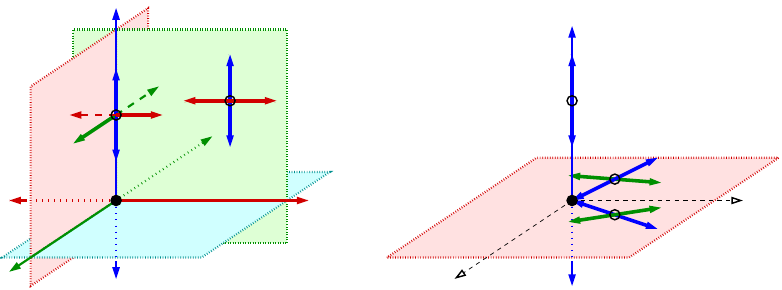}

\caption{\label{fbasis}
Fiber prebases at five points on the fibers from
Figures~\ref{strat1111} and~\ref{strat112}.
Note that we have drawn each prebasis translated to
the point where it is defined
(to clarify how the subspaces are related to the strata), but
all the subspaces {\em actually} pass through the origin, and 
not necessarily through the points~$\theta$, $\theta'$, etc.
}
\end{figure}

To visualize this, see Figure~\ref{fbasis}, which depicts
the fibers from Figure~\ref{strat1111} and~\ref{strat112}:
the variety of solutions to $W_3 W_2 W_1 = [0]$ where
$W_3, W_2, W_1 \in \R^{1 \times 1}$ (left) and
the variety of solutions to $W_2 W_1 = [0]$ where
$W_2 \in \R^{1 \times 1}$ and $W_1 \in \R^{1 \times 2}$ (right).
Figure~\ref{fbasis} also shows the fiber prebases at
five points $\theta$, $\theta'$, $\theta''$, $\theta'''$, and $\theta''''$.
At any point $\theta \in S_{010}$, the fiber prebasis is
a set of three lines, namely, the three coordinate axes
(which are the one-matrix subspaces
$\phi_{20110}$, $\phi_{22211}$, and $\phi_{32331}$ at $\theta$).
Two of these lines span $T_\theta \bar{S}_{011}$,
two of them span $T_\theta \bar{S}_{110}$,
and one of them spans $T_\theta S_{010}$,
accounting for all three strata that meet at $\theta$.
The lines reflect three degrees of freedom by which paths on the fiber can leave
$\theta$, though no single path can exploit more than two degrees of freedom. 

The same three lines form the fiber prebasis for any point on
$S_{001}$, $S_{100}$, or the origin $S_{000}$
(albeit with different indices for the one-matrix subspaces).
But at a point $\theta' \in S_{011}$, the fiber prebasis has only two lines,
one parallel to~$S_{001}$ and one parallel to~$S_{010}$
($\phi_{22100}$ and $\phi_{22200}$).
On the second fiber (Figure~\ref{fbasis}, right),
at the origin ${\bf 0}$ on the stratum $S_{00}$,
the fiber prebasis (not illustrated) is
a set containing one plane ($\phi_{10110}$, parallel to $S_{01}$) and
one line ($\phi_{21221}$, parallel to $S_{10}$).

In general, the fiber prebasis is different at different points, even
on the same stratum.
For example, on a curved stratum $S$, $T_\theta S$ changes as
one moves along $S$, and the one-matrix subspaces change accordingly.
A more subtle example appears at right in Figure~\ref{fbasis}, where
the points $\theta'', \theta''' \in S_{01}$ have different fiber prebases.
At both points, the fiber prebasis contains
a radial line $\phi_{11100}$ that passes through both the origin $S_{00}$ and
the defining point ($\theta''$ or $\theta'''$), so
$\phi_{11100}$ is not the same at $\theta''$ and~$\theta'''$.
At each point on $S_{01}$, the fiber basis also contains
a second line $\phi_{10100}$, which we can choose to be
any line on $S_{01}$ except $\phi_{11100}$ (because
$o_{10100} = a_{110} \otimes b_{000}$ and
$b_{000}$ can be any one-dimensional subspace of $\R^2$ except $\row W_1$).


In this section, we also study the subspace spanned by the fiber prebasis.
This subspace is the nullspace of the differential map of $\mu(\theta)$,
which we define and explain in Section~\ref{diffmap}.
This differential map is written $\dmu(\theta)$ and
its nullspace is written $\Null \dmu(\theta)$.
The significance of the differential map is that every smooth path on the fiber
leaving~$\theta$ is necessarily tangent to $\Null \dmu(\theta)$.
However, {\em not} every direction in $\Null \dmu(\theta)$ is  
necessarily tangent to a smooth path on the fiber; only some directions are.  
For example, in Figure~\ref{fbasis}~(left), $\Null \dmu(\theta) = \R^3$ 
for every~$\theta \in S_{010}$, reflecting the three degrees of freedom, but
many directions in $\R^3$ are not tangent to
any path leaving $\theta$ on the fiber.

But if we take all the vectors tangent to smooth paths leaving $\theta$
on the fiber, their vector sum is $\Null \dmu(\theta)$
(see Corollary~\ref{freedomspan2} in Section~\ref{proofbasis}).
This implies that $T_\theta S \subseteq \Null \dmu(\theta)$
(Lemma~\ref{TSnulldmu});
it also implies that if $S'$ is a stratum such that $\theta \in \bar{S}'$
(equivalently, $S \subseteq \bar{S}'$) and $T_\theta \bar{S}'$~exists, then
$T_\theta \bar{S}' \subseteq \Null \dmu(\theta)$.
For example, at the point $\theta$ in Figure~\ref{fbasis},
$T_\theta S_{010} + T_\theta \bar{S}_{011} + T_\theta \bar{S}_{110} = \R^3 =
\Null \dmu(\theta)$, whereas at the point $\theta'$,
$\Null \dmu(\theta')$ is the plane parallel to $S_{011}$ and
$T_{\theta'} S_{011} = \Null \dmu(\theta')$.  

We say that the number of {\em degrees of freedom} on the fiber at~$\theta$ is
the dimension of $\Null \dmu(\theta)$, because
it is the maximum number of linearly independent directions along which
smooth paths on the fiber can leave $\theta$ (though
a path cannot necessarily use all these degrees of freedom simultaneously).
There are three degrees of freedom at every $\theta \in S_{001}$, but
only two at every $\theta' \in S_{011}$.

At a weight vector $\theta$ on a stratum $S$,
we will construct three prebases (besides the one-matrix prebasis):
one that spans $T_\theta S$, and
two that span $\Null \dmu(\theta)$.
(Unlike the one-matrix prebasis, they usually do not span
the entire weight space~$\R^{d_\theta}$.)
\begin{itemize}
\item
The {\em freedom prebasis} spans $\Null \dmu(\theta)$.
The freedom prebasis
contains all the one-matrix subspaces that lie in $\Null \dmu(\theta)$
(i.e., subspaces whose moves stay on the fiber) and
excludes all the one-matrix subspaces that do not
(i.e., their moves leave the fiber).
It also contains some two-matrix subspaces that represent
directions tangent to curved paths on $S$,
as discussed in Section~\ref{2matrix}.

\item
The {\em stratum prebasis} spans $T_\theta S$.
The stratum prebasis
contains all the one-matrix subspaces that lie in $T_\theta S$
(i.e., subspaces whose moves stay on $\bar{S}$) and
excludes all the one-matrix subspaces that do not
(i.e., their moves leave $\bar{S}$).
Thus it omits all the one-matrix subspaces that the freedom prebasis omits, and
it also omits all the subspaces associated with barcode moves.
It contains all the two-matrix subspaces in the freedom prebasis and adds
some more.

\item
The {\em fiber prebasis} is a superset of the stratum prebasis
that spans $\Null \dmu(\theta)$.
The fiber prebasis has fewer one-matrix subspaces and more two-matrix subspaces
than the freedom prebasis, so it is a bit less elegant.
But because it includes the stratum prebasis,
the fiber prebasis is better aligned with the fiber:
for every stratum $S'$ whose closure contains $\theta$,
if $T_\theta \bar{S}'$ exists then
some subset of the fiber prebasis spans $T_\theta \bar{S}'$.
Thus the fiber prebasis is a single prebasis whose subspaces can be used to span
the tangent spaces of the strata that meet at $\theta$.
It clarifies how the strata meet each other at~$\theta$.
\end{itemize}

Although $T_\theta S \subseteq \Null \dmu(\theta)$ and
the freedom prebasis spans $\Null \dmu(\theta)$, the freedom prebasis does not,
in general, have a subset whose span is $T_\theta S$.
The fiber prebasis corrects that deficiency.

Before we can formally define the three prebases (in Section~\ref{fiberbasis}),
we must introduce the differential map $\dmu$ (in Section~\ref{diffmap}) and
the two-matrix subspaces
(in Sections~\ref{2matrixsubspaces}--\ref{2matrixsubspacesuseful}).

\subsection{The Differential Map $\dmu$ and its Nullspace}
\label{diffmap}

Imagine you are standing at a point $\theta$ on a fiber $\mu^{-1}(W)$.
As $\mu$ is a polynomial function of $\theta$, $\mu$ is smooth, but
the fiber might not be locally manifold at $\theta$.
Nevertheless, if you walk on a path on the fiber starting from $\theta$,
your initial direction of motion $\Delta \theta$ is necessarily one along which
the directional derivative $\mu'_{\Delta \theta}(\theta)$ is zero.
(Note that this directional derivative is a vector;
to say it is zero is to say all of its components are zero.)
But the converse does not hold---not every direction with derivative zero
necessarily is associated with some path on the fiber!
For example, if you are standing at $\theta$ in
Figure~\ref{fbasis}, the directional derivative of $\mu$ is zero
for {\em every} direction in weight space, but
only some directions stay on the fiber.

To better understand these derivatives,
Trager, Kohn, and Bruna~\cite{trager20} use differential geometry.
Given a neural network architecture
$\mu : \R^{d_\theta} \rightarrow \R^{d_L \times d_0}$ and
a specified weight vector $\theta \in \R^{d_\theta}$, the {\em differential map}
$\dmu(\theta) : \R^{d_\theta} \rightarrow \R^{d_L \times d_0}$ is a linear map
from weight space to the space of $d_L \times d_0$ matrices.
We emphasize the linearity; think of the differential map as
the linear term in a Taylor expansion of~$\mu$ about~$\theta$.
Usually we will write its argument as $\Delta \theta$, and apply the map as
$\Delta W = \dmu(\theta)(\Delta \theta)$.
The notations $\Delta W$ and $\Delta \theta$ reflect
a natural interpretation in terms of perturbations:
if you are at a point $\theta$ in weight space,
yielding a matrix $W = \mu(\theta)$, then
you perturb $\theta$ by an infinitesimal displacement $\Delta \theta$,
the matrix $W$ is perturbed by an infinitesimal $\Delta W$.

The bare form $\dmu$ denotes a map from a weight vector $\theta$ to
a linear map.
This might be confusing if you haven't seen it before---a map that produces
a map---and it accounts for the odd notation $\dmu(\theta)(\Delta \theta)$.

Let $\Delta \theta = (\Delta W_L, \Delta W_{L - 1}, \ldots, \Delta W_1) \in
\R^{d_\theta}$ be a weight displacement.
By the product rule of calculus,
the value of the differential map for $\mu$ at a fixed weight vector
$\theta = (W_L, W_{L - 1}, \ldots, W_1) \in \R^{d_\theta}$ is
\begin{equation}
\dmu(\theta)(\Delta \theta) =
\sum_{j=1}^L W_{L \ssq j} \Delta W_j W_{j-1 \ssq 0} =
\Delta W_L W_{L-1 \ssq 0} + W_L \Delta W_{L-1} W_{L-2 \ssq 0} + \ldots +
W_{L \ssq 1} \Delta W_1.
\label{dmu}
\end{equation}

With $\Delta \theta$ fixed, this is the directional derivative
$\mu'_{\Delta \theta}(\theta) = \dmu(\theta)(\Delta \theta)$.
But for our purposes, we like to think of $\theta$ as fixed and
$\Delta \theta$ as varying.
Observe that~(\ref{dmu}) is
linear in $\Delta \theta$ but not linear in $\theta$.

As $\mu$ is continuous and smooth,
if you walk from $\theta$ along a smooth path on the fiber $\mu^{-1}(W)$,
your initial direction of motion $\Delta \theta$ has
directional derivative zero; so
your initial direction is in the nullspace of $\dmu(\theta)$, defined to be
\[
\Null \dmu(\theta) =
\{ \Delta \theta \in \R^{d_\theta} : \dmu(\theta)(\Delta \theta) = 0 \}.
\]

The nullspace $\Null \dmu(\theta)$ is
the tangent space of $\mu^{-1}(W)$ at $\theta$ {\em if} the latter is defined.
The tangent space is defined at $\theta$ if
the fiber is locally a smooth manifold at $\theta$
(i.e., if some open neighborhood $N \subset \mu^{-1}(W)$ of $\theta$ is
homeomorphic to a ball).
Typically, there are singular points on the fiber where
the tangent space is not defined.
Think of $\Null \dmu(\theta)$ as a generalization of the tangent space
that is defined everywhere on $\mu^{-1}(W)$, even at singular points.
It encompasses all tangencies of smooth paths on the fiber at $\theta$.


\begin{lemma}
\label{TSnulldmu}
Let $\theta \in \R^{d_\theta}$ be a weight vector in a stratum $S$
in the rank stratification of a fiber $\mu^{-1}(W)$.
Then every smooth path containing~$\theta$ on the fiber is
tangent to $\Null \dmu(\theta)$ at $\theta$.
Moreover, $T_\theta S \subseteq \Null \dmu(\theta)$.
\end{lemma}

\begin{proof}
Let $P \subset \mu^{-1}(W)$ be a smooth path on the fiber.
Every point $\zeta \in P$ satisfies $\mu(\zeta) = W$.
As $\mu$~is an analytic function of $\R^{d_\theta}$,
any vector $\Delta \zeta$ tangent to $P$ at $\zeta$ is
a direction in which $\mu$ has a directional derivative of zero---that is,
$\mu'_{\Delta \zeta}(\zeta) = \dmu(\zeta)(\Delta \zeta) = 0$.
Hence $\Delta \zeta \in \Null \dmu(\zeta)$, verifying that
every smooth path containing~$\theta$ on the fiber is
tangent to $\Null \dmu(\theta)$ at $\theta$.

As $S$ is a smooth manifold by Theorem~\ref{stratummanifold},
for every nonzero vector $\Delta \theta \in T_\theta S$,
there is a smooth path leaving~$\theta$ on $S$ that is
tangent to $\Delta \theta$ at $\theta$.
Therefore, $T_\theta S \subseteq \Null \dmu(\theta)$.
%
\end{proof}

Recall from Section~\ref{1matrix} that $\ThetaO^\fiber$ contains
all the one-matrix subspaces whose displacements stay on the fiber, and
$\ThetaO^\off$ contains
all the one-matrix subspaces whose displacements do not stay on the fiber.
The displacements in $\ThetaO^\fiber$ lie in $\Null \dmu(\theta)$, as
a direct corollary to the first claim of Lemma~\ref{TSnulldmu}.
We now prove it a second way, which also permits us to show that
the nonzero displacements in $\ThetaO^\off$ are not in $\Null \dmu(\theta)$:
by plugging the displacements into the formula~(\ref{dmu}).

\begin{lemma}
\label{1dmu}
Every subspace $\phi_{lkjih} \in \ThetaO^\fiber$ satisfies
$\phi_{lkjih} \subseteq \Null \dmu(\theta)$.
Moreover, $\Span \ThetaO^\fiber \subseteq \Null \dmu(\theta)$.
By~contrast, every subspace $\phi_{lkjih} \in \ThetaO^\off$ satisfies
$\phi_{lkjih} \cap \Null \dmu(\theta) = \{ {\bf 0} \}$.
(That is, $\phi_{lkjih}$ and $\Null \dmu(\theta)$ are linearly independent.)
\end{lemma}

\begin{proof}
Consider a displacement
$\Delta \theta \in \phi_{lkjih} = a_{lji} \otimes b_{k,j-1,h}$.
We can write $\Delta\theta = (0, \ldots, 0, \Delta W_j, 0, \ldots, 0)$.

If $\phi_{lkjih} \in \ThetaO^\fiber$, then
either $L > l$ or $h > 0$ by the definition of~$\ThetaO^\fiber$.
In the former case, $W_{L \ssq j} \Delta W_j = 0$ because
$\col \Delta W_j \subseteq a_{lji} \subseteq A_{lji}
\subseteq \Null W_{l+1 \ssq j} \subseteq \Null W_{L \ssq j}$.
In~the latter case, $\Delta W_j W_{j-1 \ssq 0} = 0$ because
$\row \Delta W_j \subseteq b_{k,j-1,h} \subseteq B_{k,j-1,h}
\subseteq \Null W_{j-1 \ssq h-1}^\top \subseteq \Null W_{j-1 \ssq 0}^\top$.
In both cases, by the formula~(\ref{dmu}),
$\dmu(\theta)(\Delta \theta) = W_{L \ssq j} \Delta W_j W_{j-1 \ssq 0} = 0$.
Hence $\phi_{lkjih} \subseteq \Null \dmu(\theta)$.

It follows that $\Span \ThetaO^\fiber \subseteq \Null \dmu(\theta)$ because
$\Null \dmu(\theta)$ is a subspace of $\R^{d_\theta}$ and
every subspace in $\ThetaO^\fiber$ is a subset of $\Null \dmu(\theta)$.

By contrast, if $\phi_{lkjih} \in \ThetaO^\off$, then
$L = l$ and $h = 0$ by the definition of~$\ThetaO^\off$, so
$\Delta \theta \in a_{Lji} \otimes b_{k,j-1,0}$.
By Lemma~\ref{dimpreserved}, $W_{L \ssq j} a_{Lji}$ has
the same dimension as $a_{Lji}$; therefore,
$W_{L \ssq j}$ induces a linear bijection from $a_{Lji}$ to~$a_{LLi}$.
Also by Lemma~\ref{dimpreserved}, $W_{j-1 \ssq 0}^\top b_{k,j-1,0}$ has
the same dimension as $b_{k,j-1,0}$, thus
$W_{j-1 \ssq 0}^\top$ induces a linear bijection from $b_{k,j-1,0}$ to $b_{k00}$.
Hence $\Delta W_j \mapsto W_{L \ssq j} \Delta W_j W_{j-1 \ssq 0}$ is
a linear bijection from $a_{Lji} \otimes b_{k,j-1,0}$ to $a_{LLi} \otimes b_{k00}$
that maps zero to zero.
It follows that if $\Delta \theta \neq {\bf 0}$, then
$\dmu(\theta)(\Delta \theta) = W_{L \ssq j} \Delta W_j W_{j-1 \ssq 0} \neq 0$.
Hence $\phi_{lkjih} \cap \Null \dmu(\theta) = \{ {\bf 0} \}$ as claimed.
\end{proof}

Compare Lemma~\ref{1dmu} to Corollary~\ref{ofiber}.
Our partition of $\ThetaO$ into $\ThetaO^\fiber$ and $\ThetaO^\off$ separates
the one-matrix subspaces into those in $\Null \dmu(\theta)$ and
those linearly independent of $\Null \dmu(\theta)$ (Lemma~\ref{1dmu});
it also separates the one-matrix subspaces into
those whose displacements stay on the fiber and
those whose displacements move off the fiber (Corollary~\ref{ofiber}).
This is worth remarking on, because as we have said,
not all displacements in $\Null \dmu(\theta)$ are tangent to
paths leaving $\theta$ on the fiber.
But all displacements {\em in all one-matrix subspaces} in $\Null \dmu(\theta)$
(i.e., the subspaces in $\ThetaO^\fiber$)
are tangent to straight paths leaving $\theta$ on the fiber,
as Figure~\ref{fbasis} shows.

\subsection{The Hidden Tangent Subspace}
\label{hidden}

Recall the {\em two-matrix paths} we defined in Section~\ref{2matrix}, which
start at a weight vector $\theta$ and
lie on the stratum~$S$ that contains~$\theta$.
Two-matrix paths are often curved, occasionally straight, but always smooth.
Hence the direction by which a two-matrix path leaves $\theta$ is
tangent to~$S$.
The {\em hidden tangent subspace} $\tau \subset \R^{d_\theta}$ is
the subspace spanned by
the vectors tangent to all the two-matrix paths at $\theta$.
The significance of $\tau$ is that $\tau \subseteq T_\theta S$, and
we will see that $T_\theta S = \tau + \Span \ThetaO^\strat$ and
$\Null \dmu(\theta) = \tau + \Span \ThetaO^\fiber$, enabling us to write
explicit expressions for $T_\theta S$ and $\Null \dmu(\theta)$
in Section~\ref{expressions} (also summarized in Table~\ref{resulttable}).

Recall that a two-matrix move can be interpreted as
a change in the coordinate system of some layer of hidden units.
We call $\tau$ the hidden tangent subspace because
it represents all the tangencies to $S$ at $\theta$ associated with
infinitesimal perturbations of the coordinate systems of the hidden layers.

Let $P$ be a two-matrix path~(\ref{2matrixpath})
(specified by some $H \in \R^{d_j \times d_j}$)
with endpoint $\theta = (W_L, W_{L - 1}, \ldots, W_1)$ on a stratum~$S$.
Recall from~(\ref{TPtheta}) that the line tangent to $P$ at $\theta$ is
\[
T_\theta P =
\{ (0, 0, \ldots, 0, \gamma W_{j+1} H, - \gamma H W_j, 0, \ldots, 0) :
   \gamma \in \R \}.
\]
As $P \subset S$, $T_\theta P$ is tangent to $S$ at $\theta$.

The hidden tangent subspace $\tau$ is the vector sum of
the tangent lines of all the two-matrix paths one can define at $\theta$.
Thus
\begin{eqnarray}
\tau & = & \sum_{j=1}^{L-1}
\left\{ (0, \ldots, 0, \underbrace{W_{j+1} H_j}_{\Delta W_{j+1}},
         \underbrace{- H_j W_j}_{\Delta W_j}, 0, \ldots, 0) :
H_j \in \R^{d_j \times d_j} \right\}  \nonumber  \\
& = &
\left\{ (W_L H_{L-1}, W_{L-1} H_{L-2} - H_{L-1} W_{L-1}, \ldots,
         W_j H_{j-1} - H_j W_j, \ldots, W_2 H_1 - H_2 W_2, - H_1 W_1) : \right.
         \nonumber  \\
&   &    \left. H_j \in \R^{d_j \times d_j} \right\}.
\label{tauexplicit}
\end{eqnarray}

\begin{lemma}
\label{tauinTS}
Let $\theta$ be a weight vector on a stratum $S$, and
let $\tau$ be the hidden tangent subspace at $\theta$.
Then $\tau \subseteq T_\theta S$.
\end{lemma}

\begin{proof}
Every two-matrix path $P$ containing $\theta$ is smooth and satisfies
$P \subset S$.
Hence every tangent line $T_\theta P$ satisfies $T_\theta P \subseteq T_\theta S$.
The subspace $\tau$ is a vector sum of such tangent lines, so
it is a subspace of $T_\theta S$.
\end{proof}

An immediate corollary of Lemmas~\ref{tauinTS} and~\ref{TSnulldmu} is that
$\tau \subseteq \Null \dmu(\theta)$.
We now prove it a second way
(as we did for $\Span \ThetaO^\fiber$ in Lemma~\ref{1dmu}):
by plugging the displacements into the formula~(\ref{dmu}) and showing that
the result is always zero.
We see in the following lemma that
two terms in the summation~(\ref{dmu}) are nonzero and cancel each other out.
Contrast this with Lemma~\ref{1dmu}:  every term is zero for
a one-matrix move on the fiber.

\begin{lemma}
\label{2dmu}
Every line $T_\theta P$ tangent to a two-matrix path $P$ at
a point $\theta \in P$ satisfies $T_\theta P \subseteq \Null \dmu(\theta)$.
Hence $\tau \subseteq \Null \dmu(\theta)$.
\end{lemma}

\begin{proof}
Substituting a displacement $\Delta \theta \in T_\theta P$ into
the formula~(\ref{dmu}) gives
\[
\dmu(\theta)(\Delta \theta) = \gamma W_{L \ssq j+1} \Delta W_{j+1} W_{j \ssq 0} +
\gamma W_{L \ssq j} \Delta W_j W_{j-1 \ssq 0} =
\gamma W_{L \ssq j+1} W_{j+1} H W_{j \ssq 0} -
\gamma W_{L \ssq j} H W_j W_{j-1 \ssq 0} = 0.
\]
Hence $\Delta \theta \in \Null \dmu(\theta)$
for all $\Delta \theta \in T_\theta P$, hence
$T_\theta P \subseteq \Null \dmu(\theta)$.
The subspace $\tau$ is a vector sum of such tangent lines, so
it is a subspace of $\Null \dmu(\theta)$.
\end{proof}

\subsection{Two-Matrix Subspaces}
\label{2matrixsubspaces}

Usually $\ThetaO^\fiber$ does not suffice to span $\Null \dmu(\theta)$, and
$\ThetaO^\strat$ does not suffice to span $T_\theta S$.
To give a complete prebasis for $\Null \dmu(\theta)$ and
a complete prebasis for $T_\theta S$, we add some {\em two-matrix subspaces}.
Every two-matrix subspace is a subspace of the hidden tangent subspace $\tau$,
and together they span $\tau$.

Like $\tau$ and the one-matrix subspaces,
the two-matrix subspaces are defined at
a particular weight vector $\theta = (W_L, W_{L - 1}, \ldots, W_1)$ and
may be different at other weight vectors.

Recall the matrices $H_j \in \R^{d_j \times d_j}$ in
the formula~(\ref{tauexplicit}) for $\tau$.
Recall that $\mathcal{A}_j$ and $\mathcal{B}_j$ are prebases for $R^{d_j}$.
A~fruitful way to construct a prebasis for $\R^{d_j \times d_j}$ is to pair
subspaces in $\mathcal{A}_j$ with subspaces in $\mathcal{B}_j$.
That prebasis is
\[
\{ a_{lji} \otimes b_{kjh} : l, k \in [j, L], i, h \in [0, j] \},
\]
where $a_{lji}$ and $b_{kjh}$ are flow prebasis subspaces,
as defined in Section~\ref{flowrelations}.
Recall that $H_j \in a_{lji} \otimes b_{kjh}$ means that
$\col H_j \subseteq a_{lji}$ and $\row H_j \subseteq b_{kjh}$.

For all $j \in [1, L - 1]$, $l, k \in [j, L]$, and $i, h \in [0, j]$,
we define a {\em two-matrix subspace}
\begin{equation}
\tau_{lkjih} =
\{ (0, \ldots, 0, \underbrace{W_{j+1} H}_{\Delta W_{j+1}},
   \underbrace{- H W_j}_{\Delta W_j}, 0, \ldots, 0) :
   H \in a_{lji} \otimes b_{kjh} \}.
\label{2matrixsubspace}
\end{equation}
We choose flow prebasis subspaces (rather than, say, the standard prebases)
so that the following properties hold.

\begin{lemma}
\label{2to1}
If we use flow prebases (associated with a point $\theta$),
each displacement $\Delta \theta \in \tau_{lkjih}$ satisfies
\begin{eqnarray*}
\Delta W_{j+1} & = & W_{j+1} H \in
\left\{ \begin{array}{ll}
a_{l,j+1,i} \otimes b_{kjh} = o_{l,k,j+1,i,h}, & l > j  \\
\{ 0 \},                                   & l = j
\end{array} \right.
\hspace{.2in}  \mbox{and}  \\
\Delta W_j & = & - H W_j \in
\left\{ \begin{array}{ll}
a_{lji} \otimes b_{k,j-1,h} = o_{lkjih}, & j > h  \\
\{ 0 \},                              & j = h.
\end{array} \right.
\end{eqnarray*}
for some $H \in a_{lji} \otimes b_{kjh}$.
Moreover,
\begin{icompact}
\item  if $l = j = h$, then $\tau_{lkjih} = \{ {\bf 0} \}$.
\item  If $l > j = h$, then $\tau_{lkjih} = \phi_{l,k,j+1,i,h}$.
\item  If $l = j > h$, then $\tau_{lkjih} = \phi_{lkjih}$.
\item  If $l > j > h$, then no two displacements in $\tau_{lkjih}$ have
  the same value of $\Delta W_{j+1}$, nor the same value of $\Delta W_j$.
  Hence, each displacement in $\tau_{lkjih}$ is a sum of
  a unique member of $\phi_{l,k,j+1,i,h}$ and a unique member of $\phi_{lkjih}$.
  Moreover, in every nonzero $\Delta \theta \in \tau_{lkjih}$,
  both $\Delta W_{j+1}$ and $\Delta W_j$ are nonzero.
\end{icompact}

The dimension of $\tau_{lkjih}$ is
$(\dim a_{lji}) \cdot (\dim b_{kjh}) = \omega_{li} \, \omega_{kh}$
in all cases except $l = j = h$, for which $\dim \tau_{lkjih} = 0$.
\end{lemma}

\begin{proof}
If $l = j$ then $W_{j+1} H = 0$
because $\col H \subseteq a_{jji} \subseteq A_{jji} \subseteq \Null W_{j+1}$.
Whereas if $l > j$, then $W_{j+1} H \in a_{l,j+1,i} \otimes b_{kjh}$ because
$H \in a_{lji} \otimes b_{kjh}$ and
we are using flow prebases, so $W_{j+1} a_{lji} = a_{l,j+1,i}$.
Symmetrically, if $j = h$ then $- H W_j = 0$
because $\row H \subseteq b_{kjj} \subseteq B_{kjj} \subseteq \Null W_j^\top$.
Whereas if $j > h$, then $- H W_j \in a_{lji} \otimes b_{k,j-1,h}$ because
for flow prebases, $W_j^\top b_{kjh} = b_{k,j-1,h}$.

It follows immediately that
$\tau_{lkjih} = \{ {\bf 0} \}$ if $l = j = h$ (as claimed),
$\tau_{lkjih} \subseteq \phi_{l,k,j+1,i,h}$ if $l > j = h$, and
$\tau_{lkjih} \subseteq \phi_{lkjih}$ if $l = j > h$.

If $l > j$, then $W_{j+1} a_{lji}$ has the same dimension as $a_{lji}$
by Lemma~\ref{dimpreserved}; therefore,
$H \mapsto W_{j+1} H$ is a linear bijection from
$a_{lji} \otimes b_{kjh}$ to $a_{l,j+1,i} \otimes b_{kjh}$, and
the map used in~(\ref{2matrixsubspace}) is a linear bijection
from matrices in $a_{lji} \otimes b_{kjh}$ to displacements in $\tau_{lkjih}$.
Hence no two displacements in $\tau_{lkjih}$ have  
the same value of $\Delta W_{j+1}$, as claimed; and
if $l > j = h$, then $\tau_{lkjih} = \phi_{l,k,j+1,i,h}$ as claimed.

Symmetrically, if $j > h$, then $W_j^\top b_{kjh}$ has
the same dimension as $b_{kjh}$ by Lemma~\ref{dimpreserved}; therefore,
$H \mapsto -H W_j$ is a linear bijection
from $a_{lji} \otimes b_{kjh}$ to $a_{lji} \otimes b_{k,j-1,h}$, and again
the map used in~(\ref{2matrixsubspace}) is a linear bijection
from matrices in $a_{lji} \otimes b_{kjh}$ to displacements in $\tau_{lkjih}$.
Thus no two displacements in $\tau_{lkjih}$ have  
the same value of~$\Delta W_j$, as claimed; and
if $l = j > h$, then $\tau_{lkjih} = \phi_{lkjih}$, as claimed.

If $l > j > h$, it follows that $\Delta W_{j+1} = 0$ only for $H = 0$, and
likewise that $\Delta W_j = 0$ only for $H = 0$.
This verifies that in every nonzero $\Delta \theta \in \tau_{lkjih}$,
both $\Delta W_{j+1}$ and $\Delta W_j$ are nonzero.

If $l > j$ or $j > h$, the linear bijection in~(\ref{2matrixsubspace}) implies
that $\tau_{lkjih}$ has the same dimension as $a_{lji} \otimes b_{kjh}$, which is
$(\dim a_{lji}) \cdot (\dim b_{kjh})$, as claimed.
\end{proof}

Lemma~\ref{2to1} shows that some of the two-matrix subspaces are
one-matrix subspaces or the trivial subspace~$\{ {\bf 0} \}$.
We are primarily interested in the other ones---the two-matrix subspaces
with $l > j > h$, typically having two nonzero matrices.
Sometimes a subspace $\tau_{lkjih}$ with $l > j > h$ can be
the trivial subspace~$\{ {\bf 0} \}$, but
only if $a_{lji} = \{ {\bf 0} \}$ or $b_{kjh} = \{ {\bf 0} \}$.
We define two sets:
the set $\ThetaTO$ of all two-matrix subspaces except~$\{ {\bf 0} \}$
(typically including some one-matrix subspaces), and
the set $\ThetaT$ of two-matrix subspaces for which $l > j > h$
except~$\{ {\bf 0} \}$ (excluding all the one-matrix subspaces).
\begin{eqnarray*}
\ThetaT  & = & \{ \tau_{lkjih} \neq \{ {\bf 0} \} :
                  L \geq l > j > h \geq 0  \mbox{~and~}
                  L \geq k \geq j \geq i \geq 0 \}.  \\
\ThetaTO & = & \{ \tau_{lkjih} \neq \{ {\bf 0} \} :
                  L > j > 0, L \geq l \geq j \geq h \geq 0,  \mbox{~and~}
                  L \geq k \geq j \geq i \geq 0 \} \supseteq \ThetaT.
\end{eqnarray*}

For any subspace $\tau_{lkjih} \in \ThetaT$,
Lemma~\ref{2to1} shows that each displacement in $\tau_{lkjih}$ is
a sum of two one-matrix displacements in $\phi_{l,k,j+1,i,h}$ and $\phi_{lkjih}$.
More simply,
\[
\tau_{lkjih} \subseteq \phi_{lkjih} + \phi_{l,k,j+1,i,h}.
\]
We find this elegant.
This fact motivates why we choose~(\ref{2matrixsubspace}) to define
the two-matrix subspaces:
it will make it easy to prove the linear independence of the subspaces in
the freedom, stratum, and fiber prebases.

\begin{lemma}
\label{tauThetaTO}
$\tau = \Span \ThetaTO$.
\end{lemma}

\begin{proof}
As $\{ a_{lji} \otimes b_{kjh} : l, k \in [j, L], i, h \in [0, j] \}$ is
a prebasis for $\R^{d_j \times d_j}$, by~(\ref{tauexplicit}),
\begin{eqnarray*}
\tau & = & \sum_{j=1}^{L-1}
\left\{ (0, \ldots, 0, \underbrace{W_{j+1} H_j}_{\Delta W_{j+1}},
         \underbrace{- H_j W_j}_{\Delta W_j}, 0, \ldots, 0) :
H_j \in \sum_{l=j}^L \sum_{k=j}^L \sum_{i=0}^j \sum_{h=0}^j a_{lji} \otimes b_{kjh}
\right\}  \\
& = & \sum_{j=1}^{L-1} \sum_{l=j}^L \sum_{k=j}^L \sum_{i=0}^j \sum_{h=0}^j
\tau_{lkjih}  \\
& = & \Span \ThetaTO.
\qedhere
\end{eqnarray*}
\end{proof}

\begin{cor}
\label{2strat}
Every two-matrix subspace $\tau_{lkjih} \in \ThetaTO$ satisfies
$\tau_{lkjih} \subseteq T_\theta S \subseteq \Null \dmu(\theta)$.
Moreover, $\Span \ThetaT \subseteq \Span \ThetaTO \subseteq T_\theta S
\subseteq \Null \dmu(\theta)$.
\end{cor}

\begin{proof}
Follows immediately from
Lemmas~\ref{tauThetaTO}, \ref{tauinTS}, and~\ref{TSnulldmu}.
\end{proof}

Note that $\Span \ThetaTO$ does not depend on
our specific choices of the prebasis subspaces $a_{lji}$ and~$b_{kjh}$
(it is always $\tau$), but $\Span \ThetaT$ does depend on those choices, and
we cannot write a formula for $\Span \ThetaT$ that is independent of them.

In Section~\ref{proofbasis}, we will prove that
the set of directions by which a smooth path can leave $\theta$
on the fiber $\mu^{-1}(W)$ spans $\Null \dmu(\theta)$.
For that purpose, we will need the following lemma.

\begin{lemma}
\label{2stratpath}
For every two-matrix subspace $\tau_{lkjih} \in \ThetaTO$ and
every nonzero displacement $\Delta \theta \in \tau_{lkjih}$, there exists
a smooth two-matrix path $P \subset S$ that leaves $\theta$
in the direction $\Delta \theta$.
\end{lemma}

\begin{proof}
By the definition~(\ref{2matrixsubspace}) of $\tau_{lkjih}$,
$\Delta \theta = (0, \ldots, 0, W_{j+1} H, - H W_j, 0, \ldots, 0)$
for some $H \in a_{lji} \otimes b_{kjh}$.
Consider the two-matrix path
\[
P = \{ \theta'(\epsilon) : \epsilon \in [0, \hat{\epsilon}] \}
\hspace{.2in}  \mbox{where}  \hspace{.2in}
\theta'(\epsilon) = (W_L, \ldots, W_{j+2}, W_{j+1} (I + \epsilon H),
                  (I + \epsilon H)^{-1} W_j, W_{j-1}, \ldots, W_1)
\]
and $\hat{\epsilon} > 0$ is sufficiently small that
$I + \epsilon H$ is invertible for all $\epsilon \in [0, \hat{\epsilon}]$.
The path $P$ is connected and smooth, and $\theta$ is one of its endpoints.
It satisfies $P \subset S$, as all points $\theta' \in P$
satisfy $\mu(\theta') = \mu(\theta)$ and
have the same rank list as $\theta$.
Recall from Section~\ref{2matrix} that
$(\di\theta'/\di\epsilon)|_{\epsilon = 0} = \Delta \theta$.
The result follows.
\end{proof}

\subsection{The Two-Matrix Subspaces We Care about Most}
\label{2matrixsubspacesuseful}

Which two-matrix subspaces are useful in a basis for $T_\theta S$?
Consider a two-matrix subspace $\tau_{lkjih} \in \ThetaT$.
Recall again that Lemma~\ref{2to1} shows that
$\tau_{lkjih} \subseteq \phi_{lkjih} + \phi_{l,k,j+1,i,h}$.
If $\phi_{lkjih} \subseteq T_\theta S$ and
$\phi_{l,k,j+1,i,h} \subseteq T_\theta S$, then
$\tau_{lkjih}$ adds nothing useful to the one-matrix subspaces.
But if $\phi_{lkjih} \not\subseteq T_\theta S$ and
$\phi_{l,k,j+1,i,h} \not\subseteq T_\theta S$, then
$\tau_{lkjih}$ is interesting and useful, because
$\tau_{lkjih} \subseteq T_\theta S$ (by Corollary~\ref{2strat}).
If, moreover, $\phi_{lkjih} \not\subseteq \Null \dmu(\theta)$ and
$\phi_{l,k,j+1,i,h} \not\subseteq \Null \dmu(\theta)$,
$\tau_{lkjih}$ is even more interesting and useful, because
$\tau_{lkjih} \subseteq \Null \dmu(\theta)$. 

As every two-matrix subspace satisfies
$\tau_{lkjih} \subseteq T_\theta S \subseteq \Null \dmu(\theta)$,
we will use $\tau_{lkjih}$ in the freedom, stratum, and fiber prebases if
$\phi_{lkjih}$ and $\phi_{l,k,j+1,i,h}$ do not lie in $\Null \dmu(\theta)$; and
we will use $\tau_{lkjih}$ in the stratum and fiber prebases if
$\phi_{lkjih}$ and $\phi_{l,k,j+1,i,h}$ do not lie in $T_\theta S$.
Recall from Corollary~\ref{ofiber} in Section~\ref{1matrixprebasis} that
$\phi_{lkjih} \not\subseteq \Null \dmu(\theta)$ if and only if
$\phi_{lkjih} \in \ThetaO^\off$; equivalently, $l = L$ and $h = 0$.
Recall from Sections~\ref{onofffiber} and~\ref{barcmoves} that
$\phi_{lkjih} \not\subseteq T_\theta S$ if and only if
$\phi_{lkjih} \in \ThetaO^\barc$ or $\phi_{lkjih} \in \ThetaO^\off$; and
$\phi_{lkjih} \in \ThetaO^\barc$ if and only if $l > k$ and $i > h$.
Hence we define the sets
\begin{eqnarray*}
\ThetaT^\off & = & \{ \tau_{lkjih} \in \ThetaT : l = L \mbox{~and~} h = 0 \}
                =   \{ \tau_{Lkji0} \neq \{ {\bf 0} \} :
                        L > j > 0 \mbox{~and~} L \geq k \geq j \geq i \geq 0 \}
\hspace{.2in}  \mbox{and}  \\
\ThetaT^\barc & = & \{ \tau_{lkjih} \in \ThetaT : l > k \mbox{~and~} i > h \}
                 = \{ \tau_{lkjih} \neq \{ {\bf 0} \} :
                      L \geq l > k \geq j \geq i > h \geq 0 \}.
\end{eqnarray*}
For example, in the two-matrix case ($L = 2$),
\begin{eqnarray*}
\ThetaT^\off      =
\ThetaT        & = & \{ \tau_{21100}, \tau_{21110}, \tau_{22100}, \tau_{22110} \}
                        \setminus \{ \{ {\bf 0} \} \}
\hspace{.2in}  \mbox{and}  \\
\ThetaT^\barc  & = & \{ \tau_{21110} \} \setminus \{ \{ {\bf 0} \} \}.
\end{eqnarray*}

The freedom, stratum, and fiber prebases will include $\ThetaT^\off$
(as a subset).
The stratum and fiber prebases will also include $\ThetaT^\barc$.
We use the notations $\ThetaT^\off$ and $\ThetaT^\barc$ because
they are replacements (of a sort) for $\ThetaO^\off$ and~$\ThetaO^\barc$, but
these notations are misnomers, as
$\ThetaT^\barc$ does not encode barcode moves and
$\ThetaT^\off$ is not related to moves off the fiber.

\begin{figure}
\centerline{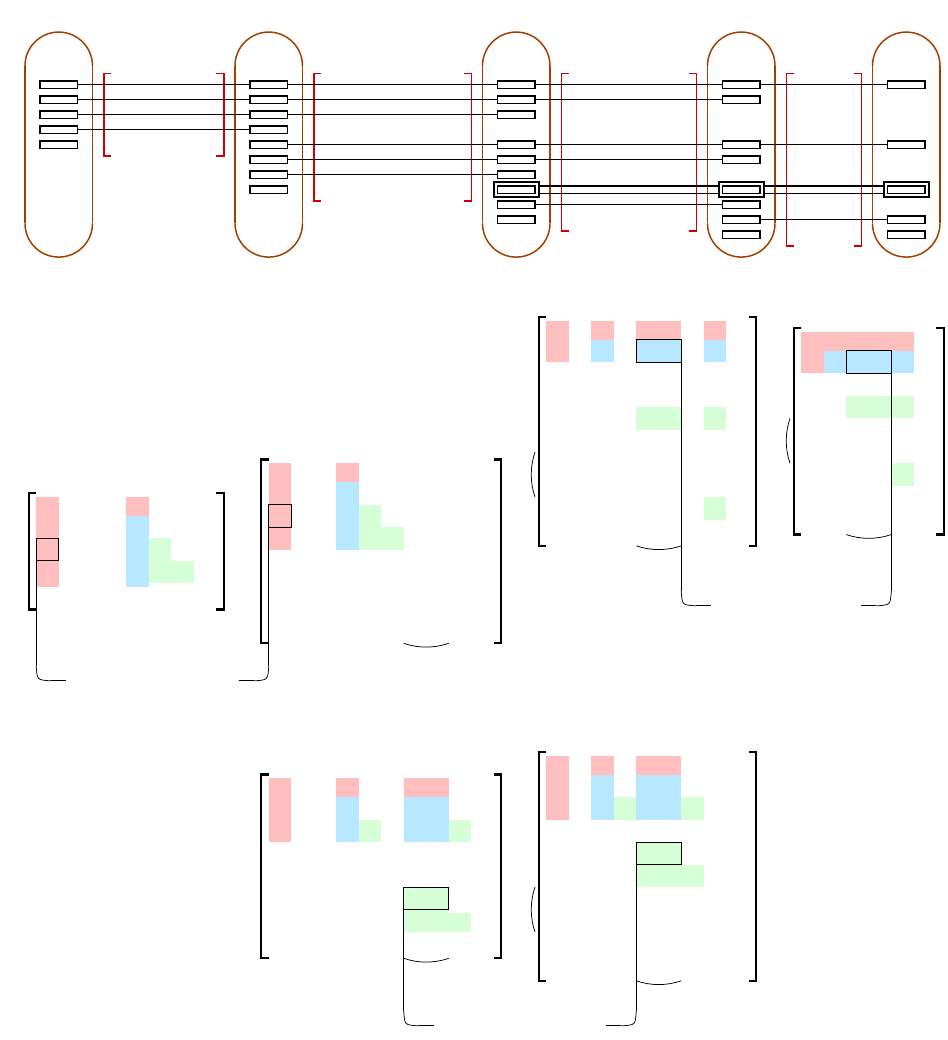}

\caption{\label{ident2moves}
Two-matrix subspaces in $\ThetaT^\off \cup \ThetaT^\barc$ at
the canonical weight vector
$\tilde{\theta} = (\tilde{I}_4, \tilde{I}_3, \tilde{I}_2, \tilde{I}_1)$
when every $\omega_{ki} = 1$ except $\omega_{20} = 2$.
Each two-matrix subspace $\tau_{lkjih} \in \ThetaT$ has nonzero components in
two blocks within two consecutive factor matrices.
Pink backgrounds indicate subspaces in $\ThetaT^\off$;
green backgrounds indicate subspaces in $\ThetaT^\barc$; and
blue backgrounds indicate subspaces in both $\ThetaT^\off$ and $\ThetaT^\barc$.
Subspaces in $\ThetaT^\off \cup \ThetaT^\barc$ (all three colors) represent
directions along which the stratum is curved.
Compare with Figure~\ref{identfactor}, which illustrates
the one-matrix subspaces at $\tilde{\theta}$.
}
\end{figure}

Figure~\ref{ident2moves} illustrates the two-matrix subspaces in
$\ThetaT^\off \cup \ThetaT^\barc$ for a canonical weight vector
$\tilde{\theta} = (\tilde{I}_4, \tilde{I}_3, \tilde{I}_2, \tilde{I}_1)$
by reprising part of Figure~\ref{ident} with some background colors added.
A displacement in the two-matrix subspace $\tau_{32210}$, illustrated at bottom,
is the sum of a displacement in~$\phi_{32310}$ (which modifies~$\tilde{I}_3$) and
a displacement in~$\phi_{32210}$ (which modifies~$\tilde{I}_2$), and
it is tangent to the stratum at $\tilde{\theta}$
(but the stratum is curved, so a nonzero displacement
$\Delta \tilde{\theta} \in \tau_{32210}$ from $\tilde{\theta}$ moves
off the stratum).

As $\tau_{lkjih} \subseteq \phi_{lkjih} + \phi_{l,k,j+1,i,h}$
for each $\tau_{lkjih} \in \ThetaT$, we have
$\Span \ThetaT^\off \subseteq \Span \ThetaO^\off$ and
$\Span \ThetaT^\barc \subseteq \Span \ThetaO^\barc$.
However, although
$\Span \ThetaT^\off \subseteq T_\theta S \subseteq \Null \dmu(\theta)$ and
$\Span \ThetaT^\barc \subseteq T_\theta S \subseteq \Null \dmu(\theta)$,
no subspace in $\ThetaO^\off$ is a subset of $\Null \dmu(\theta)$ and
no subspace in $\ThetaO^\off$ nor $\ThetaO^\barc$ is a subset of $T_\theta S$.

To understand a little more deeply the relationship between
$\ThetaT^\barc$ and $\ThetaO^\barc$, consider
a subspace $\tau_{lkjih} \in \ThetaT^\barc$.
A nonzero displacement $\Delta \theta \in \tau_{lkjih}$ is a sum of
a displacement in $\phi_{lkjih}$ and a displacement in~$\phi_{l,k,j+1,i,h}$,
both of which are subspaces in $\ThetaO^\barc$.
By the definition of $\ThetaT^\barc$, $\tau_{lkjih}$ has $l > k \geq i > h$, so
the displacement in $\phi_{lkjih}$ corresponds to a swapping move
(never a connecting move) that moves from
the stratum~$S$ that contains $\theta$ to a higher-dimensional stratum~$S'$, and
the displacement in $\phi_{l,k,j+1,i,h}$ corresponds to
a {\em different} swapping move from $S$ to the {\em same} stratum~$S'$---so
it is the same {\em abstract} swapping move, but
it changes $W_{j+1}$ instead of~$W_j$.
The displacement in $\phi_{lkjih}$ and the displacement in $\phi_{l,k,j+1,i,h}$
are balanced so that $\Delta \theta$~is tangent to $S$.
We can think of $\ThetaT^\barc$ as a set of degrees of freedom along $S$ that
cannot be expressed as one-matrix subspaces because changing only one matrix
would trigger a swapping move.
Similarly, we can think of $\ThetaT^\off$ as a set of degrees of freedom
along $S$ that cannot be expressed as one-matrix subspaces because changing
only one matrix would change the value of $W$.


\subsection{The Freedom, Stratum, and Fiber Prebases}
\label{fiberbasis}

The {\em freedom prebasis} at $\theta$ is
a set of linearly independent subspaces of $\R^{d_\theta}$
that spans $\Null \dmu(\theta)$, namely,
\begin{eqnarray*}
\Theta^\free
& = & \ThetaO^\fiber \cup \ThetaT^\off
  =   \left( \ThetaO \setminus \ThetaO^\off \right) \cup \ThetaT^\off  \\
& = & \left( \ThetaO \setminus
      \{ \phi_{lkjih} \in \ThetaO : l = L \mbox{~and~} h = 0 \} \right) \cup
      \{ \tau_{lkjih} \in \ThetaT : l = L \mbox{~and~} h = 0 \}.
\end{eqnarray*}
The freedom prebasis contains every one-matrix subspace satisfying
$\phi_{lkjih} \subseteq \Null \dmu(\theta)$ and $\phi_{lkjih} \neq \{ {\bf 0} \}$,
plus some additional two-matrix subspaces as needed so that
$\Theta^\free$ spans $\Null \dmu(\theta)$ (as we will show).

The definition of $\Theta^\free$ above
takes the set $\ThetaO$ of one-matrix subspaces,
discards the subspaces in $\ThetaO^\off$
(because they do not lie in $\Null \dmu(\theta)$), and
replaces them with the two-matrix subspaces in $\ThetaT^\off$
(which lie in $\Null \dmu(\theta)$ by Corollary~\ref{2strat}).
Specifically, for each pair $k, i \in [0, L]$ satisfying $k + 1 \geq i$,
the definition discards the subspaces $\phi_{Lkji0}$ for $j \in [i, k + 1]$, and
replaces them with the subspaces $\tau_{Lkji0}$ for $j \in [i, k]$.
Thus it discards $k - i + 2$ one-matrix subspaces
of dimension $\omega_{Li} \, \omega_{k0}$ and replaces them with
$k - i + 1$ two-matrix subspaces, also of dimension~$\omega_{Li} \, \omega_{k0}$.

Let $W = \mu(\theta)$ and
let $S$ be the stratum of $\mu^{-1}(W)$ that contains $\theta$.
The {\em stratum prebasis} at~$\theta$ is
a set of linearly independent subspaces of $\R^{d_\theta}$ that spans
the tangent space $T_\theta S$, namely,
\begin{eqnarray*}
\Theta^\strat
& = & \ThetaO^\strat \cup \ThetaT^\off \cup \ThetaT^\barc
  =   \left( \ThetaO \setminus \ThetaO^\off \setminus \ThetaO^\barc \right) \cup
      \ThetaT^\off \cup \ThetaT^\barc  \\
& = &
\left( \ThetaO \setminus
\{ \phi_{lkjih} \in \ThetaO : (l = L \mbox{~and~} h = 0) \mbox{~or~}
                             (l > k \mbox{~and~} i > h) \} \right) \cup  \\
& &
\{ \tau_{lkjih} \in \ThetaT : (l = L \mbox{~and~} h = 0) \mbox{~or~}
                             (l > k \mbox{~and~} i > h) \}.
\end{eqnarray*}
The stratum prebasis contains every one-matrix subspace satisfying
$\phi_{lkjih} \subseteq T_\theta S$ and $\phi_{lkjih} \neq \{ {\bf 0} \}$,
plus some additional two-matrix subspaces as needed so that
$\Theta^\strat$ spans $T_\theta S$.

The definition of $\Theta^\strat$ can be interpreted as taking
the freedom prebasis $\Theta^\free$, discarding the subspaces in~$\ThetaO^\barc$
(because they do not lie in $T_\theta S$), and
replacing them with the two-matrix subspaces in $\ThetaT^\barc$
(which lie in $T_\theta S$ by Corollary~\ref{2strat}).
Specifically, for each choice of indices $l, k, i, h$ satisfying
$L \geq l \geq k + 1 \geq i > h \geq 0$,
the definition discards the subspaces $\phi_{lkjih}$ for $j \in [i, k + 1]$, and
replaces them with the subspaces $\tau_{lkjih}$ for $j \in [i, k]$.
Thus it discards $k - i + 2$ one-matrix subspaces
of dimension $\omega_{li} \, \omega_{kh}$ and replaces them with
$k - i + 1$ two-matrix subspaces, also of dimension~$\omega_{li} \, \omega_{kh}$.

The {\em fiber prebasis} at $\theta$ is
a superset of the stratum prebasis that spans $\Null \dmu(\theta)$, namely,
\begin{eqnarray*}
\Theta^\fiber & = &
\Theta^\strat \cup
\{ \phi_{l,k,k+1,i,h} \in \ThetaO : (L > l \mbox{~or~} h > 0) \mbox{~and~}
                                  l > k \mbox{~and~} i > h \}  \\
& = &
\left( \ThetaO \setminus
\{ \phi_{lkjih} \in \ThetaO : (l = L \mbox{~and~} h = 0) \mbox{~or~}
                             (l > k \geq j \geq i > h) \} \right) \cup  \\
& &
\{ \tau_{lkjih} \in \ThetaT : (l = L \mbox{~and~} h = 0) \mbox{~or~}
                             (l > k \geq j \geq i > h) \}.
\end{eqnarray*}
Recall our goal for the fiber prebasis:
for every stratum $S'$ whose closure contains $\theta$,
if $T_\theta \bar{S}'$ exists,
we want some subset of $\Theta^\fiber$ to be a prebasis for $T_\theta \bar{S}'$.
As $T_\theta \bar{S}' \supseteq T_\theta S$,
that subset of $\Theta^\fiber$ is a superset of $\Theta^\strat$.
The fiber prebasis gives us a compact way to express
the tangent spaces of all the strata meeting at a point $\theta$.

We think of $\Theta^\fiber$ as being obtained from $\Theta^\free$ as follows:
for each choice of indices $l, k, i, h$ satisfying
$L \geq l \geq k + 1 \geq i > h \geq 0$, and for each $j \in [i, k]$,
we discard $\phi_{lkjih}$ and replace it with $\tau_{lkjih}$.
This replacement does not change the span---that is,
$\Span \Theta^\fiber = \Span \Theta^\free = \Null \dmu(\theta)$.
The difference between $\Theta^\fiber$ and~$\Theta^\strat$ is that
$\Theta^\fiber$ retains $\phi_{lkjih}$ for $j = k + 1$, whereas
$\Theta^\strat$ discards it.\footnote{
This is an arbitrary choice---for each index set $l, k, i, h$, we
could have chosen any single $j \in [i, k + 1]$ for retention---but
choosing $j = k + 1$ makes the indices nice.
}

In the two-matrix case ($L = 2$),
\begin{eqnarray*}
\Theta^\free    =
\Theta^\fiber & = & \{ \phi_{10100}, \phi_{10110}, \phi_{11100}, \phi_{11110},
                       \phi_{12100}, \phi_{12110},  \\
              &   &    \phi_{21201}, \phi_{21211}, \phi_{21221}, \phi_{22201},
                       \phi_{22211}, \phi_{22221},
                       \tau_{21100}, \tau_{21110}, \tau_{22100}, \tau_{22110} \}
                       \setminus \{ \{ {\bf 0} \} \},  \\
\Theta^\strat & = & \{ \phi_{10100}, \phi_{11100}, \phi_{11110},
                       \phi_{12100}, \phi_{12110},  \\
              &   &    \phi_{21201}, \phi_{21211}, \phi_{22201},
                       \phi_{22211}, \phi_{22221},
                       \tau_{21100}, \tau_{21110}, \tau_{22100}, \tau_{22110} \}
                       \setminus \{ \{ {\bf 0} \} \}.
\end{eqnarray*}

\begin{figure}
\centerline{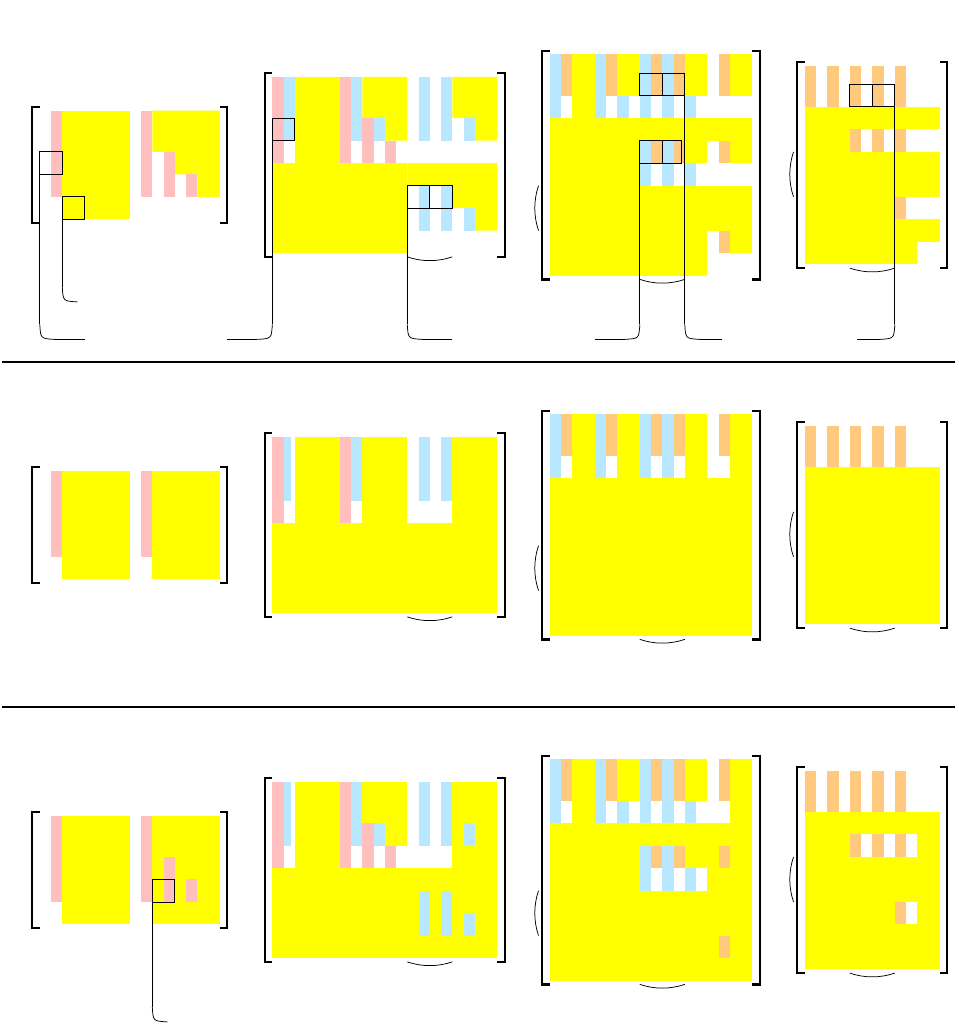}

\caption{\label{dstratum}
Visualizing prebases at the canonical weight vector
$\tilde{\theta} = (\tilde{I}_4, \tilde{I}_3, \tilde{I}_2, \tilde{I}_1)$ when
every $\omega_{ki} = 1$ except $\omega_{20} = 2$.
Top:  visualizing the prebasis $\Theta^\strat$,
the tangent space $T_\theta S = \Span \Theta^\strat$, and
its dimension $D^\strat$.
Each component with a yellow background indicates a dimension in
a one-matrix subspace in $\Theta^\strat$ (such as $\phi_{44441}$);
this component varies freely in $T_\theta S$.
Other background colors indicate dimensions in
two-matrix subspaces in~$\Theta^\strat$ (such as $\tau_{44320}$),
dimensions that linearly link two components in weight space:
pink for a direction that changes one component of $\tilde{I}_4$ and
one component of $\tilde{I}_3$;
blue for $\tilde{I}_3$ and~$\tilde{I}_2$; and
orange for $\tilde{I}_2$ and~$\tilde{I}_1$.
You can see the dimension $D^\strat$ of $T_\theta S$ by
counting dimensions of all four colors.
Center:  visualizing the prebasis $\Theta^\free$,
the nullspace $\Null \dmu(\theta) = \Span \Theta^\free$, and
its dimension~$D^\free$ at $\tilde{\theta}$.
Bottom:  visualizing the prebasis $\Theta^\fiber$.
Observe that for $\Theta^\fiber$ only, some components participate in both
a one-matrix subspace and a two-matrix subspace, such as
$\phi_{43431}$ and $\tau_{43331}$.
}
\end{figure}

Figure~\ref{dstratum} illustrates the prebases $\Theta^\strat$, $\Theta^\free$,
and $\Theta^\fiber$ at the canonical weight vector for
the same network depicted in Figures~\ref{identfactor} and~\ref{ident2moves}.

\subsection{Proof that the Freedom, Stratum, and Fiber Prebases are Prebases}
\label{proofbasis}

In this section, we prove that
$\Theta^\free$ and $\Theta^\fiber$ are prebases for $\Null \dmu(\theta)$ and
we partly prove that $\Theta^\strat$ is a prebasis for~$T_\theta S$.
Unfortunately, the proof of the latter will not be truly complete until
the end of Section~\ref{normal}, as
it requires some reasoning about the dimension of $T_\theta S$ that we defer to
Sections~\ref{counting} and~\ref{normal2basis}.
But the incomplete proof we give here motivates
the difficult work of determining the dimension of $T_\theta S$.

There are three steps to each proof.
The first step is to observe, as we have in
Lemmas~\ref{1strat} and~\ref{1dmu} and Corollary~\ref{2strat}, that
every subspace in $\Theta^\strat$ is a subset of $T_\theta S$, and
every subspace in $\Theta^\free$ or $\Theta^\fiber$ is
a subset of $\Null \dmu(\theta)$.
Hence $\Span \Theta^\strat \subseteq T_\theta S$,
$\Span \Theta^\free \subseteq \Null \dmu(\theta)$, and
$\Span \Theta^\fiber \subseteq \Null \dmu(\theta)$.
The second step is to show that
$\Theta^\free$ is a set of linearly independent subspaces, and
so are $\Theta^\strat$ and~$\Theta^\fiber$.
This step is Lemma~\ref{prebaseslinind}, immediately below.
The third and final step is to
add up the dimensions of the subspaces in $\Theta^\free$ (or $\Theta^\fiber$) and
see that their total dimension is the dimension of $\Null \dmu(\theta)$; hence
$\Span \Theta^\free = \Null \dmu(\theta)$ and
$\Span \Theta^\fiber = \Null \dmu(\theta)$.
The third step is more complicated for $\Theta^\strat$,
as we haven't yet derived the dimension of $T_\theta S$.
The subspace orthogonal to $S$ at $\theta$, denoted $N_\theta S$, is
the orthogonal complement of $T_\theta S$ in $\R^{d_\theta}$, so
$\dim N_\theta S + \dim T_\theta S = d_\theta$.
In Section~\ref{normal} we will identify some linearly independent subspaces
in~$N_\theta S$, then
we will find that the sum of the dimensions of those subspaces plus
the sum of the dimensions of the subspaces in $\Theta^\strat$ is $d_\theta$.
Therefore, $\Span \Theta^\strat = T_\theta S$.

\begin{lemma}
\label{prebaseslinind}
The subspaces in $\Theta^\free$ are linearly independent.
Likewise for $\Theta^\strat$ and $\Theta^\fiber$.
\end{lemma}

\begin{proof}
Let $\Theta$ be one of $\Theta^\free$, $\Theta^\strat$, or $\Theta^\fiber$.
Suppose for the sake of contradiction that $\Theta$ is not linearly independent.
Then we can choose one weight vector from each subspace in $\Theta$---call
them {\em canceling vectors}---such that
the sum of all the canceling vectors is zero, and
at least two canceling vectors are nonzero.
By Lemma~\ref{ThetaOprebasis},
the one-matrix subspaces in $\ThetaO$ are linearly independent.
Therefore, at least one canceling vector from a two-matrix subspace is nonzero.
Let $\tau_{lkjih} \in \Theta$ be
a two-matrix subspace whose canceling vector $\zeta \in \tau_{lkjih}$ is nonzero
such that the index $j$ is minimal---there is
no nonzero canceling vector from a two-matrix subspace with a smaller $j$.
The fact that $\tau_{lkjih} \in \Theta$ implies that $\phi_{lkjih} \not\in \Theta$
(by the definition of $\Theta^\free$, $\Theta^\strat$, or $\Theta^\fiber$).

Given a weight vector $\xi = (X_L, X_{L-1}, \ldots, X_1)$,
we define an operator that picks one matrix out of the weight vector:
let $M_j(\xi) = X_j$.
Given a subspace $\sigma$ of weight vectors,
let $M_j(\sigma) = \{ M_j(\xi) : \xi \in \sigma \}$.
It is clear from the definition of the one-matrix subspaces that
$M_j(\phi_{vujts}) = o_{vujts}$ and
$M_j(\phi_{vuj'ts}) = \{ 0 \}$ for $j' \neq j$.
By Lemma~\ref{2to1}, $M_j(\tau_{vujts}) = o_{vujts}$,
$M_j(\tau_{v,u,j-1,t,s}) = o_{vujts}$, and
$M_j(\tau_{vuj'ts}) = \{ 0 \}$ for $j' \not\in \{ j - 1, j \}$.
Every subspace in $\Theta$ falls into one of these five cases.

The only subspaces $\sigma \in \Theta$ satisfying $M_j(\sigma) = o_{lkjih}$ are
$\tau_{lkjih}$ and $\tau_{l,k,j-1,i,h}$
(recall that $\phi_{lkjih} \not\in \Theta$),
and the latter might not be in~$\Theta$.
If $\tau_{l,k,j-1,i,h} \in \Theta$, its canceling vector is zero
(otherwise we would have made a different choice of $\tau_{lkjih}$).
As $\zeta \in \tau_{lkjih} \setminus \{ {\bf 0} \}$, by Lemma~\ref{2to1},
$M_j(\zeta) \in o_{lkjih} \setminus \{ 0 \}$.
Every other canceling vector $\xi$ satisfies $M_j(\xi) = 0$ or
$M_j(\xi) \in o_{vujts}$ for some subspace $o_{vujts} \neq o_{lkjih}$.
The one-matrix subspaces are linearly independent by Lemma~\ref{Ojprebasis}, so
the sum $\sum_\xi M_j(\xi)$ is nonzero,
where $\xi$ varies over all the canceling vectors including $\zeta$.
But by assumption the canceling vectors sum to zero, so
$\sum_\xi M_j(\xi) = 0$, a contradiction.
It follows that the subspaces in $\Theta$ are linearly independent.
\end{proof}

\begin{lemma}
\label{Dfree}
The sum of the dimensions of the subspaces in~$\Theta^\free$ is
\begin{equation}
\label{Dfreedomrank}
D^\free = d_\theta
- \sum_{j=1}^L \rank W_{L \ssq j} \cdot \rank W_{j-1 \ssq 0}
+ \sum_{j=1}^{L-1} \rank W_{L \ssq j} \cdot \rank W_{j \ssq 0}.
\end{equation}
The sum of the dimensions of the subspaces in~$\Theta^\fiber$ is $D^\free$ too.
\end{lemma}

\begin{proof}
Recall that $\Theta^\free = (\ThetaO \setminus \ThetaO^\off) \cup \ThetaT^\off$.
The subspaces in~$\ThetaO$ have total dimension $d_\theta$.
The subspaces in~$\ThetaO^\off$ have total dimension
\[
\sum_{j=1}^L \sum_{i=0}^j \sum_{k=j-1}^L \omega_{Li} \, \omega_{k0} =
\sum_{j=1}^L \rank W_{L \ssq j} \cdot \rank W_{j-1 \ssq 0}.
\]
The subspaces in~$\ThetaT^\off$ have total dimension
\[
\sum_{j=1}^{L-1} \sum_{i=0}^j \sum_{k=j}^L \omega_{Li} \, \omega_{k0} =
\sum_{j=1}^{L-1} \rank W_{L \ssq j} \cdot \rank W_{j \ssq 0}.
\]
The expression~(\ref{Dfreedomrank}) follows.

We have seen that $\Theta^\fiber$ can be obtained from $\Theta^\free$
by discarding some one-matrix subspaces of the form $\phi_{lkjih}$ and
replacing them with two-matrix subspaces of the form $\tau_{lkjih}$ such that
$\dim \tau_{lkjih} = \dim \phi_{lkjih}$.
Hence the sum of the dimensions of the subspaces in $\Theta^\fiber$ is
the same as for $\Theta^\free$.
\end{proof}

\begin{lemma}
\label{dimnulldmu}
The dimension of $\Null \dmu(\theta)$ is $D^\free$,
as written in identity~(\ref{Dfreedomrank}).
\end{lemma}

\begin{proof}
Trager, Kohn, and Bruna~\cite{trager20} show that
the rank of $\dmu(\theta)$ is $d_\theta - D^\free$.
(See Appendix~\ref{funddmu} for details and a proof.)
It follows from the Rank-Nullity Theorem that
$\dim \Null \dmu(\theta) = D^\free$.
\end{proof}

\begin{theorem}
\label{freedomspan}
$\Theta^\free$ is a prebasis for $\Null \dmu(\theta)$.
So is $\Theta^\fiber$.
In particular, $\Null \dmu(\theta) = \Span \Theta^\free = \Span \Theta^\fiber$.
\end{theorem}

\begin{proof}
By Lemma~\ref{1dmu} and Corollary~\ref{2strat},
each subspace in $\Theta^\free$ or $\Theta^\fiber$ is
a subspace of $\Null \dmu(\theta)$, so
$\Span \Theta^\free \subseteq \Null \dmu(\theta)$ and
$\Span \Theta^\fiber \subseteq \Null \dmu(\theta)$.

By Lemma~\ref{prebaseslinind},
the subspaces in~$\Theta^\free$ are linearly independent, so
the dimension of $\Span \Theta^\free$ is
the sum of the dimensions of the subspaces in $\Span \Theta^\free$,
which is $D^\free$ by Lemma~\ref{Dfree}.
Likewise, by the same two lemmas, $\dim \Span \Theta^\fiber = D^\free$.
By Lemma~\ref{dimnulldmu},
the dimension of $\Null \dmu(\theta)$ is also $D^\free$.
Hence $\Null \dmu(\theta) = \Span \Theta^\free = \Span \Theta^\fiber$, and
both $\Theta^\free$ and $\Theta^\fiber$ are prebases for $\Null \dmu(\theta)$.
\end{proof}

\begin{cor}
\label{freedomspan2}
The set of directions by which a smooth path can leave $\theta$
on the fiber $\mu^{-1}(W)$ spans $\Null \dmu(\theta)$.
(Informally, we say $D^\free$ is the number of {\em degrees of freedom}
along which a smooth path can leave $\theta$ on the fiber.)
\end{cor}

\begin{proof}
If there is a smooth path on the fiber leaving $\theta$ tangent to
some direction $\Delta \theta$, then by Lemma~\ref{TSnulldmu},
$\Delta \theta \in \Null \dmu(\theta)$.

By Corollary~\ref{ofiber}, for every nonzero displacement $\Delta \theta$ in
every one-matrix subspace $\phi_{lkjih} \in \Theta^\free$,
a ray originating at~$\theta$ in the direction $\Delta \theta$ is
a smooth path on the fiber.
By Lemma~\ref{2stratpath}, for every displacement $\Delta \theta$ in
every two-matrix subspace $\tau_{lkjih} \in \Theta^\free$,
there is a smooth two-matrix path on the fiber that leaves $\theta$ in
the direction~$\Delta \theta$.
By Theorem~\ref{freedomspan}, $\Null \dmu(\theta) = \Span \Theta^\free$; hence
these directions suffice to span $\Null \dmu(\theta)$.
\end{proof}

\begin{theorem}
\label{stratspan}
$\Theta^\strat$ is a prebasis for $T_\theta S$.
In particular, $T_\theta S = \Span \Theta^\strat$.
Moreover, the dimension of~$S$ and the dimension of~$T_\theta S$ is
\begin{equation}
\label{Dstratba}
D^\strat = d_\theta - \rank W \cdot (d_L + d_0 - \rank W)
- \sum_{L \geq k+1 \geq i > 0} \beta_{k+1,i,i} \, \alpha_{k,k,i-1}.
\end{equation}
\end{theorem}

\begin{proof}
Let $D^\strat$ be
the sum of the dimensions of all the subspaces in $\Theta^\strat$.
In Section~\ref{counting},
we will show that this sum is the expression~(\ref{Dstratba}).
By Lemma~\ref{prebaseslinind},
the subspaces in $\Theta^\strat$ are linearly independent, so
the dimension of $\Span \Theta^\strat$ is $D^\strat$.

By Lemma~\ref{1strat} and Corollary~\ref{2strat},
each subspace in $\Theta^\strat$ is a subspace of $T_\theta S$, so
$\Span \Theta^\strat \subseteq T_\theta S$.
Hence the dimension of $T_\theta S$ is at least $D^\strat$.
Let $N_\theta S$ denote the orthogonal complement of $T_\theta S$
in the space $\R^{d_\theta}$;
$N_\theta S$ is the highest-dimensional subspace normal to $S$ at $\theta$.
We will show in Lemma~\ref{psistratdim}
that the dimension of $N_\theta S$ is at least
\[
\rank W \cdot (d_L + d_0 - \rank W)
+ \sum_{L \geq k+1 \geq i > 0} \beta_{k+1,i,i} \, \alpha_{k,k,i-1}.
\]
The sum of these lower bounds on the dimensions of $T_\theta S$ and $N_\theta S$
is $d_\theta$, so both bounds must be tight.
Hence, $\dim T_\theta S = D^\strat$ and $T_\theta S = \Span \Theta^\strat$.
Therefore, $\Theta^\strat$ is a prebasis for $T_\theta S$.
As $S$ has the same dimension as $T_\theta S$, $\dim S = D^\strat$.
\end{proof}

\subsection{Expressions for the Tangent Space of $S$ and
            the Nullspace of d$\mu(\theta)$}
\label{expressions}

In this section,
we write explicit expressions for $T_\theta S$ and $\Null \dmu(\theta)$.
We begin by writing expressions for
$\Span \ThetaO^\strat$ and $\Span \ThetaO^\fiber$, as
$\Span \ThetaO^\strat \subseteq T_\theta S$ (Lemma~\ref{1strat}) and
$\Span \ThetaO^\fiber \subseteq \Null \dmu(\theta)$ (Lemma~\ref{1dmu}).
Recall the flow subspaces defined in Section~\ref{flowspaces},
$A_{kji} = \Null W_{k+1 \ssq j} \cap \col W_{j \ssq i}$ and
$B_{kji} = \row W_{k \ssq j} \cap \Null W_{j \ssq i-1}^\top$.

\begin{lemma}
\label{spanthetaOstrat}
\begin{eqnarray}
\Span \ThetaO^\strat
& = &
\left\{ (\Delta W_L, \Delta W_{L-1}, \ldots, \Delta W_1) :  \rule{0pt}{16pt}
\right.  \nonumber  \\
&   &
\Delta W_j \in
\sum_{h=1}^{j-1} \col W_{j \ssq h} \otimes \Null W_{j-1 \ssq h-1}^\top +
(\Null W_{L \ssq j} \cap \col W_{j \ssq 0}) \otimes \R^{d_{j-1}} +
\nonumber  \\
&   &
\left. \sum_{l=j}^{L-1} \Null W_{l+1 \ssq j} \otimes \row W_{l \ssq j-1} +
\R^{d_j} \otimes (\row W_{L \ssq j-1} \cap \Null W_{j-1 \ssq 0}^\top) \right\}.
\label{spanthetaOstrateq}
\end{eqnarray}
\end{lemma}

\begin{proof}
By the definition of $\ThetaO^\strat$,
\begin{eqnarray}
\Span \ThetaO^\strat
& = &
\Span \{ \phi_{lkjih} \in \ThetaO : (L > l \mbox{~or~} h > 0) \mbox{~and~}
(l \leq k \mbox{~or~} i \leq h) \}  \nonumber  \\
& = &
\left\{ (\Delta W_L, \Delta W_{L-1}, \ldots, \Delta W_1) :  \rule{0pt}{10pt}
\right.  \nonumber  \\
&   & \left. \Delta W_j \in \Span \{ o_{lkjih} \in \mathcal{O}_j :
(L > l \mbox{~or~} h > 0) \mbox{~and~} (l \leq k \mbox{~or~} i \leq h) \}
\right\}.
\label{ThetaOstrat}
\end{eqnarray}

We rewrite the condition in~(\ref{ThetaOstrat}).
As $k$ cannot exceed $L$ and $i$ cannot be negative,
\begin{eqnarray*}
(L > l \mbox{~or~} h > 0) \mbox{~and~} l \leq k & \Leftrightarrow &
(L > l \mbox{~and~} l \leq k) \mbox{~or~}
(h > 0 \mbox{~and~} L \leq l \mbox{~and~} l \leq k) \\
& \Leftrightarrow &
(L > l \mbox{~and~} l \leq k) \mbox{~or~} (h > 0 \mbox{~and~} k = L).  \\
(L > l \mbox{~or~} h > 0) \mbox{~and~} i \leq h & \Leftrightarrow &
(L > l \mbox{~and~} h \leq 0 \mbox{~and~} i \leq h) \mbox{~or~}
(h > 0 \mbox{~and~} i \leq h) \\
& \Leftrightarrow &
(L > l \mbox{~and~} i = 0) \mbox{~or~} (h > 0 \mbox{~and~} i \leq h).
\end{eqnarray*}
Thus we can rewrite part of~(\ref{ThetaOstrat}) as
\begin{eqnarray}
&   & \hspace*{-.4in}
\Span \{ o_{lkjih} \in \mathcal{O}_j :
(L > l \mbox{~or~} h > 0) \mbox{~and~} (l \leq k \mbox{~or~} i \leq h) \}
\nonumber  \\
& = &
\Span \{ o_{lkjih} \in \mathcal{O}_j :
(L > l \mbox{~or~} h > 0) \mbox{~and~} l \leq k \} +
\Span \{ o_{lkjih} \in \mathcal{O}_j :
(L > l \mbox{~or~} h > 0) \mbox{~and~} i \leq h \}  \nonumber  \\
& = &
\Span \{ o_{lkjih} \in \mathcal{O}_j : L > l \mbox{~and~} l \leq k \} +
\Span \{ o_{lkjih} \in \mathcal{O}_j : h > 0 \mbox{~and~} k = L \} +
\nonumber  \\
&   &
\Span \{ o_{lkjih} \in \mathcal{O}_j : h > 0 \mbox{~and~} i \leq h \} +
\Span \{ o_{lkjih} \in \mathcal{O}_j : L > l \mbox{~and~} i = 0 \}.
\label{span4}
\end{eqnarray}

To relate the spans in~(\ref{span4}) to
the tensor products in~(\ref{spanthetaOstrateq}),
we use Lemma~\ref{bases} to write
\begin{eqnarray}
\sum_{h=1}^{j-1} A_{Ljh} \otimes B_{j-1,j-1,h} & = &
\sum_{h=1}^{j-1} \left( \sum_{l=j}^L \sum_{i=0}^h a_{lji} \otimes
\sum_{k=j-1}^L \sum_{h'=h}^{j-1} b_{k,j-1,h'} \right)
= \sum_{h=1}^{j-1} \sum_{l=j}^L \sum_{i=0}^h
\sum_{k=j-1}^L \sum_{h'=h}^{j-1} o_{lkjih'}  \nonumber  \\
& = &
\sum_{h=1}^{j-1} \sum_{l=j}^L \sum_{i=0}^h \sum_{k=j-1}^L o_{lkjih} =
\Span \{ o_{lkjih} \in \mathcal{O}_j : h > 0  \mbox{~and~}  i \leq h \},
\label{tensorspan1}  \\
A_{L-1,j,0} \otimes B_{j-1,j-1,0} & = &
\sum_{l=j}^{L-1} a_{lj0} \otimes \sum_{k=j-1}^L \sum_{h=0}^{j-1} b_{k,j-1,h}
= \sum_{l=j}^{L-1} \sum_{k=j-1}^L \sum_{h=0}^{j-1} o_{lkj0h}  \nonumber  \\
& = & \Span \{ o_{lkjih} \in \mathcal{O}_j : L > l  \mbox{~and~}  i = 0 \},
\label{tensorspan2}  \\
\sum_{l=j}^{L-1} A_{ljj} \otimes B_{l,j-1,0} & = &
\sum_{l=j}^{L-1} \left( \sum_{l'=j}^l \sum_{i=0}^j a_{l'ji} \otimes
\sum_{k=l}^L \sum_{h=0}^{j-1} b_{k,j-1,h} \right)
= \sum_{l=j}^{L-1} \sum_{l'=j}^l \sum_{i=0}^j
\sum_{k=l}^L \sum_{h=0}^{j-1} o_{l'kjih}  \nonumber  \\
& = &
\sum_{l=j}^{L-1} \sum_{i=0}^j \sum_{k=l}^L \sum_{h=0}^{j-1} o_{lkjih} =
\Span \{ o_{lkjih} \in \mathcal{O}_j : L > l  \mbox{~and~}  l \leq k \},
\hspace{.2in}  \mbox{and}  \label{tensorspan3}  \\
A_{Ljj} \otimes B_{L,j-1,1} & = &
\sum_{l=j}^L \sum_{i=0}^j a_{lji} \otimes \sum_{h=1}^{j-1} b_{L,j-1,h}
= \sum_{l=j}^L \sum_{i=0}^j \sum_{h=1}^{j-1} o_{lLjih}  \nonumber  \\
& = & \Span \{ o_{lkjih} \in \mathcal{O}_j : h > 0  \mbox{~and~}  k = L \}.
\label{tensorspan4}
\end{eqnarray}

It follows from~(\ref{ThetaOstrat})--(\ref{tensorspan4}) that
\begin{align*}
\Span \ThetaO^\strat = &
\left\{ (\Delta W_L, \Delta W_{L-1}, \ldots, \Delta W_1) :  \rule{0pt}{16pt}
\right.  \\
&
\Delta W_j \in
\left.
\sum_{h=1}^{j-1} A_{Ljh} \otimes B_{j-1,j-1,h} + A_{L-1,j,0} \otimes B_{j-1,j-1,0} +
\sum_{l=j}^{L-1} A_{ljj} \otimes B_{l,j-1,0} + A_{Ljj} \otimes B_{L,j-1,1}
\right\}
\end{align*}
The claim~(\ref{spanthetaOstrateq}) follows from
the flow subspace definitions~(\ref{A}),~(\ref{B}).
\end{proof}

For example, in the two-matrix case ($L = 2$),
\begin{eqnarray*}
\Span \ThetaO^\strat
& = & \left\{ (\Delta W_2, \Delta W_1) : \rule{0pt}{9pt} \right.  \\
&   &
\Delta W_2 \in \col W_2 \otimes \Null W_1^\top +
\R^{d_2} \otimes (\row W_2 \cap \Null W_1^\top),  \\
&   & \left.
\Delta W_1 \in (\Null W_2 \cap \col W_1) \otimes \R^{d_0} +
\Null W_2 \otimes \row W_1 \right\}.
\end{eqnarray*}

The formula~(\ref{spanthetaOstrateq}) gives
the most explicit and concise expression we know how to write for
$\Span \ThetaO^\strat$, but the expression does not readily reveal
the dimension of $\Span \ThetaO^\strat$ nor a basis for it
(such as $\ThetaO^\strat$), because
the expression is a vector sum of subspaces that are not linearly independent
(nor even close).
We derive the dimension of $\Span \ThetaO^\strat$ in Section~\ref{counting}.

Next, we seek an expression for $\Span \ThetaO^\fiber$.
As a first step toward that, recall that in Section~\ref{1matrix} we defined
the subspace $N_j = \Null W_{L \ssq j} \otimes \R^{d_{j-1}} +
\R^{d_j} \otimes \Null W_{j-1 \ssq 0}^\top$,
the set of displacements $\Delta W_j$ that satisfy
$W_{L \ssq j} \Delta W_j W_{j-1 \ssq 0} = 0$.

\begin{lemma}
\label{Ofiberjprebasis}
$\mathcal{O}^\fiber_j$ is a prebasis for $N_j$.
Therefore,
\begin{equation}
\Span \ThetaO^\fiber =
\{ (\Delta W_L, \Delta W_{L-1}, \ldots, \Delta W_1) : \Delta W_j \in N_j \}.
\label{spanthetaOfiber}
\end{equation}
\end{lemma}

\begin{proof}
We can rewrite each of the four factors defining $N_j$ as
a flow subspace.
By Lemma~\ref{bases},
\begin{eqnarray*}
\R^{d_j} & = & A_{Ljj} = \sum_{l=j}^L \sum_{i=0}^j a_{lji},  \\
\Null W_{L \ssq j} & = & A_{L-1,j,j} =
\sum_{l=j}^{L-1} \sum_{i=0}^j a_{lji},  \\
\R^{d_{j-1}} & = & B_{j-1,j-1,0} = \sum_{k=j-1}^L \sum_{h=0}^{j-1} b_{k,j-1,h},
\hspace{.2in}  \mbox{and}  \\
\Null W_{j-1 \ssq 0}^\top & = & B_{j-1,j-1,1} =
\sum_{k=j-1}^L \sum_{h=1}^{j-1} b_{k,j-1,h}.
\end{eqnarray*}

Therefore,
\begin{eqnarray*}
\Null W_{L \ssq j} \otimes \R^{d_{j-1}} & = &
\sum_{l=j}^{L-1} \sum_{i=0}^j \sum_{k=j-1}^L \sum_{h=0}^{j-1}
a_{lji} \otimes b_{k,j-1,h} =
\Span \{ o_{lkjih} \in \mathcal{O}_j : L > l \},  \\
\R^{d_j} \otimes \Null W_{j-1 \ssq 0}^\top & = &
\sum_{l=j}^L \sum_{i=0}^j \sum_{k=j-1}^L \sum_{h=1}^{j-1}
a_{lji} \otimes b_{k,j-1,h} =
\Span \{ o_{lkjih} \in \mathcal{O}_j : h > 0 \},
\hspace{.2in}  \mbox{and}  \\
N_j & = & \Null W_{L \ssq j} \otimes \R^{d_{j-1}} +
          \R^{d_j} \otimes \Null W_{j-1 \ssq 0}^\top  \\
    & = & \Span \{ o_{lkjih} \in \mathcal{O}_j : L > l  \mbox{~or~}  h > 0 \}  \\
    & = & \Span \mathcal{O}^\fiber_j.
\end{eqnarray*}

By Lemma~\ref{Ojprebasis},
the subspaces in $\mathcal{O}_j$ are linearly independent; hence
so are the subspaces in $\mathcal{O}^\fiber_j$.
Therefore, $\mathcal{O}^\fiber_j$ is a prebasis for $N_j$.

The second claim follows immediately.
\end{proof}

Recall identity~(\ref{dimNj}) from Section~\ref{1matrix}:
$\dim N_j = d_j d_{j-1} - \rank W_{L \ssq j} \cdot \rank W_{j-1 \ssq 0}$.
It follows that
\begin{equation}
\dim \Span \ThetaO^\fiber =
\sum_{j=1}^L (d_j d_{j-1} - \rank W_{L \ssq j} \cdot \rank W_{j-1 \ssq 0}) =
d_\theta - \sum_{j=1}^L \rank W_{L \ssq j} \cdot \rank W_{j-1 \ssq 0}.
\label{dimspanthetaOfiber}
\end{equation}

In the two-matrix case ($L = 2$),
\begin{eqnarray*}
\Span \ThetaO^\fiber & = &
\{ (\Delta W_2, \Delta W_1) : \Delta W_2 \in \R^{d_2} \otimes \Null W_1^\top,
\Delta W_1 \in \Null W_2 \otimes \R^{d_0} \}  \hspace{.2in}  \mbox{and}  \\
\dim \Span \ThetaO^\fiber & = &
d_\theta - d_2 \cdot \rank W_1 - d_0 \cdot \rank W_2.
\end{eqnarray*}

Now we arrive at our goal:
explicit expressions for $T_\theta S$ and $\Null \dmu(\theta)$.
By Theorem~\ref{stratspan}, $T_\theta S = \Span \Theta^\strat$;
by Theorem~\ref{freedomspan}, $\Null \dmu(\theta) = \Span \Theta^\free$;
by Lemma~\ref{tauinTS}, $\tau \subseteq T_\theta S$; and
by Lemma~\ref{2dmu}, $\tau \subseteq \Null \dmu(\theta)$, so
from~(\ref{spanthetaOstrateq}), (\ref{spanthetaOfiber}),
and~(\ref{tauexplicit}) we have
\begin{eqnarray}
\Null \dmu(\theta) & = &
\Span \Theta^\free = \tau + \Span \Theta^\free =
\tau + \Span \ThetaO^\fiber  \nonumber  \\
& = &
\left\{ (\Delta W_L + W_L H_{L-1},
         \Delta W_{L-1} + W_{L-1} H_{L-2} - H_{L-1} W_{L-1}, \ldots,
\rule{0pt}{10pt}  \right.  \nonumber  \\
&   &    \Delta W_j + W_j H_{j-1} - H_j W_j, \ldots,
         \Delta W_2 + W_2 H_1 - H_2 W_2, \Delta W_1 - H_1 W_1) :  \nonumber  \\
&   &    \left. H_j \in \R^{d_j \times d_j},
         \Delta W_j \in \Null W_{L \ssq j} \otimes \R^{d_{j-1}} +
         \R^{d_j} \otimes \Null W_{j-1 \ssq 0}^\top \right\},
\label{nulldmueq}
\\
T_\theta S & = &
\Span \Theta^\strat = \tau + \Span \Theta^\strat =
\tau + \Span \ThetaO^\strat  \nonumber  \\
& = &
\left\{ (\Delta W_L + W_L H_{L-1},
         \Delta W_{L-1} + W_{L-1} H_{L-2} - H_{L-1} W_{L-1}, \ldots,
\rule{0pt}{10pt}  \right.  \nonumber  \\
&   &    \Delta W_j + W_j H_{j-1} - H_j W_j, \ldots,
         \Delta W_2 + W_2 H_1 - H_2 W_2, \Delta W_1 - H_1 W_1) :
\nonumber  \\
&   &    H_j \in \R^{d_j \times d_j},
\nonumber  \\
&   &    \Delta W_j \in
\sum_{h=1}^{j-1} \col W_{j \ssq h} \otimes \Null W_{j-1 \ssq h-1}^\top +
(\Null W_{L \ssq j} \cap \col W_{j \ssq 0}) \otimes \R^{d_{j-1}} +
\nonumber  \\
&   &
         \left. \sum_{l=j}^{L-1} \Null W_{l+1 \ssq j} \otimes \row W_{l \ssq j-1} +
         \R^{d_j} \otimes (\row W_{L \ssq j-1} \cap \Null W_{j-1 \ssq 0}^\top)
\right\}.
\label{TthetaSeq}
\end{eqnarray}
Recall the conventions that $\Null W_{L \ssq L} = \{ {\bf 0} \}$ and
$\Null W^\top_{0 \ssq 0} = \{ {\bf 0} \}$,
both of which appear in~(\ref{nulldmueq}) and~(\ref{TthetaSeq}).

For a more readable example, consider the two-matrix case ($L = 2$),
\begin{eqnarray*}
\Null \dmu(\theta)
& = & \{ (\Delta W_2 + W_2 H, \Delta W_1 - H W_1) :  \\
&   & H \in \R^{d_1 \times d_1},  
      \Delta W_2 \in \R^{d_2} \otimes \Null W_1^\top,
      \Delta W_1 \in \Null W_2 \otimes \R^{d_0} \},  \\
T_\theta S
& = & \left\{ (\Delta W_2 + W_2 H, \Delta W_1 - H W_1) :
\rule{0pt}{10pt}  \right.  \nonumber  \\
&   & H \in \R^{d_1 \times d_1},  
      \Delta W_2 \in \col W_2 \otimes \Null W_1^\top +
      \R^{d_2} \otimes (\row W_2 \cap \Null W_1^\top),  \\
&   & \left.
      \Delta W_1 \in (\Null W_2 \cap \col W_1) \otimes \R^{d_0} +
      \Null W_2 \otimes \row W_1 \right\}.
\end{eqnarray*}

We note in passing that
in both~(\ref{nulldmueq}) and~(\ref{TthetaSeq}),
we can replace ``$H_j \in \R^{d_j \times d_j}$'' with
``$H_j \in \row W_{j+1} \otimes \col W_j$''; and
in~(\ref{nulldmueq}), we can further narrow the range to
``$H_j \in \row W_{L \ssq j} \otimes \col W_{j \ssq 0}$.''
It is possible to narrow the range in~(\ref{TthetaSeq}) too
(though not as narrow), but the expressions become ugly.\footnote{
There is a unique split of $H_j$ into a sum of two parts,
$H_j = H'_j + H''_j$, with
$H'_j \in \row W_{j+1} \otimes \col W_j$ and
$H''_j \in \Null W_{j+1} \otimes \R^{d_j} + \R^{d_j} \otimes \Null W_j^\top$.
Observe that
\begin{eqnarray*}
\{ (W_{j+1} H_j, - H_j W_j) : H_j \in \R^{d_j \times d_j} \}
& = &
\{ (W_{j+1} H''_j + W_{j+1} H'_j, - H''_j W_j - H'_j W_j) :
H'_j \in \row W_{j+1} \otimes \col W_j,  \\
&   &
H''_j \in \Null W_{j+1} \otimes \R^{d_j} + \R^{d_j} \otimes \Null W_j^\top \}
\\
& = &
\{ (\Delta W_{j+1} + W_{j+1} H'_j, \Delta W_j - H'_j W_j) :
H'_j \in \row W_{j+1} \otimes \col W_j,  \\
&   &
\Delta W_{j+1} \in \col W_{j+1} \otimes \Null W_j^\top,
\Delta W_j \in \Null W_{j+1} \otimes \row W_j \}.
\end{eqnarray*}
The contribution due to $H''_j$ is already subsumed by
the contributions of $\Delta W_{j+1}$ and $\Delta W_j$
in~(\ref{nulldmueq}) and~(\ref{TthetaSeq}), so $H''_j$ is not needed.

Both ranges can be narrowed further.
For~(\ref{nulldmueq}), the splitting
$H'_j \in \row W_{L \ssq j} \otimes \col W_{j \ssq 0}$ and
$H''_j \in \Null W_{L \ssq j} \otimes \R^{d_j} +
\R^{d_j} \otimes \Null W_{j \ssq 0}^\top$ works.
For~(\ref{TthetaSeq}), one can use the splitting in which
$H''_j \in
\sum_{h=1}^{j-1} \col W_{j \ssq h} \otimes \Null W_{j \ssq h-1}^\top +
(\Null W_{L \ssq j} \cap \col W_{j \ssq 0}) \otimes \R^{d_j} +
\sum_{l=j+1}^{L-1} \Null W_{l+1 \ssq j} \otimes \row W_{l \ssq j} +
\R^{d_j} \otimes (\row W_{L \ssq j} \cap \Null W_{j \ssq 0}^\top)$ and
$H'_j$ varies over the orthogonal complement of that subspace.
(We choose not to write out the ugly orthogonal complement.)
Again, $H''_j$ is not needed.
%
}

\subsection{Counting Dimensions}
\label{counting}

Let $\DO$, $\DO^\off$, $\DO^\fiber$, $\DO^\barc$, $\DO^\strat$, $\DT^\off$,
$\DT^\barc$, $D^\free$, $D^\strat$, or $D^\fiber$ (and so forth) denote
the dimension of the subspace (of $\R^{d_\theta}$) spanned by the prebasis
$\ThetaO$, $\ThetaO^\off$, $\ThetaO^\fiber$, $\ThetaO^\barc$, $\ThetaO^\strat$,
$\ThetaT^\off$, $\ThetaT^\barc$, $\Theta^\free$, $\Theta^\strat$, or
$\Theta^\fiber$, respectively, at a specified weight vector $\theta$.
Table~\ref{subspacesets} collects
the definitions of these prebases and several others, and
the dimensions of the subspaces they span.
In this section we derive those dimensions.
(See Appendix~\ref{countingmore} for some additional prebases.)

\begin{table}
\begin{center}
\begin{tabular}{l}
\hline
$\ThetaO = \{ \phi_{lkjih} \neq \{ {\bf 0} \} :
               L \geq l \geq j \geq i \geq 0$ and
               $L \geq k \geq j - 1 \geq h \geq 0 \}$  \\
\begin{minipage}{6.3in}
\vspace*{-.18in}
\begin{equation}  \lm
\DO = d_\theta = \sum_{j=1}^L d_j d_{j-1}
\end{equation}
\end{minipage}  \\
$\ThetaO^\off = \{ \phi_{Lkji0} \in \ThetaO \}
               = \{ \phi_{Lkji0} \neq \{ {\bf 0} \} :
                    L \geq j > 0, L \geq k \geq j - 1,$ and
                    $j \geq i \geq 0 \}$
\\
\begin{minipage}{6.3in}
\vspace*{-.16in}
\begin{equation}  \lm
\DO^\off = \sum_{j=1}^L \left( \sum_{i=0}^j \omega_{Li} \right)
                       \left( \sum_{k=j-1}^L \omega_{k0} \right)
         = \sum_{j=1}^L \rank W_{L \ssq j} \cdot \rank W_{j-1 \ssq 0}
\label{DOoff}
\end{equation}
\end{minipage}  \\
$\ThetaO^\fiber = \ThetaO \setminus \ThetaO^\off
                = \{ \phi_{lkjih} \in \ThetaO : L > l$ or $h > 0 \}$  \\
\begin{minipage}{6.3in}
\vspace*{-.16in}
\begin{equation}  \lm
\DO^\fiber = \DO - \DO^\off =
d_\theta - \sum_{j=1}^L \rank W_{L \ssq j} \cdot \rank W_{j-1 \ssq 0}
\label{DOfreedom}
\end{equation}
\end{minipage}  \\
$\ThetaO^\barc = \{ \phi_{lkjih} \in \ThetaO : l > k$ and $i > h \}
                = \{ \phi_{lkjih} \neq \{ {\bf 0} \} :
                     L \geq l \geq k + 1 \geq j \geq i > h \geq 0 \}$  \\
\begin{minipage}{6.3in}
\vspace*{-.14in}
\begin{equation}  \lm
\DO^\barc =
\sum_{L \geq k+1 \geq i > 0} (k - i + 2) \,
  \underbrace{(\rank W_{k+1 \ssq i} - \rank W_{k+1 \ssq i-1})}_{\beta_{k+1,i,i}} \,
  \underbrace{(\rank W_{k \ssq i-1} - \rank W_{k+1 \ssq i-1})}_{\alpha_{k,k,i-1}}
\label{DOcomb}
\end{equation}
\end{minipage}  \\
$\ThetaO^{\off,\neg\barc} = \ThetaO^\off \setminus \ThetaO^\barc
                       = \{ \phi_{Lkji0} \in \ThetaO : L = k$ or $i = 0 \}$  \\
\begin{minipage}{6.3in}
\vspace*{-.14in}
\begin{equation}  \lm
\DO^{\off,\neg\barc} =
\rank W \cdot
\sum_{j=1}^L \left( \rank W_{L \ssq j} + \rank W_{j-1 \ssq 0} - \rank W \right)
\rule[-18pt]{0pt}{11pt}
\label{DOoffnotcomb}
\end{equation}
\end{minipage}  \\
$\ThetaO^\strat = \ThetaO^\fiber \setminus \ThetaO^\barc
                = \ThetaO \setminus \ThetaO^\off \setminus \ThetaO^\barc
                = \{ \phi_{lkjih} \in \ThetaO :
                     (L > l$ or $h > 0)$ and
                     $(l \leq k \mbox{~or~} i \leq h) \}$
\rule[-4pt]{0pt}{14pt}  \\
\begin{minipage}{6.3in}
\vspace*{-.14in}
\begin{equation}  \lm
\DO^\strat = \DO - \DO^\barc - \DO^{\off,\neg\barc}
\rule[-6pt]{0pt}{10pt}
\label{DOstrat}
\end{equation}
\end{minipage}  \\
\hline
$\ThetaTO = \{ \tau_{lkjih} \neq \{ {\bf 0} \} :
               L > j > 0$, $L \geq l \geq j \geq h \geq 0$, and
               $L \geq k \geq j \geq i \geq 0 \}$  \\
$\ThetaT = \{ \tau_{lkjih} \neq \{ {\bf 0} \} :
               L \geq l > j > h \geq 0$ and
               $L \geq k \geq j \geq i \geq 0 \}$  \\
$\ThetaT^\off = \{ \tau_{Lkji0} \in \ThetaT \}
               = \{ \tau_{Lkji0} \neq \{ {\bf 0} \} :
                    L > j > 0$ and $L \geq k \geq j \geq i \geq 0 \}$  \\
\begin{minipage}{6.3in}
\vspace*{-.16in}
\begin{equation}  \lm
\DT^\off = \sum_{j=1}^{L-1} \left( \sum_{i=0}^j \omega_{Li} \right)
                           \left( \sum_{k=j}^L \omega_{k0} \right)
         = \sum_{j=1}^{L-1} \rank W_{L \ssq j} \cdot \rank W_{j \ssq 0}
\label{DToff}
\end{equation}
\end{minipage}  \\
$\ThetaT^\barc = \{ \tau_{lkjih} \in \ThetaT : l > k$ and $i > h \}
                = \{ \tau_{lkjih} \neq \{ {\bf 0} \} :
                     L \geq l > k \geq j \geq i > h \geq 0 \}$  \\
\begin{minipage}{6.3in}
\vspace*{-.14in}
\begin{equation}  \lm
\DT^\barc =
\sum_{L > k \geq i > 0} (k - i + 1) \,
  \underbrace{(\rank W_{k+1 \ssq i} - \rank W_{k+1 \ssq i-1})}_{\beta_{k+1,i,i}} \,
  \underbrace{(\rank W_{k \ssq i-1} - \rank W_{k+1 \ssq i-1})}_{\alpha_{k,k,i-1}}
\label{DTcomb}
\end{equation}
\end{minipage}  \\
$\ThetaT^{\off,\neg\barc} = \ThetaT^\off \setminus \ThetaT^\barc
                       = \{ \tau_{Lkji0} \in \ThetaT : L = k$ or $i = 0 \}$  \\
\begin{minipage}{6.3in}
\vspace*{-.14in}
\begin{equation}  \lm
\DT^{\off,\neg\barc} =
\rank W \cdot
\sum_{j=1}^{L-1} \left( \rank W_{L \ssq j} + \rank W_{j \ssq 0} - \rank W \right)
\rule[-18pt]{0pt}{11pt}
\label{DToffnotcomb}
\end{equation}
\end{minipage}  \\
\hline
$\Theta^\free =
\ThetaO^\fiber \cup \ThetaT^\off =
\left( \ThetaO \setminus \ThetaO^\off \right) \cup \ThetaT^\off$
\rule{0pt}{13pt}  \\
\begin{minipage}{6.3in}
\vspace*{-.16in}
\begin{equation}  \lm
D^\free = D^\fiber = 
d_\theta - \sum_{k,i \in [0,L], k+1 \geq i} \omega_{Li} \, \omega_{k0} =
d_\theta - \sum_{j=1}^L \rank W_{L \ssq j} \cdot \rank W_{j-1 \ssq 0}
+ \sum_{j=1}^{L-1} \rank W_{L \ssq j} \cdot \rank W_{j \ssq 0}
\label{Dfreedom}
\end{equation}
\end{minipage}  \\
$\Theta^\strat =
\ThetaO^\strat \cup \ThetaT^\off \cup \ThetaT^\barc =
\left( \ThetaO \setminus \ThetaO^\off \setminus \ThetaO^\barc \right) \cup
\ThetaT^\off \cup \ThetaT^\barc$  \\
\begin{minipage}{6.3in}
\vspace*{-.12in}
\begin{equation}  \lm
D^\strat
= d_\theta - \rank W \cdot (d_L + d_0 - \rank W)
- \hspace*{-.18in}
\sum_{L \geq k+1 \geq i > 0} 
  \underbrace{(\rank W_{k+1 \ssq i} - \rank W_{k+1 \ssq i-1})}_{\beta_{k+1,i,i}} \,
  \underbrace{(\rank W_{k \ssq i-1} - \rank W_{k+1 \ssq i-1})}_{\alpha_{k,k,i-1}}
\rule[-20pt]{0pt}{11pt}
\label{Dstrat}
\end{equation}
\end{minipage}  \\
$\Theta^\fiber =
\Theta^\strat \cup
\{ \phi_{l,k,k+1,i,h} \in \ThetaO : (L > l \mbox{~or~} h > 0)$ and
                                  $l > k$ and $i > h \}$  \\
\hline
\end{tabular}
\end{center}

\caption{\label{subspacesets}
Sets of subspaces of $\R^{d_\theta}$ and their total dimensions.
See also Table~\ref{subspacesets2} in Appendix~\ref{countingmore}.
}

\end{table}

The most important of these numbers are $D^\strat$,
the dimension of $\Span \Theta^\strat = T_\theta S$ and therefore
the dimension of the stratum $S$ that contains $\theta$; and
$D^\free$, the dimension of $\Span \Theta^\free = \Null \dmu(\theta)$,
which we interpret as the number of degrees of freedom of motion along
all the smooth paths leaving $\theta$ on the fiber.
We have already used these dimensions to prove Theorems~\ref{freedomspan}
(showing that $\Span \Theta^\free = \Null \dmu(\theta)$) and~\ref{stratspan}
(showing that $\Span \Theta^\strat = T_\theta S$) in Section~\ref{proofbasis}.

Admittedly, this section makes for mind-numbing reading and
can be safely skipped.
It serves as a reference for anyone who wants
to know the dimensions of specific subspaces, to check our proofs, or
to extend the results in this paper.


Recall that a one-matrix subspace $\phi_{lkjih} \in \ThetaO$ or
a two-matrix subspace $\tau_{lkjih} \in \ThetaT$ has
dimension $\omega_{li} \, \omega_{kh}$.
If the subspaces in a set $\Theta$ are linearly independent,
then the dimension of the space they span is equal to
the sum of the dimensions of the subspaces in the set.
That is, the dimension is
\[
D =
\sum_{\phi_{lkjih} \in \Theta} \omega_{li} \, \omega_{kh} +
\sum_{\tau_{lkjih} \in \Theta} \omega_{li} \, \omega_{kh}.
\]

We can apply this formula to
the one-matrix prebasis $\ThetaO$ or any subset of $\ThetaO$, because
its members, the one-matrix subspaces, are
linearly independent by Lemma~\ref{ThetaOprebasis}.
As $\ThetaO$ spans the entire weight space $\R^{d_\theta}$
(Lemma~\ref{ThetaOprebasis} again), $\DO = \dim \Span \ThetaO = d_\theta$.
By Lemma~\ref{prebaseslinind},
each of $\Theta^\free$, $\Theta^\strat$,  $\Theta^\fiber$ is
a set of linearly independent subspaces, so
we can also apply the formula to any subset of one of them.

In Section~\ref{1matrixprebasis} we defined $\ThetaO^\off \subseteq \ThetaO$,
representing
the one-matrix moves that move off the fiber (change~$\mu(\theta)$), as
a prelude to defining $\ThetaO^\fiber = \ThetaO \setminus \ThetaO^\off$,
representing the one-matrix moves that stay on the fiber
(don't change~$\mu(\theta)$).
$\ThetaO^\off$ contains all the one-matrix subspaces $\phi_{lkjih}$ such that
$l = L$ and $h = 0$, and
$\ThetaO^\fiber$~contains all the other one-matrix subspaces.
In Table~\ref{subspacesets},
two formulae for $\DO^\off = \dim \Span \ThetaO^\off$ appear in~(\ref{DOoff}).
The first formula follows immediately from
the second definition for $\ThetaO^\off$ above it.
The second formula follows from the identity~(\ref{ranksum}).\footnote{
Perhaps it is worth noting that the triple summation in~(\ref{DOoff})
can be simplified to a double summation, because
the term $\omega_{Li} \, \omega_{k0}$ occurs once for each
$j \in [ \max \{ i, 1 \}, \min \{ k + 1, L \} ]$.
In~(\ref{DToff}), the term $\omega_{Li} \, \omega_{k0}$ occurs once for each
$j \in [ \max \{ i, 1 \}, \min \{ k, L - 1 \} ]$.
Hence, we can write
\[
\DO^\off = \sum_{i=0}^L \sum_{k = \max \{ i - 1, 0 \}}^L
(\min \{ k + 1, L \} - \max \{ i - 1, 0 \}) \, \omega_{Li} \, \omega_{k0}
\mbox{~and~}
\DT^\off = \sum_{i=0}^{L-1} \sum_{k = \max \{ i, 1 \}}^L
(\min \{ k, L - 1 \} - \max \{ i - 1, 0 \}) \, \omega_{Li} \, \omega_{k0}.
\]
}

In Section~\ref{2matrixsubspacesuseful}
we defined $\ThetaT^\off \subseteq \ThetaT$, which contains
all the two-matrix subspaces $\tau_{lkjih}$ such that $l = L$ and $h = 0$.
As $\ThetaT^\off \subseteq \Theta^\free$,
the subspaces in $\ThetaT^\off$ are linearly independent
(by Lemma~\ref{prebaseslinind}).
In Table~\ref{subspacesets}, two formulae for $\DT^\off$ appear in~(\ref{DToff}).
The first follows immediately from
the second definition for $\ThetaT^\off$ above it, and
the second follows from the identity~(\ref{ranksum}).

The space spanned by $\ThetaO^\fiber = \ThetaO \setminus \ThetaO^\off$ has
dimension $\DO^\fiber = \DO - \DO^\off$; see the formula~(\ref{DOfreedom}).
Recall that we already derived this another way as~(\ref{dimspanthetaOfiber}).



Two particularly important prebases are $\Theta^\free$ and $\Theta^\fiber$, which
we have already counted.
Lemma~\ref{Dfree} gives us the right-hand side of the formula~(\ref{Dfreedom}).
We include in~(\ref{Dfreedom}) an alternative formula that expresses
$D^\free$ in terms of the interval counts,
\[
D^\free = d_\theta - \sum_{k,i \in [0,L], k+1 \geq i} \omega_{Li} \, \omega_{k0}.
\]
This formula arises from an alternative formula for $\rank \dmu(\theta)$
we derive in Appendix~\ref{funddmu} (Lemma~\ref{rkdmuomega}).
We interpret the summation as a count of
the number of pairs of intervals of the form $([i, L], [0, k])$ that could
undergo a barcode move.
(Any such move would change the value of $\mu(\theta)$, so
it corresponds to a dimension of weight space that is
not in the nullspace of the differential map.)

Recall from Section~\ref{barcmoves}
the set of subspaces that specify barcode moves,
$\ThetaO^\barc = \{ \phi_{lkjih} \neq \{ {\bf 0} \} :
L \geq l \geq k + 1 \geq j \geq i > h \geq 0 \}$.
%
To derive the dimension of the space spanned by $\ThetaO^\barc$, we use
the identity~(\ref{alphabetaid}) and
the fact that in the first summation below,
the term $\omega_{li} \, \omega_{kh}$ appears
$k - i + 2$ times---once for each $j \in [ i, k + 1 ]$.
\begin{eqnarray*}
\DO^\barc
& = & \sum_{L \geq l \geq k+1 \geq j \geq i > h \geq 0}
      \omega_{li} \, \omega_{kh}
  =   \sum_{L \geq l \geq k+1 \geq i > h \geq 0}
      (k - i + 2) \, \omega_{li} \, \omega_{kh}  \\
& = & \sum_{L \geq k+1 \geq i > 0} (k - i + 2) \,
      \left( \sum_{l = k+1}^L \omega_{li} \right) \,
      \left( \sum_{h = 0}^{i-1} \omega_{kh} \right)
  =   \sum_{L \geq k+1 \geq i > 0} (k - i + 2) \, \beta_{k+1,i,i} \, \alpha_{k,k,i-1}.
\end{eqnarray*}

Now consider $\ThetaT^\barc$.
As $\ThetaT^\barc \subseteq \Theta^\strat$, the subspaces in $\ThetaT^\barc$ are
linearly independent (by Lemma~\ref{prebaseslinind}).
We derive $\DT^\barc = \dim \Span \ThetaT^\barc$ as we just derived $\DO^\barc$,
but the indices have the constraint $k \geq j$ (rather than $k + 1 \geq j$).
Hence the term $\omega_{li} \, \omega_{kh}$ appears $k - i + 1$ times---once
for each $j \in [ i, k ]$---and
\[
\DT^\barc =
\sum_{L \geq l > k \geq j \geq i > h \geq 0} \omega_{li} \, \omega_{kh} =
\sum_{L \geq l > k \geq i > h \geq 0} (k - i + 1) \, \omega_{li} \, \omega_{kh} =
\sum_{L > k \geq i > 0} (k - i + 1) \, \beta_{k+1,i,i} \, \alpha_{k,k,i-1}.
\]
When we derive $D^\strat$,
we will use the difference
\[
\DO^\barc - \DT^\barc =
\sum_{L \geq k+1 \geq i > 0} \beta_{k+1,i,i} \, \alpha_{k,k,i-1}.
\]

To derive $D^\strat$ we must address the fact that
$\ThetaO^\off$ and $\ThetaO^\barc$ are not disjoint, nor are
$\ThetaT^\off$ and $\ThetaT^\barc$.
Hence we define
$\ThetaO^{\off,\neg\barc} = \ThetaO^\off \setminus \ThetaO^\barc
= \{ \phi_{Lkji0} \in \ThetaO : L = k$ or $i = 0 \}$,
a set of one-matrix subspaces representing
moves off the fiber that are not barcode moves.
(That is, moves that change $\mu(\theta)$ but
do not increase the rank of any subsequence matrix.)
With the identity~(\ref{ranksum}),
we find that the subspace $\Span \ThetaO^{\off,\neg\barc}$ has dimension
\begin{eqnarray*}
\DO^{\off,\neg\barc}
& = & \sum_{\phi_{Lkji0} \in \ThetaO^{\off,\neg\barc}} \omega_{Li} \, \omega_{k0}
  =   \sum_{\phi_{LLji0} \in \ThetaO} \omega_{Li} \, \omega_{L0} +
      \sum_{\phi_{Lkj00} \in \ThetaO} \omega_{L0} \, \omega_{k0} -
      \sum_{\phi_{LLj00} \in \ThetaO} \omega_{L0} \, \omega_{L0}  \\
& = & \omega_{L0} \, \sum_{j=1}^L \left( \sum_{i=0}^j \omega_{Li} +
      \sum_{k=j-1}^L \omega_{k0} - \omega_{L0} \right)  \\
& = & \rank W \cdot \sum_{j=1}^L
      \left( \rank W_{L \ssq j} + \rank W_{j-1 \ssq 0} - \rank W \right),
\end{eqnarray*}
giving us the formula~(\ref{DOoffnotcomb}).
Analogously, let $\ThetaT^{\off,\neg\barc} = \ThetaT^\off \setminus \ThetaT^\barc
= \{ \tau_{Lkji0} \in \ThetaT : L = k$ or $i = 0 \}$.
The space spanned by $\ThetaT^{\off,\neg\barc}$ has dimension
\begin{eqnarray*}
\DT^{\off,\neg\barc}
& = & \sum_{\tau_{Lkji0} \in \ThetaT^{\off,\neg\barc}} \omega_{Li} \, \omega_{k0}
  =   \sum_{\tau_{LLji0} \in \ThetaT} \omega_{Li} \, \omega_{L0} +
      \sum_{\tau_{Lkj00} \in \ThetaT} \omega_{L0} \, \omega_{k0} -
      \sum_{\tau_{LLj00} \in \ThetaT} \omega_{L0} \, \omega_{L0}  \\
& = & \omega_{L0} \, \sum_{j=1}^{L-1} \left( \sum_{i=0}^j \omega_{Li} +
      \sum_{k=j}^L \omega_{k0} - \omega_{L0} \right)  \\
& = & \rank W \cdot \sum_{j=1}^{L-1}
      \left( \rank W_{L \ssq j} + \rank W_{j \ssq 0} - \rank W \right),
\end{eqnarray*}
giving us the formula~(\ref{DToffnotcomb}).
To derive $D^\strat$,
we will use the difference
\begin{eqnarray*}
\DO^{\off,\neg\barc} - \DT^{\off,\neg\barc}
& = & \rank W \cdot
      \left( \rank W_{L \ssq L} + \rank W_{0 \ssq 0} - \rank W \right)  \\
& = & \rank W \cdot (d_L + d_0 - \rank W).
\end{eqnarray*}


At last we can derive $D^\strat$.

\begin{lemma}
\label{dimstrat}
The dimension $D^\strat$ of the subspace spanned by
the stratum prebasis $\Theta^\strat$ is specified by the formula~(\ref{Dstrat}).
\end{lemma}

\begin{proof}
Recall that the subspaces in $\Theta^\strat$ are linearly independent
(Lemma~\ref{prebaseslinind});
that the subspaces in $\ThetaO$ are linearly independent
(Lemma~\ref{Ojprebasis});
that $\ThetaO^\barc \subseteq \ThetaO$ and $\ThetaO^\off \subseteq \ThetaO$;
that $\ThetaO^{\off,\neg\barc} = \ThetaO^\off \setminus \ThetaO^\barc$ and
$\ThetaT^{\off,\neg\barc} = \ThetaT^\off \setminus \ThetaT^\barc$; and
that $\ThetaO$~is disjoint from $\ThetaT^\barc \cup \ThetaT^\off$.
Therefore,
\begin{eqnarray*}
D^\strat
& = & \dim \Span \Theta^\strat  \\
& = & \dim \Span \left( \left( \ThetaO \setminus \ThetaO^\barc \setminus
                               \ThetaO^\off \right) \cup
                        \ThetaT^\barc \cup \ThetaT^\off \right)  \\
& = & \dim \Span \left( \left( \ThetaO \setminus \ThetaO^\barc \setminus
                               \ThetaO^{\off,\neg\barc} \right) \cup
                        \ThetaT^\barc \cup \ThetaT^{\off,\neg\barc} \right)  \\
& = & \DO - \DO^\barc - \DO^{\off,\neg\barc} + \DT^\barc + \DT^{\off,\neg\barc}  \\
& = & \DO - \left( \DO^{\off,\neg\barc} - \DT^{\off,\neg\barc} \right) -
      \left( \DO^\barc - \DT^\barc \right)  \\
& = & d_\theta - \rank W \cdot (d_L + d_0 - \rank W) -
      \sum_{L \geq k+1 \geq i > 0} \beta_{k+1,i,i} \, \alpha_{k,k,i-1}.
\qedhere
\end{eqnarray*}
\end{proof}

In Section~\ref{normal2basis}, we will confirm that $D^\strat$ is also
the dimension of the stratum that contains $\theta$, by showing that
the subspace normal to that stratum at $\theta$ has
dimension (at least) $d_\theta - D^\strat$.

\clearpage

\section{The Normal Space, the Rowspace of $\mathrm{d}\mu(\theta)$, and
         Prebases that Span Them}
\label{normal}

This section is devoted to determining
two subspaces of the weight space $\R^{d_\theta}$:
the {\em normal space of~$S$ at~$\theta$}, written $N_\theta S$, which is
the maximal subspace perpendicular to a stratum $S$
at a weight vector $\theta \in S$; and
the rowspace of the differential map at $\theta$, written $\row \dmu(\theta)$.
These two subspaces are the orthogonal complements of
$T_\theta S$ and $\Null \dmu(\theta)$, respectively.
Note that $N_\theta S \supseteq \row \dmu(\theta)$, as
$T_\theta S \subseteq \Null \dmu(\theta)$.
The rowspace is of particular interest because
gradient descent directions are usually chosen from $\row \dmu(\theta)$.

One could say that by deriving $T_\theta S$ and $\Null \dmu(\theta)$
in Section~\ref{weightprebases},
we have already derived $N_\theta S$ and $\row \dmu(\theta)$, but in fact
we have not finished deriving $T_\theta S$.
In Sections~\ref{fiberbasis} and~\ref{proofbasis}, we specified
two sets of subspaces, $\Theta^\free$ and $\Theta^\strat$, and
we showed that $\Null \dmu(\theta) = \Span \Theta^\free$ and
$T_\theta S \supseteq \Span \Theta^\strat$.
In Section~\ref{normal2basis} we will complete the proof that
$T_\theta S = \Span \Theta^\strat$, and thereby complete the proof that
$S$ and $T_\theta S$ both have dimension $D^\strat = \dim \Span \Theta^\strat$,
specified by the formula~(\ref{Dstrat}) in Table~\ref{subspacesets}.

A second motivation for directly deriving $\row \dmu(\theta)$ and $N_\theta S$
is to specify two prebases, $\Psi^\free$ and $\Psi^\strat$, that span
$\row \dmu(\theta)$ and $N_\theta S$.
The number of basis vectors needed to express
$\row \dmu(\theta)$ and $N_\theta S$ is
substantially smaller than the number of basis vectors needed to express
$\Null \dmu(\theta)$ and $T_\theta S$.
We anticipate that some computations can be more easily done with
$\Psi^\free$ and $\Psi^\strat$ than with $\Theta^\free$ and $\Theta^\strat$.

Our conclusions are that
$\row \dmu(\theta) = \Span \Psi^\free$ and
$N_\theta S = \Span \Psi^\strat$ where
the {\em freedom normal prebasis} $\Psi^\free$ and
the {\em stratum normal prebasis} $\Psi^\strat$ are defined to be
\begin{eqnarray}
\Psi^\free & = & \{ \psi_{Lki0} \neq \{ {\bf 0} \} :
                     k, i \in [0, L] \mbox{~and~} k + 1 \geq i \}
\hspace{.2in}  \mbox{and}  \label{freenormal}  \\
\Psi^\strat & = &
\Psi^\free \cup
\{ \psi_{lkih} \neq \{ {\bf 0} \}: L \geq l \geq k + 1 \geq i > h \geq 0 \},
\label{psistrat}
\end{eqnarray}
and each $\psi_{lkih}$ satisfying $L \geq l \geq i \geq 0$,
$L \geq k \geq h \geq 0$, and $l > h$ is a subspace of $\R^{d_\theta}$ of the form
\begin{eqnarray}
\psi_{lkih}
& = & \{ (X_L, X_{L-1}, \ldots, X_1) : M \in b_{lli} \otimes a_{khh} \}
\hspace{.2in}  \mbox{where~}
\label{psilkih}  \\
X_j & = &
\left\{ \begin{array}{rl}
W_{l \ssq j}^\top M W_{j-1 \ssq h}^\top, & j \in [ i, k + 1 ],  \\
0, & j \not\in [ i, k + 1 ],
\end{array} \right.  \nonumber
\end{eqnarray}
and $b_{lli}$ and $a_{khh}$ are flow prebasis subspaces,
as described in Section~\ref{flowrelations} and Theorem~\ref{fundnn}.
For example, in the two-matrix case ($L = 2$),
\begin{eqnarray*}
\Psi^\free & = & \{ \psi_{2000}, \psi_{2010}, \psi_{2100}, \psi_{2110},
                    \psi_{2120}, \psi_{2200}, \psi_{2210}, \psi_{2220} \}
\hspace{.2in}  \mbox{and}  \\
\Psi^\strat & = & \{ \psi_{2000}, \psi_{2010}, \psi_{2100}, \psi_{2110},
                    \psi_{2120}, \psi_{2200}, \psi_{2210}, \psi_{2220},
                    \psi_{1010}, \psi_{2121} \}.
\end{eqnarray*}

Note that for $k + 1 \geq i$, the dimension of the subspace $\psi_{lkih}$ is
$\dim b_{lli} \cdot \dim a_{khh} = \omega_{li} \, \omega_{kh}$
(see Lemma~\ref{psilktsih}) and
we can easily construct a basis for $\psi_{lkih}$ given
$\omega_{li}$ basis vectors for $b_{lli}$ and
$\omega_{kh}$ basis vectors for $a_{khh}$.

\begin{figure}
\centerline{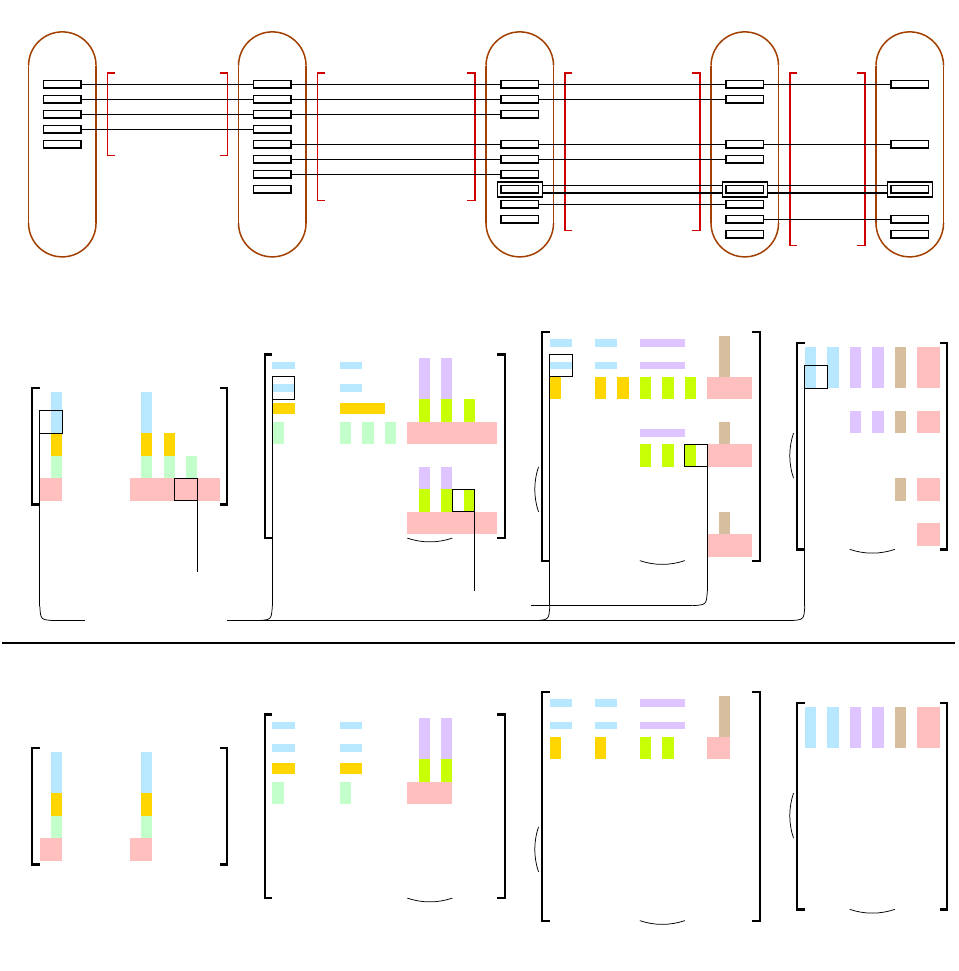}

\caption{\label{nstratum}
Top:  visualizing the prebasis $\Psi^\strat$,
the normal space $N_\theta S = \Span \Psi^\strat$, and
its dimension \mbox{$d_\theta - D^\strat$} at the canonical weight vector
$\tilde{\theta} = (\tilde{I}_4, \tilde{I}_3, \tilde{I}_2, \tilde{I}_1)$ when
every $\omega_{ki} = 1$ except $\omega_{20} = 2$.
Each component with a pink background is
a dimension that varies freely in $N_\theta S$;
it also varies freely in some subspace in $\Psi^\strat$ (such as $\psi_{4342}$).
Blue backgrounds represent a dimension in which four components vary together,
one in each of the four matrices
(such as the single degree of freedom in $\psi_{4410}$).
Gold and purple backgrounds represent dimensions that linearly link
components in three of the four matrices.
Turquoise, green, and tan backgrounds represent dimensions that linearly link
components in two of the four matrices.
You can see the dimension of $N_\theta S$ by counting dimensions of all colors.
Bottom:  visualizing the prebasis $\Psi^\free$,
the rowspace $\row \dmu(\theta) = \Span \Psi^\free$, and
its dimension~$\rank \dmu(\theta) = d_\theta - D^\free$ at $\tilde{\theta}$.
Gradient descent algorithms for training neural networks choose
a descent direction in $\row \dmu(\theta)$.
}
\end{figure}

Figure~\ref{nstratum} illustrates the prebases $\Psi^\strat$ and $\Psi^\free$
at the canonical weight vector for
the network depicted in Figures~\ref{identfactor} and~\ref{ident2moves}.
Compare it with Figure~\ref{dstratum}, which illustrates
the complementary prebases $\Theta^\strat$ and~$\Theta^\free$.

Recall that Theorem~\ref{stratspan} establishes that
$T_\theta S = \Span \Theta^\strat$ by the following logic.
We have $T_\theta S \supseteq \Span \Theta^\strat$
from Lemma~\ref{1strat} and Corollary~\ref{2strat}.
We will see that $N_\theta S \supseteq \Span \Psi^\strat$
in Section~\ref{normal2basis}.
We determined the value of $D^\strat = \dim \Span \Theta^\strat$ in
Lemma~\ref{dimstrat}.
We will determine the value of $\dim \Span \Psi^\strat$ in
Section~\ref{normal2basis} and see that
$\dim \Span \Theta^\strat + \dim \Span \Psi^\strat = d_\theta$.
It follows that $T_\theta S = \Span \Theta^\strat$ and
$N_\theta S = \Span \Psi^\strat$.
From the former it follows that the dimension of $S$ is $D^\strat$.

Sections~\ref{normal1}--\ref{normal2basis} can be safely skipped by
readers who don't want to know how these results were obtained.

\subsection{The Rowspace of the Differential Map}
\label{normal1}

Here we write an explicit expression for $\row \dmu(\theta)$,
the rowspace of the differential map.

What {\em is} the rowspace of a differential map?
Recall that the differential map $\dmu(\theta)$ is a linear map
from a weight displacement $\Delta \theta \in \R^{d_\theta}$
to a $d_L \times d_0$ matrix $\Delta W$.
One could represent $\dmu(\theta)$ as $(d_L d_0) \times d_\theta$ matrix $H$,
and apply it to a weight displacement in the form $H \Delta \theta$.
However, we prefer to write the linear transformation in the form~(\ref{dmu}),
\[
\dmu(\theta)(\Delta \theta) =
\sum_{j=1}^L W_{L \ssq j} \Delta W_j W_{j-1 \ssq 0}
\hspace{.2in}  \mbox{where}  \hspace{.2in}
\Delta \theta = (\Delta W_L, \Delta W_{L-1}, \ldots, \Delta W_1).
\]
Nevertheless, the rowspace, nullspace, and rank of $\dmu(\theta)$ are
the rowspace, nullspace, and rank of the matrix $H$.
See Appendix~\ref{funddmu} for a more detailed explanation.

Just like a matrix, the linear map $\dmu(\theta)$ has a transpose,
analogous to $H^\top$; we write it as $\dmu^\top(\theta)$.
Given a matrix $\Delta W \in \R^{d_L \times d_0}$,
the analogue of applying $H^\top$ to the components of $\Delta W$ is
\begin{equation}
\dmu^\top(\theta)(\Delta W) = (\Delta W W_{L-1 \ssq 0}^\top, \ldots,
W_{L \ssq j}^\top \Delta W W_{j-1 \ssq 0}^\top, \ldots, W_{L \ssq 1}^\top \Delta W).
\label{dmuT}
\end{equation}

Now we can write $\row \dmu(\theta) = \dmu^\top(\theta)(\R^{d_L \times d_0})$,
which is also called the {\em image} of $\dmu^\top(\theta)$ because
it is the set found by applying $\dmu^\top(\theta)$ to
every point in the domain $\R^{d_L \times d_0}$.
In more detail,
\begin{eqnarray}
\row \dmu(\theta)
& = & \{ (X_L, X_{L-1}, \ldots, X_1) : M \in \R^{d_L \times d_0} \}
\hspace{.2in}  \mbox{where~}  \label{rowdmueq}  \\
X_j & = & W_{L \ssq j}^\top M W_{j-1 \ssq 0}^\top.  \nonumber
\end{eqnarray}
For example, in the three-matrix case ($L = 3$),
\[
\row \dmu(\theta) =
\{ (M W_1^\top W_2^\top, W_3^\top M W_1^\top, W_2^\top W_3^\top M) :
M \in \R^{d_3 \times d_0} \}.
\]
This is the most explicit expression we will write for $\row \dmu(\theta)$, and
it is interesting to compare it with~(\ref{nulldmueq}),
our explicit expression for $\Null \dmu(\theta)$
(the orthogonal complement of $\row \dmu(\theta)$).
As an enlightening exercise for the reader, we suggest verifying that
every vector in~(\ref{rowdmueq}) is orthogonal to
every vector in~(\ref{nulldmueq}).
One unfortunate thing~(\ref{rowdmueq}) and~(\ref{nulldmueq}) have in common is
that they reveal very little about the dimension of either subspace, nor
about how to construct a basis for either subspace.

\subsection{A Prebasis for the Rowspace of the Differential Map}
\label{normal1basis}

Here we confirm that the rowspace of the differential map is
$\row \dmu(\theta) = \Span \Psi^\free$, where
$\Psi^\free$ is the freedom normal prebasis, specified by~(\ref{freenormal}).

To decompose $\row \dmu(\theta) = \dmu^\top(\theta)(\R^{d_L \times d_0})$ into
a prebasis, we decompose $\R^{d_L \times d_0}$ into
a prebasis (a direct sum decomposition of $\R^{d_L \times d_0}$), then
apply $\dmu^\top(\theta)$ to each subspace in the latter prebasis.
We will define the subspaces in this prebasis with some extra generality that
we will need in Section~\ref{normal2basis}.
The prebasis will contain some of the {\em subsequence subspaces}
\begin{equation}
e_{lktsih} = b_{lti} \otimes a_{ksh},  \hspace{.2in}
L \geq l \geq t \geq i \geq 0, L \geq k \geq s \geq h \geq 0,  \mbox{~and~}
t > s.
\label{elktsih}
\end{equation}
Our results below require that all the subspaces of the form
$b_{lti}$ and $a_{ksh}$ are flow prebasis subspaces, as described in
Section~\ref{flowrelations} and Theorem~\ref{fundnn}.
The dimension of $e_{lktsih}$ is
$\dim b_{lti} \cdot \dim a_{ksh} = \omega_{li} \, \omega_{kh}$.

We group these subspaces into {\em subsequence prebases}.
Given $t > s$, we define a subsequence prebasis that spans $\R^{d_t \times d_s}$.
\begin{equation}
\mathcal{E}_{ts} = \{ e_{lktsih} \neq \{ 0 \} :
                      l \in [t, L], k \in [s, L], i \in [0, t], h \in [0, s] \}.
\label{Ets}
\end{equation}
The prebasis $\mathcal{E}_{ts}$ pairs every subspace $b_{lti} \in \mathcal{B}_t$
with every subspace $a_{ksh} \in \mathcal{A}_s$.
As $\mathcal{B}_t$ is a prebasis for $\R^{d_t}$ and
$\mathcal{A}_s$ is a prebasis for $\R^{d_s}$ by Lemma~\ref{bases},
it is clear that $\mathcal{E}_{ts}$ is a prebasis for $\R^{d_t \times d_s}$.
Hence we can uniquely represent a subsequence matrix $W_{t \ssq s}$ as
a vector sum of matrices, one from each subspace in $\mathcal{E}_{ts}$.

In this section, we will use only $\mathcal{E}_{L0}$, which contains
subspaces of the form $e_{LkL0i0} = b_{LLi} \otimes a_{k00}$.
But in Section~\ref{normal2basis} we will use all the subsequence prebases.

Returning to the rowspace, we have
\begin{equation}
\row \dmu(\theta) = \dmu^\top(\theta)(\R^{d_L \times d_0}) =
\Span \{ \dmu^\top(\theta)(e_{LkL0i0}) : e_{LkL0i0} \in \mathcal{E}_{L0} \}.
\label{rowdmu}
\end{equation}
This motivates our defining, for indices satisfying
$L \geq l \geq i \geq 0$, $L \geq k \geq h \geq 0$, and $l > h$, the subspaces
\begin{eqnarray*}
\psi_{lkih}
& = & \dmu^\top(\theta)(e_{lklhih})  \\
& = & \{ (X_L, X_{L-1}, \ldots, X_1) : M \in e_{lklhih} \}
\hspace{.2in}  \mbox{where~}  \\
X_j & = & W_{l \ssq j}^\top M W_{j-1 \ssq h}^\top.
\end{eqnarray*}
Then $\row \dmu(\theta) = \Span \{ \psi_{Lki0} : k, i \in [0, L] \}$.
Next, we show that this subspace is $\Span \Psi^\free$.

\begin{lemma}
\label{psilkihform}
For indices satisfying
$L \geq l \geq i \geq 0$, $L \geq k \geq h \geq 0$, and $l > h$,
\begin{icompact}
\item
$X_j = 0$ for $j \not\in [i, k + 1]$---hence
the subspace $\psi_{lkih}$ has the form~(\ref{psilkih});
\item
if the $a_{kji}$'s and $b_{kji}$'s are flow prebasis subspaces,
$X_j \in b_{lji} \otimes a_{k,j-1,h}$ for $j \in [i, k + 1]$; and
\item
if $k + 1 < i$ then $\psi_{lkih} = \{ {\bf 0} \}$.
\end{icompact}
\end{lemma}

\begin{proof}
As $M \in e_{lklhih} = b_{lli} \otimes a_{khh}$,
$X_j \in (W_{l \ssq j}^\top b_{lli}) \otimes (W_{j-1 \ssq h} a_{khh})$.
Recall that for flow subspaces,
$W_{l \ssq j}^\top b_{lli} = b_{lji}$ if $j \geq i$; otherwise,
$W_{l \ssq j}^\top b_{lli} = \{ {\bf 0} \}$.
Symmetrically, $W_{j-1 \ssq h} a_{khh} = a_{k,j-1,h}$ if $k \geq j - 1$; otherwise,
$W_{j-1 \ssq h} a_{khh} = \{ {\bf 0} \}$.
Therefore, for any $M \in b_{lli} \otimes a_{khh}$,
$X_j = 0$ if $j \not\in [ i, k + 1 ]$---hence
$\psi_{lkih}$ has the form~(\ref{psilkih})---whereas
$X_j \in b_{lji} \otimes a_{k,j-1,h}$ if $j \in [i, k + 1]$.

If $k + 1 < i$ then the range $[i, k + 1]$ is empty, so
$\psi_{lkih} = \{ {\bf 0} \}$.
\end{proof}


\begin{cor}
\label{rowdmuspanfree}
$\row \dmu(\theta) = \Span \Psi^\free$.
\end{cor}

\begin{proof}
The identity~(\ref{rowdmu}) implies that
$\row \dmu(\theta) = \Span \{ \psi_{Lki0} : k, i \in [0, L] \}$.
By Lemma~\ref{psilkihform}, if $k + 1 < i$ then $\psi_{Lki0} = \{ {\bf 0} \}$.
Hence the only difference between $\Psi^\free$ as defined by~(\ref{freenormal})
and $\{ \psi_{Lki0} : k, i \in [0, L] \}$ is that
$\Psi^\free$ omits the trivial subspace $\{ {\bf 0} \}$, which does not change
the span.
Therefore, $\row \dmu(\theta) = \Span \Psi^\free$.
\end{proof}

In Section~\ref{normal2basis}, we will see that
the subspaces in $\Psi^\free$ are linearly independent
(Lemma~\ref{psistratlinind}).
Hence $\Psi^\free$ is a prebasis for $\row \dmu(\theta)$.
We will also see (Lemma~\ref{psilktsih}) that
\[
\dim \psi_{Lki0} = \dim e_{LkL0i0} = \omega_{Li} \, \omega_{k0},
\hspace{.2in}  \mbox{for all~} k, i \in [0, L] \mbox{~such that~} k + 1 \geq i.
\]


\subsection{The Normal Space of a Stratum}
\label{normal2}

Here we write an explicit expression for $N_\theta S$,
the normal space of the stratum~$S$ at $\theta$.
To begin, recall from Section~\ref{sec:detmanifolds} that
the stratum associated with a rank list $\underline{r}$ in the fiber of $W$ is
\[
S_{\underline{r}}^W =
\mu^{-1}(W) \cap \bigcap_{L \geq t > s \geq 0} \WDS_{\underline{r}}^{t \ssq s}.
\]
We use the abbreviation $S = S_{\underline{r}}^W$.
Consider a weight vector $\theta \in S$
(thus $\theta$ has rank list $\underline{r}$).
As $S$ is a smooth manifold by Theorem~\ref{stratummanifold},
it has a well-defined tangent space $T_\theta S$ and normal space $N_\theta S$
at $\theta$.
If the fiber $\mu^{-1}(W)$ and
all of the weight-space determinantal sets $\WDS_{\underline{r}}^{t \ssq s}$ are
also smooth manifolds, then $T_\theta S$ is
a subset of those sets' tangent spaces and we can write
\[
T_\theta S \subseteq T_\theta \, \mu^{-1}(W) \cap
\bigcap_{L \ge t > s \ge 0} T_\theta \WDS_{\underline{r}}^{t \ssq s}.
\]
But the fiber and the weight-space determinantal sets are
not necessarily manifolds and might not have tangent spaces at $\theta$.
A generalized idea of ``tangent space'' can circumvent that.
Recall that $T_\theta S \subseteq \Null \dmu(\theta)$ by Lemma~\ref{TSnulldmu},
so we can replace the sometimes undefined $T_\theta \, \mu^{-1}(W)$ with
the everywhere defined $\Null \dmu(\theta)$.
We make an analogous observation for the weight-space determinantal sets.

Recall that $\WDS_{\underline{r}}^{t \ssq s}$ is
the set of weight vectors for which $\rank W_{t \ssq s} = r_{t \ssq s}$.
We will write $\WDS_{\underline{r}}^{t \ssq s}$ as a semi-algebraic set
by introducing two polynomial functions.
The first function, $\mu_{t \ssq s}$, simply takes a weight vector~$\theta$ and
returns the corresponding subsequence matrix $W_{t \ssq s}$.
\[
\mu_{t \ssq s}(\theta) = W_t W_{t-1} \cdots W_{s+1}.
\]

The second function, $\chi_r$, maps a matrix $M$ to
a vector that is zero if and only if $M$ has rank $r$ or less.
This vector lists the determinants of
all the $(r + 1) \times (r + 1)$ minors of $M$.
(Note that this vector might be {\em very} long---the number of minors can be
exponential in the dimensions of $M$.)
A $p \times q$ matrix $M$ lies in the determinantal variety $\DV_r^{p \times q}$
(has rank $r$ or less) if and only if
the determinant of every $(r + 1) \times (r + 1)$ minor of $M$ is zero.
That is, $\DV_r^{p \times q} = \{ M \in \R^{p \times q}: \chi_r(M) = {\bf 0} \}$.
For $r > 0$, $M$ lies in the determinantal manifold $\DM_r^{p \times q}$
(has rank exactly $r$) if and only if
$\chi_r(M) = {\bf 0}$ and $\chi_{r-1}(M) \neq {\bf 0}$.

Each weight-space determinantal set is like
a determinantal manifold, but the domain is weight space and
we fix the rank of one subsequence matrix.
Given a rank list $\underline{r}$ and indices $t$ and $s$,
let $r = r_{t \ssq s}$.
Then
\[
\WDS_{\underline{r}}^{t \ssq s} = \{ \zeta \in \R^{d_\theta}:
\chi_r(\mu_{t \ssq s}(\zeta)) = {\bf 0}  \mbox{~and~}
(r = 0  \mbox{~or~}  \chi_{r-1}(\mu_{t \ssq s}(\zeta)) \neq {\bf 0}) \}.
\]
The differential map of $\chi_r \circ \mu_{t \ssq s}$ at $\theta$ is written
$\di(\chi_r \circ \mu_{t \ssq s})(\theta)$, and  
it is a linear map from a displacement $\Delta \theta$ in weight space to
a vector of reals (in the range of $\chi_r$).
The nullspace of $\di(\chi_r \circ \mu_{t \ssq s})(\theta)$ is
the set of displacements $\Delta \theta$ such that
$\di(\chi_r \circ \mu_{t \ssq s})(\theta)(\Delta \theta) = 0$.

There is a special case to consider:  
$\chi_r(M)$ makes no sense for a matrix $M$ of width $r$ or height $r$,
as $M$~has no $(r + 1) \times (r + 1)$ minors, so
for that case we establish the convention that $\chi_r(M) = (0)$ and thus
\mbox{$\Null \di(\chi_r \circ \mu_{t \ssq s})(\theta) = \R^{d_\theta}$}.


\begin{lemma}
\label{TSnulldchidmu}
Let $\theta \in \R^{d_\theta}$ be a weight vector in a stratum $S$
in the rank stratification of a fiber $\mu^{-1}(W)$.
Let $t$ and $s$ be indices satisfying $L \geq t > s \geq 0$,
let $W_{t \ssq s} = \mu_{t \ssq s}(\theta)$, and
let $r$ be the rank of $W_{t \ssq s}$.
Then $T_\theta S \subseteq
\Null \di(\chi_r \circ \mu_{t \ssq s})(\theta)$.
\end{lemma}

\begin{proof}
Let $P \subset S$ be a smooth path on the stratum.
Every point $\zeta \in P$ satisfies $\chi_r(\mu_{t \ssq s}(\zeta)) = {\bf 0}$.
As $\chi_r \circ \mu_{t \ssq s}$ is
an analytic function of $\R^{d_\theta}$,
any vector $\Delta \zeta$ tangent to $P$ at $\zeta$ is
a direction in which $\chi_r \circ \mu_{t \ssq s}$ has
a directional derivative of zero---that is,
$(\chi_r \circ \mu_{t \ssq s})'_{\Delta \zeta}(\zeta) =
\di(\chi_r \circ \mu_{t \ssq s})(\zeta)(\Delta \zeta) = {\bf 0}$.
Hence $\Delta \zeta \in \Null \di(\chi_r \circ \mu_{t \ssq s})(\zeta)$.
Hence every smooth path containing~$\theta$ on $S$ is tangent to
$\Null \di(\chi_r \circ \mu_{t \ssq s})(\theta)$ at~$\theta$.

As $S$ is a smooth manifold,
for every nonzero vector $\Delta \theta \in T_\theta S$,
there is a smooth path leaving~$\theta$ on $S$ that is
tangent to $\Delta \theta$ at $\theta$; hence
$\Delta \theta \in 
\Null \di(\chi_r \circ \mu_{t \ssq s})(\theta)$.
Therefore, $T_\theta S \subseteq
\Null \di(\chi_r \circ \mu_{t \ssq s})(\theta)$.
\end{proof}

By Lemmas~\ref{TSnulldmu} and~\ref{TSnulldchidmu},
\begin{equation}
T_\theta S \subseteq \Null \dmu(\theta) \cap
\bigcap_{L \ge t > s \ge 0}
\Null \di(\chi_{r_{t \ssq s}} \circ \mu_{t \ssq s})(\theta).
\label{TS}
\end{equation}
Given subspaces satisfying $\sigma \subseteq \bigcap_j \sigma_j$,
their orthogonal complements are related by
$\sigma^\perp \supseteq \sum_j \sigma_j^\perp$.
Hence
\begin{equation}
N_\theta S \supseteq \row \dmu(\theta) +
\sum_{L \ge t > s \ge 0}
\row \di(\chi_{r_{t \ssq s}}\circ \mu_{t \ssq s})(\theta),
\label{NS}
\end{equation}
where the sums are vector sums of subspaces of $\R^{d_\theta}$.
We will show that the right-hand side of~(\ref{NS}) is $\Span \Psi^\strat$, and
that we can replace the inclusion relations in~(\ref{TS}) and~(\ref{NS})
with equality (see the forthcoming Theorem~\ref{dimS}).

To evaluate~(\ref{NS}), observe that
by the chain rule, $\di(\chi_r \circ \mu_{t \ssq s})(\theta) =
\di\chi_r(W_{t \ssq s}) \circ \dmu_{t \ssq s}(\theta)$, where 
$W_{t \ssq s} = \mu_{t \ssq s}(\theta)$.
As the notation is becoming a bit intimidating here,
recall that $\di\chi_r(W_{t \ssq s})$ and $\dmu_{t \ssq s}(\theta)$ are
linear transformations, so you may think of the composition
$\di\chi_r(W_{t \ssq s}) \circ \dmu_{t \ssq s}(\theta)$ as  
a matrix multiplication.
The rowspace of $\di\chi_r(W_{t \ssq s}) \circ \dmu_{t \ssq s}(\theta)$ is
the image of its transpose
$\dmu^\top_{t \ssq s}(\theta) \circ \di\chi_r^\top(W_{t \ssq s})$, hence
\[
\row \di(\chi_r \circ \mu_{t \ssq s})(\theta)
= \dmu_{t \ssq s}^\top(\theta)(\row \di\chi_r(W_{t \ssq s})).
\]
Thus we need to know $\row \di\chi_r(W_{t \ssq s})$ where $r = \rank W_{t \ssq s}$.
It is well known~\cite{harris95} that the subspaces tangent and normal to
the determinantal variety $\DV_r^{p \times q}$ at
a rank-$r$ matrix $M$ are
\begin{eqnarray*}
T_M \DV_r^{p \times q}
& = & \col M \otimes \R^q + \R^p \otimes \row M
\hspace{.2in}  \mbox{and}  \\
N_M \DV_r^{p \times q}
& = & \Null M^\top \otimes \Null M.
\end{eqnarray*}
Recall that $\DV_r^{p \times q}$ is the set of matrices $H$ such that
$\chi_r(H) = 0$.
Hence the nullspace of~$\di\chi_r(M)$ is
the tangent space of $\DV_r^{p \times q}$ at $M$; that is,
\begin{eqnarray*}
\Null \di\chi_r(W_{t \ssq s})
& = & T_{W_{t \ssq s}} \DV_r^{d_t \times d_s}
  =   \col W_{t \ssq s} \otimes \R^{d_s} + \R^{d_t} \otimes \row W_{t \ssq s},  \\
\row \di\chi_r(W_{t \ssq s})
& = & N_{W_{t \ssq s}} \DV_r^{d_t \times d_s}
  =   \Null W_{t \ssq s}^\top \otimes \Null W_{t \ssq s},
\hspace{.2in}  \mbox{and}  \\
\row \di(\chi_r \circ \mu_{t \ssq s})(\theta)
& = & \dmu_{t \ssq s}^\top(\theta)
      (\Null W_{t \ssq s}^\top \otimes \Null W_{t \ssq s}).
\end{eqnarray*}

We need an expression for $\dmu_{t \ssq s}^\top(\theta)$.
Analogous to~(\ref{dmu}) and~(\ref{dmuT}), by the product rule we have
\begin{eqnarray}
\dmu_{t \ssq s}(\theta)(\Delta \theta)
& = & \sum_{j=s+1}^t W_{t \ssq j} \Delta W_j W_{j-1 \ssq s}
\hspace{.2in}  \mbox{and therefore}  \nonumber  \\
\dmu_{t \ssq s}^\top(\theta)(M)
& = & (X_L, X_{L-1}, \ldots, X_1)
\hspace{.2in}  \mbox{where}  \nonumber  \\
X_j & = &
\left\{ \begin{array}{rl}
W_{t \ssq j}^\top M W_{j-1 \ssq s}^\top, & j \in [ s + 1, t ],  \\
0, & j \not\in [ s + 1, t ].
\end{array} \right.
\label{Xj}
\end{eqnarray}
Note that $X_j$ is defined differently in~(\ref{Xj}) than it was in
Section~\ref{normal1}; it depends on the indices $t$ and $s$.
Now
\begin{equation}
\row \di(\chi_r \circ \mu_{t \ssq s})(\theta) =
\{ (X_L, X_{L-1}, \ldots, X_1) :
   M \in \Null W_{t \ssq s}^\top \otimes \Null W_{t \ssq s} \}.
\label{NthetaWDS}
\end{equation}

From~(\ref{NS}), (\ref{rowdmueq}), and~(\ref{NthetaWDS}),
we can write the space normal to the stratum $S$ as
\begin{eqnarray}
N_\theta S & = &
\left\{ (X^{(L0)}_L, X^{(L0)}_{L-1}, \ldots, X^{(L0)}_1) :
        M^{(L0)} \in \R^{d_L \times d_0} \right\} +  \nonumber  \\
& &
\sum_{L \ge t > s \ge 0} \left\{ (X^{(ts)}_L, X^{(ts)}_{L-1}, \ldots, X^{(ts)}_1) :
M^{(ts)} \in \Null W_{t \ssq s}^\top \otimes \Null W_{t \ssq s} \right\}
\hspace{.2in}  \mbox{where}  \label{NthetaSeq}  \\
X^{(ts)}_j & = &
\left\{ \begin{array}{rl}
W_{t \ssq j}^\top M^{(ts)} W_{j-1 \ssq s}^\top, & j \in [ s + 1, t ],  \\
0, & j \not\in [ s + 1, t ].
\end{array} \right.
\nonumber
\end{eqnarray}
For example, in the two-matrix case ($L = 2$),
\[
N_\theta S = \{ (M W_1^\top + M'', W_2^\top M + M') :
M \in \R^{d_2 \times d_0}, M'' \in \Null W_2^\top \otimes \Null W_2,
M' \in \Null W_1^\top \otimes \Null W_1 \}.
\]

Beware that in~(\ref{NthetaSeq}) we have not yet justified the ``$=$'' sign,
only a ``$\supseteq$'' relation, as~(\ref{NthetaSeq}) follows from (\ref{NS});
but as we will prove equality in Theorem~\ref{dimS},
we write~(\ref{NthetaSeq}) as an identity now.
This is the most explicit expression we will write for $N_\theta S$;
compare it with~(\ref{TthetaSeq}), our explicit expression for $T_\theta S$.
The expressions~(\ref{TthetaSeq}) and~(\ref{NthetaSeq})
reveal little about the dimension of either subspace, nor
about how to construct a basis for either subspace.

\subsection{A Prebasis for the Subspace Orthogonal to a Stratum}
\label{normal2basis}

Here we confirm that the normal space of the stratum~$S$ at $\theta$ is
$N_\theta S = \Span \Psi^\strat$, where
$\Psi^\strat$ is the stratum normal prebasis, specified by~(\ref{psistrat}).
We also show that the dimension of $N_\theta S$ is $d_\theta - D^\strat$.
As a~consequence, we verify that our formula~(\ref{Dstrat}) for $D^\strat$ is
the dimension of $S$.

To begin, we use the flow subspaces
$B_{t,t,s+1} = \Null W_{t \ssq s}^\top$ and $A_{t-1,s,s} = \Null W_{t \ssq s}$
from Section~\ref{flowspaces},
their decomposition into flow prebasis subspaces $b_{lti}$ and $a_{ksh}$
from Lemma~\ref{fundnn}, and
the subsequence subspaces $e_{lktsih} = b_{lti} \otimes a_{ksh}$
defined by~(\ref{elktsih}) in Section~\ref{normal1basis}.
We can write the subspace in~(\ref{NthetaWDS})
as a vector sum of subsequence subspaces,
\[
\Null W_{t \ssq s}^\top \otimes \Null W_{t \ssq s} =
B_{t,t,s+1} \otimes A_{t-1,s,s}
= \left( \sum_{l=t}^L \sum_{i=s+1}^t b_{lti} \right) \otimes
  \left( \sum_{k=s}^{t-1} \sum_{h=0}^s a_{ksh} \right)
= \sum_{l=t}^L \sum_{k=s}^{t-1} \sum_{i=s+1}^t \sum_{h=0}^s e_{lktsih}.
\]
The summands $e_{lktsih}$ are members of
the subsequence prebasis $\mathcal{E}_{ts}$,
defined by~(\ref{Ets}) in Section~\ref{normal1basis}.
This expression implies that each subspace $e_{lktsih} \in \mathcal{E}_{ts}$ is
a subset of $\Null W_{t \ssq s}^\top \otimes \Null W_{t \ssq s}$
if $t > k$ and $i > s$.
By substituting this identity into~(\ref{NthetaWDS}),
we can write
\[
\row \di(\chi_r \circ \mu_{t \ssq s})(\theta) =
\sum_{l=t}^L \sum_{k=s}^{t-1} \sum_{i=s+1}^t \sum_{h=0}^s
\{ (X_L, X_{L-1}, \ldots, X_1) : M \in e_{lktsih} \},
\]
where $r = \rank W_{t \ssq s}$ and each $X_j$ is defined by~(\ref{Xj}).
This motivates our defining the subspaces
\[
\psi_{lktsih} =
\{ (X_L, X_{L-1}, \ldots, X_1) : M \in e_{lktsih} \},  \hspace{.2in}
L \geq l \geq t \geq i \geq 0, L \geq k \geq s \geq h \geq 0,  \mbox{~and~}
t > s,
\]
so we can write
$ \row \di(\chi_r \circ \mu_{t \ssq s})(\theta) =
\sum_{l=t}^L \sum_{k=s}^{t-1} \sum_{i=s+1}^t \sum_{h=0}^s \psi_{lktsih}$.

The following lemma elucidates the relationship between
a matrix $M \in e_{lktsih}$ and each corresponding matrix $X_j$ in~(\ref{Xj}).
Two of its claims use the flow bases described in
Sections~\ref{flowbases} and~\ref{basestranspose}.
Recall that each $K_{lti}$ is a $d_t \times \omega_{li}$ matrix
whose columns form a basis for $b_{lti}$,
each $J_{ksh}$ is a $d_s \times \omega_{kh}$ matrix
whose columns form a basis for $a_{ksh}$, and
the flow relationships state that
$K_{lji} = W_{t \ssq j}^\top K_{lti}$ and $J_{k,j-1,h} = W_{j-1 \ssq s} J_{ksh}$.

\begin{lemma}
\label{MXj}
Define $X_j$ by~(\ref{Xj}) where $M \in e_{lktsih}$.
Then $X_j = 0$
for each $j \not\in [ \max \{ i, s + 1 \}, \min \{ k + 1, t \} ]$, and
the following claims hold
for each $X_j$ with $j \in [ \max \{ i, s + 1 \}, \min \{ k + 1, t \} ]$.
\begin{enumerate}[(a)]
\item  $X_j$ has the same rank as $M$.
       In particular, $X_j = 0$ if and only if $M = 0$.
\item  $X_j \in e_{l,k,j,j-1,i,h} = b_{lji} \otimes a_{k,j-1,h}$.
\item  The identity~(\ref{Xj}), mapping $M$ to $X_j$, is
       a bijection from $e_{lktsih}$ to $e_{l,k,j,j-1,i,h}$.
       Its inverse is the identity
       $M = K_{lti} K_{lji}^+ X_j J_{k,j-1,h}^{+\top} J_{ksh}^\top$ where
       $H^+$ denotes the Moore--Penrose pseudoinverse of a matrix $H$ and
       $H^{+\top}$~denotes the transpose of $H^+$.
\item  For every $z \in [ \max \{ i, s + 1 \}, \min \{ k + 1, t \} ]$,
       $X_z = K_{lzi} K_{lji}^+ X_j J_{k,j-1,h}^{+\top} J_{k,z-1,h}^\top$.
       Hence $X_j$ uniquely determines every other $X_z$ in the range.
\item  For every $X_j \in e_{l,k,j,j-1,i,h}$, $\psi_{lktsih}$ contains
       exactly one weight vector having that value of~$X_j$.
\end{enumerate}
\end{lemma}

\begin{proof}
Given some $M \in e_{lktsih} = b_{lti} \otimes a_{ksh}$ and $j \in [ s + 1, t ]$,
the identity~(\ref{Xj}) sets $X_j = W_{t \ssq j}^\top M W_{j-1 \ssq s}^\top$.
As the $\mathcal{B}_t$'s are flow prebases,
$W_{t \ssq j}^\top b_{lti} = b_{lji}$ if $j \geq i$; otherwise,
$W_{t \ssq j}^\top b_{lti} = \{ {\bf 0} \}$
(as $b_{lti} \subseteq B_{lti} \subseteq \Null W_{t \ssq i-1}^\top$).
Symmetrically, as the $\mathcal{A}_s$'s are flow prebases,
$W_{j-1 \ssq s} a_{ksh} = a_{k,j-1,h}$ if $k \geq j - 1$; otherwise,
$W_{j-1 \ssq s} a_{ksh} = \{ {\bf 0} \}$
(as $a_{ksh} \subseteq A_{ksh} \subseteq \Null W_{k+1 \ssq s}$).
Hence if $j \not\in [ \max \{ i, s + 1 \}, \min \{ k + 1, t \} ]$, then
$X_j = 0$ as claimed.
Otherwise, $X_j \in b_{lji} \otimes a_{k,j-1,h}$, confirming claim~(b).

Any member of $b_{lti} \otimes a_{ksh}$ can be written in the form
$M = K_{lti} C J_{ksh}^\top$ where $C$ is an $\omega_{li} \times \omega_{kh}$
matrix of real coefficients.
(Each coefficient scales the outer product of
one basis vector in $K_{lti}$ and one basis vector in~$J_{ksh}$.)
As $K_{lti}$ has rank $\omega_{li}$ and $J_{ksh}$ has rank $\omega_{kh}$,
$M$ has the same rank as $C$.

For any $j \in [ \max \{ i, s + 1 \}, \min \{ k + 1, t \} ]$,
$X_j = W_{t \ssq j}^\top M W_{j-1 \ssq s}^\top
= W_{t \ssq j}^\top K_{lti} C J_{ksh}^\top W_{j-1 \ssq s}^\top
= K_{lji} C J_{k,j-1,h}^\top$.
As every vector in $e_{l,k,j,j-1,i,h}$ can be written in the form
$K_{lji} C J_{k,j-1,h}^\top$ for some $C$,
the identity~(\ref{Xj}) is a surjection onto $e_{l,k,j,j-1,i,h}$.
As $K_{lji}$ has rank $\omega_{li}$ and $J_{k,j-1,h}$ has rank $\omega_{kh}$,
$X_j$~has the same rank as $C$ and
we can recover $C$ from $X_j$ by writing $C = K_{lji}^+ X_j J_{k,j-1,h}^{+\top}$.
Therefore $\rank X_j = \rank C = \rank M$, confirming claim~(a), and
$M = K_{lti} K_{lji}^+ X_j J_{k,j-1,h}^{+\top} J_{ksh}^\top$, confirming claim~(c).
Then for any $z \in [ \max \{ i, s + 1 \}, \min \{ k + 1, t \} ]$,
\begin{eqnarray*}
X_z & = & W_{t \ssq z}^\top M W_{z-1 \ssq s}^\top  \\
    & = & W_{t \ssq z}^\top K_{lti} K_{lji}^+ X_j J_{k,j-1,h}^{+\top} J_{ksh}^\top
          W_{z-1 \ssq s}^\top  \\
    & = & K_{lzi} K_{lji}^+ X_j J_{k,j-1,h}^{+\top} J_{k,z-1,h}^\top,
\end{eqnarray*}
confirming claim~(d).

Claim~(c) implies that for every matrix $X_j \in e_{l,k,j,j-1,i,h}$,
some weight vector in $\psi_{lktsih}$ has that value of~$X_j$.
Claim~(d) implies that no other weight vector in $\psi_{lktsih}$ has
that value of~$X_j$, confirming claim~(e).
\end{proof}

\begin{lemma}
\label{psilktsih}
Consider a subspace $\psi_{lktsih}$ with valid indices (i.e.,
$L \geq l \geq t \geq i \geq 0$, $L \geq k \geq s \geq h \geq 0$, and $t > s$).
If $k + 1 < i$ then $\psi_{lktsih} = \{ {\bf 0} \}$; whereas
if $k + 1 \geq i$ then $\psi_{lktsih}$ has the same dimension as $e_{lktsih}$:
\[
\dim \psi_{lktsih} = \dim e_{lktsih} = \omega_{li} \, \omega_{kh},
\hspace{.2in}
L \geq l \geq t \geq i \geq 0, L \geq k \geq s \geq h \geq 0,
t > s,  \mbox{~and~}  k + 1 \geq i.
\]
Moreover, $\psi_{lktsih} = \psi_{lkt's'ih}$ if
$t > k$, $t' > k$, $i > s$, and $i > s'$
(assuming $\psi_{lkt's'ih}$ has valid indices too).
\end{lemma}

\begin{proof}
If $k + 1 < i$ then $\psi_{lktsih} = \{ {\bf 0} \}$ by Lemma~\ref{MXj}, because
the range $[ \max \{ i, s + 1 \}, \min \{ k + 1, t \} ]$ is empty.
Whereas if $k + 1 \geq i$, then there exists at least one index
$j \in [ \max \{ i, s + 1 \}, \min \{ k + 1, t \} ]$,
because $t \geq i$, $k \geq s$, and $t > s$ by assumption.
As claim~(c) of Lemma~\ref{MXj} applies to $X_j$, $\psi_{lktsih}$ has
the same dimension as $e_{lktsih}$, namely, $\omega_{li} \, \omega_{kh}$.

If $t > k$, $t' > k$, $i > s$, and $i > s'$ then $\psi_{lktsih} = \psi_{lkt's'ih}$
because of the following three observations:
\begin{itemize}
\item
$\min \{ k + 1, t \} = k + 1 = \min \{ k + 1, t' \}$ and
$\max \{ i, s + 1 \} = i = \max \{ i, s' + 1 \}$, so
both subspaces have nonzero $X_j$ only in the positions
$j \in [i, k + 1]$ by Lemma~\ref{MXj};
\item
by claim~(d) of Lemma~\ref{MXj},
each $X_j$ in these positions determines the others through
a formula independent of $s$ and~$t$; and
\item
by claims~(e) and~(b), both $\psi_{lktsih}$ and $\psi_{lkt's'ih}$ have
a weight vector for every $X_j \in e_{l,k,j,j-1,i,h}$ and
no weight vector with $X_j \not\in e_{l,k,j,j-1,i,h}$.
\qedhere
\end{itemize}
\end{proof}

We can write some subspaces with fewer indices by defining
\[
\psi_{lkih} = \psi_{lklhih},
\hspace{.2in}
L \geq l \geq i \geq 0, L \geq k \geq h \geq 0,  \mbox{~and~}  l > h.
\]

\begin{cor}
\label{psilklhih}
If $\psi_{lktsih}$ has valid indices satisfying $t > k$ and $i > s$,
then $\psi_{lktsih} = \psi_{lklhih} = \psi_{lkih}$.
\end{cor}

\begin{proof}
Valid indices imply that $l \geq t$ and $s \geq h$,
so $l > k$ and $i > h$.
By Lemma~\ref{psilktsih}, $\psi_{lktsih} = \psi_{lklhih}$.
\end{proof}

Recall that
$\row \di(\chi_r \circ \mu_{t \ssq s})(\theta) =
\sum_{l=t}^L \sum_{k=s}^{t-1} \sum_{i=s+1}^t \sum_{h=0}^s \psi_{lktsih}$
(assuming $r = \rank W_{t \ssq s}$).
The indices in the summations always satisfy $t > k$ and $i > s$, so
by Corollary~\ref{psilklhih}, we can shorten the indices to $\psi_{lkih}$.
We can discard the subspaces with $k + 1 < i$, as
they are all $\{ {\bf 0} \}$ by Lemma~\ref{psilktsih}.
Thus
\begin{align*}
& \row \di(\chi_r \circ \mu_{t \ssq s})(\theta) =
\Span \Psi_{t \ssq s},  \hspace{.2in}  \mbox{where}  \\
& \Psi_{t \ssq s} =
\{ \psi_{lkih} \neq \{ {\bf 0} \} :
   l \in [t, L], k \in [s, t - 1], i \in [s + 1, t], h \in [0, s],
   \mbox{~and~}  k + 1 \geq i \}.
\end{align*}

We return to determining a prebasis for $N_\theta S$.
It follows from~(\ref{NS}) and Corollary~\ref{rowdmuspanfree} that
$N_\theta S \supseteq \Span \Psi^\strat$ where
\begin{eqnarray*}
\Psi^\strat & = &
\Psi^\free \cup \bigcup_{L \geq t > s \geq 0} \Psi_{t \ssq s}  \nonumber  \\
& = &
\{ \psi_{Lki0} \neq \{ {\bf 0} \} :
   k, i \in [0, L] \mbox{~and~} k + 1 \geq i \} \cup
\{ \psi_{lkih} \neq \{ {\bf 0} \} :
   L \geq l \geq k + 1 \geq i > h \geq 0 \}.
\end{eqnarray*}
This reiterates our definition~(\ref{psistrat}) of $\Psi^\strat$
from the beginning of Section~\ref{normal}.
In this definition, some subspaces appear twice.
For example, $\psi_{L,L-1,1,0}$ is in both the former and the latter set.
The only subspaces unique to the first set are those of the form
$\psi_{LLi0}$ or $\psi_{Lk00}$.
We rewrite $\Psi^\strat$ so that each subspace appears in exactly one set,
which will help us count the dimension of $\Span \Psi^\strat$.
\begin{eqnarray}
\Psi^\strat & = &
\{ \psi_{LLi0} \neq \{ {\bf 0} \} : i \in [0, L] \} \cup
\{ \psi_{Lk00} \neq \{ {\bf 0} \} : k \in [0, L - 1] \} \cup  \nonumber  \\
& &
\{ \psi_{lkih} \neq \{ {\bf 0} \}: L \geq l \geq k + 1 \geq i > h \geq 0 \}.
\label{psistrat2}
\end{eqnarray}

The next two lemmas prove the linear independence of $\Psi^\strat$ and
derive the dimension of $\Span \Psi^\strat$.

\begin{lemma}
\label{psistratlinind}
The subspaces in $\Psi^\strat$ are linearly independent.
\end{lemma}

\begin{proof}
Given a weight vector $\xi = (X_L, X_{L-1}, \ldots, X_1)$,
we define an operator that picks one matrix out of the weight vector:
let $M_j(\xi) = X_j$.
Given a subspace $\sigma$ of weight vectors,
let $M_j(\sigma) = \{ M_j(\xi) : \xi \in \sigma \}$.

Suppose for the sake of contradiction that $\Psi^\strat$ is
not linearly independent.
Then there is a subspace $\psi_{lkih} \in \Psi^\strat$ and
a nonzero weight vector $\zeta \in \psi_{lkih}$ such that
$\zeta$ is a linear combination of weight vectors taken from
the subspaces in $\Psi^\strat \setminus \{ \psi_{lkih} \}$.
Let $j$ be an index such that $M_j(\zeta) \neq 0$.
By claim~(b) of Lemma~\ref{MXj}, $M_j(\zeta) \in e_{l,k,j,j-1,i,h}$.

By claims~(b) and~(e) of Lemma~\ref{MXj},
for every subspace $\psi_{qpnm} \in \Psi^\strat$,
either $M_j(\psi_{qpnm}) = e_{q,p,j,j-1,n,m}$ or
$M_j(\psi_{qpnm}) = \{ 0 \}$.  
Hence, distinct members of $\Psi^\strat$ are mapped to $\{ 0 \}$ or
to distinct members of the prebasis~$\mathcal{E}_{j,j-1}$,
whose members are linearly independent.
Therefore, $M_j(\zeta) \in e_{l,k,j,j-1,i,h}$ cannot be written as
a linear combination of matrices taken from
subspaces in $\mathcal{E}_{j,j-1} \setminus \{ e_{l,k,j,j-1,i,h} \}$, so
$\zeta$ cannot be written as a linear combination of weight vectors taken from
$\Psi^\strat \setminus \{ \psi_{lkih} \}$, a~contradiction.
It follows that the subspaces in $\Psi^\strat$ are linearly independent.
\end{proof}

\begin{lemma}
\label{psistratdim}
The dimension of $\Span \Psi^\strat$ is $d_\theta - D^\strat$.
\end{lemma}

\begin{proof}
The subspaces in $\Psi^\strat$ are linearly independent by
Lemma~\ref{psistratlinind}, so the dimension of $\Span \Psi^\strat$ is
the sum of the dimensions of the subspaces in $\Psi^\strat$.
If we use the formula~(\ref{psistrat2}) as a list of those subspaces, and
recall the formulae~(\ref{intervallayer}), (\ref{Wrank}),
(\ref{alphabetaid}), and~(\ref{Dstrat}), this sum is
\begin{eqnarray*}
\sum_{\psi_{lkih} \in \Psi^\strat} \omega_{li} \, \omega_{kh}
& = &
\sum_{i=0}^L \omega_{Li} \, \omega_{L0} +
\sum_{k=0}^{L-1} \omega_{L0} \, \omega_{k0} +
\sum_{L \geq l \geq k+1 \geq i > h \geq 0}
\omega_{li} \, \omega_{kh}  \\
& = &
\omega_{L0}
\left(\sum_{i=0}^L \omega_{Li} +\sum_{k=0}^L \omega_{k0} - \omega_{L0} \right) +
\sum_{L \geq k+1 \geq i > 0}
\left( \sum_{l=k+1}^L \omega_{li} \right)
\left( \sum_{h=0}^{i-1} \omega_{kh} \right)  \\
& = &
\rank W \cdot (d_L + d_0 - \rank W) +
\sum_{L \geq k+1 \geq i > 0} \beta_{k+1,i,i} \, \alpha_{k,k,i-1}  \\
& = &
d_\theta - D^\strat.
\qedhere
\end{eqnarray*}
\end{proof}

At last we complete our proof of the identities of $T_\theta S$ and $N_\theta S$
and of our formula for the dimension of a stratum.

\begin{theorem}
\label{dimS}
Consider a matrix $W$, its fiber $\mu^{-1}(W)$, and
a stratum $S = S_{\underline{r}}^W$ in $W$'s fiber
with rank list~$\underline{r}$.
\begin{icompact}
\item
The dimension of $S$ is~$D^\strat$.
\item
$\Theta^\strat$ is a prebasis for $T_\theta S$.
(In particular, $T_\theta S = \Span \Theta^\strat$.)
\item
$\Psi^\strat$ is a prebasis for $N_\theta S$.
(In particular, $N_\theta S = \Span \Psi^\strat$.)
\end{icompact}
\end{theorem}

\begin{proof}
By Lemma~\ref{1strat} and Corollary~\ref{2strat},
$\Span \Theta^\strat \subseteq T_\theta S$.
In this section we see that $\Span \Psi^\strat \subseteq N_\theta S$.
In Section~\ref{counting} we saw that
$\Span \Theta^\strat$ has dimension $D^\strat$.
By Lemma~\ref{psistratdim},
$\Span \Psi^\strat$ has dimension $d_\theta - D^\strat$.
As $N_\theta S$ is the orthogonal complement of $T_\theta S$,
$\dim N_\theta S + \dim T_\theta S = d_\theta$.
The sum of the dimensions of $\Span \Theta^\strat$ and $\Span \Psi^\strat$ is
$d_\theta$, so
$\Span \Theta^\strat = T_\theta S$ and $\Span \Psi^\strat = N_\theta S$.
Hence the dimension of $T_\theta S$ is $D^\strat$ and thus
the dimension of $S$ is $D^\strat$.
As $\Theta^\strat$ is linearly independent by Lemma~\ref{prebaseslinind},
$\Theta^\strat$ is a prebasis for $T_\theta S$.
As $\Psi^\strat$ is linearly independent by Lemma~\ref{psistratlinind},
$\Psi^\strat$ is a prebasis for $N_\theta S$.
\end{proof}

\clearpage

\section{Computing the Stratum Dag}
\label{computedag}

Here we describe an algorithm for generating the stratum dag
that represents the rank stratification of a fiber~$\mu^{-1}(W)$---for instance,
the stratum dag in Figure~\ref{strat564}---given
the integers $d_0, d_1, \ldots, d_L$, and $\rank W$ as input.

First we specify what information is stored with
each vertex and directed edge of the dag
(but we do not specify how the directed graph data structure is implemented).
Each dag vertex $v$ represents
a nonempty stratum $S_{\underline{r}}$ in the rank stratification.
The vertex data structure stores four fields:
the integers $v$.dim, $v$.dof, and $v$.rdof introduced in
Section~\ref{foretaste}
(recall Figures~\ref{strat112}, \ref{strat1111}, and~\ref{strat564}),
and $v$.ranklist stores the stratum's rank list~$\underline{r}$.
The dimension of $S_{\underline{r}}$ is
$v$.dim $= D^\strat$ from formula~(\ref{Dstrat});
the number of degrees of freedom of motion on the fiber from
a point on $S_{\underline{r}}$ (i.e., the dimension of $\Null \dmu(\theta)$) is
$v$.dof $= D^\free$ from formula~(\ref{Dfreedom}); and
the number of rank-increasing degrees of freedom is
$v$.rdof $=$ $v$.dof $-$ $v$.dim.
Each dag edge $e$ represents a rank-$1$ barcode move.
The edge data structure stores three fields:
$e$.origin and $e$.dest are the origin and destination vertices of
the directed edge, and $e$.label is an index tuple $(l, k, i, h)$,
listing the indices associated with the barcode move.

Our first task is to find a way to list
all the strata in the rank stratification---that is,
to enumerate all the valid rank lists that contain
the specified values of the $d_j$'s and $\rank W$.
This is not as easy as you might think.
Section~\ref{enumeratesec} provides an algorithm that  
creates one vertex in the stratum dag for each valid rank list, thus
one vertex per stratum.
(Appendix~\ref{valid} shows that for every valid rank list
$\underline{r}$, the stratum $S_{\underline{r}}$ is nonempty.)
Section~\ref{enumeratesec2} suggests an alternative algorithm.

Our second task is to create the edges of the stratum dag.
Section~\ref{dagedges} gives an algorithm for
identifying the rank-$1$ abstract barcode moves and
creating the edges and labels that represent them.

\subsection{Enumerating the Valid Rank Lists}
\label{enumeratesec}

Given specified values for the $d_j$'s and $\rank W$,
this section describes a procedure to enumerate
all the valid rank lists that include those values.
We assume that $\rank W \leq d_j$ for all $j \in [0, L]$; otherwise,
there is no valid rank list with those values and
the fiber $\mu^{-1}(W)$ is the empty set.

\begin{figure}
\centerline{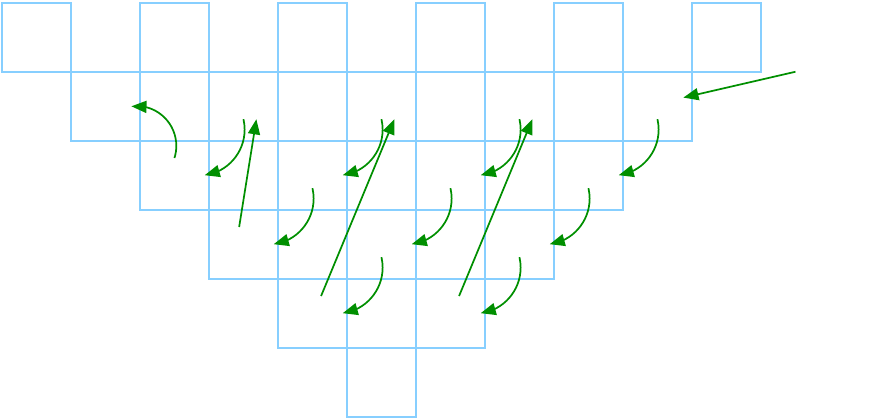}

\caption{\label{enumerate}
Green arrows indicate the order in which our algorithm for
enumerating valid rank lists assigns
ranks to the subsequence matrices.
The $d_j$'s and $\rank W$ are fixed (black text);
only the ranks in red can vary from stratum to stratum.
The $\geq$ and $\leq$ symbols indicate some of the constraints that
the ranks must satisfy.
There are additional constraints:  each interval count
$\omega_{ki} = \rank W_{k \ssq i} - \rank W_{k \ssq i-1} -
\rank W_{k+1 \ssq i} + \rank W_{k+1 \ssq i-1}$ must be nonnegative
(by the Frobenius rank inequality).
For example, $\rank W_2 \leq d_1 - \rank W_1 + \rank W_2W_1$.
}
\end{figure}

Our procedure selects one subsequence matrix and iterates through
all the valid choices of the rank of that subsequence matrix.
For each valid rank, it recursively finds
all the rank lists consistent with that choice. 
The order in which the subsequence matrix ranks are selected is depicted in
Figure~\ref{enumerate}.
Recall that for a specified fiber~$\mu^{-1}(W)$,
$\rank W_{L \ssq 0} = \rank W$ is fixed and so is each $\rank W_{j \ssq j} = d_j$.
The other ranks $\rank W_{k \ssq i}$ are chosen in order of increasing $i$, and
for ranks with equal $i$, in order of increasing $k$.

For each rank $\rank W_{k \ssq i}$ with $i = 0$,
the procedure tries all values in the range
\[
\rank W_{k \ssq 0} \in [ \rank W, \min \{ \rank W_{k-1 \ssq 0}, d_k \} ],
\hspace{.2in}  k \in [1, L - 1].
\]
For the other ranks ($i \neq 0$), the procedure tries all values in the range
\[
\rank W_{k \ssq i} \in [ \rank W_{k \ssq i-1},
\min \{ \rank W_{k-1 \ssq i} + \rank W_{k \ssq i-1} - \rank W_{k-1 \ssq i-1}, d_k \}
],  \hspace{.2in}  L \geq k > i \geq 1.
\]
The constraints that $\rank W_{k \ssq 0} \geq \rank W$,
$\rank W_{k \ssq i} \geq \rank W_{k \ssq i-1}$ for $i \geq 1$, and
$\rank W_{k \ssq i} \leq d_k$ for all $i$ are best understood from
gazing at the $\geq$ and $\leq$ symbols in Figure~\ref{enumerate}.
The constraint that $\rank W_{k \ssq i} \leq
\rank W_{k-1 \ssq i} + \rank W_{k \ssq i-1} - \rank W_{k-1 \ssq i-1}$ comes from
the requirement that the interval count $\omega_{k-1,i}$ is
nonnegative.

The advantage of assigning ranks in this order is that,
after we assign some of the ranks in the ranges indicated above,
there is always at least one assignment of the remaining, unassigned ranks
that satisfies the constraints and yields a valid rank list.
We omit a proof.\footnote{
One way to prove it is to show that
if we complete an incomplete rank list by always choosing
the smallest rank in each range
(i.e., $\rank W_{k \ssq 0} \leftarrow \rank W$ and
$\rank W_{k \ssq i} \leftarrow \rank W_{k \ssq i-1}$ for $i \geq 1$),
these choices satisfy the upper bound constraints and yield
a valid rank list.
}

Figure~\ref{enumeratealg} lists pseudocode for
a procedure, {\sc EnumerateVertices}, that enumerates
the valid rank lists (and the strata in the rank stratification)
for a fiber $\mu^{-1}(W)$.
For each valid rank list $\underline{r}$,
the algorithm creates a vertex in the stratum dag which represents
both $\underline{r}$ and the stratum $S_{\underline{r}}$.
{\sc EnumerateVertices} creates no edges;
see Section~\ref{dagedges} for the procedure that creates them.

\begin{figure}
\centerline{\mbox{
\begin{minipage}{6in}
\begin{tabbing}
MMM \= MM \= MM \= MM \= MM \= MM \= MM \= \kill
{\sc EnumerateVertices}$(L, d[], R)$  \\
// Enumerates all the valid rank lists that include specified values of
   $d_0, d_1, \ldots, d_L$ and $\rank W = R$.  \\
// Creates one dag vertex for each valid rank list.
   $L$ is the number of matrices (i.e., layers of edges).  \\
1  \> $\underline{r} \leftarrow$ rank list in which every rank $r_{k \ssq i}$ is
      $R$  \hspace{.2in}
      // in particular, $r_{L \ssq 0} \leftarrow R = \rank W$  \\ 
2  \> {\bf for} $j \leftarrow 0$ {\bf to} $L$  \\
3  \> \> $r_{j \ssq j} \leftarrow d_j$  \\
4  \> {\sc EnumerateVerticesRecurse}$(L, d[], \underline{r}, 1, 0)$
      \hspace{.2in}
      // start the recursion with $r_{1 \ssq 0}$, representing $\rank W_1$
\\
\\
{\sc EnumerateVerticesRecurse}$(L, d[], \underline{r}, k, i)$  \\
// Iterates through all the valid values of $r_{k \ssq i}$, then recursively does
   the same for the remaining ranks.  \\
5  \> {\bf if} $i = 0$  \\
   \> \> // There is at least one valid rank list for each $\rank W_{k \ssq 0}
            \in [\rank W, \min \{ \rank W_{k-1 \ssq 0}, d_k \}]$.  \\
6  \> \> low $\leftarrow r_{L \ssq 0}$  \hspace{.2in}
         // note:  $r_{L \ssq 0} = \rank W$  \\
7  \> \> high $\leftarrow \min \{ r_{k-1 \ssq 0}, d_k \}$  \\
8  \> {\bf else}  \hspace{.2in}  // $i \geq 1$  \\
   \> \> // \ldots for each $\rank W_{k \ssq i} \in [\rank W_{k \ssq i-1},
             \min \{ \rank W_{k-1 \ssq i} + \rank W_{k \ssq i-1} -
             \rank W_{k-1 \ssq i-1}, d_k \}]$.  \\
9  \> \> low $\leftarrow r_{k \ssq i-1}$  \\
10 \> \> high $\leftarrow \min \{ r_{k-1 \ssq i} + r_{k \ssq i-1} -
         r_{k-1 \ssq i-1}, d_k \}$  \\
11 \> {\bf if} $k < L - 1$ {\bf or} ($k < L$ {\bf and} $i > 0$)  \\
12 \> \> for $r_{k \ssq i} \leftarrow$ low {\bf to} high  \\
13 \> \> \> {\sc EnumerateVerticesRecurse}$(L, d[], \underline{r},
            k + 1, i)$  \hspace{.2in}  // increment $k$  \\
14 \> {\bf else if} $i < L - 1$  \\
15 \> \> for $r_{k \ssq i} \leftarrow$ low {\bf to} high  \\
16 \> \> \> {\sc EnumerateVerticesRecurse}$(L, d[], \underline{r},
            i + 2, i + 1)$  \hspace{.2in}  // increment $i$; restart $k$  \\
17 \> {\bf else}  \hspace{.2in}
      // $k = L$ and $i = L - 1$; base case of recursion  \\
18 \> \> for $r_{k \ssq i} \leftarrow$ low {\bf to} high  \\
\> \> \> // All the ranks in $\underline{r}$ are filled in now; create
            a new vertex to represent the stratum $S_{\underline{r}}$.  \\
19 \> \> \> $v \leftarrow$ {\sc CreateDAGVertex}$()$  \hspace{.2in}
            // creates new vertex in dag data structure  \\
20 \> \> \> $v$.ranklist $\leftarrow \underline{r}$  \hspace{.2in}
            // important:  copy the rank list $\underline{r}$;
               don't just store a reference to $\underline{r}$  \\
   \> \> \> // $v$.dim is the dimension of the stratum~$S_{\underline{r}}$.  \\
21 \> \> \> $v$.dim $\displaystyle \leftarrow \sum_{j=1}^L d_j d_{j-1} -
            r_{L \ssq 0} \cdot (d_L + d_0 - r_{L \ssq 0}) - \hspace*{-.18in}
            \sum_{L \geq k+1 \geq i > 0} \hspace*{-.1in}
            (r_{k+1 \ssq i} - r_{k+1 \ssq i-1})
            (r_{k \ssq i-1} - r_{k+1 \ssq i-1})$  \\
   \> \> \> // $v$.dof is the number of degrees of freedom of motion on the
               fiber from a point on $S_{\underline{r}}$.  \\
22 \> \> \> $v$.dof $\displaystyle \leftarrow \sum_{j=1}^L d_j d_{j-1} -
            \sum_{j=1}^L r_{L \ssq j} \cdot r_{j-1 \ssq 0} +
            \sum_{j=1}^{L-1} r_{L \ssq j} \cdot r_{j \ssq 0}$  \\
   \> \> \> // $v$.rdof is the number of rank-increasing degrees of freedom
            (moving off $S_{\underline{r}}$ on fiber).  \\
23 \> \> \> $v$.rdof $\leftarrow v$.dof $- v$.dim  \\
   \> \> \> // Maintain a map from rank lists to dag vertices so
               the record $v$ can be found later.  \\
24 \> \> \> {\sc AddToMap}$(\underline{r}, v)$
\end{tabbing}
\end{minipage}
}}

\caption{\label{enumeratealg}
Algorithm to enumerate all the strata in the rank stratification of
a fiber $\mu^{-1}(W)$---equivalently, to enumerate all the valid rank lists
that include specified values of $d_0, d_1, \ldots, d_L$, and~$\rank W$.
For each such rank list $\underline{r}$, the algorithm creates
a vertex in the stratum dag to represent the stratum~$S_{\underline{r}}$.
A dag vertex $v$ stores four fields:
the numbers $v$.dim, $v$.dof, and $v$.rdof introduced in
Section~\ref{foretaste},
and $v$.ranklist stores a rank list $\underline{r}$.
}
\end{figure}

Our algorithm maintains a map from rank lists to (references to) graph vertices.
This map is not part of the directed graph data structure,
but it is used to find vertices in the graph data structure
knowing only their rank lists, and it will be needed in Section~\ref{dagedges}
to help compute the edges of the dag.
The procedure {\sc AddToMap} (called in Line~25) adds
a rank list (the key) and a vertex (the target) to this map.
There are several ways the map could be implemented.
The simplest method is to store the vertices in a multidimensional array
indexed by the ranks in the rank list.
If we omit the fixed ranks (the $d_j$'s and $\rank W$),
this array has $(L^2 + L) / 2 - 1$ dimensions, and
the index associated with $r_{k \ssq i}$ can range
from $\rank W$ to $\min \{ d_i, d_{i+1}, \ldots, d_k \}$.
However, if $L$ is large, most of the entries in this array are never used,
as they correspond to invalid rank lists, and
the array might be too large to store.
In that case, it makes more sense to use a hash table to implement the map.

\subsection{Computing the Edges of the Stratum Dag}
\label{dagedges}

In Section~\ref{dagdetails}, we introduced the convention that
the stratum dag has a directed edge $(S_{\underline{r}}, S_{\underline{z}})$ if
there exists a rank-$1$ abstract barcode move that transforms
the rank list $\underline{r}$ to the rank list $\underline{z}$.
This idea is partly justified by Theorem~\ref{rankclosureequiv},
which establishes that, assuming $S_{\underline{r}}$ and $S_{\underline{z}}$ are
both nonempty strata in the rank stratification of $\mu^{-1}(W)$,
the following three statements are equivalent:
$S_{\underline{r}} \subseteq \bar{S}_{\underline{z}}$;
$\underline{r} \leq \underline{z}$; and
there exists a sequence of rank-$1$ abstract barcode moves
that proceed from $\underline{r}$ to~$\underline{z}$, with
all the intermediate rank lists being valid.
The last statement is equivalent to there being a directed path
from $S_{\underline{r}}$ to~$S_{\underline{z}}$ in the stratum dag.

Figure~\ref{edgealg} lists pseudocode for a procedure, {\sc EnumerateEdges},
that creates the directed edges of the stratum dag for a fiber $\mu^{-1}(W)$,
connecting the vertices already created by
the procedure {\sc EnumerateVertices}.
We consider one stratum $S_{\underline{r}}$ at a time, and enumerate
all the rank-$1$ abstract barcode moves from $\underline{r}$;
these correspond to directed edges out of $S_{\underline{r}}$ in the stratum dag.
From $\underline{r}$, there is an abstract move
with index tuple $(l, k, i, h)$ if $L \geq l \geq k + 1 \geq i > h \geq 0$,
$\omega_{li} > 0$, and $\omega_{kh} > 0$ (see Section~\ref{abstractmoves}).
Recall from Lemma~\ref{rankincrease}
(and Section~\ref{barcmoves} and Figure~\ref{matchange}) that
a rank-$1$ barcode move with these indices increases
each rank~$r_{t \ssq s}$ by~$1$
for all $t \in [k + 1, l]$ and $s \in [h, i - 1]$.
This tells us how to compute the rank list $\underline{z}$ reached by
an abstract move, whereupon we can create
a directed edge $(S_{\underline{r}}, S_{\underline{z}})$ in the dag.

\begin{figure}
\centerline{\mbox{
\begin{minipage}{6in}
\begin{tabbing}
MMM \= MM \= MM \= MM \= MM \= MM \= MM \= MM \= MM \= \kill
{\sc EnumerateEdges}$(L)$  \\
// Creates the edges of the stratum dag, connecting the vertices created by
   {\sc EnumerateVertices}.  \\
// $L$ is the number of matrices (i.e., layers of neural network edges).  \\
1  \> {\bf for} each vertex $v$ in the stratum dag  \\
2  \> \> $\underline{r} = v$.ranklist  \\
   \> \> // Compute the interval counts for $\underline{r}$.  \\
3  \> \> {\bf for} $t \leftarrow 0$ {\bf to} $L$  \\
4  \> \> \> {\bf for} $s \leftarrow 0$ {\bf to} $t$  \\
   \> \> \> \> // For the next line, use the conventions
                  $r_{L+1 \ssq s} = 0$ and $r_{t \ssq -1} = 0$.  \\
5  \> \> \> \> $\omega_{ts} \leftarrow r_{t \ssq s} - r_{t \ssq s-1} - r_{t+1 \ssq s}
               + r_{t+1 \ssq s-1}$  \\
   \> \> // Identify all possible rank-$1$ abstract discrete moves from
            $\underline{r}$.  \\
6  \> \> {\bf for} $l \leftarrow 1$ {\bf to} $L$  \\
7  \> \> \> {\bf for} $k \leftarrow 0$ {\bf to} $l - 1$  \\
8  \> \> \> \> {\bf for} $i \leftarrow 1$ {\bf to} $k + 1$  \\
9  \> \> \> \> \> {\bf for} $h \leftarrow 0$ {\bf to} $i - 1$  \\
10 \> \> \> \> \> \> {\bf if} $\omega_{li} > 0$ {\bf and} $\omega_{kh} > 0$  \\
   \> \> \> \> \> \> \> // There is a move connecting or swapping intervals
                           $[i, l]$ and $[h, k]$.  \\
11 \> \> \> \> \> \> \> $\underline{z} \leftarrow \underline{r}$  \hspace{.2in}
                        // copy the rank list $\underline{r}$;
                           don't just store a reference to $\underline{r}$  \\
12 \> \> \> \> \> \> \> {\bf for} $t \leftarrow k + 1$ {\bf to} $l$  \\
13 \> \> \> \> \> \> \> \> {\bf for} $s \leftarrow h$ {\bf to} $i - 1$  \\
14 \> \> \> \> \> \> \> \> \> $z_{t \ssq s} \leftarrow r_{t \ssq s} + 1$  \\
   \> \> \> \> \> \> \> // Add directed edge representing
                           $(S_{\underline{r}}, S_{\underline{z}})$
                           to dag data structure.  \\
15 \> \> \> \> \> \> \> $w \leftarrow$
                        {\sc MapRankListToVertex}$(\underline{z})$  \\
16 \> \> \> \> \> \> \> $e \leftarrow$ {\sc CreateDAGEdge}$(v, w)$
                        \hspace{.2in}
                        // sets $e$.origin to $v$ and $e$.dest to $w$  \\
17 \> \> \> \> \> \> \> $e$.label $\leftarrow (l, k, i, h)$  \hspace{.2in}
                        // the index tuple of the abstract move
\end{tabbing}
\end{minipage}
}}

\caption{\label{edgealg}
Algorithm to create the edges of the stratum dag.
The procedure {\sc MapRankListToVertex} invoked in Line~15 performs
a lookup in the same map maintained by
the procedure {\sc AddToMap} in Line~24 of {\sc EnumerateVerticesRecurse}.
}
\end{figure}

Line~1 of {\sc EnumerateEdges} presumes that
the directed graph data structure that represents the stratum dag
permits us to loop through the complete set of graph vertices
(which were originally created by {\sc Enumerate\-Vertices}).
Line~15 uses the map maintained by Line~24 of {\sc Enumerate\-Vertices} to
find the data structure representing a dag vertex,
knowing only its rank list $\underline{z}$.
Lines~16 and~17 add an edge to the graph data structure,
labeled with the index tuple $(l, k, i, h)$.

\subsection{An Alternative Method of Enumerating the Valid Rank Lists}
\label{enumeratesec2}

We briefly mention an alternative to the algorithm {\sc Enumerate\-Vertices}.
Our alternative builds on the structure of {\sc EnumerateEdges}.
Recall that every fiber's rank stratification has one minimal element:
the stratum for which every rank (except the $d_j$'s) is equal to $\rank W$.
Every other stratum in the rank stratification can be reached from
the minimal stratum by a sequence of rank-$1$ discrete moves.
Hence we can efficiently find all the vertices with a procedure that is
essentially depth-first search, except that the graph is
not explicitly stored as a data structure
(until the search is complete; we create the data structure as we go).

Therefore,
we can modify {\sc EnumerateEdges} to create all the dag vertices as well.
Line~15 is modified to check whether a dag vertex exists to represent
the rank list $\underline{z}$; if no such vertex exists,
one is created and added to a queue of vertices whose outbound edges have
not yet been created.
Line~1 is modified to iterate through that queue.
The algorithm begins by creating one dag vertex to represent
the minimal valid rank list, then creating a queue that contains that vertex.
As the algorithm runs, newly created vertices are added to the queue, and
the modified loop in Line~1 repeatedly takes a vertex off the queue and
determines its outbound edges.

\clearpage

\section{Linear Neural Networks with Additive Terms}
\label{additive}

Here we extend our study to the fibers of linear neural networks that have
additive terms, sometimes called {\em bias terms}.
We will see that once we understand a fiber of the matrix multiplication map,
it takes little extra effort to understand
a fiber of a fully-connected linear neural network with bias terms.

Consider a linear neural network whose weights (parameters) include bias terms
$g_1 \in \R^{d_1}, g_2 \in \R^{d_2}, \ldots, \allowbreak g_L \in \R^{d_L}$
as well as the weight matrices $W_1 \in \R^{d_1 \times d_0},
W_2 \in \R^{d_2 \times d_1}, \ldots, W_L \in \R^{d_L \times d_{L-1}}$.
The network takes an input vector $x \in \R^{d_0}$ and returns an output vector
\begin{eqnarray*}
y & = &
g_L + W_L(g_{L-1} + W_{L-1}(g_{L-2} + W_{L-2}(g_{L-3} + \ldots (g_1 + W_1 x))))  \\
& = &
g_L + W_L g_{L-1} + W_{L \ssq L-2} g_{L-2} + W_{L \ssq L-3} g_{L-3} + \ldots +
W_{L \ssq 1} g_1 + W_{L \ssq 0} x
\\
& = & W_{L \ssq 0} x + g,
\hspace{.2in}  \mbox{where~}  \hspace{.2in}
g = \sum_{j=1}^L W_{L \ssq j} g_j.
\end{eqnarray*}

We write weight vectors for this network in the form
\[
\vartheta = (W_L, W_{L-1}, \ldots, W_2, W_1, g_L, g_{L-1}, \ldots, g_1)
\in \R^{d_{\vartheta}}
\]
where $d_{\vartheta} = \sum_{j=1}^L d_j (d_{j-1} + 1)$.
As an analog to the matrix multiplication map $\mu$, we define
a {\em linear network map} whose range is an ordered pair $(W, g)$,
which represents a network that computes the output $y = Wx + g$ for every $x$.
The linear network map is
\[
\check{\mu}(\vartheta) =
\left( W_{L \ssq 0}, \sum_{j=1}^L W_{L \ssq j} g_j \right).
\]

Given a matrix $W \in \R^{d_L \times d_0}$ and a vector $g \in \R^{d_L}$,
the fiber of $(W, g)$ in the linear network map is
\[
\check{\mu}^{-1}(W, g) =
\left\{ \left( \theta, g_1, g_2, \ldots, g_{L-1},
        g - \sum_{j=1}^{L-1} W_{L \ssq j} g_j \right) :
        \theta \in \mu^{-1}(W), g_1 \in \R^{d_1}, g_2 \in \R^{d_2}, \ldots,
        g_{L-1} \in \R^{d_{L-1}} \right\},
\]
where each $W_{L \ssq j}$ in the summation is a function of $\theta$.
The set on the right-hand side is the image of a bijective function that maps
$\mu^{-1}(W) \times \R^{d_1} \times \R^{d_2} \times \cdots \times \R^{d_{L-1}}$
to weight space.
The bijection is continuous and has a continuous inverse
(just drop $g_L$ from the weight vector), so it is a homeomorphism.
Thus the topology of the fiber $\check{\mu}^{-1}(W, g)$ is
\[
\mu^{-1}(W) \times \R^{d_1 + d_2 + \ldots + d_{L-1}},
\]
if we treat $\mu^{-1}(W)$ as a topological space.

A natural next question is whether we can better understand
the fiber $\check{\mu}^{-1}(W, g)$ by finding a stratification of it.
Indeed, we can turn a rank stratification $\mathbb{S}$ of $\mu^{-1}(W)$ into
a rank stratification of $\check{\mu}^{-1}(W, g)$ by extruding each stratum in
$\mathbb{S}$.
That is, for each stratum $S^W_{\underline{r}} \in \mathbb{S}$,
the rank stratification of $\check{\mu}^{-1}(W, g)$ contains a stratum
\[
S^{(W,g)}_{\underline{r}} =
\left\{ \left( \theta, g_1, g_2, \ldots, g_{L-1},
        g - \sum_{j=1}^{L-1} W_{L \ssq j} g_j \right) :
        \theta \in S^W_{\underline{r}}, g_1 \in \R^{d_1}, g_2 \in \R^{d_2}, \ldots,
        g_{L-1} \in \R^{d_{L-1}} \right\}
\]
with topology
$S^W_{\underline{r}} \times \R^{d_1 + d_2 + \ldots + d_{L-1}}$.

\clearpage

\section{Application to Critical Points}
\label{critical}

We mentioned in the Introduction that this paper was inspired by
a paper of Trager, Kohn, and Bruna~\cite{trager20}.
Here we apply our understanding of the fiber $\mu^{-1}(W)$ to
the topic of Trager et al.:
the critical points in the cost function used to train the neural network.

\subsection{Cost Functions and their Differential Maps}
\label{cost}

Consider a neural network that takes an input vector $x \in \R^{d_0}$ and
produces an output vector $\hat{y}(x) \in \R^{d_L}$.
The goal of training is to find a function $\hat{y}$
(representable by the network) that locally minimizes some
real-valued functional $\ell(\hat{y})$, which we call a {\em cost function}.
For linear neural networks, we can write $\hat{y}(x) = Wx$ and
write the cost function as $\ell(W)$; then
we find some $W$ that minimizes $\ell(W)$.
We assume $\ell : \R^{d_L \times d_0} \rightarrow \R$ is continuous and smooth,
and gradient descent methods are appropriate.

In neural network training, we find a locally optimal weight vector $\theta$,
rather than directly finding $W$.
Thus we minimize another cost function,\footnote{
Note:  Trager, Kohn, and Bruna call $\ell$ and $\mathcal{L}$
{\em loss functions}; but we prefer to use
{\em loss function} for the loss associated with a single training point, and
{\em cost function} for the objective function that the neural network
optimizes, which aggregates the loss functions over all the training points.
}
\[
\mathcal{L}(\theta) = \ell(\mu(\theta)).
\]
Here we assume $\mu$ is the matrix multiplication map
(giving us a linear neural network), so $\mathcal{L}(\theta)$ is
continuous and smooth.
By far the most commonly used optimization algorithms for neural networks are
stochastic gradient descent and batch gradient descent.
The composition of $\ell$ and $\mu$ introduces some difficulties that are
absent when we perform gradient descent on~$\ell(W)$ directly.
Among these difficulties are {\em spurious critical points}, as explored by
Trager, Kohn, and Bruna; we define these in Sections~\ref{spuriousfilling}
and~\ref{spuriousnonfilling}.

We are interested in the critical points of $\mathcal{L}$---the
points where $\nabla_\theta \mathcal{L} = {\bf 0}$---for two reasons.
First, obviously, we seek a local minimum of $\mathcal{L}$.
Second, gradient descent can become stuck or slow down at or near
saddle points of $\mathcal{L}$.
Here we try to understand the second phenomenon.

The most common cost function in neural network training is
the {sum of squared errors},
\[
\ell(W) = \sum_{i=1}^n \| \hat{y}(x_i) - y_i \|^2,
\]
where we are given $n$ {\em training points}
$x_0, x_1, \ldots, x_n \in \R^{d_0}$ with
corresponding {\em labels} $y_0, y_1, \ldots, y_n \in \R^{d_L}$.
For linear neural networks, we can rewrite this cost function as
\begin{equation}
\ell(W) = \| WX - Y \|_F^2,
\label{leastsquares}
\end{equation}
where $X$ is
a $d_0 \times n$ {\em design matrix} whose columns are the training points,
$Y$ is a $d_L \times n$ matrix whose $i$th column is
the vector of training labels for the $i$th training point
($i$th column of $X$), and
\[
\| M \|_F = \sqrt{\sum_{i=1}^{d_L} \sum_{j=1}^n M_{ij}^2}
\]
is the Frobenius norm.
This is a classic least-squares linear regression problem, easily solved by
many different numerical methods.
When we substitute a linear neural network $W = \mu(\theta)$ into $\ell(W)$ and
optimize the weights in $\theta$ instead of directly optimizing
the components of $W$, we are handicapping ourselves; but
our goal is to understand the problems this substitution introduces in
the simple context of a linear neural network, in hopes of gaining insights
that apply to practical neural networks that use nonlinear activation functions
such as rectified linear units (ReLUs).

Recall the discussion of differential maps such as $\dmu(\theta)$ from
Section~\ref{diffmap}.
The cost function $\ell$ has a differential map
$\dell(W) : \R^{d_L \times d_0} \rightarrow \R$.
Its most intuitive interpretation is that,
for an infinitesimal perturbation $\Delta W$ of~$W$, $\dell(W)(\Delta W)$ is
the corresponding infinitesimal perturbation of $\ell(W)$.
Another interpretation is that the differential map produces
the directional derivative $\ell'_{\Delta W}(W) = \dell(W)(\Delta W)$.
But note that the differential map is a linear map over $\Delta W$, whereas
the directional derivative is a map over $W$, not necessarily linear
(though for the quadratic cost function~(\ref{leastsquares}),
it is linear in $W$).

If we were to minimize $\ell(W)$ by batch gradient descent,
we would use the gradient of $\ell$ with respect to~$W$, which we write
\[
G = \nabla_W \ell(W),
\]
the $d_L \times d_0$ matrix with $G_{ij} = \partial \ell / \partial W_{ij}$.
To avoid writing ``$G(W)$,''
we find it useful to fix $W$ and pretend $G$ is a constant, but
the reader should remember that $G$ is a function of $W$.
This gradient is closely related to our differential map $\dell(W)$, because
by definition the directional derivative of $\ell$ is
\[
\ell'_{\Delta W}(W) = \langle \nabla_W \ell(W), \Delta W \rangle_F
\]
where $\langle S, T \rangle_F = \trace (S T^\top)$
is the Frobenius inner product (recall Section~\ref{notation}).
Hence
\[
\dell(W)(\Delta W) = \ell'_{\Delta W}(W) = \langle G, \Delta W \rangle_F.
\]

For example, for the sum-of-squared-errors cost function~(\ref{leastsquares}),
$G = 2 (WX - Y) X^\top$ and
\[
\dell(W)(\Delta W) = 2 \, \langle (WX - Y) X^\top, \Delta W \rangle_F.
\]

The composite cost function $\mathcal{L} = \ell \circ \mu$ has
a differential map $\dL(\theta) : \R^{d_\theta} \rightarrow \R$.
Consider a fixed $\theta \in \R^{d_\theta}$, $W = \mu(\theta)$, and
$G = \left. \nabla_W \ell \right|_{W = \mu(\theta)}$.
By the chain rule, identity~(\ref{dmu}), and the cyclic property of the trace,
\begin{eqnarray}
\dL(\theta) & = &
\di(\ell \circ \mu)(\theta) = \dell(\mu(\theta)) \circ \dmu(\theta)
\hspace{.2in}  \mbox{~and~}  \nonumber  \\
\dL(\theta)(\Delta \theta) & = &
\dell(\mu(\theta))(\dmu(\theta)(\Delta \theta))  \nonumber  \\
& = & \left\langle G, \dmu(\theta)(\Delta \theta) \right\rangle_F  \nonumber  \\
& = & \left\langle G, \sum_{j=1}^L W_{L \sim j} \Delta W_j W_{j-1 \sim 0}
\right\rangle_F  \nonumber  \\
& = & \sum_{j=1}^L
\trace \left( G W_{j-1 \sim 0}^\top \Delta W_j^\top W_{L \sim j}^\top \right)
\nonumber  \\
& = & \sum_{j=1}^L
\trace \left( W_{L \sim j}^\top G W_{j-1 \sim 0}^\top \Delta W_j^\top \right)
\nonumber  \\
& = & \sum_{j=1}^L
\left\langle W_{L \sim j}^\top G W_{j-1 \sim 0}^\top, \Delta W_j \right\rangle_F.
\label{dL}
\end{eqnarray}

Recall again that a directional derivative is defined in terms of a gradient;
in this case, $\mathcal{L}'_{\Delta \theta}(\theta) =
\nabla_\theta \mathcal{L} \cdot \Delta \theta$.
Here $\nabla_\theta \mathcal{L}$ and
$\Delta \theta = (\Delta W_L, \ldots, \Delta W_1)$ are vectors in $\R^{d_\theta}$.
Hence
\[
\nabla_\theta \mathcal{L} \cdot (\Delta W_L, \ldots, \Delta W_1) =
\mathcal{L}'_{\Delta \theta}(\theta) =
\dL(\theta)(\Delta \theta) =
\sum_{j=1}^L
\left\langle W_{L \sim j}^\top G W_{j-1 \sim 0}^\top, \Delta W_j \right\rangle_F
\]
and we infer that
\[
\nabla_\theta \mathcal{L} =
(G W_{L-1 \sim 0}^\top, \ldots, W_{L \sim j}^\top G W_{j-1 \sim 0}^\top, \ldots,
W_{L \sim 1}^\top G).
\]
If we minimize $\mathcal{L}(\theta)$ by batch gradient descent,
a step of gradient descent from a weight vector $\bar{\theta}$ moves in
the direction $\nabla_\theta \mathcal{L}(\bar{\theta})$.
(Remember that $G$ is a function of $\mu(\bar{\theta})$, thus
indirectly a function of $\bar{\theta}$).

In the following sections we will be interested in
critical points of $\mathcal{L}$, which are points $\bar{\theta}$ where
$\nabla_\theta \mathcal{L}(\bar{\theta}) = {\bf 0}$, or equivalently,
$\dL(\bar{\theta})(\Delta \theta) = 0$ for all $\Delta \theta \in \R^{d_\theta}$.

But first, we point out an interesting observation about the gradient
$\nabla_\theta \mathcal{L}$ that is most relevant where it is not zero.
Recall that we have studied $\row \dmu(\theta)$ in
Sections~\ref{normal1} and~\ref{normal1basis}.
Regardless of how we choose the cost function $\ell$,
the direction of steepest ascent $\nabla_\theta \mathcal{L}(\theta)$
always lies in~$\row \dmu(\theta)$;
this gives us another reason for our interest in $\row \dmu(\theta)$.

\begin{lemma}
\label{descentrowdmu}
For any differentiable $\ell$, $\mu$, and $\mathcal{L} = \ell \circ \mu$ with
$\mathcal{L}(\theta) \in \R$,
$\nabla_\theta \mathcal{L}(\theta) \in \row \dmu(\theta)$.
\end{lemma}

\begin{proof}
Recall that $\dL(\theta) = \dell(\mu(\theta)) \circ \dmu(\theta)$, so
$\row \dL(\theta) \subseteq \row \dmu(\theta)$.
But $\dL(\theta)$ is just a linear map into~$\R$, so it has only one ``row,''
and $\row \dL(\theta)$ is just the set of all scalar multiples of
the gradient $\nabla_\theta \mathcal{L}(\theta)$.
(That is, $\row \dL(\theta)$ is a line in weight space encoding
the directions of steepest ascent and steepest descent,
or just $\{ {\bf 0} \}$ if
$\theta$ is a critical point of $\mathcal{L}$).
Hence $\nabla_\theta \mathcal{L}(\theta) \in \row \dmu(\theta)$.
\end{proof}

\subsection{Networks with and without Bottlenecks}

We say that a network has a bottleneck if
it has a layer of units that is
narrower than both the input and output layers.
We write the network's smallest layer size as
\[
d = \min_{j=0}^L d_j.
\]
Commonly, $d = \min \{ d_L, d_0 \}$, in which case
the linear neural network can express
any linear transformation $W \in \R^{d_L \times d_0}$, so
the network is said to be {\em filling}.
But if $d < \min \{ d_L, d_0 \}$, then the network can express
only linear transformations of rank $d$ or less, and
the network is said to be {\em non-filling}, or to have a {\em bottleneck}.

The significance of this distinction is that it is appropriate to use
a different definition of ``critical point of $\ell$'' for
non-filling networks.
Specifically, we restrict the domain of $\ell$ to be
the set of matrices of rank $d$ or less
(instead of $\R^{d_L \times d_0}$), and
look for critical points on this restricted domain.

Recall from Section~\ref{sec:detmanifolds} that
the set of all $p \times q$ matrices of rank $r$ or less is called
a determinantal variety and written $\DV_r^{p \times q}$.
The linear transformation computed by a linear neural network always lies on
$\DV_d^{d_L \times d_0}$.
That is, $\mu(\theta) \in \DV_d^{d_L \times d_0}$
for all $\theta \in \R^{d_\theta}$.
For a filling network, $\DV_d^{d_L \times d_0} = \R^{d_L \times d_0}$,
which is why we call it ``filling''; every $d_L \times d_0$ matrix is possible.

A linear transformation $W$ might be a local minimum of $\ell$ when
the search space is restricted to $\DV_d^{d_L \times d_0}$, and yet
not be a local minimum of $\ell$ over the domain $\R^{d_L \times d_0}$.
See Section~\ref{spuriousnonfilling} for details.
First, we discuss pure and spurious critical points for
the simpler case of a filling network.

\subsection{Pure and Spurious Critical Points for Networks without Bottlenecks}
\label{spuriousfilling}

A matrix $W$ is a {\em critical point} of $\ell$ if
$\dell(W) \equiv 0$, which means that
$\dell(W)(\Delta W) = 0$ for all $\Delta W \in \R^{d_L \times d_0}$.
A~weight vector $\theta$ is a {\em critical point} of $\mathcal{L}$ if
$\dL(\theta) \equiv 0$, which means that
$\dL(\theta)(\Delta \theta) = 0$ for all $\Delta \theta \in \R^{d_\theta}$.

If $W = \mu(\theta)$ is a critical point of $\ell$, then
$\theta$ is a critical point of $\mathcal{L}$.
But the reverse implication does not hold.
We give alternative characterizations of these two kinds of critical point.

\begin{lemma}
\label{critl}
A matrix $W$ is a critical point of $\ell$ if and only if $G = 0$, where
$G = \nabla_W \ell(W)$.
\end{lemma}

\begin{proof}
Recall that $\dell(W)(\Delta W) = \langle G, \Delta W \rangle_F$.
If $G = 0$, then $\dell(W)(\Delta W) = 0$ for all $\Delta W$, so
$W$ is a critical point of $\ell$.
If $G \neq 0$, then $\dell(W)(G) = \langle G, G \rangle_F^2 \neq 0$, so
it is not true that $\dell(W)(\Delta W) = 0$ for all $\Delta W$, and
$W$ is not a critical point of $\ell$.
\end{proof}

Note that $\Null \dmu^\top(\theta)$ is the orthogonal complement of
$\col \dmu(\theta)$; both are derived in Appendix~\ref{funddmu}.

\begin{lemma}
\label{crittheta}
A weight vector $\theta$ is a critical point of $\mathcal{L}$ if and only if
$G \in \Null \dmu^\top(\theta)$, where $G = \nabla_W \ell(W)|_{W = \mu(\theta)}$.
\end{lemma}

\begin{proof}
Recall that $\dL(\theta)(\Delta \theta) =
\left\langle G, \dmu(\theta)(\Delta \theta) \right\rangle_F$.
Observe\footnote{
If it's not clear why
$\left\langle G, \dmu(\theta)(\Delta \theta) \right\rangle_F =
\Delta \theta \cdot \dmu^\top(\theta)(G)$,
imagine expressing the $d_L \times d_0$ matrix $G$ as
a vector $g$ of length $d_L d_0$.
The linear transformation $\dmu(\theta)$ can be expressed as a
$d_L d_0 \times d_\theta$ matrix $M$ such that
$\dmu(\theta)(\Delta \theta) = M \Delta \theta$.
Then $\left\langle G, \dmu(\theta)(\Delta \theta) \right\rangle_F =
g^\top M \Delta \theta = \Delta \theta^\top M^\top g =
\Delta \theta \cdot \dmu^\top(\theta)(G)$.
}
that $\left\langle G, \dmu(\theta)(\Delta \theta) \right\rangle_F =
\Delta \theta \cdot \dmu^\top(\theta)(G)$.
If \mbox{$G \in \Null \dmu^\top(\theta)$}, then
$\dmu^\top(\theta)(G) = {\bf 0}$ and thus
$\dL(\theta)(\Delta \theta) = 0$ for all $\Delta \theta$ and
$\theta$ is a critical point of $\mathcal{L}$.
If~$G \not\in \Null \dmu^\top(\theta)$, then
$\dmu^\top(\theta)(G) \neq {\bf 0}$,
it is not true that $\dL(\theta)(\Delta \theta) = 0$ for all $\Delta \theta$
(as we could set $\Delta \theta = \dmu^\top(\theta)(G)$), and
$\theta$ is not a critical point of $\mathcal{L}$.
\end{proof}

\begin{lemma}
\label{crittheta2}
A weight vector $\theta$ is a critical point of $\mathcal{L}$
if and only if $W_{L \sim j}^\top G W_{j-1 \sim 0}^\top = 0$ for all $j \in [1, L]$,
where $G = \nabla_W \ell(W)|_{W = \mu(\theta)}$.
\end{lemma}

\begin{proof}
Recall identity~(\ref{dL}), $\dL(\theta)(\Delta \theta) = \sum_{j=1}^L
\left\langle W_{L \sim j}^\top G W_{j-1 \sim 0}^\top, \Delta W_j \right\rangle_F$.
If $W_{L \sim j}^\top G W_{j-1 \sim 0}^\top = 0$ for all $j \in [1, L]$, then
$\dL(\theta)(\Delta \theta) = 0$ for all $\Delta \theta \in \R^{d_\theta}$, so
$\theta$ is a critical point of $\mathcal{L}$.
By contrast, if $W_{L \sim j}^\top G W_{j-1 \sim 0}^\top \neq 0$ for some $j$,
we can choose $\Delta \theta$ so that
$\Delta W_j = W_{L \sim j}^\top G W_{j-1 \sim 0}^\top$ and
every other $\Delta W_i$ is zero, in which case
$\dL(\theta)(\Delta \theta) = \| W_{L \sim j}^\top G W_{j-1 \sim 0}^\top \|_F^2
\neq 0$, so $\theta$ is not a critical point of $\mathcal{L}$.
\end{proof}

A problem we face in neural network training is that
a weight vector $\theta$ can be a saddle point of $\mathcal{L}$ even if
$W = \mu(\theta)$ is not a critical point of $\ell$.
For example, suppose that $\theta = (W_2, W_1)$ where
\[
W_2 = \left[ \begin{array}{cc} 0 & 0 \\ 1 & 1 \end{array} \right],
\hspace{.2in}
W_1 =
\left[ \begin{array}{cc} 0 & 0 \\ 0 & 1 \end{array} \right],
\hspace{.2in}
W = W_2 W_1 = \left[ \begin{array}{cc} 0 & 0 \\ 0 & 1 \end{array} \right],
\hspace{.2in}
\mbox{and}
\]
\[
\dell(W)(\Delta W) = \langle G, \Delta W \rangle_F
\hspace{.2in}
\mbox{~where~}
\hspace{.2in}
G = \left[ \begin{array}{cc} 1 & 0 \\ 0 & 0 \end{array} \right].
\]
In this example, we won't explicitly specify $\ell$; it suffices to know $G$.
Here, $W$ is not a critical point of $\ell$ because $G \neq 0$
(thus $\dell(W)(G) = \langle G, G \rangle_F = 1 \neq 0$).

The differential maps for $\mu$ and $\mathcal{L}$ are
\[
\dmu(\theta)(\Delta \theta) =
\left[ \begin{array}{cc} 0 & 0 \\ 1 & 1 \end{array} \right] \Delta W_1 +
\Delta W_2 \left[ \begin{array}{cc} 0 & 0 \\ 0 & 1 \end{array} \right]
\hspace{.2in}
\mbox{and}
\hspace{.2in}
\dL(\theta)(\Delta \theta) =
\left\langle \left[ \begin{array}{cc} 1 & 0 \\ 0 & 0 \end{array} \right],
             \dmu(\theta)(\Delta \theta) \right\rangle_F = 0.
\]
The latter is zero because
the upper left component of $\dmu(\theta)(\Delta \theta)$ is necessarily zero.
Hence, $\theta$ is a critical point of $\mathcal{L}$ although $W$ is
not critical for $\ell$.

Why would $\mathcal{L}$ have a critical point at $\theta$ when
$\ell$ has no critical point at $W = \mu(\theta)$?
$W$ is not a critical point of~$\ell$ because $\ell$'s gradient $G$ is not zero;
we can decrease $\ell$ rapidly by reducing the upper left component of~$W$.
The underlying problem here is that to change the upper left component of $W$,
we must change both $W_2$ and~$W_1$.
Changing $W_2$ alone or changing $W_1$ alone has no effect on
the upper left component of $W$, so
the upper left component of $\dmu(\theta)(\Delta \theta)$ is zero, thus
$\dL(\theta)(\Delta \theta)$ is zero.
Equivalently, the first-order derivative $\nabla_\theta \mathcal{L}$ is zero.
The second-order derivatives are not zero;
$\theta$ is a saddle point of~$\mathcal{L}$ and
changing $W_2$ and $W_1$ together can decrease or increase $\mathcal{L}$.

We call $\theta$ a {\em spurious critical point} of $\mathcal{L}$.
We don't like spurious critical points, because gradient descent
on $\mathcal{L}$ slows down dramatically if we're very close to one.

Let us formally define pure and spurious critical points of $\mathcal{L}$ for
filling networks
(networks without bottlenecks, for which $d = \min \{ d_L, d_0 \}$).
\begin{itemize}
\item
A weight vector $\theta$ is a {\em pure critical point} of $\mathcal{L}$ if
$W = \mu(\theta)$ is a critical point of $\ell$---that is,
$\dell(W)(\Delta W) = 0$ for all $\Delta W \in \R^{d_L \times d_0}$.
Equivalently, $G = 0$.

\item
A weight vector $\theta$ is a {\em spurious critical point} of $\mathcal{L}$ if
$\theta$ is a critical point of $\mathcal{L}$ but $\theta$ is not pure.
Equivalently, $G \in \Null \dmu^\top(\theta)$ but $G \neq 0$.
\end{itemize}

One interesting thing about spurious critical points is that
it is always possible to move away from them without changing
the value of $W$.
In the example above, we can modify $W_2$ to give
\[
W_2 W_1 =
\left[ \begin{array}{cc} 0.01 & 0 \\ 1 & 1 \end{array} \right]
\left[ \begin{array}{cc} 0    & 0 \\ 0 & 1 \end{array} \right] =
\left[ \begin{array}{cc} 0    & 0 \\ 0 & 1 \end{array} \right] = W,
\]
whereupon $\theta$ is no longer a critical point but $W$ is unchanged.
Trager, Kohn, and Bruna~\cite[Proposition~9]{trager20} show that
from every spurious critical point,
there is an infinitesimal perturbation that yields a weight vector that is not
a critical point (for either $\mathcal{L}$ or $\ell$).

\subsection{Circumstances without Spurious Critical Points}
\label{nospuriousfilling}

Given a filling network (i.e., no bottleneck),
here we consider some circumstances in which a stratum or an entire fiber
is free of spurious critical points.
The following two lemmas are results of Trager, Kohn, and Bruna~\cite{trager20}
(combining parts of their Theorem~4 and Proposition~6), albeit
rephrased in the language of strata, rank lists, and barcodes.

\begin{lemma}
\label{nocrit1}
There is no spurious critical point of $\mathcal{L}$ on
any stratum $S_{\underline{r}}$ that satisfies, for some $j \in [1, L]$,
both $r_{L \ssq j} = d_L$ and $r_{j-1 \ssq 0} = d_0$.
(Equivalently, both $\omega_{L,j+1} = \omega_{L,j+2} = \cdots = 0$ and
$\omega_{j-2,0} = \omega_{j-3,0} = \cdots = 0$.)
\end{lemma}

\begin{proof}
Suppose for the sake of contradiction that
there is a spurious critical point $\theta$ and a $j \in [1, L]$ such that
$\rank W_{L \ssq j} = d_L$ and $\rank W_{j-1 \ssq 0} = d_0$.
As $G$ is a $d_L \times d_0$ matrix,
$\rank W_{L \ssq j}^\top G = \rank G$ and thus
$\rank W_{L \ssq j}^\top G W_{j-1 \sim 0}^\top = \rank G$.
But for $\theta$ to be a spurious critical point implies that
$G \neq 0$ and (by Lemma~\ref{crittheta2})
$W_{L \ssq j}^\top G W_{j-1 \sim 0}^\top = 0$, a contradiction.
\end{proof}

Recall that we define $d = \min_{j\in[0,L]} d_j$ and, by definition,
a filling network satisfies $d = \min \{ d_L, d_0 \}$.

\begin{lemma}
\label{rkwdfilling}
If the network is filling and $\rank W = d$, then
there is no spurious critical point of $\mathcal{L}$ on the fiber $\mu^{-1}(W)$.
\end{lemma}

\begin{proof}
Let $\theta$ be a point on $\mu^{-1}(W)$, and
let $\underline{r}$ be $\theta$'s rank list.
Every subsequence matrix $W_{k \ssq i}$ must have rank at least as great as $W$.
Either $d = d_L$, in which case $\theta$'s rank list has
$r_{L \ssq 1} = d_L$ and $r_{0 \ssq 0} = d_0$, so Lemma~\ref{nocrit1} applies; or
$d = d_0$, in which case $\theta$'s rank list has
$r_{L \ssq L} = d_L$ and $r_{L-1 \ssq 0} = d_0$, and Lemma~\ref{nocrit1} applies.
In either case, Lemma~\ref{nocrit1} implies that $\theta$ is not
a spurious critical point.
\end{proof}

\begin{figure}
\centerline{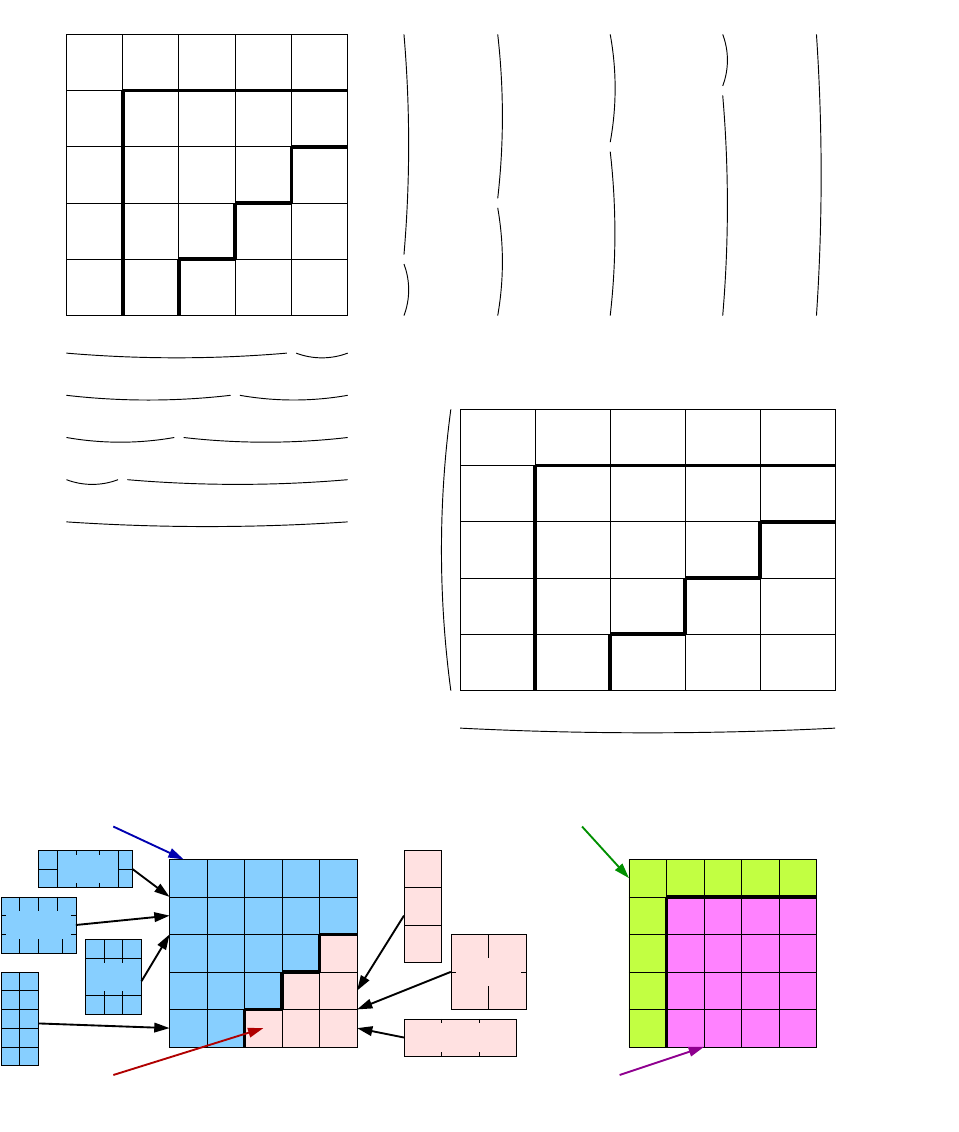}

\caption{\label{coldmufig}
A map showing the structure of $\col \dmu(\theta)$, $\Null \dmu^\top(\theta)$,
$T_W \DV_\rho$, and $N_W \DV_\rho$ for a weight vector~$\theta$,
where $W = \mu(\theta)$ and $\rho = \rank W$.
Here we use the standard prebases $\mathcal{B}_4$ and $\mathcal{A}_0$
(see Appendix~\ref{standardapp}) so that
the subspaces $a_{k00}$ are mutually orthogonal and span $\R^{d_0}$,
the subspaces $b_{44i}$ are mutually orthogonal and span~$\R^{d_4}$, and
the subspaces $e_{ki} = b_{44i} \otimes a_{k00}$ are mutually orthogonal and
span $\R^{d_4 \times d_0}$.
}
\end{figure}

To dive into a deeper understanding of spurious critical points and
their relationship to barcodes,
recall from Section~\ref{normal1basis} the subsequence subspaces
$e_{lktsih} = b_{lti} \otimes a_{ksh}$; we define the shorthand
\[
e_{ki} = e_{LkL0i0} = b_{LLi} \otimes a_{k00}.
\]
Recall also the subsequence prebasis
\[
\mathcal{E}_{L0} = \{ e_{ki} \neq \{ 0 \} : k,i \in [0, L] \},
\]
whose subspaces are linearly independent.
Figure~\ref{coldmufig} (top left) shows in graphical form
how the subsequence subspaces span $\R^{d_L \times d_0}$.

The following lemma shows that the subspace $\Null \dmu^\top(\theta)$
(which is part of the definition of critical point of $\mathcal{L}$)
can be written as a vector sum of subsequence subspaces.
Figure~\ref{coldmufig} (bottom left) gives a visual interpretation of
this lemma, showing which subsequence subspaces are included and
which are not.

\begin{lemma}
\label{nulldmute}
$\displaystyle  \Null \dmu^\top(\theta) = \sum_{k=0}^{L-2} \sum_{i=k+2}^L e_{ki}$.
\end{lemma}

\begin{proof}
Lemma~\ref{nulldmut} in Appendix~\ref{funddmu} shows that
\begin{eqnarray*}
\Null \dmu^\top(\theta)
& = & \sum_{j=1}^{L-1} \Null W_{L \ssq j}^\top \oplus \Null W_{j \ssq 0}
  =   \sum_{j=1}^{L-1} B_{L,L,j+1} \oplus A_{j-1,0,0}  \\
& = & \sum_{j=1}^{L-1} \sum_{k=0}^{j-1} \sum_{i=j+1}^L b_{LLi} \oplus a_{k00}
  =   \sum_{j=1}^{L-1} \sum_{k=0}^{j-1} \sum_{i=j+1}^L e_{ki}.
\end{eqnarray*}
It is straightforward to see that this triple summation is the vector sum of
every subsequence subspace $e_{ki}$ for which $i \geq k + 2$ (and no other), so
the result follows.
\end{proof}

As $\mathcal{E}_{L0}$ is a prebasis for $\R^{d_L \times d_0}$,
there is one unique way to write the gradient matrix $G$ as
a sum of matrices from the subsequence spaces,
\[
G = \sum_{k = 0}^L \sum_{i = 0}^L G_{ki}
\hspace{.2in}
\mbox{~where~}
\hspace{.2in}
G_{ki} \in e_{ki}.
\]

\begin{lemma}
\label{critG}
$\theta$ is a critical point of $\mathcal{L}$ if and only if
$G_{ki} = 0$ for all indices $k,i \in [0, L]$ satisfying $i \leq k + 1$.
\end{lemma}

\begin{proof}
By Lemma~\ref{crittheta}, $\theta$ is a critical point of $\mathcal{L}$
if and only if $G \in \Null \dmu^\top(\theta)$.
The result follows from Lemma~\ref{nulldmute}.
\end{proof}

\begin{cor}
\label{spuriousG}
$\theta$ is a spurious critical point of $\mathcal{L}$ if and only if
$G_{ki} = 0$ for all indices $k,i \in [0, L]$ satisfying $i \leq k + 1$, but
$G_{ki} \neq 0$ for some pair of indices $k,i \in [0, L]$ satisfying
$i > k + 1$.
\end{cor}

Let's consider an interpretation of Lemma~\ref{critG} and
Corollary~\ref{spuriousG}.
For a weight vector $\theta$, a nonzero $G_{ki}$ can be interpreted as
a prompt for the gradient descent algorithm to create or modify 
a signal propagating from the subspace $a_{k00}$ presented at the input layer to
the subspace $b_{LLi}$ presented at the output layer.
If there is a nonzero $G_{ki}$ such that $i \leq k + 1$, then
$\theta$ is not a critical point, and gradient descent can make progress
by suitably adjusting the value of $W_j$ for any $j \in [i, k + 1]$.
However, if
$G_{ki} = 0$ for all indices $k,i \in [0, L]$ satisfying $i \leq k + 1$, but
there is a nonzero $G_{ki}$ with $i > k + 1$, then
$d\mathcal{L}(\theta) \equiv 0$ and gradient descent is stuck at
a spurious critical point of $\mathcal{L}$.
The issue is that gradient descent cannot respond to the nonzero $G_{ki}$
because more than one matrix (layer of connections) must be
modified to create a signal that propagates from $a_{k00}$ to $b_{LLi}$.
(For example, one could update the matrices $W_{k+1}, W_{k+2}, \ldots, W_i$.)
One cannot see this by looking solely at the gradient of $\mathcal{L}$;
one must also look at higher-order derivatives.

\begin{figure}
\centerline{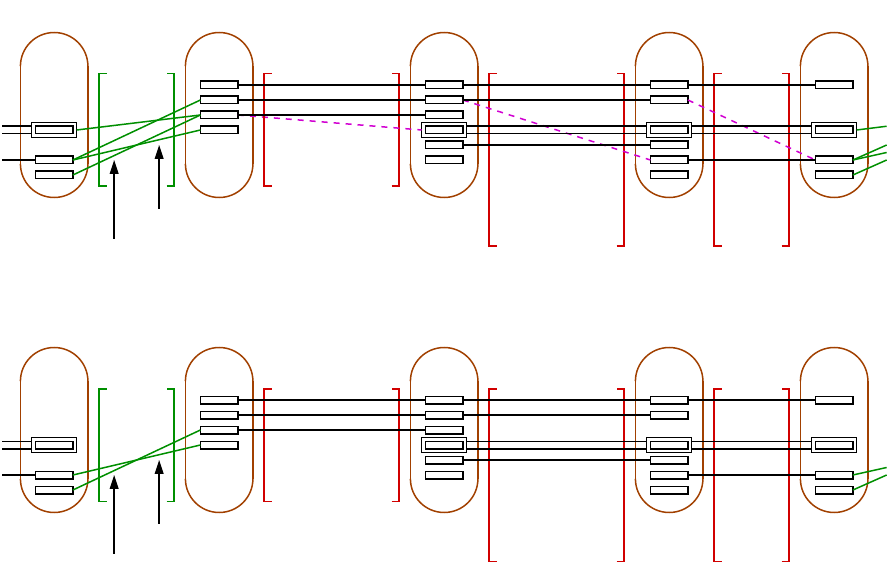}

\caption{\label{criticalloop}
To the left of the barcode, we draw the matrix $G^\top$ and
draw green links in it to indicate which components $G_{ki}$ are nonzero.
In the top example,
dashed purple edges show some of the ways by which gradient descent can
``complete a loop'' and make progress by modifying $W_3$, $W_2$, or $W_1$.
Hence this example indicates that $\theta$ is not a critical point.
In the bottom example, $G$ is nonzero but it is not possible to complete a loop
by modifying just one factor matrix.
Hence this example indicates that $\theta$ is a spurious critical point.
(Not depicted is the case of a pure critical point, which simply has $G = 0$.)
}
\end{figure}

Figure~\ref{criticalloop} shows
a visual interpretation of these observations that uses the barcode.
We need consider only the intervals (bars in the barcode) that touch
the input layer or the output layer.
Suppose some component $G_{ki}$ of the gradient of $\mathcal{L}$ is nonzero;
as $G \neq 0$, $\mu(\theta)$ is a critical point of $\ell$.
Each nonzero $G_{ki}$ is interpreted as connecting
an interval $[0, k]$ to an interval $[i, L]$.
If it is possible to complete a loop by modifying just one matrix $W_j$, then
$\theta$ is not a critical point.
But if that is not possible, then $\theta$ is a spurious critical point.

\subsection{Pure and Spurious Critical Points for Networks with Bottlenecks}
\label{spuriousnonfilling}

Recall that a non-filling network with bottleneck size
$d = \min_{j=0}^L d_j < \min \{ d_L, d_0 \}$ can compute only
rank-$d$ linear transformations of the input.
These transformations constitute
the determinantal variety $\DV_d^{d_L \times d_0}$---the set of all
$d_L \times d_0$ matrices of rank $d$ or less.
Hence we write the cost function on $W$ as
\[
\ell|_{\DV_d^{d_L \times d_0}} : \DV_d^{d_L \times d_0} \rightarrow \R.
\]
The notation $\ell|_{\DV_d^{d_L \times d_0}}$ means that $\ell$'s domain is
restricted to the set of matrices of rank $d$ or less.
For readability, we shorten this notation to
\[
\ell|_d : \DV_d \rightarrow \R.
\]

Although restricting the domain might seem like an inconsequential change
to $\ell$, it changes the notion of what is a critical point, because
a weight vector $\theta \in \DV_d$ could be a critical point of $\ell|_d$ yet
not a critical point of $\ell$.
By restricting the domain, we might have taken away all the directions
along which the directional derivative $\ell'_{\Delta \theta}(\theta)$ is nonzero.

What are the possible directions of motion on the determinantal variety $\DV_d$,
starting from a matrix $W$?
Recall that in Section~\ref{normal2},
we defined a function $\chi_d$ that maps $W$ to a vector that lists
the determinants of all the $(d + 1) \times (d + 1)$ minors of $W$.
Hence $\chi_d(W) = {\bf 0}$ if and only if $W$ has rank $d$ or less, and
$\DV_d = \{ M \in \R^{d_L \times d_0} : \chi_d(M) = {\bf 0} \}$.
As $\chi_d$~is continuous and smooth (being a polynomial),
every smooth path leaving $W$ that lies on $\DV_d$ is
tangent to $\Null \di\chi_d(W)$.
In other words, every vector $\Delta W$ tangent to
a smooth path on the determinantal variety at $W$ satisfies
$\di\chi_d(W)(\Delta W) = {\bf 0}$.
The nullspace is
\[
\Null \di\chi_d(W) = \left\{ \begin{array}{ll}
T_W \DV_d = \col W \otimes \R^{d_0} + \R^{d_L} \otimes \row W, &
\rank W = d,  \\
\R^{d_L} \otimes \R^{d_0}, & \rank W < d.
\end{array} \right.
\]

Its orthogonal complement is
\[
\row \di\chi_d(W) = \left\{ \begin{array}{ll}
N_W \DV_d = \Null W^\top \otimes \Null W, &
\rank W = d,  \\
\{ {\bf 0} \}, & \rank W < d.
\end{array} \right.
\]

When $\rank W = d$,
$W$ lies on the determinantal manifold $\DM_d^{d_L \times d_0}$
(the set of matrices of rank exactly $d$), which is a smooth manifold.
In other words, $W$ is not a singular point of $\DV_d$.
So at $W$ there is a well-defined tangent space $T_W \DV_d$, which indicates
all the possible instantaneous directions of motion from $W$.
As these are the directions in which
the instantaneous rate of change of $\chi_d(W)$ is zero,
$\Null \di\chi_d(W) = T_W \DV_d$.

When $\rank W < d$, $W$ is a singular point of the variety and
does not have a well-defined tangent space nor normal space.
In that case, $\Null \di\chi_d(W) = \R^{d_L} \otimes \R^{d_0}$.
This needs a little explanation.
First, consider the fact that if we change a single component of $W$,
yielding another matrix $W'$, the rank of $W'$ is
at most $1$ greater than the rank of $W$, so
$\rank W' \leq d$ and $W' \in \DV_d$.
It follows that if we draw the $d_L d_0$ lines through $W$ that are
each parallel to an axis of $\R^{d_L \times d_0}$,
those lines are all subsets of $\DV_d$.
All of the lines are therefore also subsets of $\Null \di\chi_d(W)$; and
as $\Null \di\chi_d(W)$ is a subspace, $\Null \di\chi_d(W) = \R^{d_L \times d_0}$.

At last we are ready to define critical points of~$\ell|_d$
(as compared with critical points of~$\ell$).
A matrix $W \in \DV_d$ is a {\em critical point} of~$\ell|_d$ if
$\dell(W)(\Delta W) = 0$ for all $\Delta W \in \Null \di\chi_d(W)$.
In other words, the directional derivative $\ell'_{\Delta W}(W)$ is zero
for every instantaneous direction of motion on the determinantal variety.
By contrast, a critical point of $\ell$ has
$\dell(W)(\Delta W) = 0$ for all $\Delta W \in \R^{d_L \times d_0}$.
Hence the two definitions differ only when $\rank W = d$.
(When $\rank W < d$, if $W$ is not a critical point of $\ell$, then
$W$ is not a critical point of $\ell|_d$ either, because
$G = \nabla_W \ell(W) \neq 0$ and thus
there exists a displacement $\Delta W$ with exactly one nonzero component
such that $\langle G, \Delta W \rangle_F < 0$,
indicating a direction on $\DV_d$ along which $\ell|_d$ is decreasing and
an opposite direction on $\DV_d$ along which $\ell|_d$ is increasing.)


Consider the following alternative characterization.

\begin{lemma}
\label{critld}
A matrix $W$ is a {\em critical point} of $\ell|_d$ if and only if
$G \in \row \di\chi_d(W)$, where $G = \nabla_W \ell(W)$.
\end{lemma}

\begin{proof}
Recall that $\dell(W)(\Delta W) = \langle G, \Delta W \rangle_F$.
By definition, $W$ is a critical point of~$\ell|_d$ if
$\dell(W)(\Delta W) = 0$ for all $\Delta W \in \Null \di\chi_d(W)$;
equivalently, $\langle G, \Delta W \rangle_F = 0$
for all $\Delta W \in \Null \di\chi_d(W)$;
equivalently, $G$ is orthogonal to the subspace $\Null \di\chi_d(W)$.
The result follows.
\end{proof}

We now revisit the definitions of
pure and spurious critical points of $\mathcal{L}$ for
non-filling networks (i.e., networks with bottlenecks).
Recall that for a weight vector $\theta$,
we define $W = \mu(\theta)$ and $G = \nabla_W \ell(W)$.
Recall also that $\mathcal{L}(\theta) = \ell(\mu(\theta))$.
We can equally well write $\mathcal{L}(\theta) = \ell|_d(\mu(\theta))$,
as $\mu$ yields a matrix in $\DV_d$.
\begin{itemize}
\item
A weight vector $\theta$ is a {\em pure critical point} of $\mathcal{L}$ if
$W = \mu(\theta)$ is a critical point of $\ell|_d$---that is,
$\dell(W)(\Delta W) = 0$ for all $\Delta W \in \Null \di\chi_d(W)$.
Equivalently, $G \in \row \di\chi_d(W)$.

\item
A weight vector $\theta$ is a {\em spurious critical point} of $\mathcal{L}$ if
$\theta$ is a critical point of $\mathcal{L}$ but $\theta$ is not pure.
Equivalently, $G \not\in \row \di\chi_d(W)$ but
$G \in \Null \dmu^\top(\theta)$.
\end{itemize}

The last sentence is clarified by the recognition that when $\rank W = d$,
there is no $G$ that satisfies both
\mbox{$G \in \Null \dmu^\top(\theta)$} and $G \not\in \row \di\chi_d(W)$.
Figure~\ref{coldmufig} shows the following lemma visually.

\begin{lemma}
\label{nonfillinginc}
Given a weight vector $\theta$, let $W = \mu(\theta)$ and $\rho = \rank W$.
Then $\Null \dmu^\top(\theta) \subseteq N_W \DV_\rho$.
\end{lemma}

\begin{proof}
Recall that $N_W \DV_\rho = \Null W^\top \otimes \Null W$.
Lemma~\ref{nulldmut} in Appendix~\ref{funddmu} shows that
$\Null \dmu^\top(\theta) =
\sum_{j=1}^{L-1} \Null W_{L \ssq j}^\top \oplus \Null W_{j \ssq 0}$.
Observe that for $j \in [1, L - 1]$,
$\Null W_{L \ssq j}^\top \subseteq \Null W^\top$ and
$\Null W_{j \ssq 0} \subseteq \Null W$.
Therefore,
$\Null W_{L \ssq j}^\top \oplus \Null W_{j \ssq 0} \subseteq
\Null W^\top \otimes \Null W$ for all $j \in [1, L - 1]$, thus
$\sum_{j=1}^{L-1} \Null W_{L \ssq j}^\top \oplus \Null W_{j \ssq 0} \subseteq
\Null W^\top \otimes \Null W$.
(Recall Figure~\ref{coldmufig}.)
The result follows.
\end{proof}

\begin{lemma}
\label{rkwdnonfilling}
If the network is non-filling and $\rank W = d$, then
there is no spurious critical point on the fiber $\mu^{-1}(W)$.
\end{lemma}

\begin{proof}
As $\rank W = d$, $\row \di\chi_d(W) = N_W \DV_d$.
Let $\theta$ be a point on $\mu^{-1}(W)$.
By Lemma~\ref{nonfillinginc},
$\Null \dmu^\top(\theta) \subseteq N_W \DV_d$.
Hence there is no $G \in \Null \dmu^\top(\theta)$ that is not also in
$\row \di\chi_d(W)$.
Hence no $\theta \in \mu^{-1}(W)$ can be a spurious critical point.
\end{proof}

Putting Lemmas~\ref{rkwdfilling} and~\ref{rkwdnonfilling} together,
we see that both filling and non-filling networks have
no spurious critical points when $\rank W = d$, albeit
for a different reason in each case.
Trager, Kohn, and Bruna~\cite{trager20} showed this
as part of their Theorem~4 and Proposition~6.

In the case where $\rank W < d$,
filling and non-filling networks both have the property that
$\theta$ is a spurious critical point if and only if
$G \in \Null \dmu^\top(\theta)$ but $G \neq 0$.
In this case, Lemma~\ref{critG} and Corollary~\ref{spuriousG} apply to
non-filling networks as well as filling networks.

\clearpage







\clearpage
\appendix

\section{Valid Rank Lists and Valid Barcodes}
\label{valid}

Consider a fully-connected linear neural network specified by
the unit layer sizes $d_0, d_1, \ldots, d_L$, with
a corresponding weight space $\R^{d_\theta}$.
Recall that the rank list $\underline{r}$ of
a weight vector $\theta \in \R^{d_\theta}$ lists
every subsequence matrix of $\theta$,
including $\rank W_{j \ssq j} = d_j$ for all $j \in [0, L]$.

Recall that a rank list $\underline{r}$ is {\em valid} if
there is some weight vector $\theta \in \R^{d_\theta}$ that has
rank list~$\underline{r}$.
Recall that a barcode, represented by
interval counts $\omega_{ki}$, $L \geq k \geq i \geq 0$, is
{\em valid} for this network if
$d_j = \sum_{t = j}^L \sum_{s = 0}^j \omega_{ts}$ for every $j \in [0, L]$.
We also require that every $\omega_{ki}$ is nonnegative, but
that's part of the definition of ``barcode'' (as a barcode is a multiset),
not part of the definition of ``valid.''

The identities~(\ref{ranksum}) and~(\ref{omegarank}) map
barcodes to rank lists and vice versa.
When the rank list and barcode are properties of
a weight vector $\theta \in \R^{d_\theta}$,
the two identities are correct by Lemma~\ref{rankomega}, so
they must define a bijection between valid rank lists and barcodes
that arise from weight vectors.
However, we did not define a barcode to be ``valid'' if
it arises from a weight vector;
we opted for a definition that is easier to check.

Here we justify that definition by showing that
a barcode is valid if and only if it arises from a weight vector.
To do that, we show that~(\ref{ranksum}) maps
every valid barcode to a valid rank list, and
(\ref{omegarank})~maps every valid rank list to a valid barcode.
Consider the latter first.

\begin{lemma}
\label{validbarcode}
Let $\underline{r}$ be a valid rank list.
Then the identity~(\ref{omegarank}) yields a valid barcode
(where each unit layer size $d_j$ in the definition of ``valid barcode'' is
taken from $\underline{r}$).
\end{lemma}

\begin{proof}
As $\underline{r}$ is valid, there exists
a weight vector $\theta = (W_L, W_{L-1}, \ldots, W_1) \in \R^{d_\theta}$ with
rank list~$\underline{r}$.
The identity~(\ref{omegarank}) is
$\omega_{ki} = \rank W_{k \ssq i} - \rank W_{k \ssq i-1} -
\rank W_{k+1 \ssq i} + \rank W_{k+1 \ssq i-1}$
with the conventions that $W_{L+1} = 0$ and $W_0 = 0$, hence
every interval count $\omega_{ki}$ is nonnegative by
the Frobenius rank inequality (Sylvester's inequality if $k = i$).
This establishes that the interval counts specify a multiset. 

Corollary~\ref{dj} states that at $\theta$,
$d_j = \sum_{t = j}^L \sum_{s = 0}^j \omega_{ts}$ for every $j \in [0, L]$, where
$\omega_{ts}$ is the dimension of $a_{tss}$ in
any valid prebasis $\mathcal{A}_s$ at $\theta$.
By Lemma~\ref{rankomega},
this definition of $\omega_{ts}$ satisfies~(\ref{omegarank}).
Therefore, the interval counts calculated by~(\ref{omegarank}) from
$\underline{r}$ yield a valid barcode.
\end{proof}

Showing that~(\ref{ranksum}) maps every valid barcode to
a valid rank list is more interesting, because
we have to exhibit a weight vector with the specified rank list.
Fortunately, we constructed one in Section~\ref{canonical}.
Recall that for any valid barcode,
Section~\ref{canonical} describes a canonical weight vector
$\tilde{\theta} = (\tilde{I}_L, \tilde{I}_{L-1}, \ldots, \tilde{I}_1)$ that has
the specified interval counts.
We will omit the details of verifying that, so
the following proof is just a proof sketch.
There are times when a formal proof that a construction is correct would be
more cryptic than simply inspecting the construction until
you see why it's correct, and we think this is one of them.
Figure~\ref{ident} is the real proof:
you can just draw the bars in the barcode, then
write down the corresponding adjacency matrices.

\begin{lemma}
\label{validranklist}
Given a network with specified unit layer sizes $d_0, d_1, \ldots, d_L$,
suppose the interval counts $\omega_{ki}$, $L \geq k \geq i \geq 0$,
represent a valid barcode.
Let $\underline{r}$ be the rank list produced by the identity~(\ref{ranksum})
from those interval counts.
Then $\underline{r}$ is a valid rank list.
\end{lemma}

\begin{proof}
The canonical weight vector $\tilde{\theta}$ described in
Section~\ref{canonical}
has the specified barcode by construction.
By Lemma~\ref{rankomega}, the rank list $\underline{r}$ produced by
identity~(\ref{ranksum}) is the rank list of $\tilde{\theta}$.
Hence $\underline{r}$ is valid.
\end{proof}

Thus both definitions of ``valid'' imply that there exists
a weight vector $\theta \in \R^{d_\theta}$ with
the specified rank list or barcode.

\begin{cor}
\label{validiff}
Consider a network with specified unit layer sizes $d_0, d_1, \ldots, d_L$.
The interval counts $\omega_{ki}$, $L \geq k \geq i \geq 0$,
represent a valid barcode if and only if
there is some weight vector $\theta \in \R^{d_\theta}$ that has
those interval counts.
\end{cor}

\begin{proof}
Lemma~\ref{validbarcode} confirms the ``if'' and  
Lemma~\ref{validranklist} confirms the ``only if.''
\end{proof}

Let us turn to a closely related question.
Until now, we have been interested in whether there exists
a weight vector in $\R^{d_\theta}$ with
a specified rank list or a specified barcode
What if we restrict our attention to $\mu^{-1}(W)$?
Does there exist a weight vector on a particular fiber $\mu^{-1}(W)$ with
a specified rank list or a specified barcode?

For a specified valid rank list $\underline{r}$,
this is possible only if $\rank W = r_{L \ssq 0}$; and
for a specified valid barcode,
it is possible only if $\rank W = \omega_{L0}$ by~(\ref{Wrank}).
But if that constraint holds, then
there always exists a weight vector $\theta \in \mu^{-1}(W)$ with
the specified rank list or barcode.
An important consequence is that for any valid rank list~$\underline{r}$ and
matrix $W \in \R^{d_L \times d_0}$ with $\rank W = r_{L \ssq 0}$,
the stratum $S_{\underline{r}}$ in the rank stratification of $\mu^{-1}(W)$ is
nonempty.
We used this fact in the statement of Theorem~\ref{rankclosureequiv}.

We now show how to construct a suitable weight vector $\theta \in \mu^{-1}(W)$.
Our construction combines a singular value decomposition of $W$ with
the canonical weight vector $\tilde{\theta}$ for $\underline{r}$.

It is well known that $W$ has a singular value decomposition $W = UDV^\top$
such that $U \in \R^{d_L \times d_L}$ and $V \in \R^{d_0 \times d_0}$ are both
orthogonal matrices ($UU^\top = I = U^\top U$ and $VV^\top = I = V^\top V$) and
$D \in \R^{d_L \times d_0}$ is a diagonal (not necessarily square) matrix,
whose diagonal components are the singular values of $W$.
Moreover, we can choose the decomposition so that
the nonzero singular values come before the zeros on the diagonal of $D$.
Of these singular values, $\omega_{L0}$ are nonzero and the rest are zero
(because $\rank W = \omega_{L0}$).

The product $\tilde{I} = \mu(\tilde{\theta}) =
\tilde{I}_L \tilde{I}_{L-1} \cdots \tilde{I}_1$ is
a diagonal $d_L \times d_0$ matrix whose first $\omega_{L0}$ diagonal components
are~$1$'s, and all the other components are $0$'s.
Hence, the nonzero diagonal components of $\tilde{I}$ and~$D$ are
in the same positions.
Let $D'$ be a diagonal $d_0 \times d_0$ matrix
whose first $\omega_{L0}$ diagonal components are the same as~$D$'s
(the nonzero singular values), and
whose remaining diagonal components are $1$'s, making $D'$ invertible.
Then $\tilde{I} D' = D$ and
\[
W = UDV^\top = U \tilde{I} D' V^\top =
U \tilde{I}_L \tilde{I}_{L-1} \cdots \tilde{I}_2 \tilde{I}_1 D' V^\top.
\]
As $U$ and $D' V^\top$ are invertible, the weight vector
\[
\theta = (U \tilde{I}_L, \tilde{I}_{L-1}, \tilde{I}_{L-2}, \ldots,
          \tilde{I}_3, \tilde{I}_2, \tilde{I}_1 D' V^\top)
\]
has the same rank list as $\tilde{\theta}$ and satisfies $\mu(\theta) = W$.

The following two lemmas use this construction to show that
for either a valid rank list $\underline{r}$ with $r_{L \ssq 0} = \rank W$ or
a valid barcode with $\omega_{L0} = \rank W$,
the fiber $\mu^{-1}(W)$ contains
a point with that rank list or barcode.

\begin{lemma}
\label{nonemptystratum2}
Given a network with specified unit layer sizes $d_0, d_1, \ldots, d_L$,
suppose the interval counts $\omega_{ki}$, $L \geq k \geq i \geq 0$,
represent a valid barcode.
Let $\underline{r}$ be the rank list produced by the identity~(\ref{ranksum})
from those interval counts.
Let $W \in \R^{d_L \times d_0}$ be a matrix of rank~$\omega_{L0}$.
Then there is
a weight vector $\theta \in \mu^{-1}(W)$ with rank list $\underline{r}$.
\end{lemma}

\begin{proof}
The canonical weight vector $\tilde{\theta}$ described in
Section~\ref{canonical}
has the specified interval counts by construction.
By Lemma~\ref{rankomega}, the rank list $\underline{r}$ produced by
identity~(\ref{ranksum}) is the rank list of $\tilde{\theta}$.
As $\rank W = \omega_{L0}$,
the procedure described above constructs a weight vector $\theta$ having
the same rank list as $\tilde{\theta}$, namely $\underline{r}$,
such that $\mu(\theta) = W$.
\end{proof}

\begin{lemma}
\label{nonemptystratum}
Let $\underline{r}$ be a valid rank list and
let $W \in \R^{d_L \times d_0}$ be a matrix of rank~$r_{L \ssq 0}$.
Then there is a weight vector $\theta \in \mu^{-1}(W)$
with rank list $\underline{r}$.
Moreover, $S_{\underline{r}}$ is a nonempty stratum in  
the rank stratification of $\mu^{-1}(W)$. 
\end{lemma}

\begin{proof}
As $\underline{r}$ is valid, by definition
there exists a weight vector $\theta_{\underline{r}} \in \R^{d_\theta}$ with
rank list $\underline{r}$.
By Lemma~\ref{validbarcode}, the identity~(\ref{omegarank}) yields
a valid barcode corresponding to $\theta_{\underline{r}}$.
In particular, $\omega_{L0} = r_{L \ssq 0}$.
By Lemma~\ref{rankomega}, identity~(\ref{ranksum}) maps that barcode back to
the rank list of $\theta_{\underline{r}}$, producing $\underline{r}$.
As $\rank W = r_{L \ssq 0} = \omega_{L0}$,
by Lemma~\ref{nonemptystratum2} there is
a weight vector $\theta \in \mu^{-1}(W)$ with rank list $\underline{r}$.
It follows immediately that $S_{\underline{r}}$ is nonempty.
\end{proof}

\clearpage

\section{The Standard Prebases}
\label{standardapp}

Recall from Section~\ref{basisspaces} that the {\em standard prebases}
$\mathcal{A}_0, \ldots, \mathcal{A}_L, \mathcal{B}_0, \ldots, \mathcal{B}_L$
are defined by setting
\[
a_{kji} = A_{kji} \cap (A_{k,j,i-1} + A_{k-1,j,i})^\perp
\hspace{.2in}  \mbox{and}  \hspace{.2in}
b_{kji} = B_{kji} \cap (B_{k,j,i+1} + B_{k+1,j,i})^\perp,
\]
and can be written explicitly as~(\ref{standarda}) and~(\ref{standardb}).
It is not generally possible to have all the subspaces in
$\mathcal{A}_j$ be pairwise orthogonal
(for a counterexample, see $a_{211}$, $a_{410}$, and $a_{310}$ in
Figure~\ref{aflow}), nor all the subspaces in~$\mathcal{B}_j$.
There are four exceptions:  the subspaces in any of
$\mathcal{A}_0$, $\mathcal{A}_L$, $\mathcal{B}_0$, or $\mathcal{B}_L$ are
pairwise orthogonal.
Subspaces that meet at oblique angles occur only in the hidden layers
($\mathcal{A}_j$ and $\mathcal{B}_j$ for $j \in [1, L - 1]$).
For those layers,
the subspaces in the standard prebasis are as close to orthogonal as possible,
which is good for the numerical stability of algorithms that use them.

The standard prebases have the property that
$\mathcal{A}_0 = \mathcal{B}_0$ and $\mathcal{A}_L = \mathcal{B}_L$, because
for all~$k, i \in [0, L]$,
\begin{align*}
& a_{k00} = A_{k00} \cap A_{k-1,0,0}^\perp
= \Null W_{k+1 \ssq 0} \cap \row W_{k \ssq 0} = B_{k+1,0,0}^\perp \cap B_{k00}
= b_{k00}  \hspace{.2in}  \mbox{and}  \\
& b_{LLi} = B_{LLi} \cap B_{L,L,i+1}^\perp
= \Null W_{L \ssq i-1}^\top \cap \col W_{L \ssq i} = A_{L,L,i-1}^\perp \cap A_{LLi}
= a_{LLi}.
\end{align*}
However, in the hidden layers,
$\mathcal{A}_j$ and $\mathcal{B}_j$ can be entirely different.

Unfortunately, the standard prebases are typically not flow prebases.
%
But at the canonical weight vector~$\tilde{\theta}$
(introduced in Section~\ref{canonical}),
the standard prebases {\em are} flow prebases.
The standard prebases at $\tilde{\theta}$ have other special properties too.
We say that a subspace of $\R^d$ is {\em axis-aligned} if it is spanned by
some of the unit coordinate vectors of $\R^d$.
Recall that the almost-identity matrices, including
the factor matrices $\tilde{I}_j$ and
the subsequence matrices $\tilde{I}_{t \ssq s}$, have the property that
every row and every column has at most one nonzero component, always a $1$.
Hence all subspaces of the forms $\row \tilde{I}_{t \ssq s}$,
$\Null \tilde{I}_{t \ssq s}$, $\col \tilde{I}_{t \ssq s}$, and
$\Null \tilde{I}_{t \ssq s}^\top$ are axis-aligned.
Hence all subspaces of the forms $A_{kji}$ and $B_{kji}$ are axis-aligned.
Hence in the standard prebases,
all subspaces of the forms $a_{kji}$ and $b_{kji}$ are axis-aligned.
As the subspaces in each standard prebasis $\mathcal{A}_j$ are
linearly independent and axis-aligned, they are pairwise orthogonal.
(Likewise for $\mathcal{B}_j$.)

\clearpage

\section{The Fundamental Subspaces of the Differential Map}
\label{funddmu}

The differential map $\dmu(\theta)$ is a linear operator that maps
a weight displacement $\Delta \theta$ in a finite vector space~$\R^{d_\theta}$
to a matrix $\Delta W = \dmu(\theta)(\Delta \theta)$
in another finite vector space $\R^{d_L \times d_0}$.
One could simply write this linear map as a matrix---a fact that is
obscured because the range of $\dmu(\theta)$ is a set of matrices.
But if we imagine that the range of $\dmu(\theta)$ is $\R^{d_L d_0}$ instead of
$\R^{d_L \times d_0}$, we can represent $\dmu(\theta)$ as
a $(d_L d_0) \times d_\theta$ matrix $M$.
Then we can apply basic ideas from linear algebra.

The four fundamental subspaces of $M$ are
its nullspace, its columnspace (also called its image),
its rowspace, and its left nullspace, defined to be
\begin{eqnarray*}
\Null M & = & \{ v \in \R^{d_\theta} : Mv = {\bf 0} \},  \\
\col M = \image M & = & \{ Mv : v \in \R^{d_\theta} \},  \\
\row M = \image M^\top & = & \{ M^\top w : w \in \R^{d_L d_0} \},
\hspace{.2in}  \mbox{and}  \\
\Null M^\top & = & \{ w \in \R^{d_L d_0} : M^\top w = {\bf 0} \}.
\end{eqnarray*}
Following our convention in this paper for multiplying a matrix by a subspace,
we can write the simpler definitions
$\col M = M \R^{d_\theta}$ and $\row M = M^\top \R^{d_L d_0}$.

The differential map $\dmu(\theta)$ has the same four fundamental subspaces,
except that its columnspace (image) and left nullspace are
subspaces of $\R^{d_L \times d_0}$ instead of $\R^{d_L d_0}$.
As $\dmu(\theta)$ is a linear map, it also has a transpose,
which we write as $\dmu^\top(\theta)$.
Considering that $M$ represents
the same linear transformation as $\dmu(\theta)$,
let $\dmu^\top(\theta)$ be the linear map from $\R^{d_L \times d_0}$ to
$\R^{d_\theta}$ represented by $M^\top$.
Recall from~(\ref{dmu}) that, given a weight displacement
$\Delta \theta = (\Delta W_L, \Delta W_{L-1}, \ldots, \Delta W_1)$,
\[
\dmu(\theta)(\Delta \theta) =
\sum_{j=1}^L W_{L \ssq j} \Delta W_j W_{j-1 \ssq 0}.
\]
Clearly this transformation is linear in~$\Delta \theta$, but
it isn't written as a multiplication by a single matrix $M$.
Nevertheless, we can write its transpose (analogous to $M^\top$),
applied to a matrix $\Delta W$, as
\[
\dmu^\top(\theta)(\Delta W) = (\Delta W W_{L-1 \ssq 0}^\top, \ldots,
W_{L \ssq j}^\top \Delta W W_{j-1 \ssq 0}^\top, \ldots, W_{L \ssq 1}^\top \Delta W).
\]

The nullspace, columnspace (image), rowspace, and left nullspace of
$\dmu(\theta)$ are
\begin{eqnarray*}
\Null \dmu(\theta) & = &
  \{ \Delta \theta \in \R^{d_\theta} : \dmu(\theta)(\Delta \theta) = 0 \}
  = \Null M,  \\
\col \dmu(\theta) = \image \dmu(\theta) & = &
  \{ \dmu(\theta)(\Delta \theta) : \Delta \theta \in \R^{d_\theta} \},  \\
\row \dmu(\theta) = \image \dmu^\top(\theta) & = &
  \{ \dmu^\top(\theta)(\Delta W) : \Delta W \in \R^{d_L \times d_0} \}
  = \row M,
\hspace{.2in}  \mbox{and}  \\
\Null \dmu^\top(\theta) & = &
  \{ \Delta W \in \R^{d_L \times d_0} : \dmu^\top(\theta)(\Delta W) = {\bf 0} \}.
\end{eqnarray*}
The only difference between $\col \dmu(\theta)$ and $\col M$ is whether
it is formatted as a set of $d_L \times d_0$ matrices or
a set of vectors of length $d_L d_0$.
Likewise for the difference between $\Null \dmu^\top(\theta)$ and $\Null M^\top$.

We now apply some basic facts from linear algebra to the differential map.
\begin{itemize}
\item
The fundamental subspaces $\col \dmu(\theta)$ and $\row \dmu(\theta)$ have
the same dimension, called the {\em rank} of $\dmu(\theta)$ and denoted
$\rank \dmu(\theta)$.
(The rank $r = \rank \dmu(\theta)$ can also be identified by the fact that
$M$ has an $r \times r$ minor with a nonzero determinant but
every $(r+1) \times (r+1)$ minor has determinant zero.)
\item
The rowspace of $\dmu^\top(\theta)$ is the columnspace of $\dmu(\theta)$ and
the columnspace of $\dmu^\top(\theta)$ is the rowspace of $\dmu(\theta)$, so
$\rank \dmu^\top(\theta) = \rank \dmu(\theta)$.
\item
$\row \dmu(\theta)$ is the orthogonal complement of $\Null \dmu(\theta)$
(in the space $\R^{d_\theta}$).
By the Rank-Nullity Theorem,
$\rank \dmu(\theta) + \dim \Null \dmu(\theta) = d_\theta$.
\item
$\col \dmu(\theta)$ is the orthogonal complement of $\Null \dmu^\top(\theta)$
(in the space $\R^{d_L \times d_0}$;
the orthogonality is in the Frobenius inner product).
By the Rank-Nullity Theorem applied to $\dmu^\top(\theta)$,
$\rank \dmu(\theta) + \dim \Null \dmu^\top(\theta) = d_L d_0$.
\end{itemize}

A major concern of this paper is to determine
$\Null \dmu(\theta)$ (Section~\ref{proofbasis}) and
$\row \dmu(\theta)$ (Section~\ref{normal1}).
Trager, Kohn, and Bruna~\cite{trager20} are more interested in
$\col \dmu(\theta)$, from which they derive $\rank \dmu(\theta)$,
which we use in Section~\ref{proofbasis} (Theorem~\ref{freedomspan}) to show
that our expression for $\Null \dmu(\theta)$ is correct---specifically,
that we did not overlook any dimensions of $\Null \dmu(\theta)$.
We reprise their derivations of
$\col \dmu(\theta)$ and $\rank \dmu(\theta)$ here (with some changes), and
for completeness we also derive $\Null \dmu^\top(\theta)$,
the orthogonal complement of $\col \dmu(\theta)$.

Recall our conventions that
$W_{j \ssq j}$ is the $d_j \times d_j$ identity matrix,
$\col W_{L \ssq L} = \R^{d_L}$, $\rank W_{L \ssq L} = d_L$,
$\row W_{0 \ssq 0} = \R^{d_0}$, and $\rank W_{0 \ssq 0} = d_0$.
Consult Figure~\ref{coldmufig} for a better understanding of
the expressions in Lemmas~\ref{coldmu}, \ref{rkdmu}, \ref{rkdmuomega},
and~\ref{nulldmut}.

\begin{lemma}
\label{coldmu}
\begin{equation}
\col \dmu(\theta) = \sum_{j=1}^L \col W_{L \ssq j} \otimes \row W_{j-1 \ssq 0}.
\label{imdmu}
\end{equation}
\end{lemma}

\begin{proof}
Recall again that $\dmu(\theta)(\Delta \theta) =
\sum_{j=1}^L W_{L \ssq j} \Delta W_j W_{j-1 \ssq 0}$.
Let $(\Delta W_j)_{st}$ denote
the component in row $s$ and column $t$ of $\Delta W_j$,
let $C_s$ be column $s$ of $W_{L \ssq j}$ (expressed as a column vector), and
let $R_t$ be row $t$ of $W_{j-1 \ssq 0}$ (expressed as a row vector).
Then $W_{L \ssq j} \Delta W_j W_{j-1 \ssq 0} =
\sum_s \sum_t (\Delta W_j)_{st} C_s R_t$, which
pairs each column of $W_{L \ssq j}$ with each row of $W_{j-1 \ssq 0}$ to form
an outer product matrix, with an independent coefficient for each pairing.
It follows that the term $W_{L \ssq j} \Delta W_j W_{j-1 \ssq 0}$ can take on
any value in $\col W_{L \ssq j} \otimes \row W_{j-1 \ssq 0}$
(independently of the other terms in the summation), so
the image of $\dmu(\theta)$ is
$\sum_{j=1}^L \col W_{L \ssq j} \otimes \row W_{j-1 \ssq 0}$.
\end{proof}

\begin{lemma}[Trager, Kohn, and Bruna~\cite{trager20}, Lemma~3]
\label{rkdmu}
\[
\rank \dmu(\theta) =
\sum_{j=1}^L \rank W_{L \ssq j} \cdot \rank W_{j-1 \ssq 0} -
\sum_{j=1}^{L-1} \rank W_{L \ssq j} \cdot \rank W_{j \ssq 0}.
\]
\end{lemma}

\begin{proof}
The dimension of each summand in~(\ref{imdmu}) is
$\rank W_{L \ssq j} \cdot \rank W_{j-1 \ssq 0}$, but the summands overlap.
For any two subspaces $Y$ and $Z$,
$\dim (Y + Z) = \dim Y + \dim Z - \dim (Y \cap Z)$.
When the summand $\col W_{L \ssq j} \otimes \row W_{j-1 \ssq 0}$ is added
(for a specified value of $j$),
its intersection with the vector sum of the previous summands
(for smaller values of $j$) is $\col W_{L \ssq j-1} \otimes \row W_{j-1 \ssq 0}$,
so by induction the dimension of $\col \dmu(\theta)$ is
\begin{eqnarray*}
\rank \dmu(\theta)
& = &
\rank W_{L \ssq 1} \cdot \rank W_{0 \ssq 0} +
\sum_{j=2}^L \left( \rank W_{L \ssq j} \cdot \rank W_{j-1 \ssq 0} -
                   \rank W_{L \ssq j-1} \cdot \rank W_{j-1 \ssq 0} \right)  \\
& = &
\sum_{j=1}^L \rank W_{L \ssq j} \cdot \rank W_{j-1 \ssq 0} -
\sum_{j=1}^{L-1} \rank W_{L \ssq j} \cdot \rank W_{j \ssq 0}.
\qedhere
\end{eqnarray*}
\end{proof}

\begin{lemma}
\label{rkdmuomega}
The rank of $\dmu(\theta)$ can also be written
\[
\rank \dmu(\theta) = \sum_{k,i \in [0,L], k+1 \geq i} \omega_{Li} \, \omega_{k0}.
\]
\end{lemma}

\begin{proof}
By Lemma~\ref{rkdmu} and the identity~(\ref{ranksum}), we have
\begin{eqnarray*}
\rank \dmu(\theta) & = &
\sum_{j=1}^L \rank W_{L \ssq j} \cdot \rank W_{j-1 \ssq 0}
- \sum_{j=1}^{L-1} \rank W_{L \ssq j} \cdot \rank W_{j \ssq 0}  \\
& = &
\rank W_{L \ssq L} \cdot \rank W_{L-1 \ssq 0}
+ \sum_{j=1}^{L-1} \rank W_{L \ssq j} \cdot
  (\rank W_{j-1 \ssq 0} - \rank W_{j \ssq 0})  \\
& = &
\rank W_{L \ssq L} \cdot (\omega_{L0} + \omega_{L-1,0})
+ \sum_{j=1}^{L-1} \rank W_{L \ssq j} \cdot
  \left( \sum_{k=j-1}^L \omega_{k0} - \sum_{k=j}^L \omega_{k0} \right)  \\
& = &
\rank W_{L \ssq L} \cdot \omega_{L0}
+ \sum_{j=1}^L \rank W_{L \ssq j} \cdot \omega_{j-1,0}  \\
& = &
\omega_{L0} \sum_{i=0}^L \omega_{Li}
+ \sum_{j=1}^L \omega_{j-1,0} \sum_{i=0}^j \omega_{Li}  \\
& = &
\omega_{L0} \sum_{i=0}^L \omega_{Li}
+ \sum_{k=0}^{L-1} \omega_{k0} \sum_{i=0}^{k+1} \omega_{Li}  \\
& = & \sum_{k,i \in [0,L], k+1 \geq i} \omega_{Li} \, \omega_{k0}.  \qedhere
\end{eqnarray*}
\end{proof}

\begin{cor}
\label{nulldimcor}
The dimension of $\Null \dmu(\theta)$ is $D^\free$,
as specified by~(\ref{Dfreedom})
\end{cor}

\begin{proof}
By the Rank-Nullity Theorem,
$\dim \Null \dmu(\theta) = d_\theta - \rank \dmu(\theta) = D^\free$.
\end{proof}

\begin{lemma}
\label{nulldmut}
\begin{equation}
\label{nulldmuteq}
\Null \dmu^\top(\theta) =
\sum_{j=1}^{L-1} \Null W_{L \ssq j}^\top \oplus \Null W_{j \ssq 0}.
\end{equation}
\end{lemma}

\begin{proof}
Lemma~\ref{coldmu} gives an expression~(\ref{imdmu}) for $\col \dmu(\theta)$;
we show that~(\ref{nulldmuteq}) is its orthogonal complement.
To see that the two expressions are orthogonal,
observe that for any $j \in [1, L-1]$ and any $j' \in [1, L]$,
$\Null W_{L \ssq j}^\top \oplus \Null W_{j \ssq 0}$ is orthogonal to
$\col W_{L \ssq j'} \otimes \row W_{j'-1 \ssq 0}$ because
$\Null W_{L \ssq j}^\top$ is orthogonal to $\col W_{L \ssq j'}$ if $j \geq j'$ and
$\Null W_{j \ssq 0}$ is orthogonal to $\row W_{j'-1 \ssq 0}$ if $j < j'$.

So~(\ref{nulldmuteq}) and~(\ref{imdmu}) are orthogonal, but
are they orthogonal complements?
Lemma~\ref{nulldmute} shows that the right-hand side of~(\ref{nulldmuteq}) is
equal to $\sum_{k=0}^{L-2} \sum_{i=k+2}^L e_{ki}$.
As the subsequence subspaces $e_{ki}$ are linearly independent and
$e_{ki}$ has dimension $\omega_{Li} \, \omega_{k0}$,
the dimension of the right-hand side of~(\ref{nulldmuteq}) is
$\sum_{k,i \in [0,L], k+1 < i} \omega_{Li} \, \omega_{k0}$.
By Lemma~\ref{rkdmuomega}, the sum of the dimension of $\col \dmu(\theta)$ and
the dimension of the right-hand side of~(\ref{nulldmuteq}) is
$\sum_{k,i \in [0,L]} \omega_{Li} \, \omega_{k0} = d_L d_0$, which is
the dimension of the ambient space $\R^{d_L \times d_0}$.
So yes, the expressions~(\ref{nulldmuteq}) and~(\ref{imdmu}) are
orthogonal complements.
Therefore, the right-hand side of~(\ref{nulldmuteq}) is
the orthogonal complement of $\col \dmu(\theta)$,
which is $\Null \dmu^\top(\theta)$.
\end{proof}

\clearpage

\section{An Affine Transformation of the Nullspace of the Differential Map}
\label{transnulldmu}

Consider a point $\theta$ on a fiber $\mu^{-1}(W)$ and
a set of matrices $J_0, J_1, \ldots, J_L$ representing a set of flow bases
at $\theta$.
Recall from Section~\ref{sec:affinestrata} the function
$\eta : (M_L, M_{L-1}, \ldots, M_1) \mapsto
(J_L M_L J_{L-1}^{-1}, J_{L-1} M_{L-1} J_{L-2}^{-1}, \ldots, J_1 M_1 J_0^{-1})$,
an invertible linear transformation of weight space that maps $\theta$ to
the canonical weight vector $\tilde{\theta}$ with
the same rank list as $\theta$.
Recall Lemma~\ref{affinestratalem}, which shows that
$\eta(\mu^{-1}(\tilde{I})) = \mu^{-1}(W)$.
We now show that $\eta$ also maps the nullspace of $\dmu(\tilde{\theta})$ to
the nullspace of $\dmu(\theta)$.
This is not very surprising, as invertible affine transformations preserve
tangencies; but it's nice to have a formal proof.

\begin{lemma}
$\eta \left(\Null \dmu(\tilde{\theta}) \right) = \Null \dmu(\theta)$.
\label{affinestratalem2}
\end{lemma}

\begin{proof}
Recall that $\tilde{I}_j = J_j^{-1} W_j J_{j-1}$ by Lemma~\ref{basesflow}, so
$\tilde{I}_{k \ssq i} = \tilde{I}_k  \tilde{I}_{k-1} \cdots \tilde{I}_{i+1} =
J_k^{-1} W_{k \ssq i} J_i$.
Recall the formula (\ref{dmu}) for $\dmu(\theta)$.
Substituting in $\tilde{\theta}$ gives
\[
\dmu(\tilde{\theta})(\Delta \theta) =
\sum_{j=1}^L \tilde{I}_{L \ssq j} \Delta W_j \tilde{I}_{j-1 \ssq 0} =
J_L^{-1} \left(
\sum_{j=1}^L W_{L \ssq j} J_j \Delta W_j J_{j-1}^{-1} W_{j-1 \ssq 0}
\right) J_0 =
J_L^{-1} \, \dmu(\theta)(\eta(\Delta \theta)) \, J_0.
\]
For any $\Delta \theta \in \R^{d_\theta}$,
$\Delta \theta \in \Null \dmu(\tilde{\theta})$ means that
$\dmu(\tilde{\theta})(\Delta \theta) = 0$, which is true if and only if
$\dmu(\theta)(\eta(\Delta \theta)) = 0$, or equivalently,
$\eta(\Delta \theta) \in \Null \dmu(\theta)$.
Therefore, $\eta \left(\Null \dmu(\tilde{\theta}) \right) = \Null \dmu(\theta)$
as claimed.
\end{proof}

\clearpage

\section{Counting More Dimensions}
\label{countingmore}

Table~\ref{subspacesets2} defines several sets of subspaces that were
not important enough to include in Section~\ref{counting}, and
the dimensions of the subspaces (of $\R^{d_\theta}$) they span.

\begin{table}
\begin{center}
\begin{tabular}{l}
\hline
$\ThetaO^\conn = \{ \phi_{l,j-1,j,j,h} \neq \{ {\bf 0} \} :
L \geq l \geq j > h \geq 0 \}$
\rule{0pt}{13pt}  \\
\begin{minipage}{6.3in}
\vspace*{-.12in}
\begin{equation}  \lm
\DO^\conn = \sum_{L \geq l \geq j > h \geq 0} \omega_{lj} \, \omega_{j-1,h}
          = \sum_{j=1}^L \underbrace{(d_j - \rank W_j)}_{\beta_{jjj}} \,
                        \underbrace{(d_{j-1} - \rank W_j)}_{\alpha_{j-1,j-1,j-1}}.
\label{DOconn}
\end{equation}
\end{minipage}  \\
$\ThetaO^\swap = \{ \phi_{lkjih} \in \ThetaO : l > k \geq i > h \}
           = \{ \phi_{lkjih} \neq \{ {\bf 0} \} :
                L \geq l > k \geq i > h \geq 0$ and $k + 1 \geq j \geq i \}$  \\
\begin{minipage}{6.3in}
\vspace*{-.14in}
\begin{equation}  \lm
\DO^\swap =
\sum_{L > k \geq i > 0} (k - i + 2) \,
  \underbrace{(\rank W_{k+1 \ssq i} - \rank W_{k+1 \ssq i-1})}_{\beta_{k+1,i,i}} \,
  \underbrace{(\rank W_{k \ssq i-1} - \rank W_{k+1 \ssq i-1})}_{\alpha_{k,k,i-1}}
\label{DOswap}
\end{equation}
\end{minipage}  \\
$\ThetaO^{\off,\barc} = \ThetaO^\off \cap \ThetaO^\barc
                    = \{ \phi_{Lkji0} \neq \{ {\bf 0} \} :
                         L \geq k + 1 \geq j \geq i > 0 \}$  \\
\begin{minipage}{6.3in}
\vspace*{-.12in}
\begin{equation}  \lm
\DO^{\off,\barc} =
\sum_{L \geq k+1 \geq i > 0} (k - i + 2) \, \omega_{Li} \, \omega_{k0}
\rule[-18pt]{0pt}{11pt}
\label{DOoffcomb}
\end{equation}
\end{minipage}  \\
$\ThetaO^{\fiber,\barc} = \ThetaO^\fiber \cap \ThetaO^\barc
                        = \ThetaO^\barc \setminus \ThetaO^\off
                        = \{ \phi_{lkjih} \in \ThetaO^\barc :
                             L > l$ or $h > 0 \}$  \\
\begin{minipage}{6.3in}
\vspace*{-.14in}
\begin{equation}  \lm
\DO^{\fiber,\barc} = \DO^{\barc} - \DO^{\off,\barc} =
\sum_{L \geq k+1 \geq i > 0} (k - i + 2) \,
(\beta_{k+1,i,i} \, \alpha_{k,k,i-1} - \omega_{Li} \, \omega_{k0})
\rule[-16pt]{0pt}{11pt}
\label{DOfibercomb}
\end{equation}
\end{minipage}  \\
$\ThetaO^{\off,\conn} = \ThetaO^\off \cap \ThetaO^\conn =
\{ \phi_{L,j-1,j,j,0} \neq \{ {\bf 0} \} : L \geq j > 0 \}$  \\
\begin{minipage}{6.3in}
\vspace*{-.12in}
\begin{equation}  \lm
\DO^{\off,\conn} =
\sum_{j=1}^L \omega_{Lj} \, \omega_{j-1,0} =
\sum_{j=1}^L
  \underbrace{(\rank W_{L \ssq j} - \rank W_{L \ssq j-1})}_{\omega_{Lj}} \,
  \underbrace{(\rank W_{j-1 \ssq 0} - \rank W_{j \ssq 0})}_{\omega_{j-1,0}}
\rule[-18pt]{0pt}{11pt}
\label{DOoffconn}
\end{equation}
\end{minipage}  \\
$\ThetaO^{\fiber,\conn} = \ThetaO^\fiber \cap \ThetaO^\conn
                        = \ThetaO^\conn \setminus \ThetaO^\off
                        = \{ \phi_{l,j-1,j,j,h} \in \ThetaO^\conn :
                             L > l$ or $h > 0 \}$  \\
\begin{minipage}{6.3in}
\vspace*{-.12in}
\begin{equation}  \lm
\DO^{\fiber,\conn} =  
\sum_{j=1}^L \left(
  \underbrace{(d_j - \rank W_j)}_{\beta_{jjj}} \,
  \underbrace{(d_{j-1} - \rank W_j)}_{\alpha_{j-1,j-1,j-1}} -
  \underbrace{(\rank W_{L \ssq j} - \rank W_{L \ssq j-1})}_{\omega_{Lj}} \,
  \underbrace{(\rank W_{j-1 \ssq 0} - \rank W_{j \ssq 0})}_{\omega_{j-1,0}} \right)
\rule[-22pt]{0pt}{11pt}
\label{DOfiberconn}
\end{equation}
\end{minipage}  \\
\hline
$\ThetaT^{\off,\barc} = \ThetaT^\off \cap \ThetaT^\barc
                    = \{ \tau_{Lkji0} \neq \{ {\bf 0} \} :
                         L > k \geq j \geq i > 0 \}$
\rule{0pt}{13pt}  \\
\begin{minipage}{6.3in}
\vspace*{-.12in}
\begin{equation}  \lm
\DT^{\off,\barc} =
\sum_{L > k \geq i > 0} (k - i + 1) \, \omega_{Li} \, \omega_{k0}
\rule[-18pt]{0pt}{11pt}
\label{DToffcomb}
\end{equation}
\end{minipage}  \\
$\ThetaT^{\fiber,\barc} = \ThetaT^\barc \setminus \ThetaT^\off
                        = \{ \tau_{lkjih} \in \ThetaT^\barc :
                             L > l$ or $h > 0 \}$  \\
\begin{minipage}{6.3in}
\vspace*{-.14in}
\begin{equation}  \lm
\DT^{\fiber,\barc} = \DT^{\barc} - \DT^{\off,\barc} =
\sum_{L > k \geq i > 0} (k - i + 1) \,
(\beta_{k+1,i,i} \, \alpha_{k,k,i-1} - \omega_{Li} \, \omega_{k0})
\rule[-16pt]{0pt}{11pt}
\label{DTfibercomb}
\end{equation}
\end{minipage}  \\
\hline
\end{tabular}
\end{center}

\caption{\label{subspacesets2}
More sets of subspaces of $\R^{d_\theta}$ and their total dimensions.
See also Table~\ref{subspacesets}.
}
\end{table}

Let's count the degrees of freedom of connecting moves.
Recall from Section~\ref{barcmoves} that
the set of subspaces associated with connecting moves is
$\ThetaO^\conn = \{ \phi_{l,j-1,j,j,h} \neq \{ {\bf 0} \} :
L \geq l \geq j > h \geq 0 \}$.
First we count the degrees of freedom associated with connecting moves
that change one particular matrix $W_j$:
\[
D_j^\conn =
\left( \sum_{l = j}^L \omega_{lj} \right) \,
\left( \sum_{h = 0}^{j-1} \omega_{j-1,h} \right) =
\beta_{jjj} \, \alpha_{j-1,j-1,j-1} =
(d_j - \rank W_j) \, (d_{j-1} - \rank W_j).
\]
The space spanned by $\ThetaO^\conn$ has dimension
$\DO^\conn = \sum_{j=1}^L D_j^\conn$.
Thus we obtain the formula~(\ref{DOconn}) in Table~\ref{subspacesets2}.

Let's count the degrees of freedom of swapping moves.
Recall from Section~\ref{barcmoves} that
the set of subspaces associated with swapping moves is
$\ThetaO^\swap = \{ \phi_{lkjih} \in \ThetaO : l > k \geq i > h \}$.
The dimension $\DO^\swap$ of the space spanned by $\ThetaO^\swap$ can be
derived exactly as we derived $\DO^\barc$ in Section~\ref{counting}, except that
we omit the subspaces where $k + 1 = i$, which indicate
connecting moves and not swapping moves.
Thus we obtain the formula~(\ref{DOswap}) (see Table~\ref{subspacesets2}).

Let's count the degrees of freedom of
the barcode moves that don't change~$\mu(\theta)$.
These discrete moves move from one stratum to a different stratum of
the same fiber.
These moves are represented by the prebasis
$\ThetaO^{\fiber,\barc} = \ThetaO^\fiber \cap \ThetaO^\barc$.
The easiest way to determine the dimension of the subspace spanned by
$\ThetaO^{\fiber,\barc}$ is to first understand the prebasis
$\ThetaO^{\off,\barc} = \ThetaO^\off \cap \ThetaO^\barc$, which represents
barcode moves that change~$\mu(\theta)$ (don't stay on the fiber).
As $\ThetaO^{\off,\barc} = \{ \phi_{Lkji0} \neq \{ {\bf 0} \} :
L \geq k + 1 \geq j \geq i > 0 \}$,
it spans a subspace of dimension
\[
\DO^{\off,\barc}
= \sum_{L \geq k+1 \geq j \geq i > 0} \omega_{Li} \, \omega_{k0}
= \sum_{L \geq k+1 \geq i > 0} (k - i + 2) \, \omega_{Li} \, \omega_{k0},
\]
because the term $\omega_{Li} \, \omega_{k0}$ appears in
the first summation once for each $j \in [ i, k + 1 ]$.
Analogously, for the set
$\ThetaT^{\off,\barc} = \ThetaT^\off \cap \ThetaT^\barc$ of two-matrix subspaces,
\[
\DT^{\off,\barc}
= \sum_{L > k \geq j \geq i > 0} \omega_{Li} \, \omega_{k0}
= \sum_{L > k \geq i > 0} (k - i + 1) \, \omega_{Li} \, \omega_{k0}.
\]

The dimension of the space spanned by $\ThetaO^{\fiber,\barc}$ is
\begin{eqnarray*}
\DO^{\fiber,\barc}
& = & \dim \ThetaO^{\fiber,\barc}  \\
& = & \dim \left( \ThetaO^\barc \setminus \ThetaO^\off \right)  \\
& = & \dim \left( \ThetaO^\barc \setminus \ThetaO^{\off,\barc} \right)  \\
& = & \DO^\barc - \DO^{\off,\barc},
\end{eqnarray*}
from which we obtain the formula~(\ref{DOfibercomb})
in Table~\ref{subspacesets2}.
Analogously, we define
$\ThetaT^{\fiber,\barc} = \ThetaT^\barc \setminus \ThetaT^\off$,
which spans a space of dimension
$\DT^{\fiber,\barc} = \DT^\barc - \DT^{\off,\barc}$,
from which we obtain the formula~(\ref{DTfibercomb}).



A connecting move that changes $W_j$ also changes $\mu(\theta)$ if and only if
$l = L$ and $h = 0$; that is,
$\Delta \theta \in \phi_{L,j-1,j,j,0} \setminus \{ {\bf 0} \}$.
The total degrees of freedom of
the connecting moves that change both $W_j$ and $\mu(\theta)$ is
\[
D_j^{\off,\conn} = \omega_{Lj} \, \omega_{j-1,0} =
(\rank W_{L \ssq j} - \rank W_{L \ssq j-1}) \,
(\rank W_{j-1 \ssq 0} - \rank W_{j \ssq 0}).
\]

The dimension of the space spanned by
$\ThetaO^{\off,\conn} = \ThetaO^\off \cap \ThetaO^\conn$ is
\[
\DO^{\off,\conn} = \sum_{j=1}^L D_j^{\off,\conn} =
      \sum_{j=1}^L \omega_{Lj} \, \omega_{j-1,0} =
      \sum_{j=1}^L (\rank W_{L \ssq j} - \rank W_{L \ssq j-1}) \,
      (\rank W_{j-1 \ssq 0} - \rank W_{j \ssq 0}).
\]

Let $\ThetaO^{\fiber,\conn} = \ThetaO^\fiber \cap \ThetaO^\conn =
\ThetaO^\conn \setminus \ThetaO^\off$ represent
the connecting moves that stay on the fiber.
The dimension of the space spanned by $\ThetaO^{\fiber,\conn}$ is
\begin{eqnarray*}
\DO^{\fiber,\conn} & = & \DO^\conn - \DO^{\off,\conn}  \\
& = & \sum_{j=1}^L \left( (d_j - \rank W_j) \, (d_{j-1} - \rank W_j) -
  (\rank W_{L \ssq j} - \rank W_{L \ssq j-1}) \,
  (\rank W_{j-1 \ssq 0} - \rank W_{j \ssq 0}) \right).
\end{eqnarray*}

\clearpage

\bibliographystyle{jrs}
\bibliography{nfactor}

\end{document}